%% file: masterdocument.tex
\documentclass[a4paper,11pt]{ucthesis}

\usepackage{xcolor}
\usepackage{soul}
\usepackage{graphicx}
\usepackage{float}
\usepackage{csquotes}
\usepackage[backend=biber,style=ieee, maxbibnames=99]{biblatex}
\usepackage{todonotes}
\usepackage{bm}

\usepackage[final]{pdfpages}
\usepackage{tikz}
\usepackage{amsmath}
\usepackage{mathtools}
\usepackage{geometry}
\usepackage{epsfig}
\usepackage{amsmath}
\usepackage{amssymb}
\usepackage{graphicx}
\usepackage{longtable}
\usepackage[format=plain,justification=justified,labelfont=bf,font=small]{caption}

\usepackage{boxedminipage}
\usepackage{fancybox}
\usepackage{color}
\usepackage{ifthen}
\usepackage{fancyhdr}
\usepackage{lscape}
\usepackage{caption}
\usepackage{subcaption}
\usepackage{float}
\usepackage{flafter}
\usepackage{appendix}
\usepackage{lastpage}
\usepackage{hyperref}
\usepackage{mathrsfs}
\usepackage{multirow}
\usepackage{adjustbox}
\usepackage{threeparttable}
\usepackage[all]{hypcap} 
\usepackage{appendix}
\usepackage[printonlyused]{acronym}
\usepackage{textcomp}
\usepackage{afterpage}
\usepackage{verbatim}
\DeclareMathOperator*{\argmax}{arg\,max}
\DeclareMathOperator*{\argmin}{arg\,min}
\usepackage{footnote} 
\usepackage{textcomp} 

\usepackage[final]{template_files/sty/listofsymbols} 
\usepackage{template_files/sty/orcidlink}
\usepackage{template_files/sty/algorithm}
\usepackage{template_files/sty/algorithmic}
\usepackage{template_files/sty/booktabs} 

\hypersetup{
    unicode=false,           
    pdftoolbar=true,         
    pdfmenubar=true,         
    pdffitwindow=true,       
    pdftitle={On margin-based generalization prediction in deep neural networks - C Mouton},    
    pdfauthor={Coenraad Mouton},     
    pdfsubject={On margin-based generalization prediction in deep neural networks},   
    pdfnewwindow=true,       
    colorlinks=true,         
    linkcolor=black,         
    citecolor=black,         
    filecolor=black,         
    urlcolor=black           
}

\renewcommand{\title}{On margin-based generalization prediction \\ in deep neural networks}
\renewcommand{\author}{C. Mouton}

\begin{document}


\pagenumbering{roman}
\cfoot{}


\setboolean{@twoside}{false}
\includepdf[pages=-, offset=0 -0]{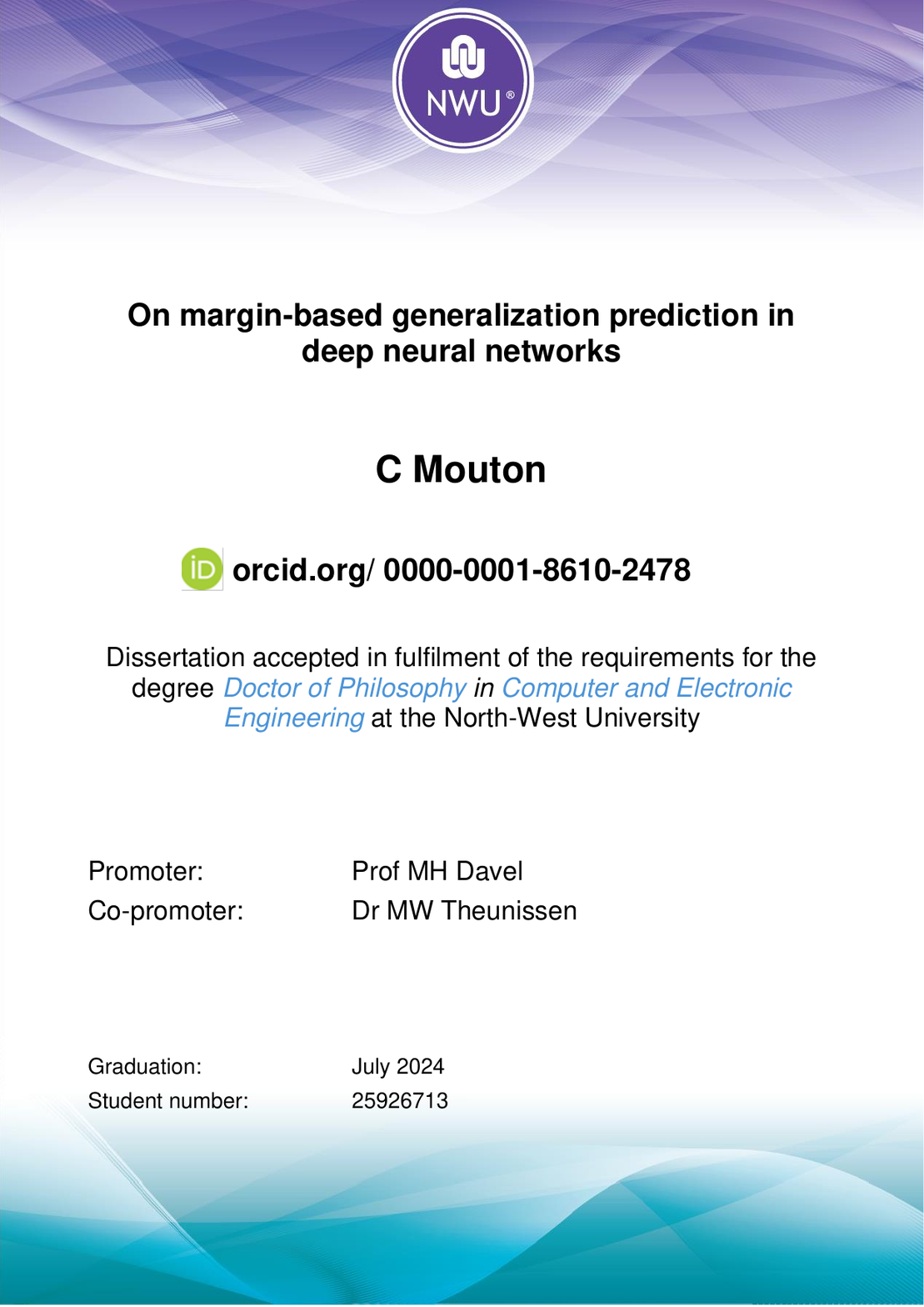}







\newpage
\restoregeometry
\cleardoublepage

\cfoot{\thepage} 


\newpage 
\cleardoublepage
\begin{flushleft}
\section*{Acknowledgements}
\begin{center}

I would like to extend my heartfelt thanks to the following individuals for their invaluable assistance and guidance throughout my PhD journey.

To my two advisors, Marelie Davel and Tian Theunissen. Marelie, thank you for keeping me on the right track and somehow finding sense in my scattered ideas. I still don't know how you do it! Tian, thank you for always challenging my assumptions and hypotheses. Both of you have made me a better, more careful scientist, and your contributions to this body of work are deeply appreciated.

To my parents. Thank you for your unwavering support and for nurturing the scientific curiosity that led me down this path. I hope that I can live up to your example.

To my girlfriend, Anna. Thank you for everything. Without your support and encouragement, this endeavor would not have been a successful one.

\end{center}
\end{flushleft}


\newpage 
\cleardoublepage




\newpage 
\cleardoublepage
\section*{Abstract}

Understanding generalization in deep neural networks is an active area of research. A promising avenue of exploration has been that of margin measurements: the shortest distance to the decision boundary for a given sample or that sample's representation internal to the network. Margin-based complexity measures have been shown to be correlated with the generalization ability of deep neural networks in some circumstances but not others. The reasons behind the success or failure of these metrics are currently unclear. In this study, we examine margin-based generalization prediction methods in different settings. We motivate why these metrics sometimes fail to accurately predict generalization and how they can be improved.

First, we analyze the relationship between margins measured in the input space and sample noise. We find that different types of sample noise can have a very different effect on the overall margin of a network that has modeled noisy data. 

Following this, we empirically evaluate how robust margins measured at different representational spaces are at predicting generalization. We find that these metrics have several limitations and that a large margin does not exhibit a strong correlation with empirical risk in many cases.

Finally, we introduce a new margin-based measure that incorporates an approximation of the underlying data manifold. It is empirically demonstrated that this measure is generally more predictive of generalization than all other margin-based measures. Furthermore, we find that this measurement also outperforms other contemporary complexity measures on a well-known generalization prediction benchmark. In addition, we analyze the utility and limitations of this approach and find that this metric is well aligned with intuitions expressed in prior work. 

\textit{\textbf{Keywords:} Margin measurements, Deep neural networks, Generalization, PGDL, Sample corruption}


\cleardoublepage 
\tableofcontents

\cleardoublepage 
\listoffigures

\cleardoublepage
\listoftables



\cleardoublepage
\include{lists/symbolslist}


\cleardoublepage
\pagenumbering{arabic}
\renewcommand{\sectionmark}[1]{\markboth{#1}{}}
\lhead{\ifthenelse{\thepage>0}
       {\it Chapter \thechapter }
      }
\rhead{\ifthenelse{\thepage>0}
       {\it \let\uppercase\relax\leftmark}
      }
\cfoot{\ifthenelse{\thepage>0}
       { \thepage}}


\input{chapters/ch1_introduction.tex}

\input{chapters/ch2_background.tex}
\input{chapters/ch3_sample_noise.tex}
\input{chapters/ch4_hidden_improv.tex}

\input{chapters/ch5_constrained.tex}

\input{chapters/ch6_conclusion.tex}


\newpage
\cleardoublepage
\lhead{}
\rhead{}
\addcontentsline{toc}{section}{References}
\printbibliography

\cleardoublepage
\include{chapters/appendix}

\end{document}

%% file: lists/symbolslist.tex





\addcontentsline{toc}{chapter}{Mathematical notation}
\chapter*{Mathematical notation}

We use the following notation for different mathematical objects throughout.

\begin{table}[H]
\centering
\caption{Notation used for different mathematical objects}
\begin{tabular}{llc}
\hline
\textbf{Object} & \textbf{Notation} & \textbf{Example} \\ \hline
Scalar & Lowercase letter & p \\
Vector/Tuple & Bold lowercase letter & \textbf{p} \\
Matrix & Bold uppercase letter & \textbf{P} \\
Set & Uppercase letter & P \\ \hline
\end{tabular}
\end{table}

There is some slight deviation from this in the case of well-known quantities, where we use the standard capital notation symbols for these scalars, for example, mutual information ($\mathcal{I}$), entropy ($\mathcal{H}$), and the coefficient of determination ($R^2$). This is made clear in the text.

%% file: chapters/ch1_introduction.tex
\let\cite\parencite

\lhead{}
\rhead{}
\chapter{Introduction} \label{chap:introduction}
\afterpage{\lhead{\ifthenelse{\thepage>0}
       {\it Chapter \thechapter }
      }
\rhead{\ifthenelse{\thepage>0}
       {\it \let\uppercase\relax\leftmark}
      }}
\underline{ \hspace{\textwidth} } 
\textit{``To the town of Agua Fria rode a stranger one fine day \\
Hardly spoke to folks around him, didn't have too much to say \\
No one dared to ask his business, no one dared to make a slip \\
For the stranger there among them had a big iron on his hip'' } \\ - Marty Robbins, \textit{Big Iron}, Verse 1\\
\underline{ \hspace{\textwidth} }

\section{Generalization in deep neural networks}
\label{sec:intro_generalization}

The use of Deep Neural Networks (DNNs) in machine learning has experienced a remarkable resurgence during the previous two decades. These neural models have revolutionized many fields and are now the state-of-the-art solution to many tasks, such as language modelling~\cite{openai_pretraining}, image recognition~\cite{imagenet_with_cnn_hinton}, image synthesis~\cite{diffusion_models}, and protein folding~\cite{deepmind_proteinfolding},  among many others.
While these deep architectures offer excellent performance, the exact mechanisms by which they achieve this, despite great over-parameterization, are unclear.   

The ability of a machine learning model to perform well on unseen data, that is, data on which it is not trained, is referred to as its \textit{generalization}~\cite[p.~110]{goodfellow_textbook}.  The mechanisms behind the generalization of DNNs are an active field of research, and many different theoretical perspectives have been proposed. These include theoretically proven bounds on the generalization error, such as Vapnik-Chervonenkis (VC) dimension~\cite{vc-dimension} and PAC-Bayesian bounds~\cite{pac_bayesian}. However, in the case of DNNs these bounds are generally not sufficiently tight to be practically useful~\cite{bounds_suck,fantastic_gen,rep_based_complexity_pgdl}. In fact, recent work~\cite{bounds_suck} suggests that in the overparameterized setting, these bounds cannot be tight without making strong assumptions about the data distribution.  A complementary approach to developing theoretical bounds is to develop empirical techniques that are able to predict the generalization ability of certain families of DNN models. That is, techniques that are able to estimate or rank the generalization performance of a set of models using only the models' training data and parameters. 

Predicting the generalization of a given model, or set of models, is important for several reasons:
\begin{itemize}
    \item An accurate measure of generalization could lead to a better theoretical understanding of the mechanisms behind generalization in DNNs.
    \item Such a measure could aid in model selection.
    \item Should a measure be predictive of generalization, it could be possible to incorporate it in the training process, e.g. as a regularization term, which could further improve the performance of DNNs.
    \item In some settings, sufficient data is not available in order to estimate the generalization ability of a model on a held-out dataset. Generalization prediction techniques can then be used to determine the model's performance.
\end{itemize}

Predicting the generalization of DNNs is generally achieved through the use of a complexity measure, where a complexity measure is defined as some measurement of the network parameters or representations that correlates well with its generalization ability~\cite{fantastic_gen}. One class of complexity measures that has shown promise is that of margin-based measures.

\section{Margin-based complexity measures}

Margin-based measures, as the name implies, rely on measuring \textit{margin}: the shortest distance to the decision boundary from a training or test sample. Margins are ubiquitous in machine learning and have long been studied in relation to generalization, such as in Support Vector Machines~\cite{boser1992training} or ensembles~\cite{margin_combined_classifiers}. However, as we will show, the link between large margins and generalization in DNNs is not as clearly established.

An interesting property of margin analysis in DNNs is that margins can be measured at multiple representational spaces throughout the network, such as in the input space (input margin), the hidden representational space (hidden margin), and the output space (output margin). Each of these measurements has distinct properties and comes with certain benefits and drawbacks when used to predict generalization (which we discuss later). 

In this study, we investigate these margin-based generalization prediction methods in DNNs. We analyze these in various settings, such as for different types of samples, architectures, datasets, and training setups. In doing so, we identify where these metrics succeed and fail, how these metrics compare to each other, and shed some light on \textit{why} they do or do not work in specific scenarios.  

In addition to the investigation of generalization, the field of margins in DNNs is also closely related to the field of adversarial examples. Adversarial examples are perturbed training or test samples that appear visually identical to their normal counterparts; however, they are consistently misclassified by DNNs.  Throughout this study, we also contribute to a better understanding of such adversarial examples.

In summary, our primary goal is to \textbf{further an understanding of margin-based complexity measures}. We aim to establish under which conditions these metrics do and do not work, how the different variants compare to one another, why they work, and what insights into generalization they can contribute. 

In the following section, we narrow down the scope of our investigations.

\section{Scope}

In this section, we specify the scope of our investigations surrounding margin-based generalization prediction methods.

\begin{itemize}
    \item \textbf{Generalization}: This term can have different meanings in different contexts. Specifically, `generalization' can refer to a model's performance on unseen data in several different scenarios in relation to its training data. These include: performance on independent identically distributed data (IID)~\cite{iid_def}, on out-of-distribution data (OOD)~\cite{ood_def}, and adversarial data~\cite{goodfellow_adversarial}. We investigate generalization in the context of its most common definition: `the empirical risk as measured on a held-out similarly distributed test set'~\cite{stronger_bounds_gen_def,neyshabur2017exploring_gen_def_2}.  Therefore, we do not make strong assumptions such as IID, but also do not investigate generalization in the broadest sense of the word.
    \item \textbf{Adversarial robustness}: In this study, our focus is on generalization (the above). However, because of the overlap between margin measurements and adversarial robustness, we do relate some of our findings to adversarial robustness as well.
    \item \textbf{Domain}: We limit our investigations to image classification tasks. This is a common setup for empirical investigations of generalization and margins in DNNs. Our study is therefore well-aligned with a large body of existing work~\cite{pgdl_overview,fantastic_gen, predict_gen_margin,large_margin_dnns, zhang2021understanding,theunissen2020benign}.
    \item \textbf{Datasets}: To further narrow down the previous point, we focus on `tiny-image' datasets such as CIFAR10~\cite{cifar}, CINIC10~\cite{CINIC}, SVHN~\cite{SVHN}, etc. These datasets have the desirable property of containing realistic, natural images and therefore accurately simulate real-world uses of DNNs.  Given their reduced (spatial) size, these datasets can provide meaningful insights into generalization without requiring tremendous computational resources to investigate. 
    \item \textbf{Architectures}: We follow the lead of existing work on margin measurements in DNNs and limit our architectures to multilayer perceptrons (MLPs) and convolutional neural networks (CNNs)~\cite{predict_gen_margin,constrained_optim_investigating,rep_based_complexity_pgdl, optimal_transport,large_margin_dnns,xu2023exploring}. We further confine the latter to $2$ - $12$ layer variants of popular architectures such as VGG~\cite{vgg} and Network-in-Network~\cite{network_in_network}. We do not investigate margins in very large architectures such as ResNet~\cite{resnet} or InceptionNet~\cite{inceptionnetv3} due to computational constraints.
\end{itemize}

\section{Overview}

In this section, we give a high-level overview of how this study is structured and highlight the key investigations of each chapter.

\begin{itemize}
    \item \textbf{Chapter 2: Background.} In this chapter, we introduce the setting in which we analyze margins, the necessary evaluation metrics, and the benchmarks that we use. In addition, we provide a comprehensive overview of existing margin-based generalization prediction methods, as well as other relevant complexity measures.
    \item \textbf{Chapter 3: Margins and sample noise.} We begin our investigation of margins in a very controlled setting. Specifically, we investigate how margin measurements relate to different types of samples across models of increasing capacity. To this end, we corrupt training datasets using different types of sample noise, which allows us to simulate real-world scenarios such as mislabeled or otherwise noisy data.  We then train multiple MLPs and CNNs of varying sizes on this corrupted data. Furthermore, we analyze these results and provide reasons for the observed behavior. This allows us to understand how different sample types influence the margin behavior and robustness properties of a trained DNN.
    \item \textbf{Chapter 4: Evaluating input and hidden margins on PGDL.} In this chapter, we turn our attention to the PGDL benchmark. We use this benchmark to do a thorough analysis of existing margin-based complexity measures for generalization prediction. 
    We compare these methods and determine in which cases these metrics succeed or fail to predict generalization. Additionally, we establish whether measuring these margins in a more accurate fashion increases their predictive performance.
    \item \textbf{Chapter 5: Input margins can predict generalization too.} In the previous chapter, we find that input margins are generally not predictive of generalization. In this chapter, we address the issues surrounding input margins. More precisely, we provide a clear intuition for why margins measured in the input space are not predictive of generalization. We then use this intuition to design a new measure that combines input margins with the principal components of the dataset. We refer to this measure as `constrained margin'. This new complexity measure is then compared to other margin-based complexity measures, as well as to the complexity measures of others. Furthermore, we also establish a link between adversarial examples and the principal components of the dataset. 
    \item \textbf{Chapter 6: Conclusion.}: Finally, in this chapter, we summarize the observations and contributions made throughout the study. Additionally, we also discuss these findings, identify directions for future work, and speculate on some of the unanswered questions.
\end{itemize}

\section{Publications}

Some of the research presented here has been published during the course of this study, specifically:

\begin{enumerate}
    \item  M. W. Theunissen\textsuperscript{*}, C. Mouton\textsuperscript{*}, and M. H. Davel, `The Missing Margin: How Sample Corruption Affects Distance to the Boundary in ANNs,' In Artificial Intelligence Research. SACAIR. Communications in Computer and Information Science, vol 1734, 2022 ~\cite{missing_margin}
    \item C. Mouton, M. W. Theunissen, and M. H. Davel, `Input margins can predict generalization too,' Proceedings of The 38th Annual AAAI Conference on Artificial Intelligence, 2024.~\cite{input_can_predict_too}
\end{enumerate}

A large part of Chapter~\ref{chap:ch3} is included in the first publication, while the majority of Chapter~\ref{chap:ch5} and a minority of Chapter~\ref{chap:ch4} are included in the second.


We elaborate on the contributions made by each author for these two publications. 

\begin{itemize}
\item \textbf{The Missing Margin: How Sample Corruption Affects Distance to the Boundary in ANNs}: This publication was a collaborative effort between C. Mouton and M.W. Theunissen, who share an equal contribution. M. H. Davel is the senior author and performed an advisory role. Mouton formulated and measured the classification margins, generated the results, assisted in analyzing the results, and trained the CNN models. Theunissen implemented the sample corruption methods, performed the analysis of the results, and trained the MLP models.
 \item \textbf{Input margins can predict generalization too}: This publication was mainly the efforts of C. Mouton. M.W. Theunissen and M.H. Davel performed an advisory role. Mouton developed and implemented the constrained margin method, generated the results, and performed the analysis. Theunissen measured the hidden margins. Davel derived Equation 5.
\end{itemize}

And now, dear reader, let us jump into the wondrous world of complexity measures and margins. In the following chapter, we start by providing the necessary background information.

%% file: chapters/ch2_background.tex
\let\cite\parencite

\lhead{}
\rhead{}
\chapter{Background} 
\label{chap:background}
\afterpage{\lhead{\ifthenelse{\thepage>0}
       {\it Chapter \thechapter }
      }
\rhead{\ifthenelse{\thepage>0}
       {\it \let\uppercase\relax\leftmark}
      }}
\underline{ \hspace{\textwidth} } 
\textit{``It was early in the morning when he rode into the town \\
He came riding from the south side slowly lookin' all around \\
He's an outlaw loose and running, came the whisper from each lip \\
And he's here to do some business with the big iron on his hip''} \\ - Marty Robbins, \textit{Big Iron}, Verse 2\\
\underline{ \hspace{\textwidth} }


\section{Overview} 
\label{sec:back_overview}

In this chapter, we provide an overview of prior work that is related to our investigation of margin measurements in DNNs. We assume prior knowledge of DNNs and therefore do not cover topics such as maximum likelihood estimation, gradient descent, architectural design, and other well-known foundations. For the unfamiliar reader, we recommend first consulting other resources that cover this in detail, such as the textbooks by Goodfellow et al.~\cite{goodfellow_textbook} and Zhang et al.~\cite{dive_into_deep_learning}.

The chapter is structured as follows. We first provide a definition and a broad overview of complexity measures in Section~\ref{sec:back_complexity}. Following this, in Section~\ref{sec:back_eval_complexity_measures}, we introduce key metrics that are used to evaluate the performance of complexity measures. We then describe the PGDL challenge, a popular benchmark for evaluating complexity measures, in Section~\ref{sec:back_pgdl}. Finally, we dive into related work surrounding classification margins and decision boundaries in Section~\ref{sec:back_margins_and_decision_boundaries}.

\section{Complexity measures}
\label{sec:back_complexity}

In Statistical Learning Theory (SLT), a `complexity measure' is defined as some measurement of the complexity of the model's hypothesis space~\cite{neyshabur2017exploring_gen_def_2}.  
We use an alternative, DNN-specific definition and define a complexity measure as any measurement of a neural network's structure (such as measurements of weights, activations, etc.) that is descriptive of its generalization ability. Specifically, for an ideal complexity measure, a lower complexity measure predicts a lower generalization gap. Here, the generalization gap is defined as the difference in performance on the training set and an unseen test set. This is similar to the definition offered by Jiang et al.~\cite{fantastic_gen}: `...a quantity that monotonically relates to some aspect of generalization. More
specifically, lower complexity should often imply smaller generalization gap.' 

This implies that a complexity measure should be able to utilize the training data and network parameters to predict or rank the generalization ability of various models~\cite{predict_gen_margin}. Numerous such measures have been proposed, with varying levels of success. These include, but are not limited to, measurements based on the geometry of the loss landscape, norms of the parameters, the complexity of the hypothesis space, the stability of the training process, and measurements based on the internal network representations, among others~\cite{fantastic_gen}.

While an overview of all these measures is outside the scope of this document, we motivate the interested reader to consult the large-scale evaluation by Jiang et al.~\cite{fantastic_gen} and other review articles~\cite{complexity_review_1,complexity_review_2,build_on_jiang_kendall} for more information. In the following sections, we instead focus on how these measurements are evaluated, the PGDL challenge, and the relevant complexity measures that have been evaluated within this framework.



\section{Evaluating complexity measures}
\label{sec:back_eval_complexity_measures}

Measuring the efficacy of a complexity measure is a difficult task.  In this section, we provide an overview of the evaluation metrics proposed by Jiang et al.~\cite{fantastic_gen, predict_gen_margin}, which we use throughout this study. We first introduce the notation we use in Section~\ref{sec:back_com_formulation_and_notation}. Following this, we describe rank correlation measures, namely Kendall's rank correlation (Section~\ref{sec:back_kendall_rank}) and granulated Kendall's coefficient (Section~\ref{sec:back_granulated_kendall}). We then explain a conditional independence test using conditional mutual information in Section~\ref{sec:back_conditional_mutual_info}. Following this, we also describe the coefficient of determination in Section~\ref{sec:back_coefficient_of_determination} for a slightly different use case. Finally, in Section~\ref{sec:back_use_of_evaluation_metrics}, we elaborate on how these various evaluation metrics are used throughout the study.

\subsection{Formulation and notation}
\label{sec:back_com_formulation_and_notation}

First, let us formalize the context within which complexity measures are investigated and introduce the necessary notation for a description of the evaluation measures. We follow the notation of Jiang et al.~\cite{fantastic_gen}, with some slight adaptations where applicable throughout this and the following sections.


We first introduce our notation surrounding hyperparameters. 
Given $n$ hyperparameters, 
let $\theta_{i,j}$ denote the value of a specific hyperparameter $i$, $i \in \{1, ... , n\}$, and let $\Theta_i$ denote the set of possible values for this hyperparameter, that is, 
$\Theta_i = \cup_j \{ \theta_{i,j} \}$.
Furthermore, let $\Theta$ denote the Cartesian product of all $n$ sets such that $\Theta = \Theta_1 \times \Theta_2 \times \ldots \times \Theta_n$, 
and let $\bm{\theta}_k$ correspond to a tuple consisting of a specific combination of these $n$ hyperparameters such that $\bm{\theta}_k = (\theta_{1,k}, \theta_{2,k}, ..., \theta_{n,k}) \in \Theta$. Here $\bm{\theta}_{i,k}$ indicates the value that hyperparameter $i$ takes for the specific $\theta_k$ combination (and {\em not} the $k^{th}$ available value of hyperparameter $i$). 
Let us consider an example. Assume that 
$\Theta_1 = \{0.01, 0.001\}$ corresponds to the learning rate of an optimizer, and 
$\Theta_2 = \{64, 128\}$ to its batch size. Then $\Theta = \{(0.01, 64), (0.01, 128), (0.001, 64), (0.001, 128)\}$. 
An example of $\bm{\theta}_k$ would then be $\bm{\theta}_1 = (0.01, 64)$, i.e. the first combination in $\Theta$. We drop the second index for a hyperparameter value $\theta_{i,j}$ going forward, as it is made clear from the context.

Now consider a set of $|\Theta|$ models trained with all these hyperparameter combinations. 
Let $g(\bm{\theta})$ describe the generalization gap of the single model trained using the hyperparameters $\bm{\theta}.$
Similarly, let $\varsigma(\bm{\theta})$ correspond to the value of a complexity measure $\varsigma$ calculated for this model. 

Given this setup and notation, we can now describe our evaluation metrics. The goal of an evaluation metric in this context is to establish the quality of the relationship between the complexity measure $g(\bm{\theta})$ and the generalization gap $\varsigma(\bm{\theta})$.



\subsection{Kendall's rank correlation coefficient}
\label{sec:back_kendall_rank}

A natural choice of evaluation metric would be simply to measure the ranking between the observed generalization gap and complexity measure for some group of models. More precisely, if a set of models are ranked according to their generalization gap, and also according to a complexity measure, how much agreement is found between these rankings. To this end, Jiang et al.~\cite{fantastic_gen} employ Kendall's rank correlation coefficient~\cite{kendall_corr}.

Kendall's rank correlation coefficient measures the consistency of a ranking.  Consider a set $T$, as defined below, where each element represents a model-specific complexity measure and generalization gap pair, such that
\begin{equation}
    T = \cup_{\bm{\theta} \in \Theta} \{(\varsigma(\bm{\theta}), g(\bm{\theta})\}
\end{equation}

The Kendall's rank correlation coefficient ($\tau$) for the set $T$ is then given by
\begin{equation}
\label{eq:kendall}
   \tau(T) = \frac{1}{|T|} \sum_{(\varsigma_1,g_1) \in T} \frac{1}{|T| - 1} \sum_{(\varsigma_2, g_2) \in T\backslash(\varsigma_1,g_1)} \operatorname{sign}(\varsigma_1 - \varsigma_2)\operatorname{sign}(g_1 - g_2)
\end{equation}

In essence, if the generalization gap of one model is larger than the other, such that $g_1 > g_2$, Kendall's $\tau$ measures whether this is also true for the complexity measure, i.e. whether $\varsigma_1 > \varsigma_2$ as well. This provides a value of $+1$ (signs agree) or $-1$ (signs disagree) for each pairwise comparison. This is then done for all pairs, and the final score is the average of these comparisons.  Kendall's $\tau$ (Equation~\ref{eq:kendall}) then gives a value between 1 (perfect agreement) and -1 (perfect disagreement). 

This metric has several desirable properties. Firstly, it provides one with an easily interpretable score of the alignment between the complexity measure and generalization gap of the set of models. For example, if one were to use a complexity measure for model selection, Kendall's correlation would indicate how accurate this selection would be in comparison to the true generalization gap of the models. Secondly, since the metric can be negative, it also indicates whether there is an inverse relationship between the complexity measure and generalization gap. 

Kendall's rank correlation is not without limitations.  Jiang et al.~\cite{fantastic_gen} argue that in this context the metric is easily susceptible to spurious correlations.
Consider the case where a complexity measure can accurately account for variation in a single hyperparameter but produces random predictions otherwise. For example, suppose that some measure can accurately rank the generalization of a group of models trained using different learning rates, but produces random predictions for a group of models trained using different batch sizes. Given that Kendall's rank correlation does not distinguish between the different hyperparameters, it is possible that the measure still achieves a high $\tau$ when both these groups of models are mixed, i.e. when considering models that vary in both batch size and learning rate. Furthermore, the poor performance in terms of variations in batch size is not indicated by the metric, which could be misleading. Jiang et al.~\cite{fantastic_gen} proposes the \textit{granulated Kendall's coefficient} to partially address this issue.   




\subsection{Granulated Kendall's coefficient}
\label{sec:back_granulated_kendall}

The granulated Kendall's coefficient is calculated by finding $\tau$ by varying a a single hyperparameter across the model distribution~\cite{fantastic_gen}, and is denoted by $\psi_i$, where $i$ is a single hyperparameter, e.g. batch size.  
This allows a more detailed view of how a complexity measure correlates with generalization gap for a specific hyperparameter.  More precisely, Jiang et al.~\cite{fantastic_gen} define this as
\begin{equation}
    \psi_i \triangleq \frac{1}{|M_i|} \sum_{\theta_1 \in \Theta_1} \ldots \sum_{\theta_{i-1} \in \Theta_{i-1}} \sum_{\theta_{i+1} \in \Theta_{i+1}} \cdots \sum_{\theta_n \in \Theta_n} \tau\left(\cup_{\theta_i \in \Theta_i}\{(\varsigma(\boldsymbol{\theta}), g(\boldsymbol{\theta}))\}\right)
\end{equation}
with $\bm{\theta}$ set to $\bm{\theta} = (\theta_1, \theta_2, ..., \theta_n)$ at each iteration and where $M_i$ is the Cartesian product of all hyperparameter sets, excluding that of the hyperparameter being investigated ($\Theta_i$). Formally,
\begin{equation}
    M_i \triangleq\Theta_1 \times \cdots \times \Theta_{i-1} \times \Theta_{i+1} \times \cdots \times \Theta_n
\end{equation}

In other words, the granulated Kendall's coefficient ($\psi_i$) calculates the Kendall's rank correlation across a group of models where the \textbf{only} variation between the models is a single hyperparameter. This is done by grouping the models so that they share all hyperparameter values except the one in question. The Kendall's rank correlation is then calculated for this small group. This is then done for all such groups, and the final $\psi_i$ is the average across these groups.

Let us borrow the example used earlier for four models that vary only in batch size and learning rate to illustrate this (recall the example in Section~\ref{sec:back_com_formulation_and_notation}). In this example,  $\theta_1 \in \Theta_1 = \{0.01, 0.001\}$ and $\theta_2 \in \Theta_2 = \{64, 128\}$ correspond to the learning rate and batch size of the optimizer, respectively. However, now we add an additional hyperparameter set, let $\theta_3 \in \Theta_3 = \{\text{SGD}, \text{Adam}\}$ correspond to the optimizer used, for a total of eight models. To calculate the granulated Kendall's coefficient for the learning rate $\theta_1$, one would first fix $\theta_2$ and $\theta_3$ to some value, say $\theta_2 = 64$ and $\theta_3 = \text{SGD}$. Following this, we then construct 
two models by only varying $\theta_1$, such that 
$\bm{\theta}_1 = (0.01, 64, \text{SGD})$ and $\bm{\theta}_2 = (0.001, 64, \text{SGD})$. The Kendall's rank correlation between the generalization gap and complexity measure for these two models is then calculated. This is then repeated by creating similar groups of two models for the other hyperparameter combinations of batch size and optimizer. This implies calculating the Kendall's rank correlation between models with the two learning rates using $(\theta_2 = 128, \theta_3 = \text{SGD})$, then $(\theta_2 = 64$, $\theta_3 = \text{Adam})$, and finally $(\theta_2 = 128$, $\theta_3 = \text{Adam})$, individually. The final score for the learning rate, $\psi_1$, is then the average Kendall's rank correlation for all four of these two-model groupings.

One could further calculate $\psi_i$ for all hyperparameters individually and then summarize the complexity measure's performance across all hyperparameters by taking the average. Formally,
\begin{equation}
    \mu(\psi) \triangleq \frac{1}{n} \sum_{i=1}^n \psi_i
\end{equation}

Jiang et al.~\cite{fantastic_gen} argue that $\mu(\psi)$ is a more reliable indicator of a complexity measure's performance than the normal Kendall's rank correlation (Equation~\ref{eq:kendall}). The intuition is that a high $\mu(\psi)$ indicates that the complexity measure is able to do well across all hyperparameter variations of the model distribution. However, the metric still relies on rank correlation, which could be problematic in certain cases. We elaborate on this in the next section.

\subsection{Conditional independence test}
\label{sec:back_conditional_mutual_info}

Although correlation is a useful measure, Jiang et al.~\cite{fantastic_gen} point out two scenarios that must be distinguished from each other when a hyperparameter is changed: 1) The hyperparameter change causes the complexity measure value to be lower, and a lower value of this complexity measure causes the generalization gap to be lower; 2) The hyperparameter change results in a lower complexity measure value and generalization gap, but the lower complexity measure value does not cause the lower generalization gap directly.  To this end, a measure of the probability of a causal relationship between the complexity measure and generalization gap must be established.  

The authors propose performing a conditional independence test by measuring the conditional mutual information between the complexity measure and generalization gap, given that a set of hyperparameters is observed.  This measure essentially indicates the likelihood of a complexity measure explaining generalization gap and assists in distinguishing between the two scenarios mentioned above. 

Allow us to first introduce some additional notation. Consider two hyperparameters, $\theta_1 \in \Theta_1$ and $\theta_2 \in \Theta_2$. Let the set $S$ correspond to the Cartesian product of these such that $S = \Theta_1 \times \Theta_2$, and let $U_S$ be a random variable representing the values of the hyperparameter combinations in this set. Furthermore, consider two binary random variables $V_\varsigma, V_g \rightarrow \{-1, +1\}$. These variables indicate if the complexity measure $\varsigma(.)$ or generalization gap $g(.)$ increases or decreases between any two unique hyperparameter combinations, respectively. That is, $V_\varsigma = \operatorname{sign}(\varsigma(\boldsymbol{\theta}) - \varsigma(\boldsymbol{\theta^{'}})$ and $V_g = \operatorname{sign}(g(\boldsymbol{\theta}) - g(\boldsymbol{\theta^{'}}))$.

To measure the mutual information between the complexity measure and generalization gap, given that hyperparameters $\theta_1$ and $\theta_2$ are observed, we must first group the models accordingly. This implies that we create groups of models that have the same values for $\theta_1$ and $\theta_2$, i.e. partitioning the models according to the values of $U_S$.
To be more precise, let us again use the example of learning rate, batch size, and optimizer. That is, consider a set of models $\Theta = \Theta_1 \times \Theta_2 \times \Theta_3$, where $\Theta_1 = \{0.01, 0.001\}$ corresponds to the learning rate, $\Theta_2 = \{64, 128\}$ batch size, and $\Theta_3 = \{\text{SGD}, \text{Adam}\}$ optimizer. If $S$ is set to $S = \{\text{learning rate}, \text{batch size}\}$ then $S = \{(0.01, 64), (0.01, 128), (0.001, 64), (0.001, 128)\}$. For each of these four combinations in $S$, we then create a group of models with these specific values for learning rate and batch size. For example, if $U_S = (0.01, 64)$, then a group can be created from the two models in $\Theta$ with these hyperparameters, i.e. $(0.01, 64, SGD)$ and $(0.01, 64, ADAM)$.    

For each group, we can then calculate the probabilities $p\left(V_g \mid U_{\mathcal{S}}\right)$, $p\left(V_\varsigma \mid U_{\mathcal{S}}\right)$, and $p\left(V_\varsigma, V_g \mid U_{\mathcal{S}}\right)$ by counting the number of instances for each of the different states of $V_g$ and $V_\varsigma$.~\footnote{See Jiang et al.~\cite[Appendix A.5]{fantastic_gen} for more information on how the probabilities are estimated}  

The conditional mutual information between the complexity measure and generalization gap is then calculated as follows:
\begin{equation}
\mathcal{I}\left(V_\varsigma, V_g \mid U_{\mathcal{S}}\right)=\sum_{U_{\mathcal{S}}} p\left(U_{\mathcal{S}}\right) \sum_{V_\varsigma \in\{+1,-1\}} \sum_{V_g \in\{+1,-1\}} p\left(V_\varsigma, V_g \mid U_{\mathcal{S}}\right) \log \left(\frac{p\left(V_\varsigma, V_g \mid U_{\mathcal{S}}\right)}{p\left(V_\varsigma \mid U_{\mathcal{S}}\right) p\left(V_g \mid U_{\mathcal{S}}\right)}\right)
\end{equation}

This conditional mutual information is then normalized by the conditional entropy of the generalization gap conditioned on the hyperparameters, as that is the maximum possible mutual information between $\varsigma$ and $g$. This conditional entropy is calculated as:
\begin{equation}
\mathcal{H}\left(V_g \mid U_{\mathcal{S}}\right)=-\sum_{U_{\mathcal{S}}} p\left(U_{\mathcal{S}}\right) \sum_{V_g \in\{+1,-1\}} p\left(V_g \mid U_{\mathcal{S}}\right) \log \left(p\left(V_g \mid U_{\mathcal{S}}\right)\right)
\end{equation}

The normalized conditional mutual information is then given by:
\begin{equation}
    \bar{\mathcal{I}}\left(V_\varsigma, V_g \mid U_{\mathcal{S}}\right)=\frac{\mathcal{I}\left(V_\varsigma, V_g \mid U_{\mathcal{S}}\right)}{\mathcal{H}\left(V_g \mid U_{\mathcal{S}}\right)}
\end{equation}

This is then repeated for all unique groups of two hyperparameters, i.e. all $|S| = 2$. The final score is then the minimum of this normalized conditional mutual information across all such groups. Formally,
\begin{equation}
\label{eq:cmi_final_score}
\mathcal{K}(\varsigma)=\min _{U_{\mathcal{S}} \text { s.t }|\mathcal{S}| = 2} \bar{\mathcal{I}}\left(V_\varsigma, V_g \mid U_{\mathcal{S}}\right) 
\end{equation}

This provides a score of between $0$ and $1$. A score of $0$ indicates that there is no relationship between the complexity measure and generalization, while $1$ indicates that there is a strong possibility of a causal relationship. In practice, we multiply Equation~\ref{eq:cmi_final_score} by $100$ when comparing complexity measures, as done in previous work~\cite{pgdl_overview}. We simply refer to the score of Equation~\ref{eq:cmi_final_score} as `Conditional Mutual Information' or `CMI' in the future, in line with others that use this metric~\cite{fantastic_gen,rep_based_complexity_pgdl,schiff2021predicting,gan_pgdl}.   

\subsection{Coefficient of determination}
\label{sec:back_coefficient_of_determination}

In this section, we consider an evaluation metric for a slightly different use case than those described earlier. In this study, we are primarily concerned with generalization ranking, and we have thus far examined metrics that establish the quality of a ranking between a complexity measure and generalization gap for a group of models. However, in a single other setting, we consider \textit{numerically} predicting the generalization gap for a group of models.  In certain practical scenarios, one would not only wish to rank models according to their generalization performance (such as for model selection), but also numerically estimate their performance. For example, if a held-out dataset is unavailable and an estimate of real-world performance is desirable. Such a setting can also be considered a complementary approach to model ranking for evaluating a complexity measure.

As evaluation metric in this setting, we make use of the coefficient of determination. We will first describe this metric in the general case, and then specify how we adapt it for our specific setting when numerically predicting generalization.

Consider any two vector $\mathbf{z}, \mathbf{\hat{z}} \in \mathbb{R}^m$. The coefficient of determination is then given by
\begin{equation}
    R^2=1-\frac{\sum_{j=1}^{m}\left(\hat{\mathbf{z}}_j-\mathbf{z}_j\right)^2}{\sum_{j=1}^{m}\left(\mathbf{z}_j-\frac{1}{m} \sum_{j=1}^{m} \mathbf{z}_j\right)^2}
\end{equation}
where $\mathbf{z}_j$ and $\mathbf{\hat{z}}_j$ corresponds to the scalar values of $\mathbf{z}$ and $\mathbf{\hat{z}}$  at index j, respectively.
Intuitively, the coefficient of determination is a ratio of how much of the variance in $\mathbf{z}$ can be explained by $\mathbf{\hat{z}}$ and the total variance in $\mathbf{z}$. 
This provides a value with a maximum of $1$, which indicates that $\mathbf{\hat{z}}$ explains all the variance in $\mathbf{z}$. There is no lower limit, but any $R^{2} < 0$ indicates that $\mathbf{\hat{z}}$ performs worse at explaining the variance in $\mathbf{z}$ than using a constant value. 

When considering numerically predicting generalization, we let $\mathbf{z}$ correspond to the measured generalization gap of some group of models, and let $\mathbf{\hat{z}}$ correspond to the predicted generalization gap for these models. More specifically, the predictions are obtained using some function $f$, such that $\mathbf{\hat{z}}_i = f(\mathbf{C})$ for $i \in \{1,..,m\}$, where $\mathbf{C}$ is some vector valued complexity measure. In this way, the coefficient of determination is used to evaluate how well the generalization gap can be predicted using the function $f$ and complexity measure $\mathbf{C}$. We elaborate more on this in Section~\ref{sec:hidden_lin_reg}.

In addition to to the numerical prediction setting, in one other case, we also use the coefficient of determination  to assess the quality of the linear relationship between other quantities than complexity measures and generalization gaps. Specifically, we measure the coefficient of determination between some measured dependent variable (which corresponds to $\mathbf{z}$) and some measured independent variable (which corresponds to $\mathbf{\hat{z}}$). The context is made clear in Section~\ref{sec:noise_analysis_label_corruption}.

\subsection{Use of evaluation metrics}
\label{sec:back_use_of_evaluation_metrics}

In this study, we make use of all of the aforementioned evaluation metrics. Each metric serves a unique purpose for different kinds of investigations. Let us describe the different use cases.

\begin{itemize}
    \item \textbf{Comparing complexity measures w.r.t. generalization gap ranking}: Throughout this study, we often compare different complexity measures in a setting where the goal is to accurately rank the generalization gap of a set of models using the complexity measure. For this purpose, we always rely on both Kendall's rank correlation and Conditional Mutual Information (CMI) as evaluation metrics. We use the former because it provides easily interpretable values and it can also indicate a negative relationship. These are desirable properties for gaining an understanding of the relation of various complexity measures to generalization. On the other hand, we also include CMI as it is more robust to spurious correlations~\cite{fantastic_gen} and allows us to compare with other scores reported in the literature. This is because CMI is generally the only metric reported on for the benchmarks we use. 
    
    We also use these metrics in slightly different ways when considering margin measures. 
    For margin measures specifically, we always use the negative of the mean margin as a complexity measure, i.e. so that a smaller value should correspond to a smaller generalization gap. This implies that for the CMI metric, we calculate the CMI between the negative of the mean margin and generalization gap. In contrast, we always calculate the Kendall's rank correlation between the positive of the mean margin and \textit{test accuracy}, so that a larger margin should correspond to a higher predicted test accuracy.
    For example, consider two models with a test accuracy of $90\%$ and $85\%$, respectively, and the same train accuracy of $100\%$. Similarly, suppose that each model has a mean margin of $5.0$ and $4.5$, respectively. The Kendall's rank correlation will then be calculated between the sets of test accuracy and mean margin for each model, meaning between the set $\{0.9, 5.0\}$ and $\{0.85, 4.5\}$. On the other hand, the CMI will be calculated between the sets of generalization gap and negative mean margin for each model, $\{0.1,-5.0\}$ and $\{0.15, -4.5\}$.
    
    Finally, in the setup in which we investigate margins all models have approximately the same train accuracy, so using either test accuracy or generalization gap results in similar performance evaluations. We elaborate on this benchmark in the following section.
    \item \textbf{Comparing generalization ranking per hyperparameter for margin measures}: In certain cases, we wish to conduct a fine-grained investigation of a specific margin measure. In this setting, the goal is to evaluate the margin measure w.r.t. specific hyperparameter variations. In these cases, we rely on the granulated Kendall's coefficient to measure ranking for each hyperparameter variation across a group of models. As for normal Kendall's rank correlation, we measure the granulated Kendall's coefficient between the positive of the mean margin and test accuracy. 
    \item \textbf{Comparing margin measures w.r.t. numerical test accuracy prediction}: In one specific case, we compare margin measurements with the goal of numerically predicting test accuracy. For this, we follow previous work~\cite{predict_gen_margin,gan_pgdl} and use the coefficient of determination. 
\end{itemize}

In the following section, we explain the benchmark on which we evaluate complexity measures using these metrics.

\section{PGDL challenge}
\label{sec:back_pgdl}

The `Predicting Generalization in Deep Learning' (PGDL) challenge is a competition that was hosted at the 2020 international \textit{Conference on Neural Information Processing Systems} (NeurIPS)~\cite{pgdl_analysis, pgdl_overview}. The objective of this challenge was to design a complexity measure to rank models according to their generalization gap. More precisely, participants only had access to various sets of trained models, along with their parameters and training data, and were tasked with ranking the models within each set according to their generalization gap. Each solution was then evaluated on how well its ranking aligns with the true ranking on a held-out set of tasks, which was unknown to the competitors.

In the following sections, we provide further details of the different model sets in the PGDL challenge, the various solutions that have been proposed, and also discuss a related generalization prediction dataset.

\subsection{Task overview}
\begin{table}[ht]
\centering
\caption[Overview of PGDL challenge tasks]{Overview of PGDL challenge tasks. Tasks $1$ to $5$ form the development set, and Tasks $6$ to $9$ form the held-out test set. There is no Task $3$.}
\label{tab:pgdl_task_overview}
\begin{adjustbox}{max width=\textwidth}
\begin{tabular}{lllllll}
\hline
\textbf{Task} & \textbf{\begin{tabular}[c]{@{}l@{}}Num\\ models\end{tabular}} & \textbf{Architecture} & \textbf{Dataset} & \textbf{\begin{tabular}[c]{@{}l@{}}Training\\ samples\end{tabular}} & \textbf{\begin{tabular}[c]{@{}l@{}}Input\\ features\end{tabular}} & \textbf{\begin{tabular}[c]{@{}l@{}}Test accuracy\\ variation (\%)\end{tabular}} \\ \hline
1 & 96 & VGG-like CNN & CIFAR-10~\cite{cifar} & 50 000 & 3 072 & 66.83 - 86.34\\
2 & 54 & Network in Network & SVHN~\cite{SVHN} & 73 257 & 3 072 & 70.32 - 95.05\\
4 & 96 & Fully Convolutional & CINIC-10~\cite{CINIC} & 36 000 & 3 072 & 53.85 - 67.87\\
5 & 64 & Fully Convolutional & CINIC-10~\cite{CINIC} & 36 000 & 3 072 & 25.33 - 66.71\\ \hline
6 & 96 & Network in Network & Oxford-flowers~\cite{oxford_flowers} & 2 040 & 3 072 & 58.43 - 73.93\\
7 & 48 & Network in Network & Oxford-pets~\cite{oxford_pets} & 3 680 & 3 072 & 44.81 - 51.81\\
8 & 64 & VGG-like CNN & FMNIST~\cite{fashion_mnist} & 60 000 & 784 & 90.38 - 93.58\\
9 & 32 & Network in Network & CIFAR-10~\cite{cifar} (augmented) & 50 000 & 3 072 & 84.36 - 93.06\\ \hline
\end{tabular}
\end{adjustbox}
\end{table}

In total, the PGDL dataset consists of $550$ trained models across eight different tasks and six different image classification datasets, where each task refers to a set of models trained on the same dataset with varying hyperparameters and subsequent test accuracy.  We show an overview of each task in the PGDL challenge in Table~\ref{tab:pgdl_task_overview}. As shown, three different architecture types are used across the eight tasks: a modified VGG~\cite{vgg} CNN architecture, which is referred to as `VGG-like', variants of Network in Network~\cite{network_in_network}, and also fully convolutional neural networks. 

Several hyperparameters are varied for the models within each task, such as dropout probability, weight decay, architectural variation (number of filters, layers, etc.), batch size, and learning rate.  This provides substantial variation in test accuracy between models for most tasks. See the `Test accuracy variation' column in Table~\ref{tab:pgdl_task_overview}, which indicates the test accuracy of the worst and best generalizing models. 
Furthermore, this `dataset' of models is especially suited for evaluating complexity measures, as all models achieve roughly the same train accuracy. Specifically, each model was trained until a specific loss threshold was reached (on the training data). These thresholds were chosen such that all models have a final train accuracy of $\geq 99\%$. In essence, the PGDL challenge asks a fundamental question: given two models that perform equally well on their training data, but vary in performance on test data, what model characteristics distinguish these models?
Note that since all models have approximately the same train accuracy, ranking the models according to generalization gap or test accuracy is equivalent. 

To ensure that participants did not specifically design complexity measures that are only applicable to the models in the dataset, only Tasks $1$ to $5$ were available to participants for development purposes, while Tasks $6$ to $9$ were used as a held-out test set. The final average score on the test set was the only value used to rank the competitors, where the aforementioned Conditional Mutual Information (CMI) (see Section~\ref{sec:back_conditional_mutual_info}) was the only evaluation metric used. However, each participant was granted three attempts on this held-out test set~\cite{pgdl_overview}. Finally, since the completion of the competition, the dataset has been made publicly available~\cite{pgdl_resources}.

\subsection{PGDL complexity measures}
\label{sec:back_pgdl_complexity_measures_explanation}

In this section, we provide an overview of the various solutions put forward during this challenge, as well as the work that has been done since the public release of the dataset. We also mention the name given to each solution for easy identification later.

\subsubsection{Developed during the challenge}

We first consider the best performing complexity measures developed during the challenge and the conclusions made by the challenge organizers.

The winning team, Natekar and Sharma~\cite{rep_based_complexity_pgdl}, developed several prediction methods based on the internal representations of each model. Their best-performing method measures clustering characteristics of hidden representations (using Davies-Bouldin Index (DBI)~\cite{davies_bouldin_index}), and combines this with the model's accuracy on Mixup-augmented~\cite{zhang2017mixup} training samples. Specifically, they define `label-wise Mixup' (LWM). This method creates a synthetic set of samples by linearly interpolating between two training samples of the same class. Formally, for training samples $\mathbf{x}_1$ and $\mathbf{x}_2$, with labels $y_1 = y_2$, the label-wise Mixup sample $\mathbf{\tilde{x}}$ and label $\tilde{y}$ are given by
\begin{equation}
\label{eq:mixup}
    \mathbf{\tilde{x}} = \lambda \mathbf{x_1} + (1 - \lambda) \mathbf{x_2}
\end{equation}
\begin{equation}
    \tilde{y} = y_1 = y_2
\end{equation}
where they set $\lambda = 0.5$ in all cases. Such label-wise Mixup samples are created for all unique combinations of training samples within the same class. The `Mixup accuracy' is then simply the model's classification accuracy on this synthetic distribution. This is then combined with their clustering measure by simply multiplying the DBI with the Mixup accuracy. Accordingly, this complexity measure is named `DBI*LWM'. Furthermore, Natekar and Sharma~\cite{rep_based_complexity_pgdl} also present two margin-based solutions, which we elaborate on later (see Section~\ref{sec:back_hidden_margins}).

Kashyap et al.~\cite{pgdl_runners_up} were the runners-up and, in a similar fashion to LWM, based their metrics on measuring the robustness of trained networks to augmentations of their training data~\cite{pgdl_runners_up}. Specifically, they argue that CNNs heavily rely on features that encode texture for classification. As such, a model that generalizes well should be robust to augmentations that alter texture. They apply several augmentations to the training data such as horizontal flips, random saturation, random crops, sobel filters, etc. They then also penalize the model more heavily if it is more vulnerable to augmentations that do not affect the texture of the image (e.g. the horizontal flips and random crops). We refer to this method as `R2A' for `Robustness to augmentations'.

Unfortunately, the top submissions all suffered from severe overfitting on the development set of tasks, and performance dramatically decreased on the held-out test set.  Even the winning solution experienced a $41$\% performance drop, and the runner-up $76$\%~\cite{pgdl_overview}.  This seems to suggest that current complexity measures are not robust enough to be applicable across different architectures and datasets.  

In a review of the challenge, Jiang et al.~\cite{pgdl_overview} also point out that solutions which measure a model's robustness to data augmentations are especially prone to overfitting.  They attribute this to the differences between tasks, where certain simulated data augmentations work well for some datasets, but not others.  Furthermore, some models are specifically trained with an augmented dataset, which further complicates the process of obtaining an accurate measure of generalization.

\subsubsection{Developed after public release}

Somewhat regrettably, solutions that have been developed since the dataset's public release also seem to mostly focus on data augmentations or creating a synthetic test set. We now briefly summarize all three of these recent works (the only ones that we are aware of).

Schiff et al.~\cite{schiff2021predicting} generate `perturbation response curves'. These curves measure the change in accuracy of the network for Mixup~\cite{zhang2017mixup} augmented samples across different ranges of $\lambda$ (recall Equation~\ref{eq:mixup}). Unlike the solution of Natekar and Sharma~\cite{rep_based_complexity_pgdl}, they further experiment with Mixup between samples of different classes in addition to those within the same class. Their complexity measures are then given by statistical measures of these curves, inspired by the Gini coefficient~\cite{gini_coeff} (named `Gi-score') and Palma ratio~\cite{palma_ratio} (named `Pal-score'). They produce eleven complexity measures with different types of sample Mixup and statistical metrics. Their best performing method is named `PCA Gi\&Mi', which combines the Gi-score and Mixup accuracy through principal component analysis.  This solution currently holds the highest average score on the test set of tasks that we are aware of. 

Zhang et al.~\cite{gan_pgdl} take the idea of a synthetic dataset to the extreme and generate an artificial test set using pre-trained generative adversarial networks (GANs). They demonstrate that simply measuring the classification accuracy on this synthetic test set is very predictive of a model's generalization on the normal test set. Furthermore, they also demonstrate good performance when numerically predicting generalization on the DEMOGEN dataset (which we elaborate on in the following section) using the same method. However, this method is not without limitations. Due to the limited number of samples of Tasks $6$ and $7$, they cannot satisfactorily fine-tune GANs on these datasets. They choose not to report any scores for these two tasks. However, they outperform the best-performing measure of Schiff et al.~\cite{schiff2021predicting} on the remaining two test set tasks (Tasks $8$ and $9$). While practically useful, this method does not make a link between any characteristics of the model and its generalization ability, as performance is purely measured on a generated test set. Therefore, it is difficult to consider this as a complexity measure. This solution is simply referred to as `Ours', so we call it `GANs'.

Finally, Chuang et al.~\cite{optimal_transport} do not rely on generating synthetic samples. Instead, they develop a cluster-aware normalization scheme for margin-based measures and demonstrated its performance on the PGDL dataset. We elaborate more on this method shortly (see Sections \ref{sec:back_hidden_margins} and \ref{sec:back_output_margins}).

One unfortunate side effect of the PGDL dataset's public release is that these subsequent works do not appear to respect the split between the development and test set of tasks. This is evident by the fact that the scores for the two sets are not presented separately~\cite{schiff2021predicting,gan_pgdl,optimal_transport}, nor is it stated that either set is isolated for the development of the measure. This brings into question how reliable the reported scores are, as they might be the result of carefully selected hyperparameters that are only applicable to these specific tasks. 
Solutions developed during the challenge are also not immune, as participants were granted three tries on the held-out set~\cite{pgdl_overview}. The runners-up used different hyperparameters for each of these three submissions~\cite{pgdl_runners_up}. On the other hand, the winners of the challenge~\cite{rep_based_complexity_pgdl} used each submission to test a unique complexity measure. This suggests that the three scores reported by Natekar and Sharma~\cite{rep_based_complexity_pgdl} are the only ones that are truly reliable. Regardless, we show the average CMI score on the development and test set of tasks for each solution in Table~\ref{tab:pgdl_task_solutions_score_summary}. Note that we explain the margin-based measures later in Sections \ref{sec:back_hidden_margins} and \ref{sec:back_output_margins}.


\begin{table}[H]
    \centering
    \begin{threeparttable}
        \caption[PGDL challenge CMI scores for various complexity measures]{Average Conditional Mutual Information (CMI) score on the PGDL development and test set for various solutions. Development set consists of Tasks $1$ to $5$. Test set consists of Tasks $6$ to $9$. Solutions above the horizontal bar were developed during the challenge and those beneath after.}
        \label{tab:pgdl_task_solutions_score_summary}
        \begin{tabular}{@{}llll@{}}
            \toprule
            \textbf{Author} & \textbf{Solution name} & \textbf{\begin{tabular}[c]{@{}l@{}}Mean dev set \\ score (CMI)\end{tabular}} & \textbf{\begin{tabular}[c]{@{}l@{}}Mean test set \\ score (CMI)\end{tabular}} \\ \midrule
            Natekar and Sharma~\cite{rep_based_complexity_pgdl} & DBI*LWM & 19.94 & 22.92 \\
            Natekar and Sharma~\cite{rep_based_complexity_pgdl} & Mixup Margin Summary & 31.56 & 13.93 \\
            Natekar and Sharma~\cite{rep_based_complexity_pgdl} & Augment Margin Summary & 30.09 & 09.29 \\
            Kashyap et al.~\cite{pgdl_runners_up} & R2A & 38.98\tnote{1} & 10.16 \\ \midrule
            Schiff et al.~\cite{schiff2021predicting} & PCA Gi\&Mi & 23.05 & \textbf{24.18} \\
            Chuang et al.~\cite{optimal_transport} & $k$V-Margin 1st & 21.51 & 06.48 \\
            Chuang et al.~\cite{optimal_transport} & $k$V-GN-Margin 1st & 27.29 & 07.55 \\
            Zhang et al.~\cite{gan_pgdl} & GANs & \textbf{50.93} & ------\tnote{2} \\ \bottomrule
        \end{tabular}
        \begin{tablenotes}
            \item[1]\textit{We find conflicting scores reported in the literature. We rely on those found in \cite{gan_pgdl}.}
            \item[2]\textit{No reported scores for Tasks $6$ and $7$ and therefore no test set average.}
        \end{tablenotes}
    \end{threeparttable}
\end{table}

\subsection{DEMOGEN}
\label{sec:back_demogen}

In this section, we briefly elaborate on a model dataset that is complementary to the PGDL challenge. Jiang et al.~\cite{predict_gen_margin} created DEMOGEN,  `\textbf{de}ep \textbf{mo}del \textbf{gen}eralization dataset'. The aim of both the PGDL challenge and DEMOGEN is to evaluate complexity measures; however, there is a key difference. The goal of the PGDL dataset is to \textit{rank} generalization gap, whereas the goal of DEMOGEN is to \textit{numerically predict} generalization gap~\cite{gan_pgdl}. This is generally accomplished by combining a complexity measure with a linear regression model, such that the linear model uses the complexity measure to provide an estimate of a model's generalization gap~\cite{predict_gen_margin, gan_pgdl}. 

The DEMOGEN dataset consists of $756$ trained models. However, only two datasets are considered, CIFAR10 and CIFAR100~\cite{cifar}. Furthermore, the models contained within are generally larger (in terms of the number of parameters and number of layers) than those in the PGDL challenge. Two different architecture families are available: Network-in-Network~\cite{network_in_network} and ResNet-32 (a deep residual network with $32$ layers)~\cite{resnet}. Furthermore, it differs from the PGDL challenge in that not all models achieve the same train accuracy, which varies from $60\%$ to $90.5\%$~\cite{demogen_online_description}.

The hyperparameter variations for these models are similar to those of the PGDL challenge and contain variations in learning rate, dropout probability, activation normalization (batch or group normalization), layer width, and data augmentation~\cite{demogen_online_description}. 

In this work we do not make use of the DEMOGEN dataset, as our focus is primarily on ranking generalization; however, we do occasionally reference other studies that have.

\section{Margins and decision boundaries}
\label{sec:back_margins_and_decision_boundaries}

We now provide an overview of prior work that has been done on classification margins and decision boundaries in DNNs. Before proceeding, let us define these terms.

\textit{Decision boundaries} are the hypersurfaces which separate some feature space into distinct classification regions, with the latter being a region in which all points are predicted to be of the same class. More formally, consider a classification model $f~:~\mathbf{x}\rightarrow \mathbb{R}^{|N|}$, $N = \{1\ldots n\}$ with classification given by $\argmax_{k \in N} f_k(\mathbf{x})$. The decision boundary of a class pair $(i,j) \in N$ can be defined as follows~\cite{large_margin_dnns}:
\begin{equation}
\label{eq:decision_boundary}
    D_{(i,j)} = \{\mathbf{x} | f_i(\mathbf{x}) = f_j(\mathbf{x})\}
\end{equation} 
where $f_i$ is the output of the model for class $i$, and similarly $f_j$ for class $j$.

Equation \ref{eq:decision_boundary} is then the hypersurface where there is a score tie between the outputs for classes $i$ and $j$ for some input $\mathbf{x}$. The \textit{margin} between classes $i$ and $j$ for some sample classified as class $i$ is then defined as the minimum distance to this decision boundary. This can be formulated as finding the smallest displacement $\mathbf{\delta}$ that results in a score tie~\cite{large_margin_dnns}. Formally, for a training sample $\mathbf{x}$ with $i = \argmax_{k \in N} f_k(\mathbf{x})$ the margin $d$ between classes $i$ and $j$ is given by:
\begin{equation}
\label{eq:minimization}
   d_{f,(i,j)}(\mathbf{x}) = \min||\bm{\delta}||_p \: s.t. \: f_i(\mathbf{x}+\bm{\delta}) = f_j(\mathbf{x}+\bm{\delta})
\end{equation}
where $||\cdot||_p$ is some $L_p$ norm. Generally, the term `margin' implies the distance to the nearest other class $j \neq i$, which is also how we use the term.
Simply put, the classification margin of a sample with regard to a specific model is the shortest distance the sample will need to move, in a given feature space, in order to change the predicted output value.

There is considerable prior work on understanding classification margins in machine learning models~\cite{boser1992training,JMLR:v10:weinberger09a}. The relation between margin and generalization is well understood for classifiers such as support vector machines (SVMs)~\cite{boser1992training,vapnik1999overview}.  However, such a theoretical foundation has not been established for deep neural networks. 
Furthermore, the decision boundaries in DNNs (assuming non-linear activation functions) are both non-convex and (typically, depending on the representation space) high-dimensional~\cite{deep_dig}, which implies that precisely measuring these margins is considered intractable~\cite{constrained_optim,yang2020boundary}. As such, throughout this section, we elaborate on several methods that have been used to approximate these distances.

It is also important to note that the term `margin' is, often confusingly, used to refer to 1) output margins~\cite{output_margin_spectral}, 2) input margins~\cite{sokolic2017robust}, and 3) hidden margins~\cite{predict_gen_margin}, interchangeably throughout the literature. Here (1) is a measure of the difference in class output values, while (2) or (3) is concerned with measuring the distance from a sample to its nearest decision boundary in either input or hidden representation space, respectively (as formulated in Equation~\ref{eq:decision_boundary}). In this work, we specifically focus on the latter two definitions. While a link can be made between the definitions of output margins and input/hidden margins, output margins have a very different character, as they are easy to measure and interpret. We elaborate more on this in Section~\ref{sec:back_output_margins}. The remainder of this section compartmentalizes relevant work along these three margin definitions.

\subsection{Hidden margins}
\label{sec:back_hidden_margins}

Hidden margins relate to measuring the minimum distance to the decision boundary in some representational space of a DNN, for example, the activation space for some hidden layer. Intuitively, hidden space margins are a measure of how well a network separates different classes at its internal representations~\cite{rep_based_complexity_pgdl} as opposed to a measure of separation in the input space. This implies that hidden margins can be difficult to interpret, as it measures distances in a mostly uninterpretable feature space.

\subsubsection{Measuring and normalizing hidden margins}

Hidden margins are generally measured by using a first-order Taylor approximation of the true distance~\cite{large_margin_dnns, predict_gen_margin, huang2016learning, deepfool}. Formally, for a classification model $f$, and sample $\mathbf{x}$ belonging to class $i$, the Taylor approximated margin w.r.t the decision boundary between classes $i$ and $j$ at layer $l$ is calculated as
\begin{equation}
\label{eq:taylor_approx_margin}
  \hat{d}_{f^{l},(i,j)}(\mathbf{x}^l) = \frac{f^{l+1}_i(\mathbf{x}^l)-f^{l+1}_j(\mathbf{x}^l)}{||\nabla_{\mathbf{x}^l} f^{l+1}_i(\mathbf{x}^l) - \nabla_{\mathbf{x}^l}  f^{l+1}_j(\mathbf{x}^l)||_p}
\end{equation}
where the $L_2$ (Euclidean) norm is usually employed for the denominator. Here, $x^l$ indicates the output activations of layer $l$. For example, $\mathbf{x}^0$ indicates the input sample, and $\mathbf{x}^1$ its activations at the first hidden layer. Furthermore, $\nabla_{\mathbf{x}^l} f_i(\mathbf{x}^l) - \nabla_{\mathbf{x}^l}  f_j(\mathbf{x}^l)$ indicates the difference between the gradients of the $i^{th}$ and $j^{th}$ output logits with respect to these activation features. Finally, $f^{l+1}$ indicates the model $f$ with layers $0$ to $l$ removed.

This estimation method is relatively inexpensive, which is a desirable property as the high dimensionality of hidden representations can make other measurement methods unfeasible (we elaborate more on other measurement methods in Section~\ref{sec:back_input_margins}). However, this computational efficiency comes at a cost, as Youzefsadeh and O'Leary~\cite{constrained_optim} show that the accuracy of this estimate can be poor in some cases. 

Hidden margins also have the additional complexity of being difficult to compare between different models and layers. For instance, consider the case of comparing the margin measured at the first hidden layer of two different networks in order to predict which model generalizes better.  The scale of each model's activations can differ, which would, of course, have an effect on the measured distance.  For example, the activation values for one model's hidden layer might be in the range of $[0,1]$, while the other model's in the range of $[0, 100]$. In fact, Jiang et al.~\cite{predict_gen_margin} point out that hidden margins can be artificially increased: such as multiplying the weights of one layer by a large constant and then dividing by the same constant at the next.\footnote{This would not change the model's classification behavior, but implies that the hidden margin can be made arbitrarily large for the hidden layer in question.} Furthermore, the dimensionality of the hidden representations can also vary across different models, which further complicates comparison of distances. 

Given the above example, it is clear that when comparing hidden margins between different layers or models a form of normalization is required. Jiang et al.~\cite{predict_gen_margin} introduce a layer-specific normalization scheme, where each layer's margin distribution is normalized by the total feature variance (henceforth referred to as total variance (TV) normalization) to partially address this issue. 

The TV normalization is defined as follows. Let $\mathbf{X}^{l} \in R^{m \times n_l}$ denote the matrix of activation values\footnote{In the case of CNNs, the activation values for each sample are flattened to a vector in order to construct this matrix} at a layer $l$ of $m$ samples with $n_l$ features (dimensions) each.  The normalization term is given by the square root of the sum of the variance (calculated over all $m$ samples) of each feature, such that
\begin{equation}
\label{eq:norm}
    \text{TV}^{l} = \sqrt{\sum_{i=1}^{n_l} \text{Var}(\mathbf{X}^{l}_{\cdot, i})}
\end{equation}
where $\mathbf{X}^{l}_{\cdot,i} \in R^{m}$ represents the vector of sample activations for the $i^{th}$ dimension, meaning the $i^{th}$ column of $\mathbf{X}^{l}$. Furthermore, note that $\text{Var}()$ indicates the sample variance.

The normalized margin for a sample at a specific layer is then simply given by the margin value divided by the total variance at that layer. The intuition of this normalization scheme is that the scale of the activation values linearly scales with the square root of the total variance. 

\subsubsection{Hidden margins and generalization}

We now elaborate on works that have investigated the link between hidden margins and generalization. 

Elsayed et al.~\cite{large_margin_dnns} use the first-order Taylor approximation of the margin to formulate a penalty term which is used during training to ensure that each sample has at least a specific (chosen hyperparameter) distance to the decision boundary at both the input, hidden, and output layers. Alternatively, they use either only the input or a combination of both the hidden and output layers. In essence, the approximated margin is used as a regularization term in combination with a standard loss function. Networks trained using this margin regularization exhibit greater adversarial example robustness, as well as better generalization when trained on data with noisy labels. However, while this results in a measurable increase in average margin, it does not result in any significant gains in test accuracy when trained on standard data without noisy labels compared to vanilla cross-entropy loss. It is also worth noting that the authors do not normalize the margin measurements in this work.

In a seminal paper, Jiang et al.~\cite{predict_gen_margin} utilize the same approximation and calculate the hidden margins for a large number of training samples for a set of trained CNNs. They then extract summary statistics from these TV normalized hidden margin distributions for three selected hidden layers. Specifically, they select three hidden layers at equally spaced locations throughout the network, for example, the first, middle, and last hidden layers. Using these layers' summary statistics, they fit a linear regression model and are then able to (numerically) predict the models' generalization gap, including that of models not used to fit the linear predictor. This is done on the DEMOGEN dataset, mentioned earlier (recall Section~\ref{sec:back_demogen}). They report very high coefficient of determination values in this setting. We discuss this method in greater detail in Section~\ref{sec:hidden_lin_reg}.

In order to use hidden margins as a complexity measure for the purpose of the PGDL dataset (that is, as a complexity measure for the purpose of \textit{ranking} generalization), Natekar and Sharma~\cite{rep_based_complexity_pgdl} simply used the mean of the TV normalized Taylor-approximated margin distributions. Specifically, they calculate the mean TV normalized hidden margin for each layer, and then average this across all of the layers. They further demonstrate that this measure can be improved if margins are measured using `manifold Mixup' representations or the representations of augmented training samples. Specifically, manifold Mixup linearly interpolates between the hidden representations of two training samples (such as in Equation~\ref{eq:mixup}, but in the activation space). The intuition here is that while two models might have similar hidden margins for standard (seen) training samples, the hidden margin of `unseen' (i.e. synthetic) samples should be larger for a better generalizing model. They refer to this solution as `Mixup margin summary'. As to the augmented training samples, their exact methods are not specified, but they refer to it as `Augment margin summary.'

Chuang et al.~\cite{optimal_transport} also apply hidden margins to the PGDL dataset. Similarly to the others mentioned, they calculate the Taylor-approximated margin distributions for a large number of training samples but they instead summarize these distributions using their medians. In this case, they rely on only the median margin of the first or last layer as the final predictor. However, instead of TV normalization, they improve thereon by proposing an alternative cluster-aware normalization scheme ($k$-variance~\cite{kvariance}). That said, the improvements in predictive performance when using this normalization scheme are rather slight and do not outperform standard TV normalization in all cases. This solution is referred to as `$k$-Variance Gradient Normalized Margin First ($k$V-GN-Margin 1st)'. The authors view the Taylor-approximated margin of Equation~\ref{eq:taylor_approx_margin} as the output margin normalized by the gradients, hence the `Gradient Normalized' in the name. The `first' refers to using the first hidden layer for this calculation. We elaborate more on these hidden margin complexity measures in Section~\ref{sec:hidden_comp_measure}.


\subsection{Input margins}
\label{sec:back_input_margins}

We use the term \textit{input margins} to refer to distance to the decision boundary measurements that are confined to the input space of DNNs. 

Input space margins and decision boundaries have been considered from many different viewpoints, and have been used as tools to probe adversarial robustness~\cite{stutz2019disentangling}, double descent~\cite{twice_dd_real_ver}, interpretability~\cite{constrained_optim}, and generalization~\cite{guan_linear_interpol} (among others). We limit our focus to the various methods that have been used to measure input margins, and the links (if any) that have been made to generalization.

\subsubsection{White-box adversarial attacks}

The most widespread use of input margins is in the field of white-box adversarial attacks and requires special consideration. The aim of an adversarial attack is to generate adversarial examples: samples that are perceptually identical to correctly classified samples but are incorrectly classified by DNNs~\cite{goodfellow_adversarial}. This is achieved by slightly perturbing some training or test sample until its classification changes, which is in effect very similar to finding a point on the decision boundary (Equation \ref{eq:minimization}). Note that we always refer to these as `adversarial examples' throughout this study, so as to prevent confusion with the training or test sample being perturbed. The term `white-box' refers to the situation where an attacker has full access to the model and its parameters, and can thus utilize gradient information. To this end, several methods have been developed, such as: The fast gradient sign method (FGSM)~\cite{goodfellow_adversarial}, projected gradient descent attack (PGD)~\cite{pgd_attack}, the Carlini\&Wagner attack (C\&W)~\cite{carlini2017towards}, DeepFool~\cite{deepfool}, and fast adaptive boundary attack (FAB)~\cite{fab_attack}, among others. The goals of these methods are 1) to find the smallest perturbation possible that fools the model, and 2) to do so in a computationally inexpensive way. As one would imagine, these goals are often opposed. 

Of particular relevance to measuring margins are adversarial attack methods that are unbounded, that is, not limited to a specific perturbation bound. As a counterexample, the PGD attack attempts to find a point that maximizes the loss within a set $\epsilon$-ball surrounding a correctly classified sample, and the attack is thus bounded within this region~\cite{pgd_attack}. When considering unbounded methods, DeepFool and FAB are the most popular.

The DeepFool algorithm is an iterative variant of the Taylor approximation described earlier (Equation~\ref{eq:taylor_approx_margin}). For a sample, the Taylor-approximated distance is calculated between the sample's true class and each other class. The sample is then perturbed in the direction of the class with the smallest approximated margin. This is then repeated until the sample's classification changes~\cite{deepfool}. We provide a thorough overview of the DeepFool algorithm in Section~\ref{sec:hidden_beyond_approximations}. The FAB attack is a more complex improvement on DeepFool, which includes an additional projection for handling bound constraints, random restarts, and biases each step towards the original point, among other improvements. They demonstrate that they are able to find smaller perturbations compared to DeepFool~\cite{fab_attack}, albeit at greater computational cost.

\subsubsection{Other measurement methods}

There are several other methods besides adversarial attacks that have been used to measure input margins.
Youzefsadeh and O'Leary~\cite{constrained_optim_investigating} formulate finding the nearest point on the decision boundary as a constrained optimization problem.  They solve this using an off-the-shelf optimizer, for example, the class of optimization methods found in the NLOpt software package~\cite{nlopt}. While this method is likely very precise, it comes at a great computational cost for high-dimensional input data (such as natural images). To alleviate this, dimensionality reduction methods are used prior to model training~\cite{constrained_optim_investigating}. We provide additional detail on the constrained optimization formulation in Section~\ref{sec:noise_margin_formulation}.

Somepalli et al.~\cite{twice_dd_real_ver} measure input margins by using a simple bisection search in $10$ random directions from each training sample. They then define a quantity called `mean margin' which is given as the average of the distance in these $10$ random directions. In a similarly approximate way, some authors simply linearly interpolate between training samples~\cite{constrained_optim_investigating, deep_dig, guan_linear_interpol} and measure the distance from the sample to the interpolated point at which the classification changes. However, this method is unlikely to be an accurate estimation~\cite{deep_dig}. Finally, Karimi et al.~\cite{deep_dig} introduce \textbf{Deep} \textbf{D}ecision boundary \textbf{I}nstance \textbf{G}eneration (DeepDIG), an autoencoder-based method to generate samples on or near the decision boundary.  This requires training an autoencoder for each class pair for a specific model, which is a computationally expensive endeavor.

\subsubsection{The link to generalization}

Unlike hidden margins, the link between large input margins and improved generalization is less clear. In fact, the opposite may be true: several authors have investigated the link between adversarial robustness and generalization and often conclude that an inherent trade-off exists~\cite{robustness_tradeoff,robustness_tradeoff_two, robustness_tradeoff_three,robustness_accuracy_tradeoff}. This is commonly referred to as the `robustness-accuracy trade-off'~\cite{robustness_accuracy_tradeoff}. That said, this conclusion is still being debated~\cite{stutz2019disentangling,robustness_good_1,robustness_good_2}. 

Some authors investigate the relationship between adversarial retraining and generalization performance~\cite{robustness_accuracy_tradeoff,robustness_tradeoff,robustness_tradeoff_three}. Adversarial retraining involves retraining a model on adversarial examples, which should result in the model being more robust against such perturbations at inference time, i.e. the retraining is expected to increase the model's margins~\cite{goodfellow_adversarial}. Tsipras et al.~\cite{robustness_tradeoff} find that the adversarial retraining of a model consistently results in decreased performance on a standard (non-adversarial) test set. They further argue that this is due to adversarially trained models learning very different feature representations than non-adversarially trained models. Furthermore, this tension between adversarial retraining and generalization has also resulted in the development of several methods that attempt to increase robustness (i.e., increase the size of the input margin) without sacrificing generalization performance~\cite{robustness_acc_GAIRT,robustness_acc_TRADES, robustness_acc_MMA,xu2023exploring,large_margin_dnns}. 

When considering the robustness and generalization of `conventionally trained' models, Su et al.~\cite{robustness_tradeoff_two} find that models with higher test accuracy (on standard data) are more vulnerable to adversarial attacks (less adversarially robust). Specifically, they perform a large evaluation on $18$ ImageNet models (such as ResNet~\cite{resnet}, InceptionNet~\cite{inceptionnetv3}, and VGG~\cite{vgg} architectures) and find a negative relationship between the model's test accuracy and its robustness. On the other hand, Rozsa et al.~\cite{robustness_good_2} compare the adversarial robustness of eight models on the ImageNet dataset, and find that better generalizing models are also more robust. Gilmer et al.~\cite{robustness_good_1} also provide theoretical motivation for the latter conclusion and empirical results on a simple dataset.

Despite this area of research, we are not aware of any study that has analyzed input margins for the purpose of generalization prediction. That is, input margins have not been evaluated as a standalone complexity measure. In this study, we fill this gap.

\subsection{Output margins}
\label{sec:back_output_margins}

Output margins are distinct from hidden and input margins in that they are simple to measure and directly correspond to the difference in class output values.  The output margin $\gamma$ for a sample $\mathbf{x}$ between two classes $i$ and $j$ for a model $f$ is given by
\begin{equation}
\label{eq:output_margin}
    \gamma = f_i(\mathbf{x}) - f_j(\mathbf{x}) 
\end{equation}

It is quite common in the literature to simply refer to this quantity as `margin' and seems to be the most common usage of the term. For example, see usages in \cite{yang2020boundary}, which easily leads to confusion. 

Output margins are often combined with norm-based measures and are frequently found in studies of generalization bounds. Bartlett et al.~\cite{output_margin_spectral} derive an output margin-based generalization bound normalized by the product of the spectral norm of the layer weights. Previous work used the path norm of the layer weights~\cite{path_norm}. Similarly, Liang et al.~\cite{fisher_rao_norm} propose using the Fisher-Rao metric~\cite{fisher_rao_measure}, which is a lower bound on the path norm. However, Jiang et al.~\cite{fantastic_gen} evaluate all of these measures in addition to other variants and show that these measures generally poorly predict generalization.

More recently, Chuang et al.~\cite{optimal_transport} introduce novel generalization bounds and normalize output margins with $k$-variance~\cite{kvariance}. Recall that the same is done for hidden margins (see Section~\ref{sec:back_hidden_margins}). They demonstrate that this outperforms the spectral normalization mentioned earlier on the PGDL dataset. This method is called `$k$V-Margin 1st', as the $k$-Variance of the first layer's activations is used to normalize the output margin. This method performs poorly in comparison to hidden margins in general; see Table~\ref{tab:pgdl_task_solutions_score_summary}.

In this work our focus is on empirical investigations of generalization as opposed to theoretically motivated bounds. Given the limitations of output margins in this regard, we restrict our analysis to input and hidden margins. 

%% file: chapters/ch3_sample_noise.tex
\let\cite\parencite

\lhead{}
\rhead{}
\chapter{Margins and sample noise}
\label{chap:ch3}
\afterpage{\lhead{\ifthenelse{\thepage>0}
       {\it Chapter \thechapter }
      }
\rhead{\ifthenelse{\thepage>0}
       {\it \let\uppercase\relax\leftmark}
      }}
\underline{ \hspace{\textwidth} } 
\textit{``Now the stranger started talking, made it plain to folks around \\
Was an Arizona ranger, wouldn't be too long in town \\
He came here to take an outlaw back alive or maybe dead \\
And he said it didn't matter he was after Texas Red''} \\ - Marty Robbins, \textit{Big Iron}, Verse 4\\
\underline{ \hspace{\textwidth} }

\section{Introduction}
\label{sec:noise_introduction}

In this chapter, we do a careful analysis of the relationship between different types of sample noise and margins. Our main focus is on input margins. Margin-based complexity measures typically use a summary statistic, such as an average. We know that not all samples are equally useful for generalization and that DNNs can easily interpolate noisy samples~\cite{zhang2021understanding}. In this chapter, we aim to establish whether there might be some characteristics of margin measurements that are not captured by such summaries. 

Adding noise to a training set and investigating the ability of DNNs to generalize in spite of this corruption is a popular technique in empirical investigations of generalization. A good example of success with such methods is the seminal paper by Zhang et al.~\cite{zhang2021understanding}, where it was shown that overparameterized models can generalize despite having enough capacity to fit per-sample random noise such as label corruption or randomized input features. Similar noise has been extensively used to experimentally probe ANNs~\cite{neyshabur2017exploring_gen_def_2,nakkiran_dd,theunissen2020benign}. However, no careful analysis has been done on the effect of these noisy samples on margin measurements.

The choice of \textit{input} margins might appear counter-intuitive for probing generalization in this setting, given that no strong link between large input margins and generalization has been previously established in DNNs (as explained in Section~\ref{sec:back_input_margins}). However, as we shall show, we analyze these margins in a controlled setting where such a relationship is present. Furthermore, we also find that these results extend to hidden margins, which have been shown to exhibit a stronger link with generalization. 

The chapter is structured as follows. First, we explain our experimental setup in Section~\ref{sec:noise_exp_setup}, and introduce the models and sample corruption that we are investigating, as well as how we measure input classification margins. Following this, in Section~\ref{sec:noise_results}, we present our experimental results and point out several interesting observations. Next, we have a further analysis of these results in Section~\ref{sec:noise_analysis}. Finally, in Section~\ref{sec:noise_hidden_margins} we extend our experiments to hidden margins and compare these to input margins.

\section{Experimental setup}
\label{sec:noise_exp_setup}

In order to investigate how sample corruption affects margin measurements, we train several networks of increasing size (e.g. number of hidden units per layer or number of convolutional channels) to the point of interpolation (close to zero training error) on the widely used classification datasets MNIST~\cite{Lecun98gradient-basedlearning} and CIFAR10~\cite{cifar}. We also corrupt the training data of some models using two types of noise, defined in Section~\ref{sec:noise_controlled_noise}, separately. By training models of varying sizes, we are in effect varying the representational capacity of the models, as such we refer to the model size as the model capacity from this point onwards. Given that capacity and generalization are strongly linked, we expect generalization to improve systematically with an increase in capacity in the overparameterized regime~\cite{nakkiran_dd}. This setup allows us to investigate how margin measurements are affected by 1) model capacity, 2) generalization, and 3) different types of sample noise. The rest of the section describes this in greater detail: First, we provide a description of the two types of sample corruption used in Section~\ref{sec:noise_controlled_noise}. Following this, in Section~\ref{sec:noise_models}, we provide an explanation of the trained models. Finally, we explain how we measure input margins in Section~\ref{sec:noise_margin_formulation}.

\subsection{Controlled noise}
\label{sec:noise_controlled_noise}

We use two specific types of noise, inspired by Zhang et al.~\cite{zhang2021understanding}: Label corruption and Gaussian input corruption. These have been designed to represent two complications that are often found in real-world data and could affect the generalization of a model fitted to them. Label corruption represents noise that comes from mislabeled training data (mislabeling is common in large real-world datasets and even in benchmark datasets~\cite{northcutt2021pervasive}), inter-class overlap, and general low separability of the underlying class manifolds. 

Gaussian input corruption, on the other hand, represents extreme examples of out-of-distribution samples. These are off-manifold samples that display a high level of randomness. Such samples do not necessarily obscure the true underlying data distribution, assuming only a small portion of the training data is corrupted, but still require sufficient capacity to fit the excessive complexity that needs to be approximated when fitting samples with few common patterns~\cite{theunissen2020benign}.

\begin{figure}[h]
    \includegraphics[width=\linewidth]{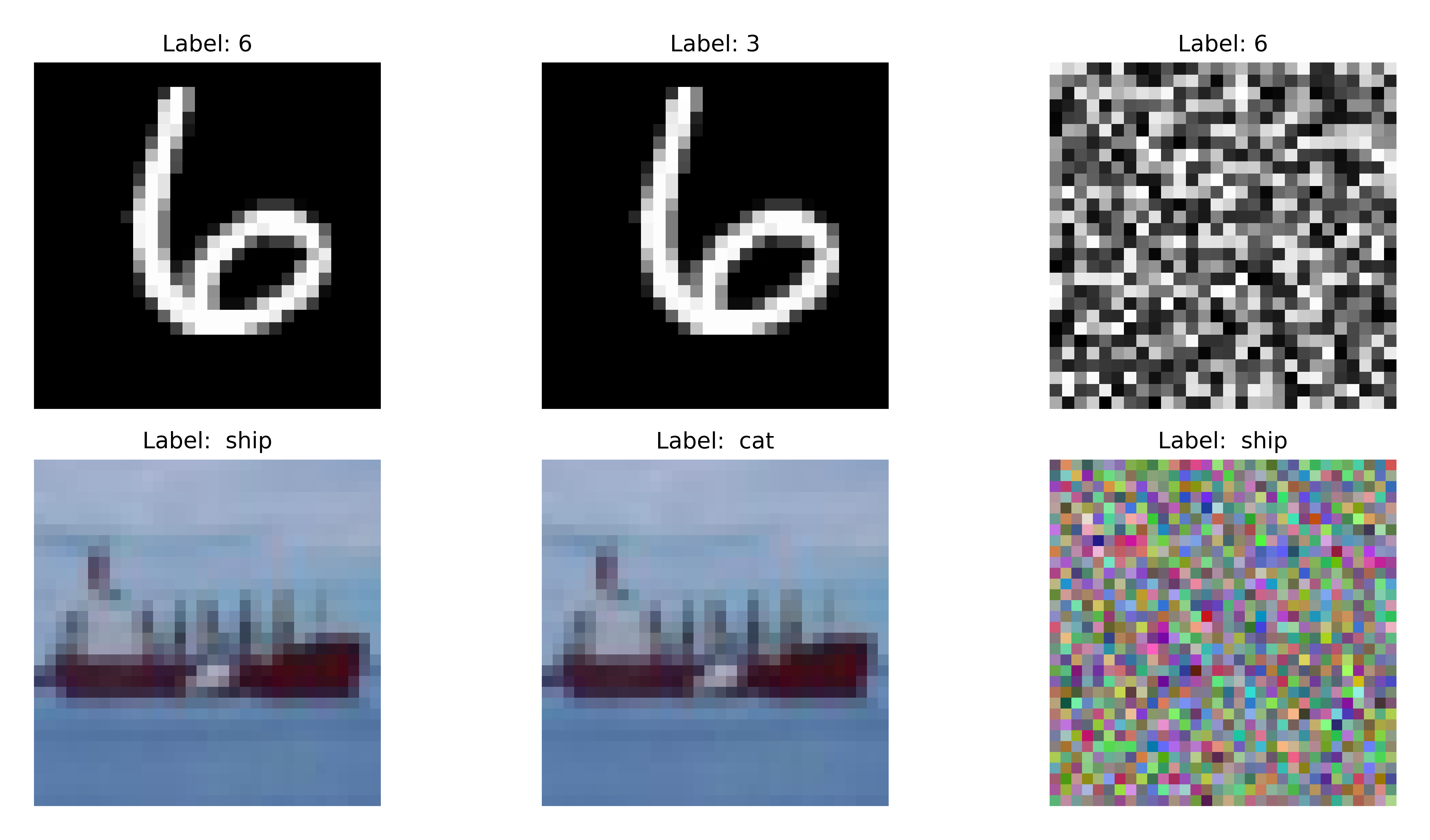}
    \caption[Example of label corruption and Gaussian input corruption]{Example of label corruption and Gaussian input corruption for MNIST (top) and CIFAR10 (bottom). Left: Original training sample. Middle: Label corrupted sample. Right: Gaussian input corrupted sample.}
    \label{fig:noise_corruption_example}
\end{figure}

Given a training sample and label $(\mathbf{x}, c)$ where $\mathbf{x} \in \mathbb{R}^{d}$ and $c \in N$ for a set of classes $N$, the corruption of a sample can be defined as follows:
\begin{itemize}
    \item \textit{Label corruption}: $(\mathbf{x}, c) \rightarrow (\mathbf{x}, \hat{c})$ where $\hat{c} \neq c, \hat{c} \in N$.
    \item \textit{Gaussian input corruption}: $(\mathbf{x}, c) \rightarrow (\mathbf{g}, c)$ where $\mathbf{g} \in \mathbb{R}^{d}$ and each value in $\mathbf{g}$ is sampled from $\mathcal{N}(\mu_{\mathbf{x}}, \sigma_{\mathbf{x}})$.
\end{itemize}
Alternative labels are selected at random and $\mathcal{N}(\mu_{\mathbf{x}}, \sigma_{\mathbf{x}})$ is a normal distribution, with $\mu_{\mathbf{x}}$ and $\sigma_{\mathbf{x}}$ the mean and standard deviation of all the features in the original sample $\mathbf{x}$. More precisely, the mean and standard deviation of the sample's entire feature vector. Figure~\ref{fig:noise_corruption_example} shows an example of both corruption types for the MNIST and CIFAR10 datasets. Also note that we sometimes drop the `Gaussian' when referring to `Gaussian input corruption' for brevity.

\subsection{Models}
\label{sec:noise_models}

For both the MNIST and CIFAR10 datasets, we train several models of increasing capacity on 1) the original unadulterated dataset, 2) a dataset where a portion of the training set samples are input-corrupted, and finally 3) where a portion of the training set samples are label-corrupted. This results in three sets of models for each dataset. What follows is a detailed explanation of this setup for the two datasets considered.

\subsubsection{MNIST}

The MNIST dataset~\cite{Lecun98gradient-basedlearning} is useful for probing generalization, as it is a simple dataset with well-separated classes. Therefore, results on this dataset are easier to interpret than on more complicated image classification benchmarks. Furthermore, MNIST can be fit by very small DNN models. However, due to the simple nature of the MNIST dataset, it is possible that it shows behaviors which are not present for more complicated tasks.  We train three distinct sets of multilayer perceptrons (MLPs) on this dataset, each set containing models of identical depth and identically varied width: 
\begin{itemize}
    \item \textbf{MNIST}: A set of clean MNIST models. These serve as baselines, showing the level of generalization and margin sizes expected should the models not have been trained on corrupted data. 
    \item  \textbf{MNISTlc} (MNIST-label-corrupted): Models with the same set of capacities as the previous model set, but where $20\%$ of the training set is label-corrupted.
    \item \textbf{MNISTgic} (MNIST-Gaussian-input-corrupted): Models with the same set of capacities as the clean models; however, $20\%$ of the training set is input-corrupted.
\end{itemize}

Note that for all three model sets, the validation set is left untouched. We only corrupt the training data.
All models for these tasks have the following hyperparameters in common. They use the same $55\ 000/5\ 000$ train-validation split of the training data. They are all single hidden layer ReLU-activated MLPs with widths ranging from $100$ to $10\ 000$ hidden layer nodes and a single bias node. Stochastic gradient descent (including momentum terms) is used to minimize the cross-entropy loss on mini-batches of size $64$ selected at random. The initial learning rate is set to $0.01$ and then multiplied by $0.99$ every $5$ epochs.
For each set, we further train each model using three different random initializations. 

Note that we train the \textbf{MNISTlc} models for $1\ 000$ epochs and the models from the other two sets for $100$ epochs. This is because the label-corrupted dataset requires more epochs to interpolate. We intentionally do not perform any per-model hyperparameter tuning or use early stopping. This is so that the only variation between models within a set is that of their representational capacity. Furthermore, we do not include any other explicit forms of regularization (e.g. weight decay or dropout), so that the models can easily interpolate these noisy samples.  We find that all models perfectly interpolate their training data (i.e. achieve $100\%$ train accuracy), except for the very smallest capacity (width=$100$) of the \textbf{MNISTlc} models.

The resulting generalization ability of all three sets is depicted in Figure~\ref{fig:noise_performance} (left). Note that, as expected for all three sets, with more capacity we see an improvement in validation set performance. Also note that only label corruption results in any noteworthy reduction in validation performance, as also previously reported by Theunissen et al.~\cite{theunissen2020benign}.

\subsubsection{CIFAR10}

We replicate these MNIST models using CNNs trained on the CIFAR10 dataset, with $10\%$ training set corruption (where applicable). The CIFAR10 dataset~\cite{cifar}, unlike MNIST, is a more complicated image classification benchmark with natural images. This dataset includes a large degree of inter-class overlap, and is considered a more difficult problem. We use a similar architecture to the `standard' CNN used by Nakkiran et al.~\cite{nakkiran_dd}. Each CNN consists of four ReLU-activated convolutional layers, with $[1k, 2k, 4k, 8k]$ output channels, respectively, where we choose various values of $k$ between $10$ and $64$ to create a group of models with varying capacity.  Each model also includes max and average pooling layers, and a final fully connected layer of $400$ nodes. The complete architectural description is shown in Table~\ref{tab:noise_cnn_architecture}. The three sets of CIFAR10 models \textbf{CIFAR10}, \textbf{CIFAR10lc}, and \textbf{CIFAR10gic}, refer to clean, label-corrupted, and input-corrupted models, respectively. We opt to use $10\%$ corruption as opposed to the $20\%$ used for MNIST models, as $20\%$ would require much larger models to interpolate the training data. As with MNIST, the validation set is left untouched and is not corrupted.

\begin{table}[h]
\centering
\caption[Convolutional neural network architecture for CIFAR10 models]{Convolutional neural network architecture for \textbf{CIFAR10}, \textbf{CIFAR10lc}, and \textbf{CIFAR10gic} models}
\label{tab:noise_cnn_architecture}
\begin{tabular}{@{}lllll@{}}
\toprule
\textbf{Layer type} & \textbf{Output Channels} & \textbf{Kernel size} & \textbf{Stride} & \textbf{Padding} \\ \midrule
Convolution & k & 3 $\times $ 3 & 1 & Same \\
Convolution & 2k & 3 $\times $ 3 & 1 & Same \\
Max pool & 2k & 2 $\times $ 2 & 2 & Valid \\
Convolution & 4k & 3 $\times $ 3 & 1 & Same \\
Max pool & 4k & 2 $\times $ 2 & 2 & Valid \\
Convolution & 8k & 3 $\times $ 3 & 1 & Same \\
Avg pool & 8k & 8 $\times $ 8 & 1 & Valid \\
Fully connected & 400 (nodes) & - & - & - \\ \bottomrule
\end{tabular}%
\end{table}

\begin{figure}[h]
    \centering
    \includegraphics[width=0.48\linewidth]{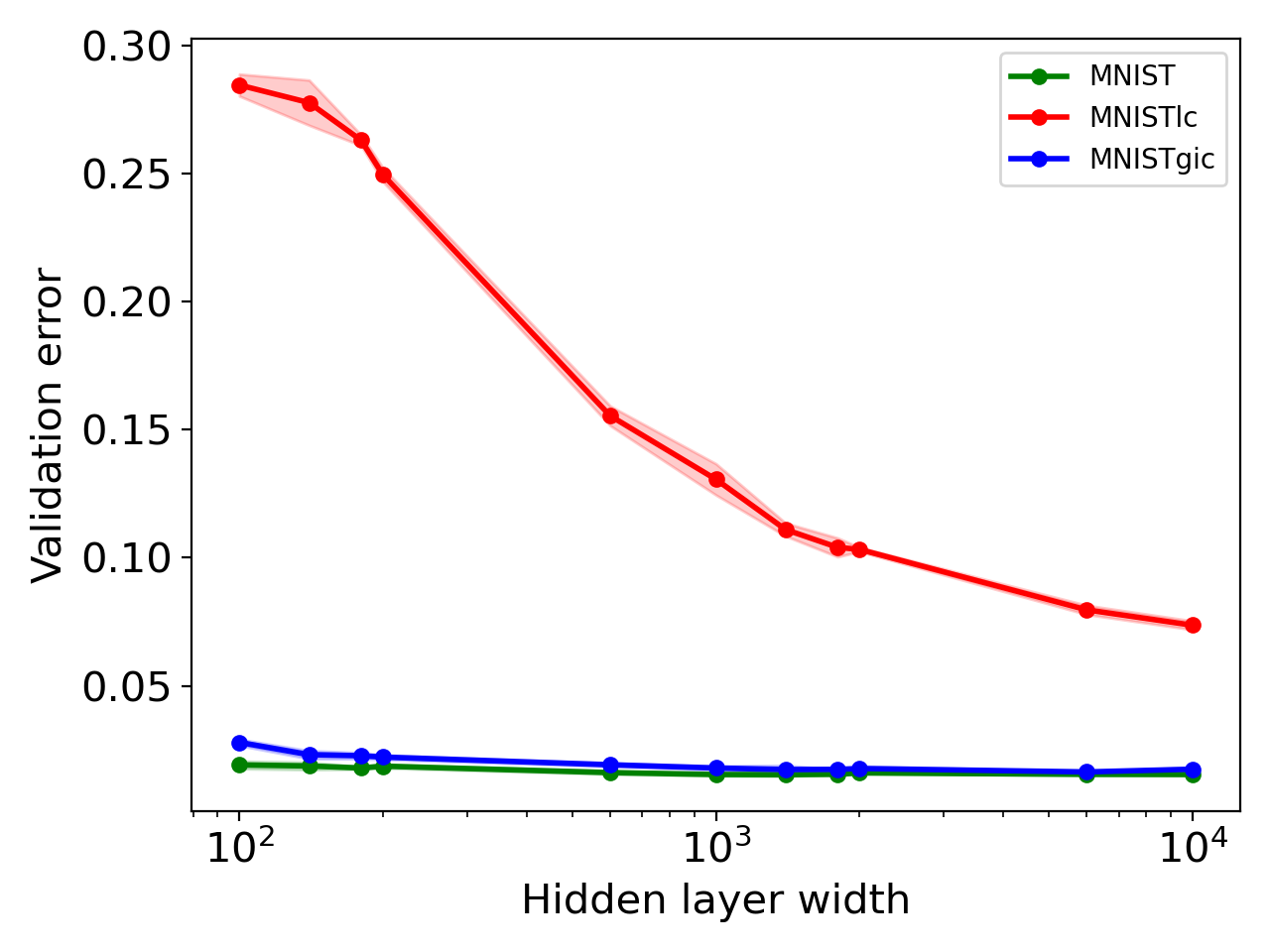}
    \includegraphics[width=0.48\linewidth]{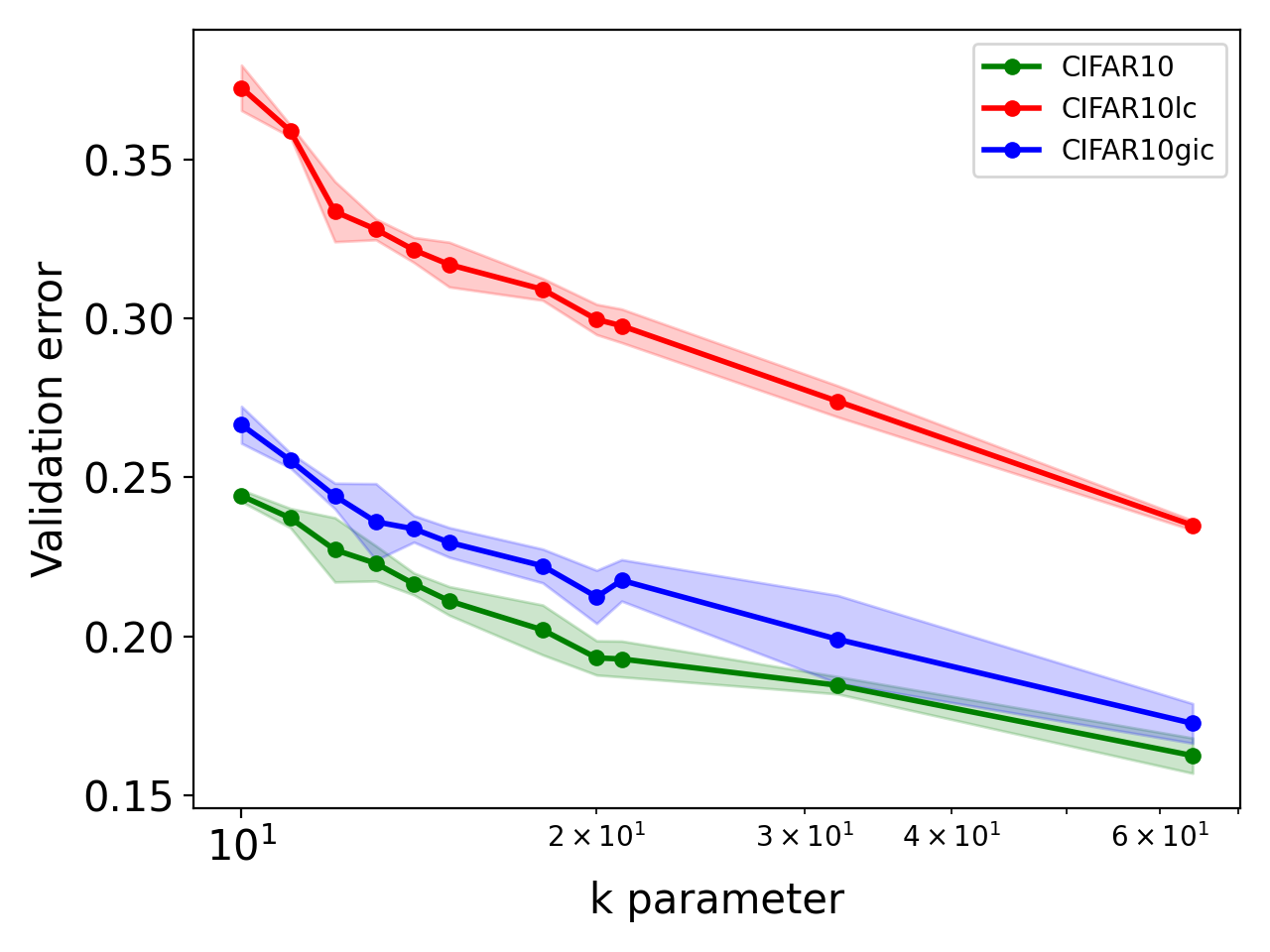}
    \caption[Validation error for MNIST and CIFAR10 models]{Validation error for MNIST models (left) and CIFAR10 models (right). Values are averaged over three random seeds and shaded areas indicate standard deviation.}
    \label{fig:noise_performance}
\end{figure}

All models are trained on a $45\ 000/5\ 000$ train-validation split of the CIFAR10 dataset. These models are trained for $500$ epochs in order to minimize a cross-entropy loss function on mini-batches of $256$ samples using the Adam optimizer. The initial learning rate of $0.001$ is multiplied by $0.99$ every $10$ epochs. We once again make use of three different random initializations for each model. Similarly to the MNIST models, we do not use any regularization or early stopping. This protocol results in all models perfectly interpolating the training data, with the exception of the smallest \textbf{CIFAR10lc} capacity model ($k = 10$). The maximum train error (across all MNIST and CIFAR10 models) is $0.0573$. From the relevant validation errors in Figure~\ref{fig:noise_performance} (right), we again observe that more capacity is accompanied by better generalization performance.  

The generalization behavior for each type of sample corruption shown by these models are expected and similar to that shown in prior work~\cite{theunissen2020benign, twice_dd_real_ver, nakkiran_dd}. This provides us with the necessary setup for the margin experimentation that follows in Section~\ref{sec:noise_results}.

In the following subsection, we explain our terminology when referring to different types of samples within these distinct model sets.

\subsubsection{Terminology}

When describing margin behavior, we refer to different subselections of margins based on the type of sample (clean or corrupted) as well as the type of model.
To avoid confusion, we always refer to these subselections using a name constructed from the type of sample and then the type of model, separated by a colon, as shown in Table~\ref{tab:noise_terminology}. 
For example, if we refer to the margins for the clean (uncorrupted) samples with regard to a label-corrupted model, we refer to them as the \textit{clean:label-corrupted} margins.  We do not specifically refer to a margin as either an input or hidden margin in the text where the context is clear. Only input margins are discussed up to Section~\ref{sec:noise_analysis}, with the first hidden margin results introduced in Section~\ref{sec:noise_hidden_margins}.
\begin{table}[H]
\centering
\caption[Sample corruption terminology]{Sample corruption terminology. Overall samples refer to all the samples types (clean + corrupt) present for the model.}
\label{tab:noise_terminology}
\begin{tabular}{@{}cccc@{}}
\toprule
 & \textbf{clean samples} & \textbf{corrupt samples} & \textbf{overall samples} \\ \midrule
\textbf{clean model} & \textit{clean:clean} & N/A & \textit{overall:clean} \\ \midrule
\textbf{\begin{tabular}[c]{@{}c@{}}label-\\ corrupted\\ model\end{tabular}} & \textit{clean:label-corrupted} & \textit{corrupt:label-corrupted} & \textit{overall:label-corrupted} \\ \midrule
\textbf{\begin{tabular}[c]{@{}c@{}}input-\\ corrupted\\ model\end{tabular}} & \textit{clean:input-corrupted} & \textit{corrupt:input-corrupted} & \textit{overall:input-corrupted} \\ \bottomrule
\end{tabular}
\end{table}

\subsection{Measuring the classification margin}
\label{sec:noise_margin_formulation}

Here, we delineate how we measure input margins for the purposes explained earlier. As was made clear in Section~\ref{sec:back_input_margins}, there are several methods available to measure margins. In this case, we wish to do a controlled comparison between the margins of different types of samples, and we therefore opt to use the most precise method available, instead of relying on more approximate methods. Therefore, we formulate measuring a margin as a constrained optimization problem~\cite{constrained_optim}:

Let $f~:~\mathbf{x}\rightarrow \mathbb{R}^{|N|}$ denote a classification model with a set of output classes $N = \{1\ldots n\}$, and $f_k(\mathbf{x})$ the output logit value of the model for input sample $\mathbf{x}$ and output class $k$.  

For a correctly classified input sample $\mathbf{x}$, the objective is to find the closest point $\mathbf{\hat{x}}$ on the decision boundary between the true class $i$ (where $i = \argmax_k (f_k(\mathbf{x}))$, i.e. the predicted class) and another class $j \neq i$, $j \in N$.
Formally, $\mathbf{\hat{x}}$ is found using some relevant distance function $d$ by solving the constrained minimization problem:
\begin{equation}
\label{eq:objective_func}
\argmin_{\mathbf{\hat{x}}} d(\mathbf{x},\mathbf{\hat{x}})
\end{equation}
such that
\begin{equation}
\label{eq:eq_constraint}
    f_i(\mathbf{\hat{x}}) = f_j(\mathbf{\hat{x}}) 
\end{equation}
\begin{equation}
    \mathbf{\hat{x}} \in [l, u]^{\dim(\hat{\mathbf{x}})}
\end{equation}
where $l$ and $u$ are the lower and upper bounds of the search space, respectively, while $i$ and $j$ are as defined above.

Finding a point that exactly meets the condition defined in Equation~\ref{eq:eq_constraint} is virtually impossible. In practice, a threshold $\epsilon$ is used, so that a point is considered valid (on the decision boundary) if $|f_i(\mathbf{\hat{x}}) - f_j(\mathbf{\hat{x}})| \leq \epsilon$. We use a default of $\epsilon = 10^{-3}$.  To find the closest point on the decision boundary for all $j$, we search over each class $j \neq i$, $j \in N$ separately for each sample and choose the one with the smallest distance. Furthermore, the search space is restricted such that $\hat{\mathbf{x}} \in [0, 1]^{\dim(\hat{\mathbf{x}})}$, as the data is normalized to these ranges prior to training. As is convention for margin measurements \cite{constrained_optim,twice_dd_real_ver}, we use Euclidean distance\footnote{In practice, we optimize for the squared Euclidean distance in order to simplify the gradient calculations, but report on the unsquared distance in all cases.} as metric, meaning that the margin between classes $i$ and $j$ is given by:
\begin{equation}
d_{i,j}(\mathbf{x}, \mathbf{\hat{x}}) =  ||\mathbf{x} - \mathbf{\hat{x}}||_{2}
\end{equation}

In order to solve this constrained minimization problem (CMP) for each sample, we follow Yousefzadeh and O'Leary~\cite{constrained_optim} and make use of an off-the-shelf constrained optimization method. We experiment with several methods available in the NLOpt (non-linear optimization) library~\cite{nlopt} in Python. We find that the augmented Lagrangian method~\cite{nlopt_lagrange_1,nlopt_lagrange_2} combined with the conservative convex separable approximation quadratic (CCSAQ) optimizer~\cite{ccsaq} for each unconstrained optimization step performs best.

\section{Results}
\label{sec:noise_results}

We calculate the margins for $10\ 000$ randomly selected training samples, for all of the models defined in Section~\ref{sec:noise_exp_setup}. Only correctly classified samples are considered. This amounts to solving $11 \text{ capacities } \times 3 \text{ random seeds } \times 3 \text{ datasets } \times 9 \text{ class pairs } \times 10k \text{ samples } = 8\ 910k$ individual CMPs for both the MNIST and CIFAR10 models.  In order to solve such a large number of CMPs we utilize $240$ CPU cores split over $10$ servers, by making use of GNU-Parallel~\cite{gnu_parallel}. 

First, we investigate the mean margins of these models for clean and corrupted samples separately in Section~\ref{sec:noise_mean_margin}. Following this, in Section~\ref{sec:noise_margin_distributions}, we take a closer look at the margin distributions. Finally, we summarize our observations in Section~\ref{sec:noise_results_summary}.

\subsection{Mean margins}
\label{sec:noise_mean_margin}

\begin{figure}[h]
    \centering
    \includegraphics[width=0.49\linewidth]{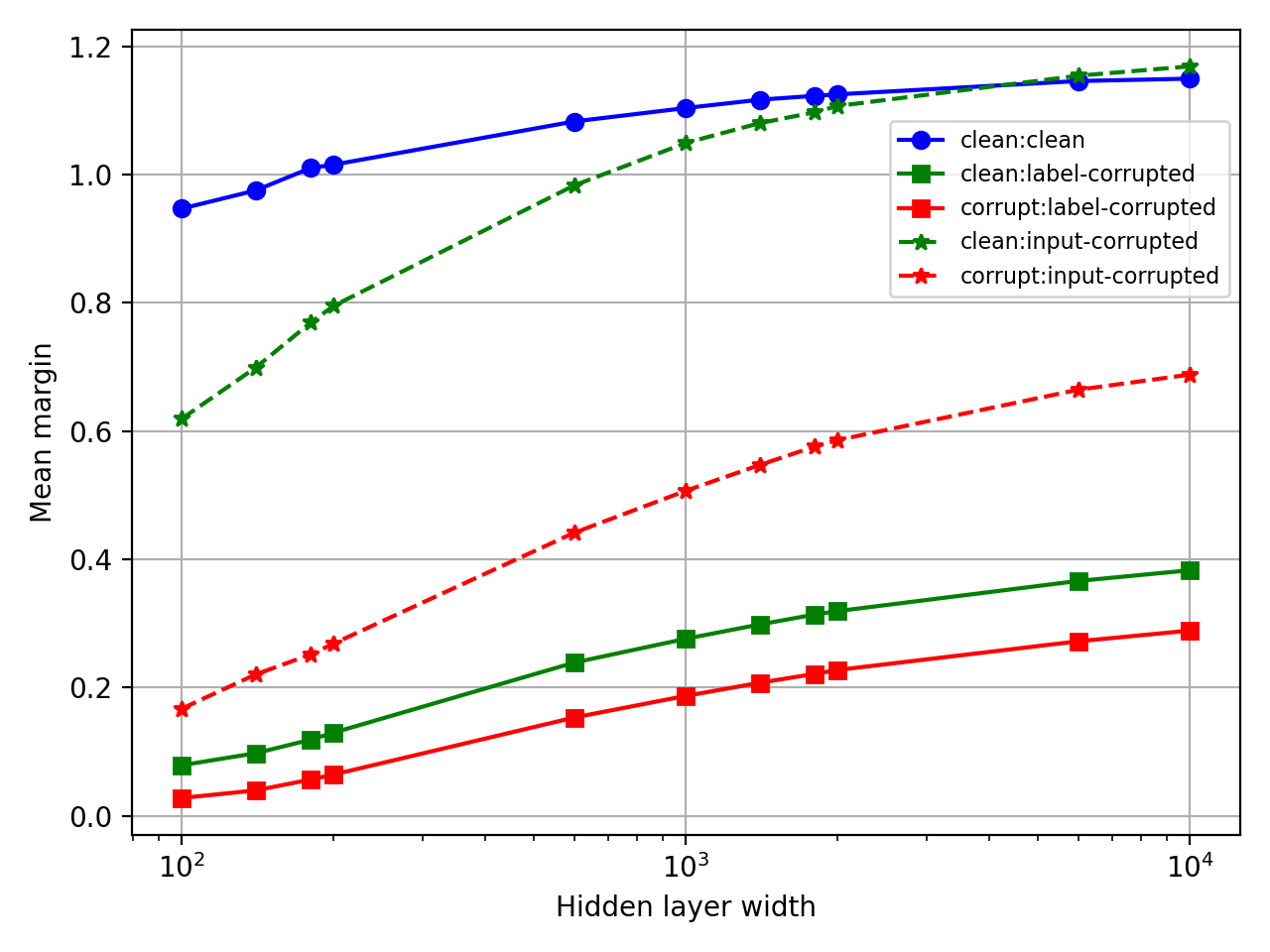}
    \includegraphics[width=0.49\linewidth]{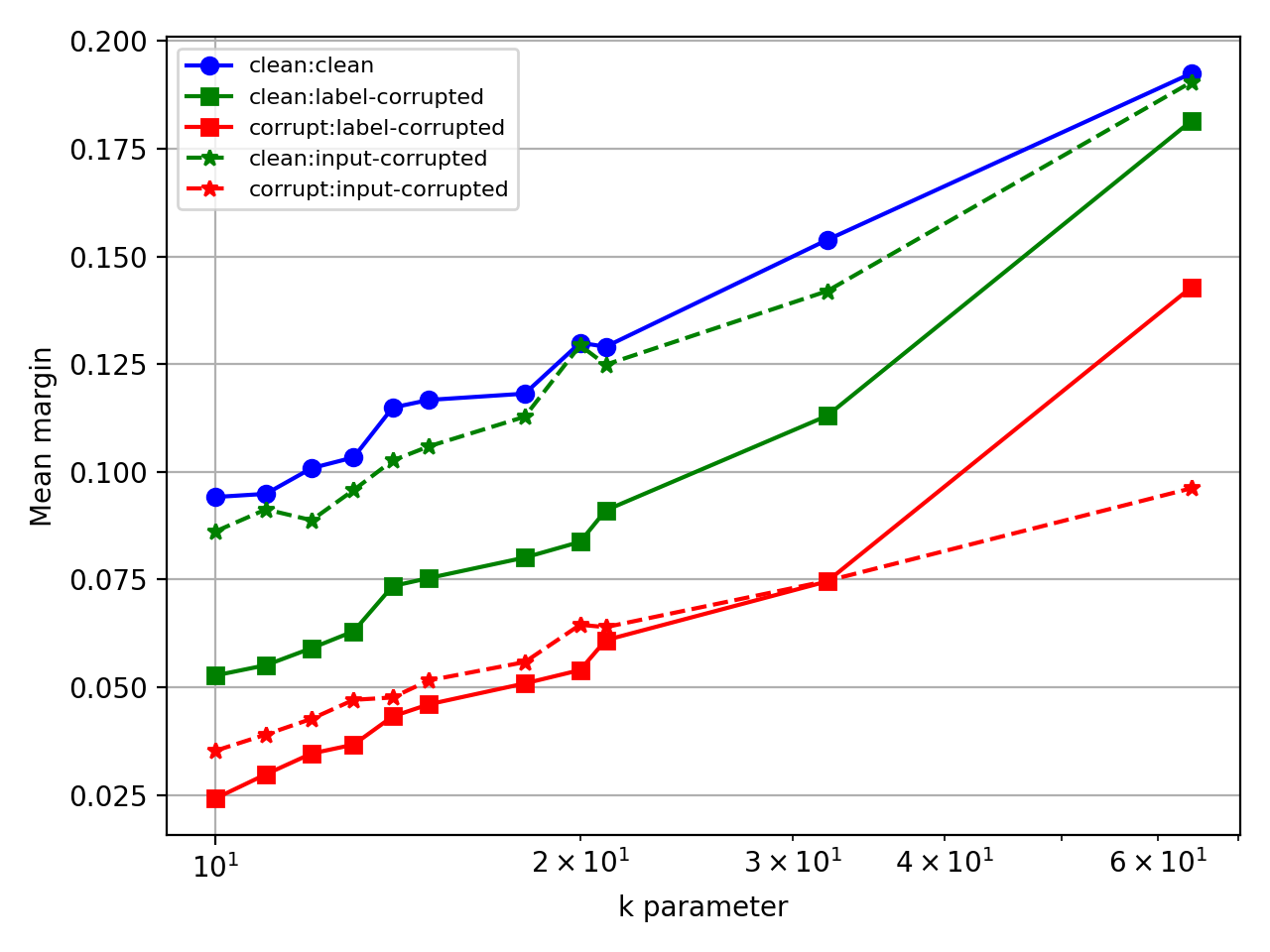}
    \caption[Mean margins for all MNIST and CIFAR10 models]{Mean margins for all MNIST (left) and CIFAR10 (right) models as a function of model capacity.}
    \label{fig:noise_margin_means_all_models}
\end{figure}

Figure~\ref{fig:noise_margin_means_all_models} shows the mean margins as a function of model capacity (for MNIST, left and CIFAR10, right), with the corrupt and clean samples from each set shown separately. 
There are several interesting observations and clear behavioral differences between the different types of samples considered. In this section, observations are listed -- these will be referred back to when we take a deeper look at the underlying mechanisms that contribute to these results in Section~\ref{sec:noise_analysis}.

Firstly, and perhaps most importantly, we note that the mean margin tends to increase along with capacity, in all cases, for all types of samples. This shows that within this controlled setting, the mean input margin appears to be a reliable indicator of generalization, as the generalization performance also increases along with capacity in each model set. Second, it is observed that the margins of corrupted samples (\textit{corrupt:input-corrupted} and \textit{corrupt:label-corrupted} indicated in red) tend to be smaller on average than those of clean samples within the same model.

Third, it is also clear that \textit{clean:input-corrupted} margins are similar in size to \textit{clean:clean} margins at large capacities. On the other hand, we observe that \textit{clean:label-corrupted} margins are much smaller than \textit{clean:clean} margins at all capacities. This implies that the introduction of label-corrupted samples leads to a large decrease in the margins of their clean counterparts, while the introduction of input corruption does not. This makes sense, since label corruption leads to a large decrease in generalization performance, while input corruption does not (see Figure~\ref{fig:noise_performance}). 

There are also discrepancies when comparing the MNIST and CIFAR10 results. 
It is interesting that the average margins, measured on CIFAR10, are smaller than the MNIST ones. The CIFAR10 samples contain $3\ 072$ features, while MNIST only $784$. We expect Euclidean distance in higher dimensions to tend to be larger. One could hypothesize that the reason for this contradiction is tied to the inherent difficulty of the CIFAR10 dataset, which contains a large degree of inherent inter-class overlap: more complicated decision boundaries are required to fit CIFAR10 than MNIST, and thus the margins are smaller. 

Another discrepancy between the margin trends of these two datasets is that the \textit{corrupt:input-corrupted} margins for CIFAR10 models are comparable to the \textit{corrupt:label-corrupted} margins. This is in contrast to the MNIST results where the \textit{corrupt:input-corrupted} margins are larger than the  \textit{corrupt:label-corrupted} margins. Furthermore, it is also clear that in the case of MNIST, the \textit{clean:input-corrupted} margins are smaller than the \textit{clean:clean} margins at smaller capacities. However, this difference disappears at larger model capacities. For CIFAR10, there is a much smaller difference between \textit{clean:input-corrupted} and \textit{clean:clean} margins. Finally, it is also interesting that the MNIST margins, compared to those of CIFAR10, appear to show asymptotic behavior (the margin sizes appear to `level off').


\subsection{Margin distributions}
\label{sec:noise_margin_distributions}

\begin{figure}[H]
    \centering
    \includegraphics[width=0.49\linewidth]{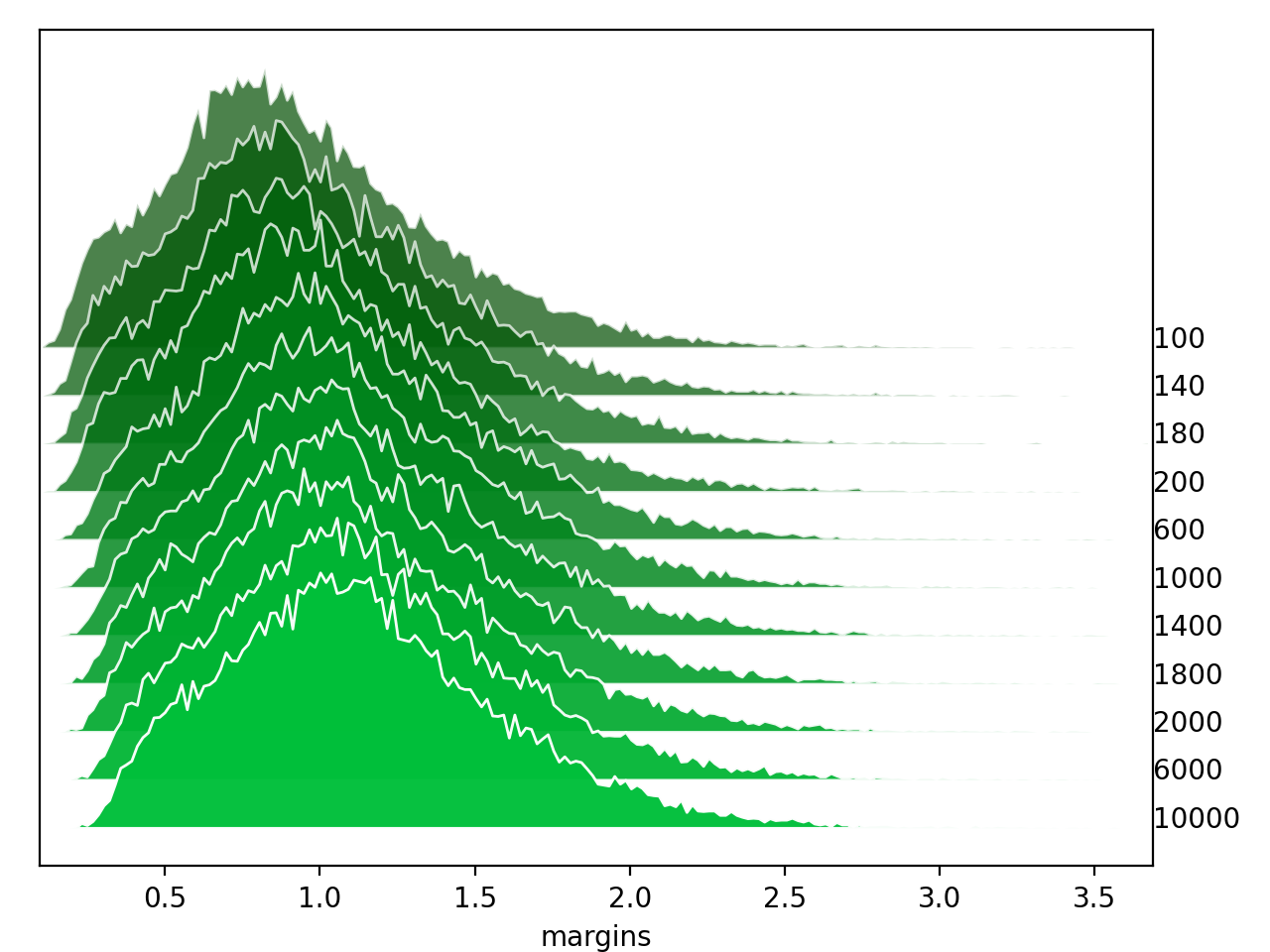}
    \includegraphics[width=0.49\linewidth]{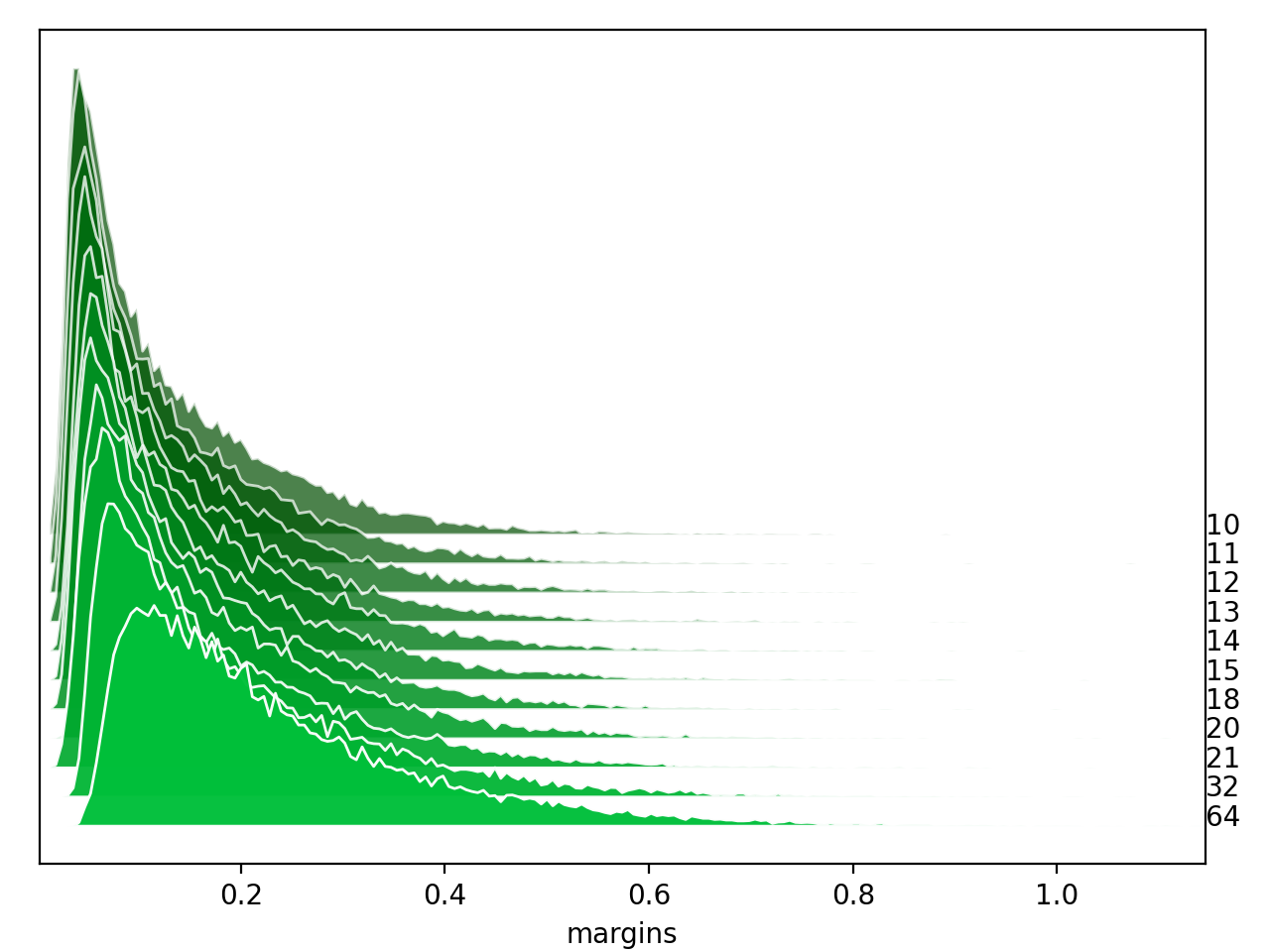}
    \includegraphics[width=0.49\linewidth]{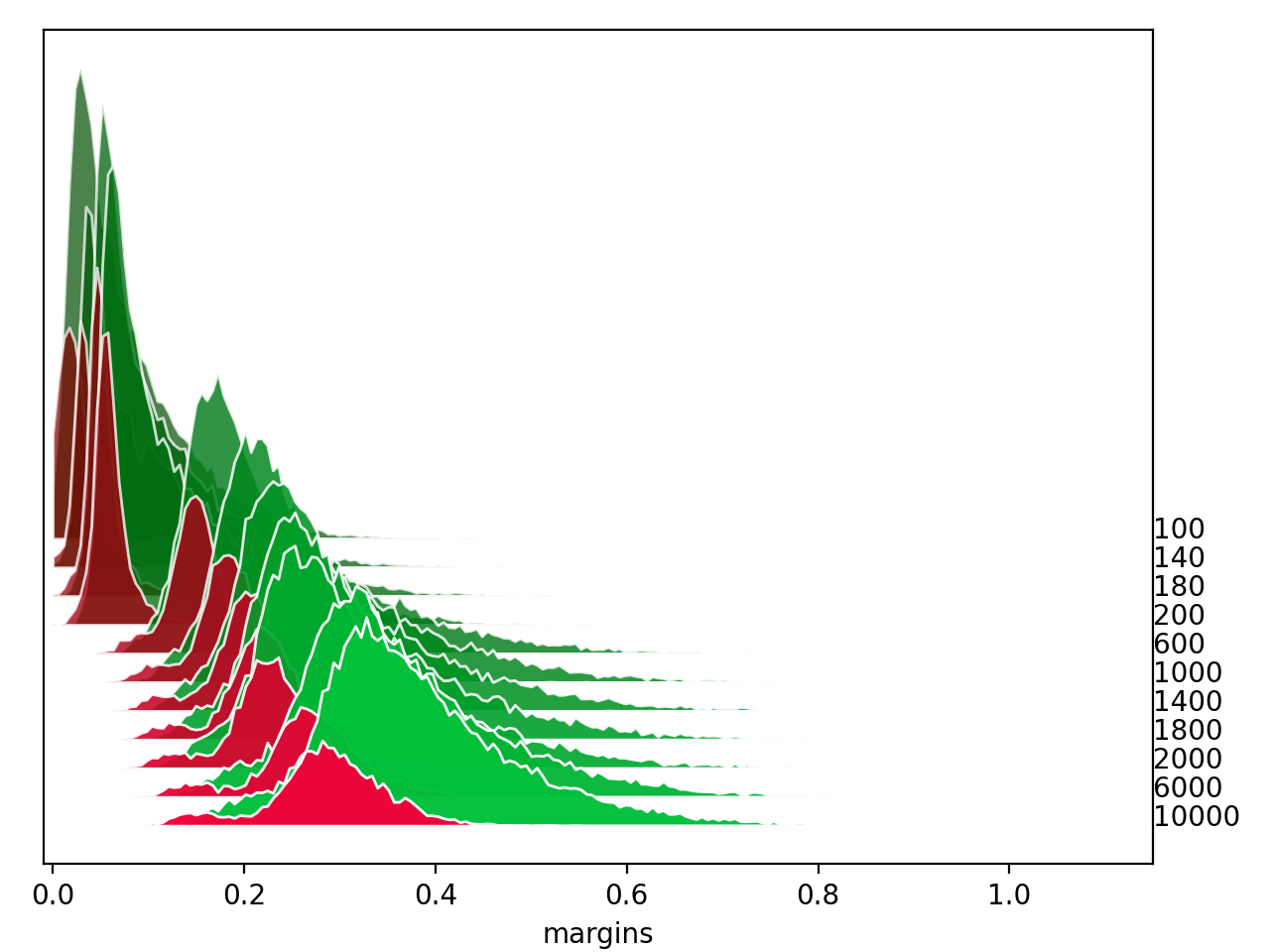}
    \includegraphics[width=0.49\linewidth]{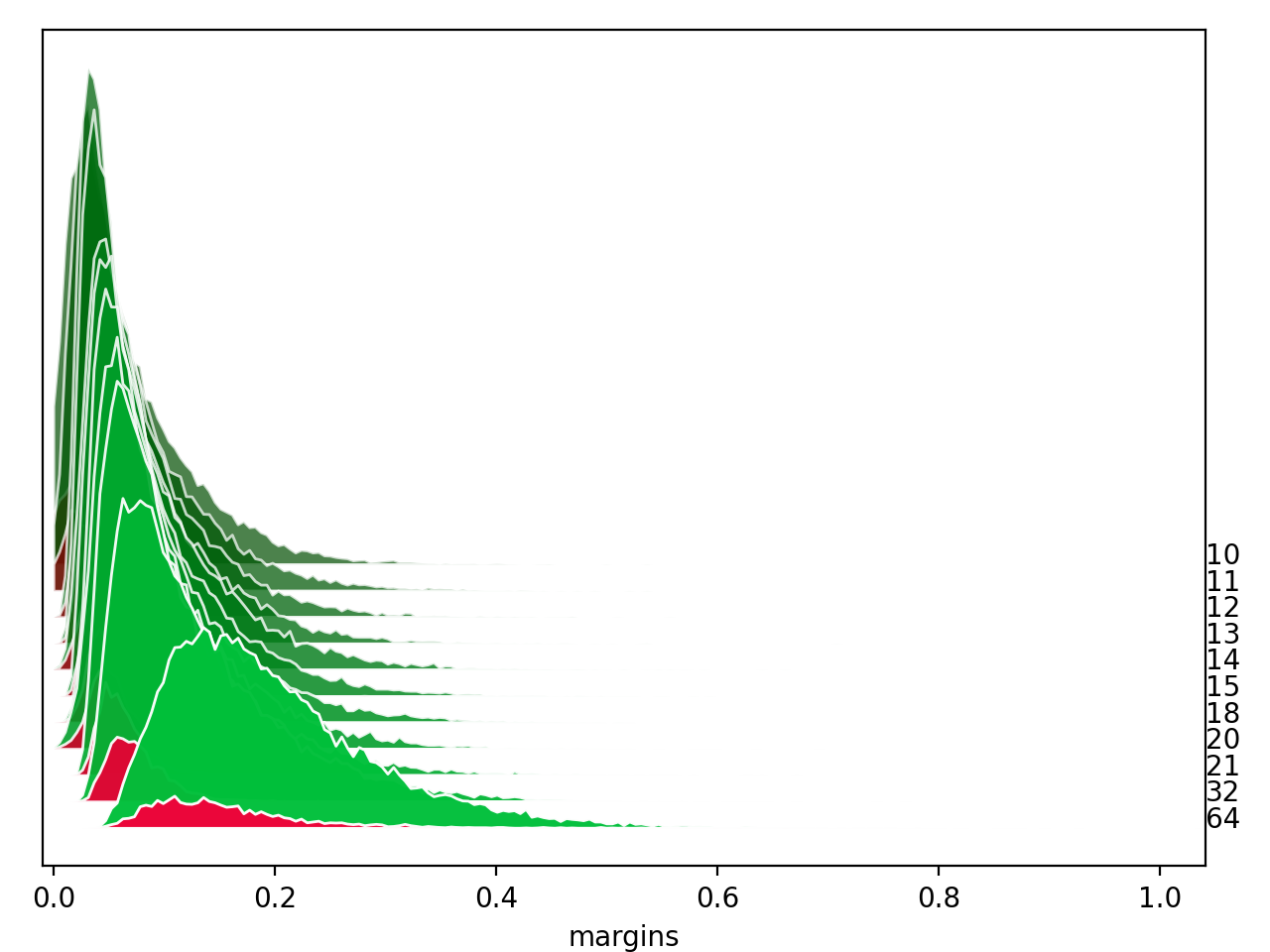}
    \includegraphics[width=0.49\linewidth]{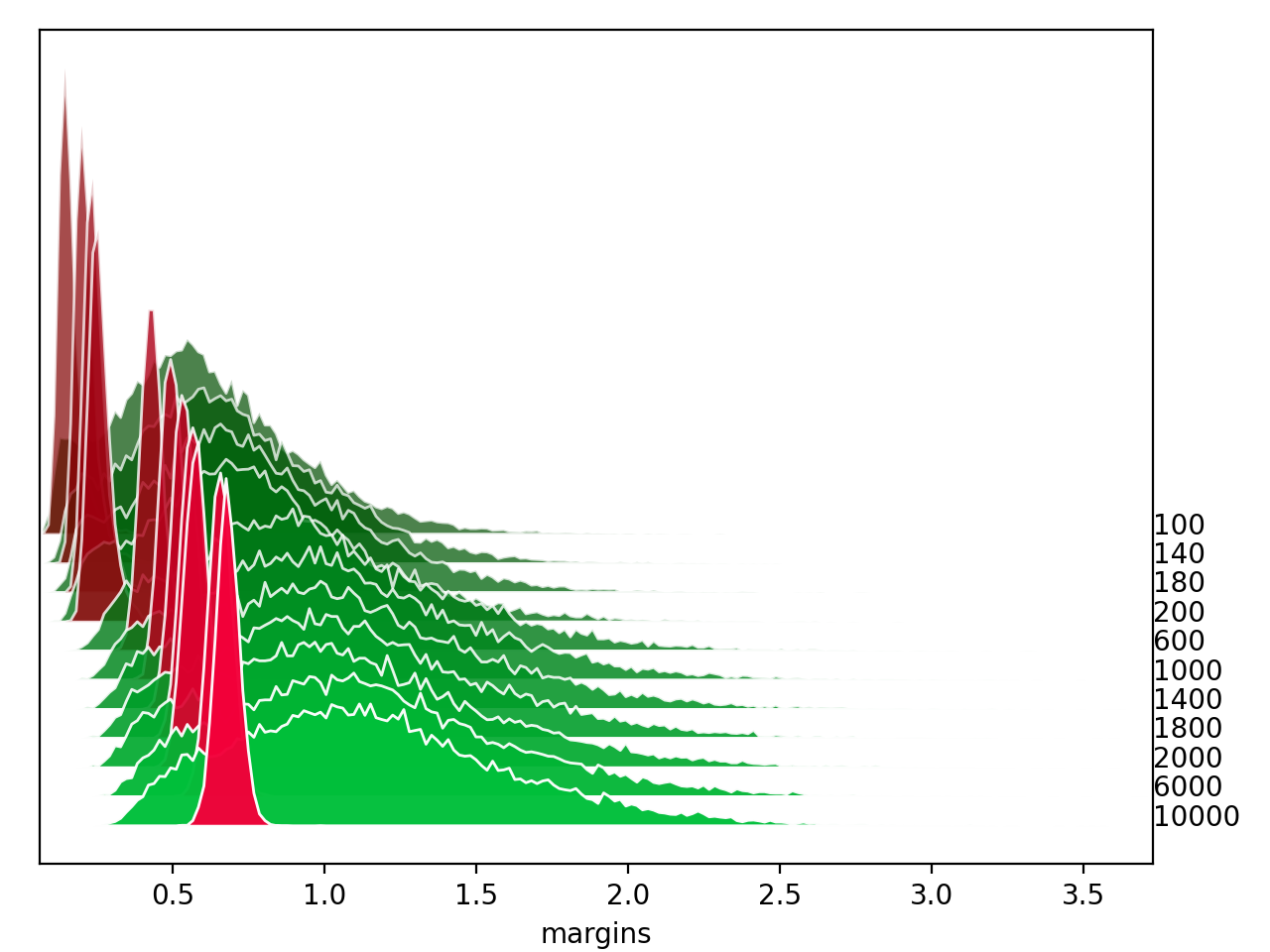}
    \includegraphics[width=0.49\linewidth]{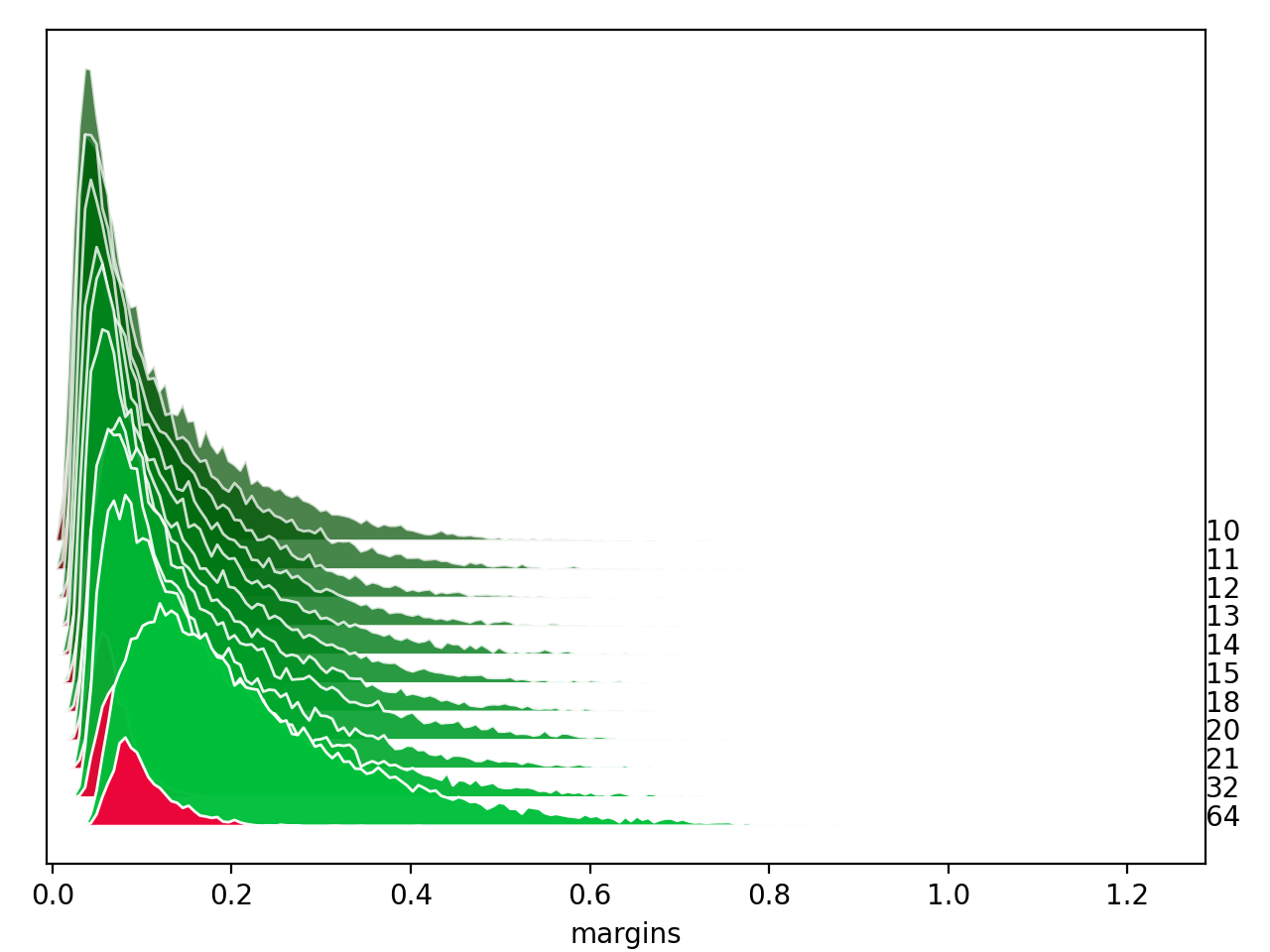}
    \caption[Margin distributions for all MNIST and CIFAR10 models]{Margin distributions for MNIST models (left) and CIFAR10 models (right) trained on clean (top), label-corrupted (middle) and input-corrupted (bottom) training sets. Within each plot, from top to bottom, the distributions are ordered by ascending model size. The relevant capacity metric is shown on the right. Green and red distributions are constructed from clean and corrupted samples, respectively. The corrupted sample distributions are also visualized separately in Figure~\ref{fig:noise_margin_distributions_noise_only}}
    \label{fig:noise_margin_distributions}
\end{figure}

Next, we visualize the distributions of margins underlying the means in Figure~\ref{fig:noise_margin_means_all_models}. Histograms are constructed of the margins measured at each capacity. These histograms share a common set of bins on the horizontal axes. These are shown in Figure~\ref{fig:noise_margin_distributions}.

One observes that most of these distributions are right-skewed distributions with a long tail containing relatively large margins, with the CIFAR10 distributions (right) more right-skewed than the MNIST (left) ones. Moreover, we observe that, in general, the distributions seem to be less peaked for models with larger capacity, i.e. the distributions appear to `flatten out' as capacity increases.

\begin{figure}[H]
    \centering
    \includegraphics[width=0.49\linewidth]{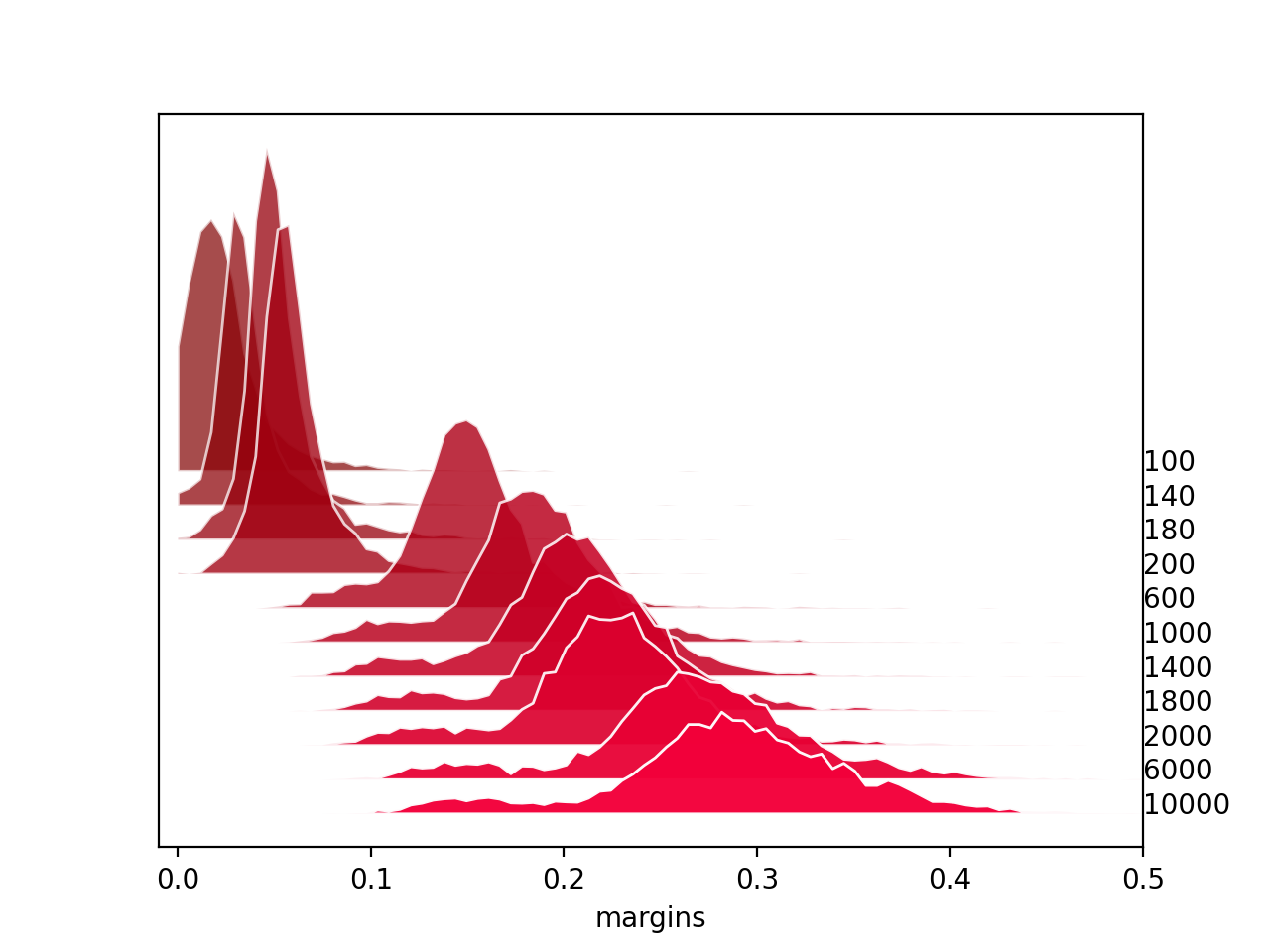}
    \includegraphics[width=0.49\linewidth]{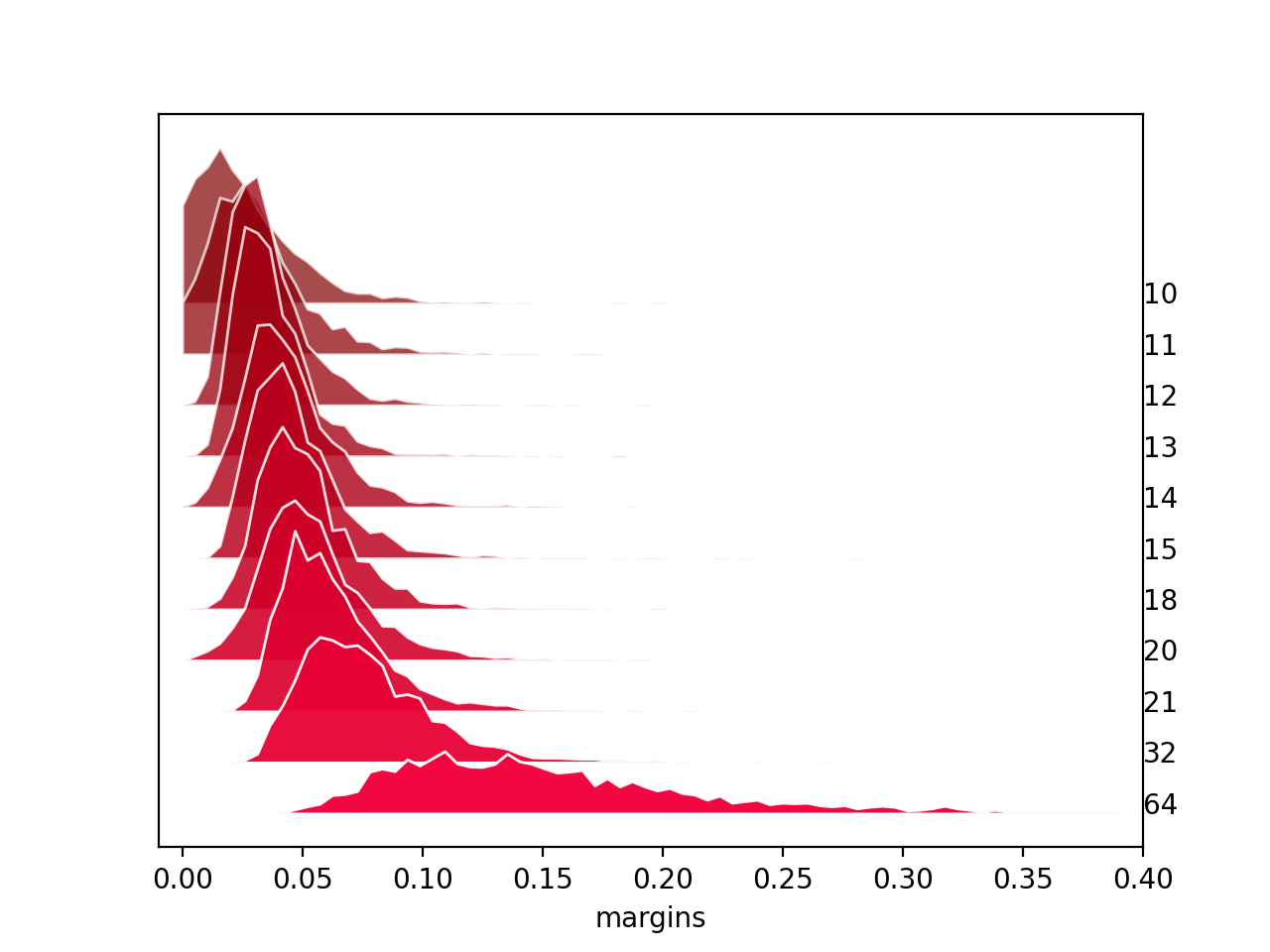}
    \includegraphics[width=0.49\linewidth]{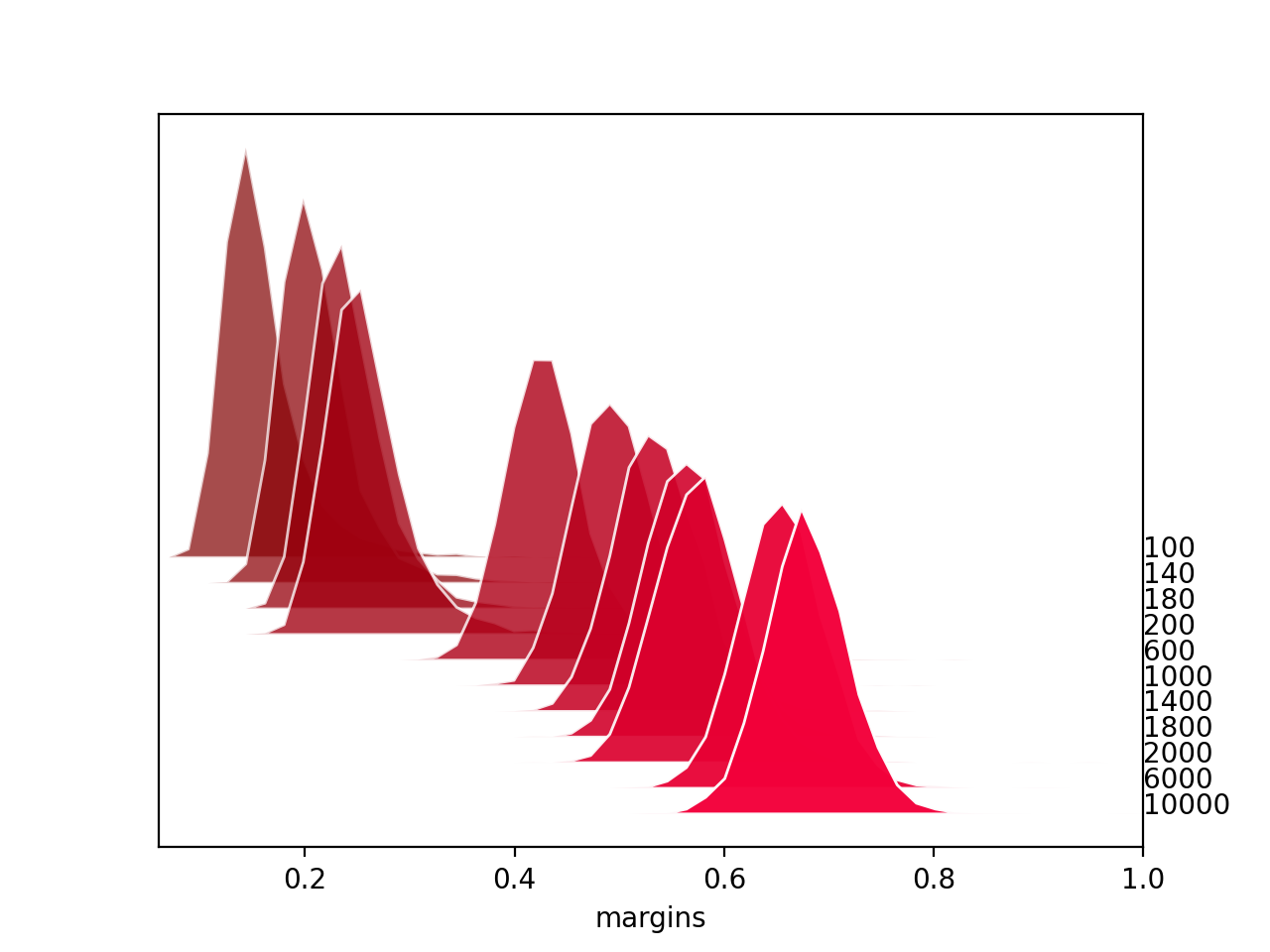}
    \includegraphics[width=0.49\linewidth]{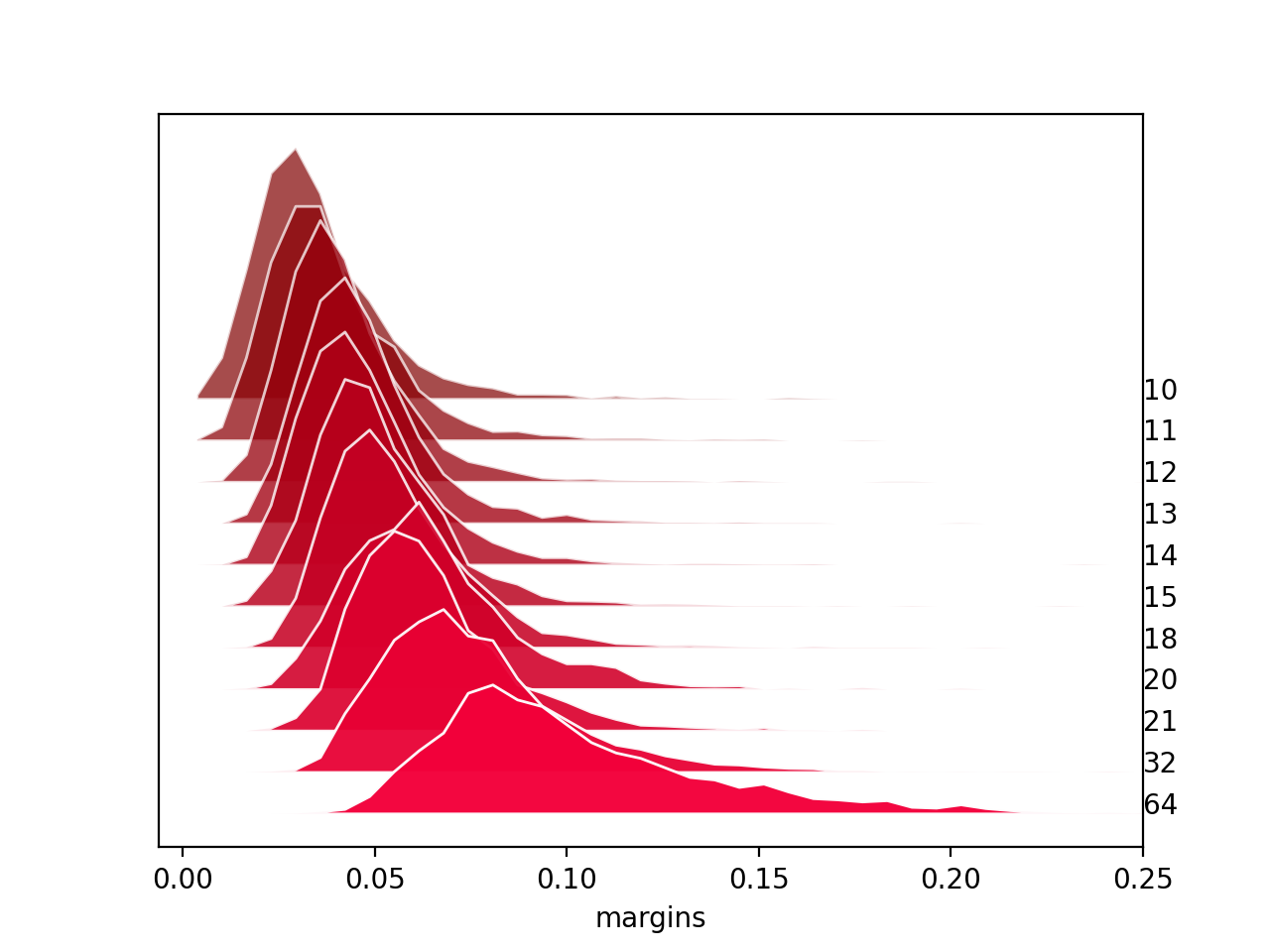}
    \caption[Corrupt sample margin distributions for  MNIST and CIFAR10 models]{Corrupt sample margin distributions for MNIST models (left) and CIFAR10 models (right) trained on label-corrupted (top) and input-corrupted (bottom) training sets. Within each plot, from top to bottom, distributions are ordered by ascending model size. The relevant capacity metric is shown on the right.}
    \label{fig:noise_margin_distributions_noise_only}
\end{figure}

Furthermore, we see that the \textit{corrupt:input-corrupted} margin distributions of the \textbf{MNISTgic} set (left-bottom) are normally distributed with relatively low variance, compared to the other distributions. This suggests that all models are constructing similar decision boundaries around \textit{corrupt:input-corrupted} samples, meaning there is not much diversity in how far these samples tend to be from their nearest point on the decision boundary.
The \textit{corrupt:label-corrupted} margins for the \textbf{MNISTlc} set (left-middle), on the other hand, show much higher diversity. The shape of the distribution changes drastically as capacity increases. At the critically small capacities, we see a distribution similar to the \textit{corrupt:input-corrupted} margin distributions. At larger capacities some \textit{corrupt:label-corrupted} samples have very small margins compared to the median of the distribution. See Figure~\ref{fig:noise_margin_distributions_noise_only}, where we only show the corrupt distributions for a clearer illustration. This phenomenon is less clearly observable in the CIFAR10 models, as we see an overall stronger tendency for the distributions to be right-skewed.

\subsection{Summary}
\label{sec:noise_results_summary}

We have made the following key observations in Sections \ref{sec:noise_mean_margin} and \ref{sec:noise_margin_distributions}:

\begin{enumerate}
    \item All margin values tend to increase with an increase in model capacity and corresponding generalization performance.
    \item The margins of corrupted samples are, on average, smaller than those of their clean counterparts within the same model. This is true for both on-manifold samples (of which label-corrupted samples are extreme examples) and off-manifold samples (of which the input-corrupted samples are extreme examples).
    \item On-manifold corrupt samples lead to a reduction in the margins of clean samples, while off-manifold corrupt samples only do so at smaller capacities and to a much lesser extent.
    \item In general, the margin distributions are not normally distributed, but right-skewed distributions with a long tail. However, there are also some variations in behavior between different types of samples and datasets in their degree of skewness, as well as sharpness of the distribution's peak.
\end{enumerate}

In the following section, we shall attempt to answer `why?'.

\section{Analysis}
\label{sec:noise_analysis}

Given the observations of the previous section, we perform further analysis to investigate the following questions:
\begin{itemize}
    \item Why are the \textit{overall:label-corrupted} margins so small compared to the margins of other samples?
    \item Why are the \textit{corrupt:input-corrupted} margins smaller than the \textit{clean:input-corrupted} margins?
    \item Why are \textit{clean:input-corrupted} margins smaller than \textit{clean:clean} margins?
\end{itemize}

\subsection{Why are label-corrupted margins so small?}
\label{sec:noise_analysis_label_corruption}

In this section, we investigate the first question: why are the \textit{overall:label-corrupted} margins so small compared to the margins of other samples?
We note that label corruption is expected to result in many samples that have different targets while being close to each other in the input space. One can think of a sample's minimum distance to another sample with a different target as its absolute maximum possible margin since a boundary needs to be drawn between them, assuming that both have been correctly classified during training. We refer to this upper bound on each sample's margin as its `max margin'.

To investigate how label corruption affects the max margin of each sample, we randomly select $10\ 000$ training samples from the data that the models in the \textbf{MNISTlc} and \textbf{CIFAR10lc} sets are trained on. For each of these samples, the max margin is measured as the minimum Euclidean distance between the sample $(\mathbf{x}_{1}, c_{1})$ and its nearest neighbor $(\mathbf{x}_{2}, c_{2})$ (selected from the entire training set). Formally, the max margin for each sample is given by:
\begin{equation}
    \min_{\mathbf{x}_{2}} ||\mathbf{x}_{1} - \mathbf{x}_{2}||_{2},~~ c_{1} \neq c_{2}
\end{equation}

\begin{figure}[ht]
    \centering
    \includegraphics[width=0.49\linewidth]{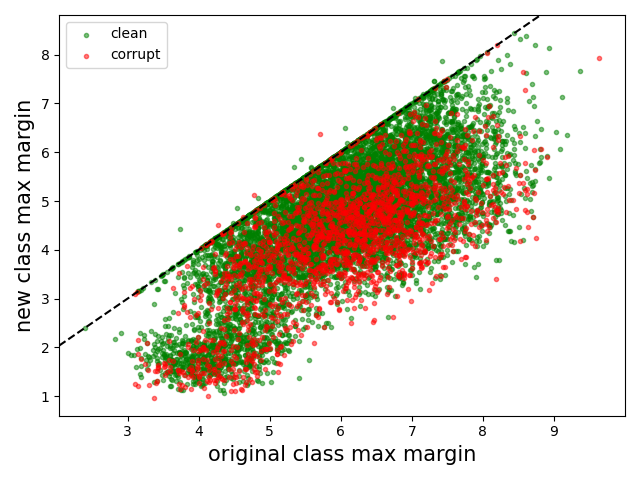}
    \includegraphics[width=0.49\linewidth]{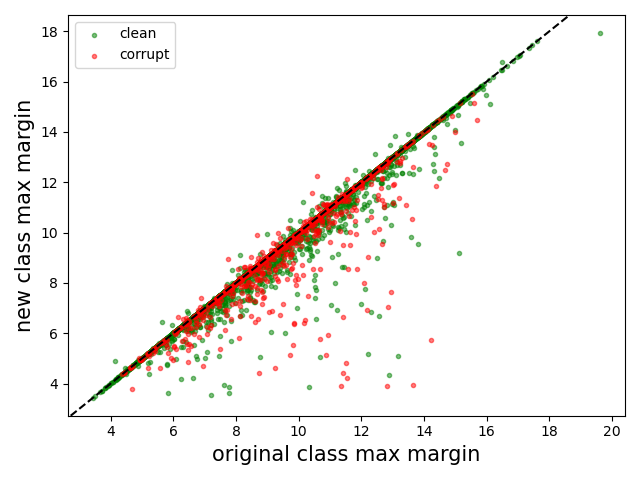}
    \caption[Maximum margins before and after label corruption for \textbf{MNISTlc} and \textbf{CIFAR10lc}]{Maximum margins before vs. after label corruption for \textbf{MNISTlc} (left) and \textbf{CIFAR10lc} (right). Green points represent clean samples and red points represent corrupt samples. The dashed line indicates $y = x$.}
    \label{fig:noise_max_margins_lc}
\end{figure}

This is done before and after the data is label corrupted. Then a scatter plot of these $10\ 000$ training samples is constructed with the distance measured with the original targets on the horizontal axis and the potentially corrupted targets on the vertical axis. The resulting scatter plots are shown in Figure~\ref{fig:noise_max_margins_lc}.
All samples below the provided identity function line had their max margin reduced due to label corruption. Note that, as expected, for MNIST the presence of label corruption causes many samples, corrupted and clean, to have drastically reduced upper bounds to their margins. In the case of CIFAR10 this is less clear, where the majority of samples only have slight max margin reductions. However, we note that the CIFAR10 models have less label corruption ($20\%$ for MNIST vs $10\%$ for CIFAR10).

\begin{figure}[h]
    \centering
    \includegraphics[width=0.49\linewidth]{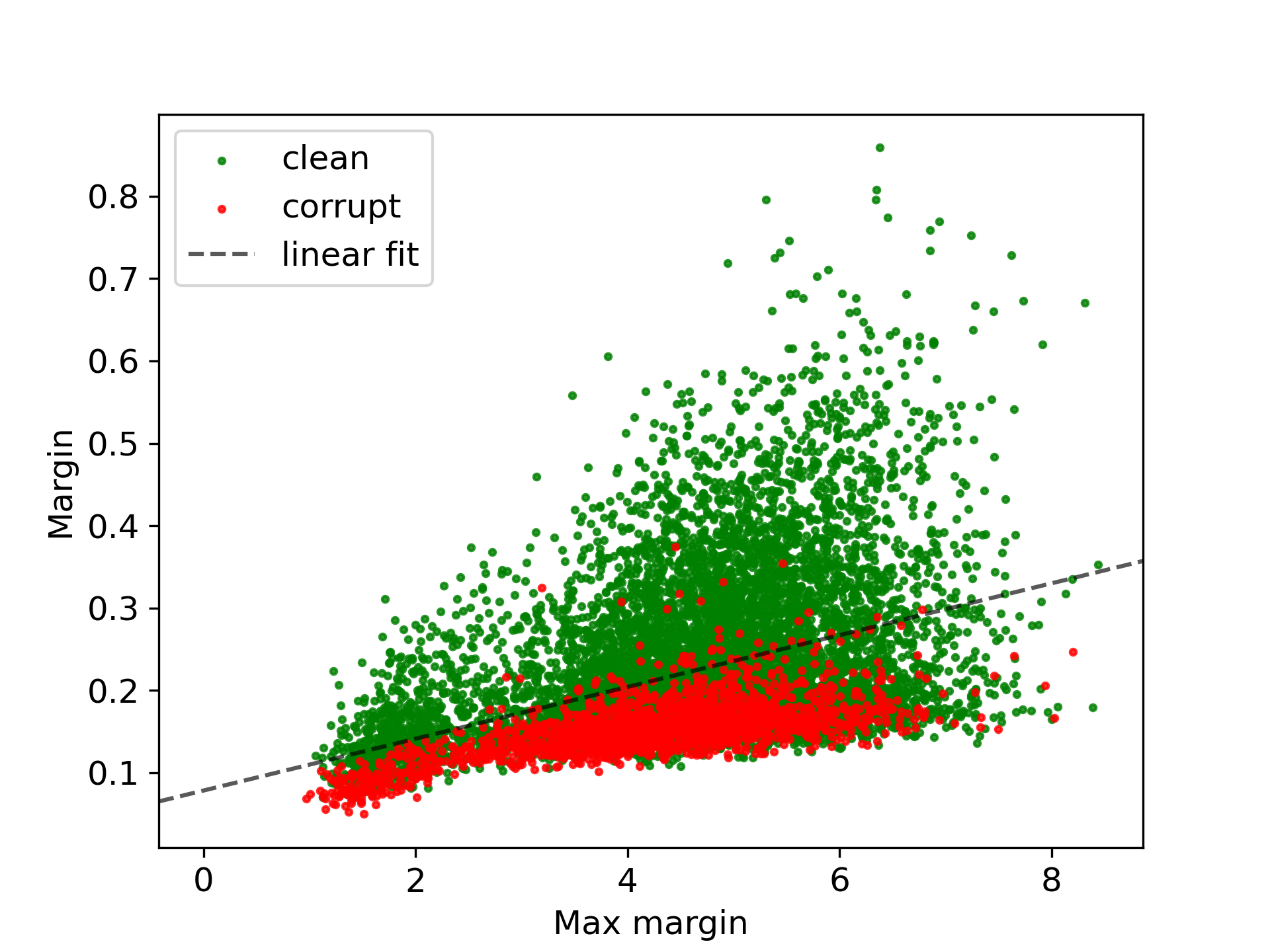}
    \includegraphics[width=0.49\linewidth]{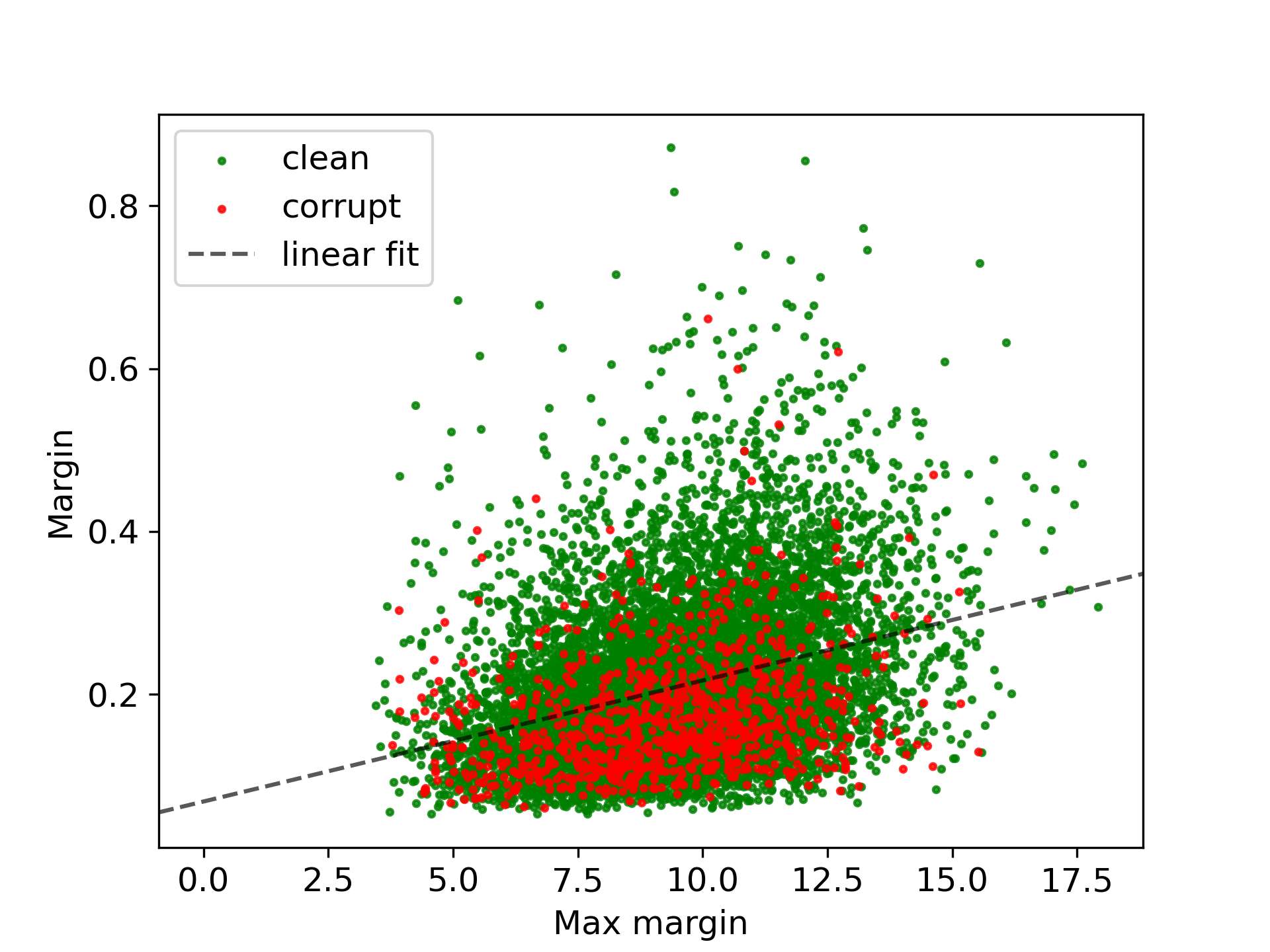}
    \caption[Maximum margin vs. measured margin for a \textbf{MNISTlc} and \textbf{CIFAR10lc} model]{Maximum margin vs. measured margin for $10\ 000$ samples for \textbf{MNISTlc} (left, width=$10\ 000$) and \textbf{CIFAR10lc} (right, k=$64$) models. Green points represent clean samples and red points represent corrupt samples. Best fit lines have $R^2$ values of $0.171$ (\textbf{MNISTlc}) and $0.106$ (\textbf{CIFAR10lc}).}
    \label{fig:noise_margin_versus_max_margin_lc}
\end{figure}

If the proximity of each sample to samples of another class within the input space is the main factor that leads to a reduction in margin, we would expect to see a clear relationship between the max margin and the measured margin. In Figure~\ref{fig:noise_margin_versus_max_margin_lc} we show a scatter plot of the measured margin of each sample versus its max margin, after label corruption. We do this for all $10\ 000$ samples for which margins have been calculated for both the largest \textbf{MNISTlc} (width=$10\ 000$) and \textbf{CIFAR10lc} (k=$64$) models and also show a best fit regression line.



Firstly, we observe that there is a weak linear relationship between the margin of each sample and its corresponding max margin. 
Notice that this relationship need not necessarily be present -- in both cases the margin is much smaller than the max margin (i.e. its upper bound) would suggest. This is an interesting observation: from an overly simplistic perspective, one would consider the `ideal margin' of each sample to be roughly half that of its max margin, but this is far from the case.

We find $R^2$ (coefficient of determination, recall Section~\ref{sec:back_coefficient_of_determination}) values of $0.171$ and $0.106$ for the best fit lines for MNIST and CIFAR10, respectively. This suggests that, while the proximity of the nearest sample of another class has an effect on the margin of each sample, it does not fully describe why some samples have smaller margins than others.

It is also important to note that again there is a discrepancy between the MNIST and CIFAR10 results. For MNIST, one observes a gradual increase in the average margin as the max margin increases. It also appears that this relationship is the strongest for samples with very small max margins ($<2$). On the other hand, the relationship is weaker for the CIFAR10 model. We suspect that the underlying reason for this is that Euclidean distance in the input space is a more suitable metric for MNIST than CIFAR10 for capturing the notion of `visual similarity' between samples. Specifically, we believe that due to the simplicity of MNIST (centered, gray scale images with no background information) Euclidean distance is a better approximation of which samples are visually related to others than for CIFAR10.

To summarize, the max margin serves as an upper bound to the input margin (by definition); however, we find that the observed margins are much smaller than this upper bound. In addition, samples with smaller upper bounds tend to adhere better to this bound (at least proportionally). Finally, proximity alone paints an incomplete picture of the margin characteristics of samples, when measured with Euclidean distance. This is especially true for CNNs trained on CIFAR10.

We conclude that the margins of label-corrupted samples are likely small on average because many of these samples are in close proximity to samples of another class (small max margins). That said, we believe that there are also other model or sample characteristics which influence the relative size of each sample's margin that we have not identified.

\subsection{Why are input-corrupted margins smaller than their clean counterparts?}
\label{sec:noise_analysis_input_corruption_one}

\begin{figure}[h]
    \centering
    \includegraphics[width=0.49\linewidth]{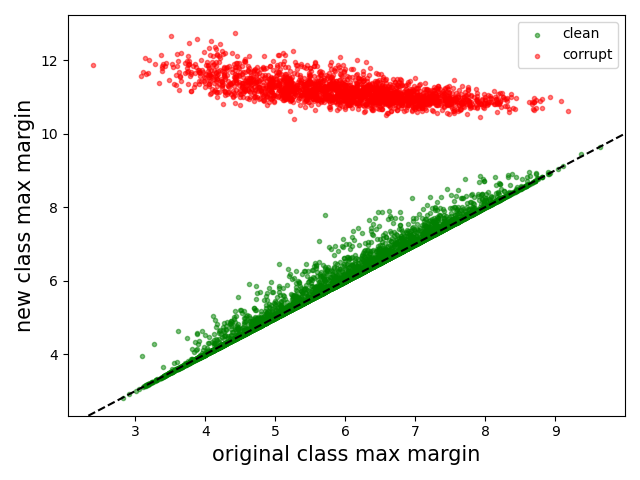}
    \includegraphics[width=0.49\linewidth]{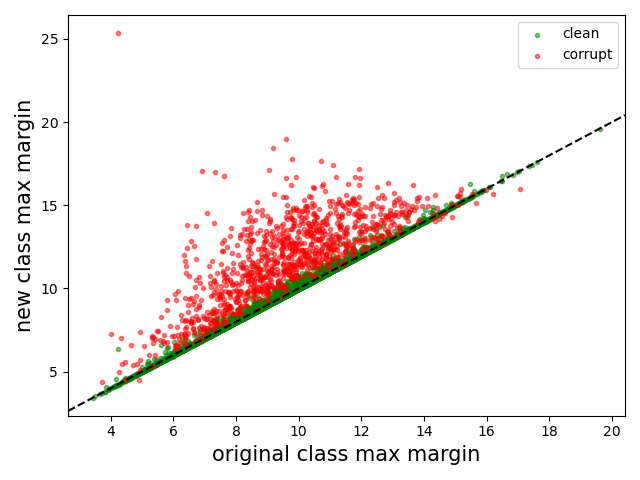}
    \caption[Maximum margins before and after input corruption for \textbf{MNISTgic} and \textbf{CIFAR10gic}]{Maximum margins before vs. after input corruption for \textbf{MNISTgic} (left) and \textbf{CIFAR10gic} (right). Green points represent clean samples and red points represent corrupt samples. The dashed line indicates $y = x$.}
    \label{fig:noise_max_margins_gaussian}
\end{figure}

In this section, we seek to investigate why \textit{corrupt:input-corrupted} margins are smaller than \textit{clean:input-corrupted} margins. To determine whether a similar phenomenon to the case of label-corrupted samples is affecting the \textit{corrupt:input-corrupted} margins, we generate a similar scatter plot to Figure~\ref{fig:noise_max_margins_lc} but for the \textbf{MNISTgic} and \textbf{CIFAR10gic} sets. This is shown in Figure~\ref{fig:noise_max_margins_gaussian}.

In contrast to the \textit{overall:label-corrupted} samples we see that the majority of the \textit{overall:input-corrupted} samples have increased max margins. We also observe that the \textit{corrupt:input-corrupted} samples have the largest max margins in comparison to the \textit{clean:input-corrupted} samples for both datasets. Furthermore, the \textit{corrupt:input-corrupted} samples, for the \textbf{MNISTgic} set, have extremely high maximum margins. In essence, we can conclude that input corruption \textit{increases} max margin, in contrast to label corruption which \textit{decreases} max margin in this setting.

Let us now consider how this results compares to that of the label-corruption case shown in the previous section. For the label-corrupted models, we argued that the close proximity of different-target samples (small max margin) reduces the average margin of \textit{overall:label-corrupted} samples. However, for the case of input corruption, we observe in Figure~\ref{fig:noise_max_margins_gaussian} that the \textit{corrupt:input-corrupted} samples are extremely distant, at least in the \textbf{MNISTgic} case, from any other sample (minimum max margin of $10$ in Figure~\ref{fig:noise_max_margins_gaussian}). On the other hand, the actual margins for these \textit{corrupt:input-corrupted} sample are very small (recall Figure~\ref{fig:noise_margin_means_all_models}).  

The two different findings between label-corruption and input-corruption present an inconsistency: if the close proximity to other samples of a different class are responsible for reduced margins, why are the \textit{corrupt:input-corrupted} margins smaller than \textit{clean:input-corrupted} samples, even though these samples are very distant from other samples? We speculate that the reason for this inconsistency is that the relatively small \textit{corrupt:input-corrupted} margins are a result of \textit{incentive}. These samples are so far off-manifold and far from each other that there is little reason to increase their respective margins beyond a certain model-specific maximum once the samples have been memorized by the network. The previously mentioned lack of variance in margins we observe for \textit{corrupt:input-corrupted} margins in Figure~\ref{fig:noise_margin_distributions} supports this notion.

These results also allow us to speculate on a second question: why are the CIFAR10 \textit{corrupt:input-corrupted} margins extremely small? Recall one of our previous observations: in Figure~\ref{fig:noise_margin_means_all_models} we observe that in the case of CIFAR10 CNNs, the mean \textit{corrupt:input-corrupted} margins are essentially as small as the \textit{corrupt:label-corrupted} mean margins. On the other hand, for the MNIST MLP models the mean \textit{corrupt:input-corrupted} margins are larger than the mean of both the \textit{clean/corrupt:label-corrupted} margins. Why?

We believe that the reason for this is that the \textit{corrupt:input-corrupted} samples in CIFAR10 are not as remote as their MNIST counterparts. Specifically, we believe that the \textit{corrupt:input-corrupted} samples in CIFAR10 are closer, in terms of Euclidean distance, to \textit{clean:input-corrupted} samples than in the case of MNIST. Notice that for the CIFAR10 samples in Figure~\ref{fig:noise_max_margins_gaussian} (right), we observe that the \textit{corrupt:input-corrupted} samples do not undergo such a large increase in max margin after input corruption. This speculation is consistent with another observation we made earlier: In Figure~\ref{fig:noise_performance}, right, we observe that the \textbf{CIFAR10gic} models achieve a noticeably worse generalization ability than the \textbf{CIFAR10} models. This is in contrast to the difference in generalization performance between \textbf{MNIST} and \textbf{MNISTgic} models (left of the same figure).


\subsection{Why are clean sample margins in input-corrupted models smaller than those in clean models?}

In the previous section we considered the discrepancy between the margins of \textit{corrupt:input-corrupted} and \textit{clean:input-corrupted} samples. Now, we turn our attention to the discrepancy between the margins of \textit{clean:input-corrupted} and \textit{clean:clean} samples.  We observe that \textit{clean:input-corrupted} samples have slightly smaller margins than \textit{clean:clean} samples. It is not obvious why this is the case, as the \textit{corrupt:input-corrupted} samples should not have a large effect on the margins of the \textit{clean:input-corrupted} samples, since these two sample types are far away from each other (recall Figure~\ref{fig:noise_max_margins_gaussian}).

We hypothesize that this is a result of the capacity it requires to fit \textit{corrupt:input-corrupted} samples. It is known that these samples are fitted later and require more capacity than clean samples~\cite{krueger2017deep,theunissen2020benign}, and margins tend to increase with capacity. It is then reasonable to conclude that the smaller margin size of \textit{clean:input-corrupted} samples is a result of the lack of available capacity. In essence, we believe the \textit{corrupt:input-corrupted} samples require some additional capacity to fit, which results in the \textit{clean:input-corrupted} samples having slightly reduced margins. This idea is rsupported by Figure~\ref{fig:noise_margin_means_all_models}, where we observe that the difference between average \textit{clean:clean} margins and average \textit{clean:input-corrupted} margins decreases with added capacity, and disappears completely when models become large enough. 

\subsection{Summary}

To summarize, our analysis suggest that there are two main mechanisms contributing to the observed differences between the margin behaviour of different samples:
\begin{enumerate}
    \item Samples that are very close to different-target samples (small max margin) will inevitably have smaller margins than those further away from different-target samples. This kind of reduced margin is indicative of poor generalization, because it is very likely to pertain to in-distribution and on-manifold regions of feature space that are difficult to model.
    \item Samples that are extremely remote, being very distant from any other sample, will also have smaller margins than less remote samples, due to a lack of incentive to increase them. This kind of reduced margin is not indicative of poor generalization because it is likely to relate to out-of-distribution and off-manifold regions of feature space.
\end{enumerate}

This is with the caveat that `on-manifold' and `off-manifold' are harder to distinguish in the case of CIFAR10. This is most evident in Figures \ref{fig:noise_max_margins_lc} and \ref{fig:noise_max_margins_gaussian}, where data corruption has a much less noticeable effect on the samples' proximity to different class samples.

\section{Extension to hidden margins}
\label{sec:noise_hidden_margins}

We now extend our analysis to hidden margins. Given that margins measured at the hidden representations are commonly used to predict the generalization of a model~\cite{predict_gen_margin,rep_based_complexity_pgdl,optimal_transport}, it is worth determining whether sample corruption has similar effects on the measurements.

Unfortunately, the large dimensionality of hidden representations makes it computationally infeasible to apply the same constrained optimization formulation (described in Section~\ref{sec:noise_margin_formulation}) to measure margins at these layers. Therefore, we opt to rather approximate these margins using the first-order Taylor approximation, described earlier in Section~\ref{sec:back_hidden_margins}, as done previously by several others~\cite{large_margin_dnns,predict_gen_margin,rep_based_complexity_pgdl,optimal_transport}. 

\begin{figure}[H]
    \centering
    \includegraphics[width=0.49\linewidth]{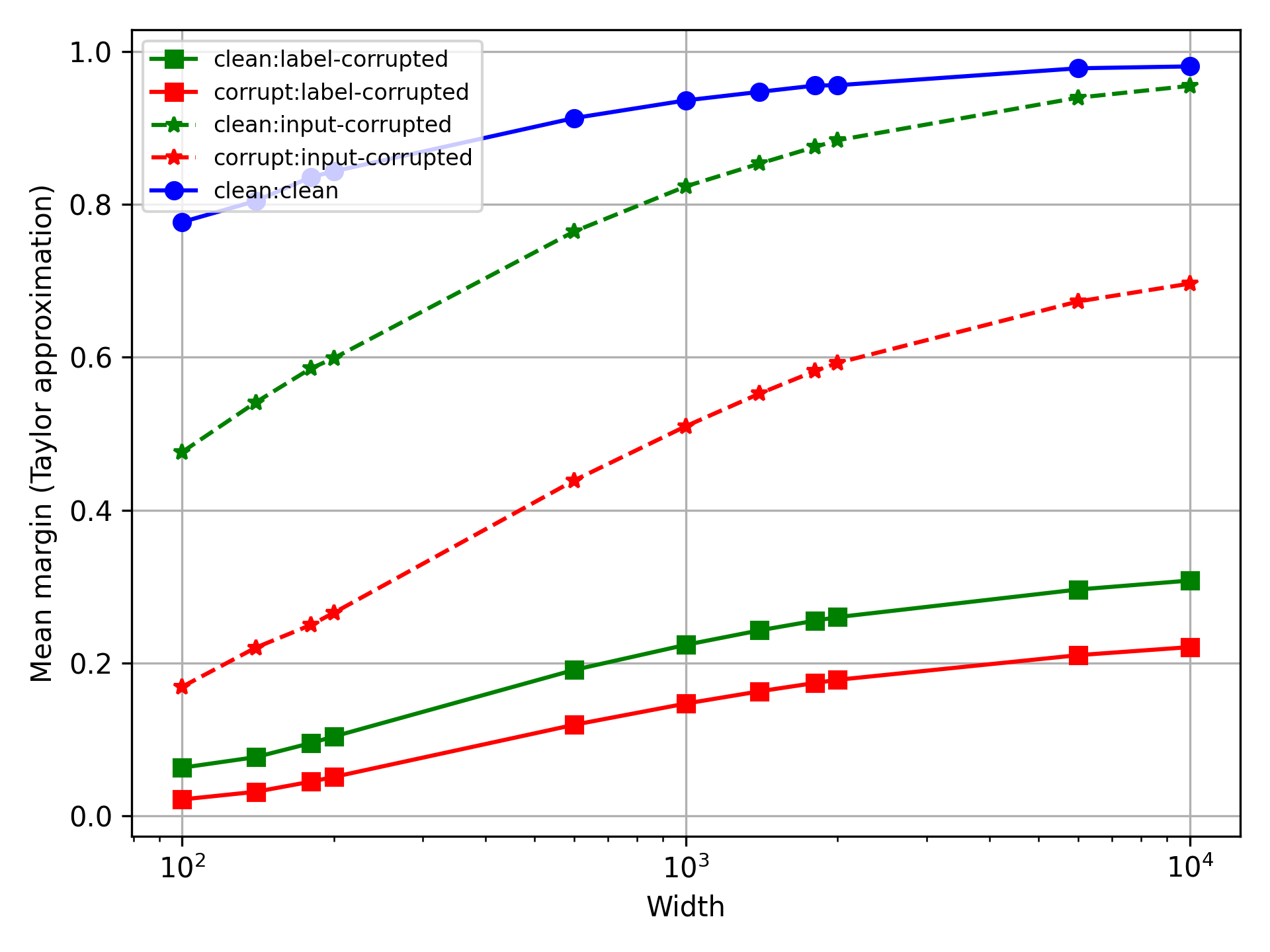}
    \includegraphics[width=0.49\linewidth]{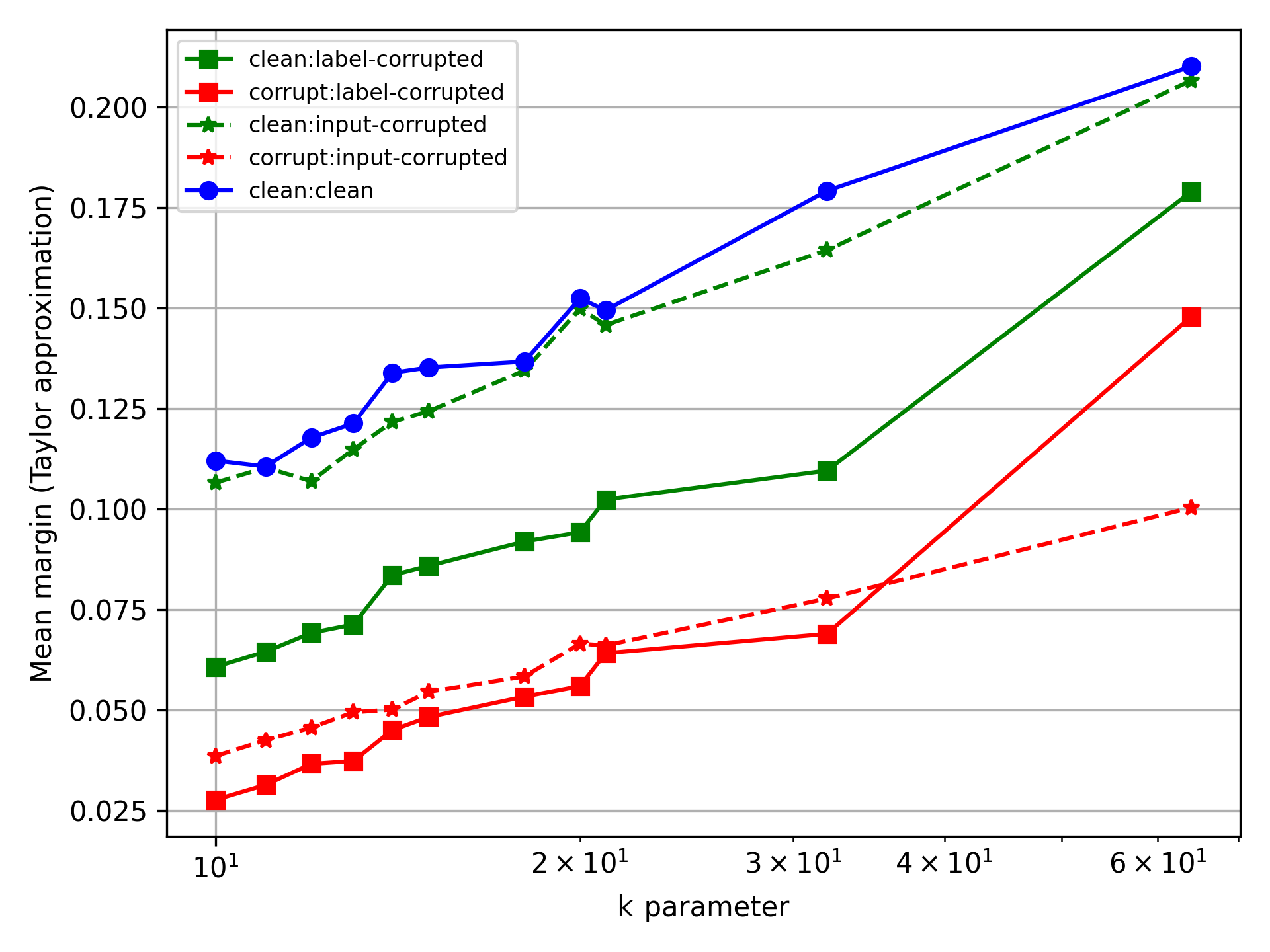}
    \caption[Mean first-order Taylor-approximated input margins for MNIST and CIFAR10 models]{Mean first-order Taylor-approximated input margins for MNIST models (left) and CIFAR10 models (right).}
    \label{fig:noise_mean_margin_taylor_input}
\end{figure}

As a first step, we measure the Taylor-approximated margins in the input space. This allows us to compare the results to those found using constrained optimization and to verify whether the Taylor approximation is suitable for this task.  Specifically, our experimental setup remains the same as described in Section~\ref{sec:noise_exp_setup}. The only difference is that we now use Equation~\ref{eq:taylor_approx_margin} to calculate the input space margin between each sample's target class and its second highest predicted class.  We show the mean Taylor-approximated input margin for all sets of models for both MNIST and CIFAR10 in Figure~\ref{fig:noise_mean_margin_taylor_input}.

The results are nearly identical to those found with constrained optimization in Figure~\ref{fig:noise_margin_means_all_models}, and the same trends are still present. For MNIST, we find that the approximation results in slightly smaller margins, while for CIFAR10 they are slightly larger. From this, one can conclude that the approximation is likely a suitable tool for this analysis. However, we confirm this through a more extensive investigation of the accuracy of the first-order Taylor approximation in the following chapter.

\begin{figure}[h]
    \centering
    \includegraphics[width=0.49\linewidth]{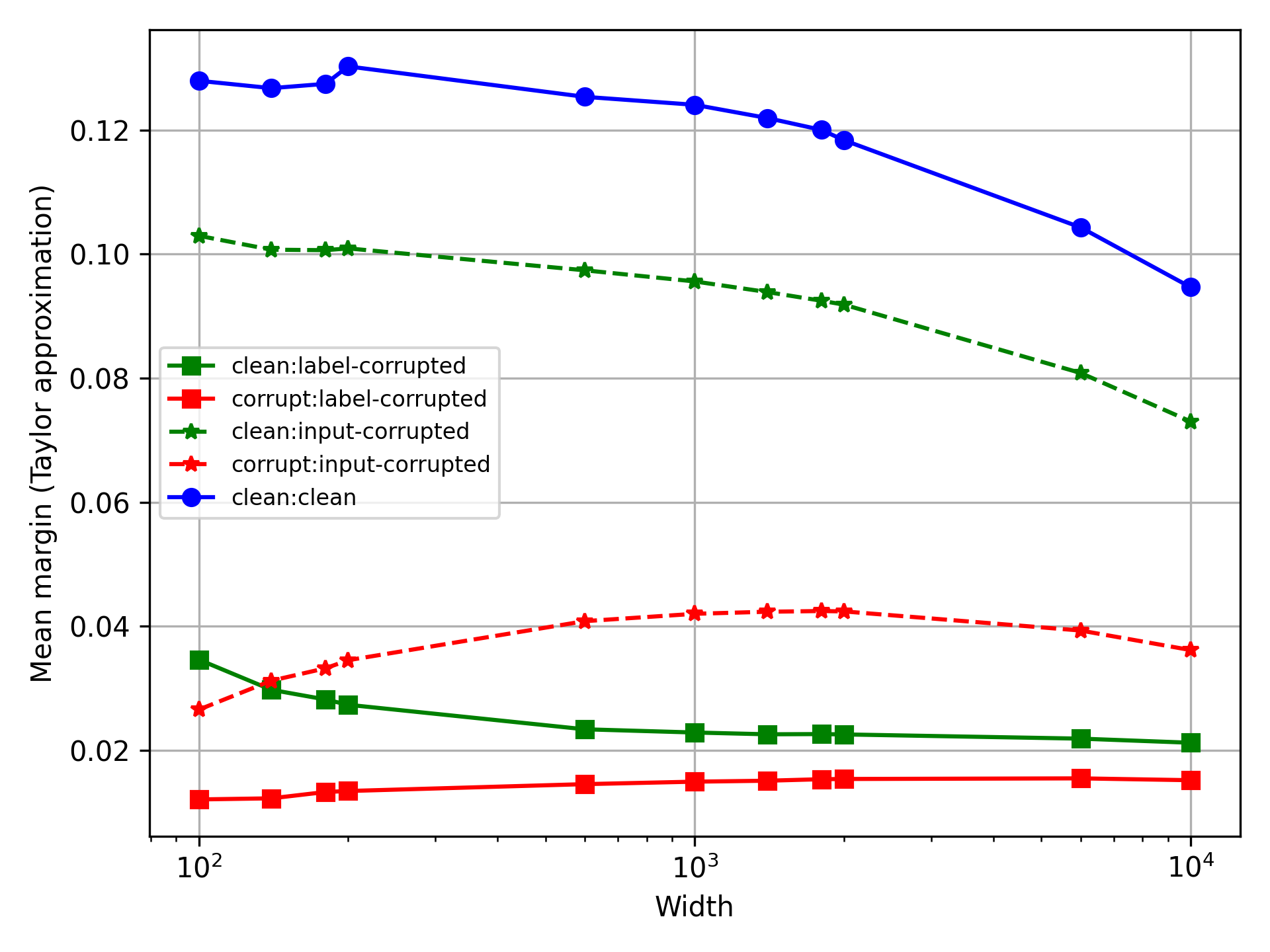}
    \includegraphics[width=0.49\linewidth]{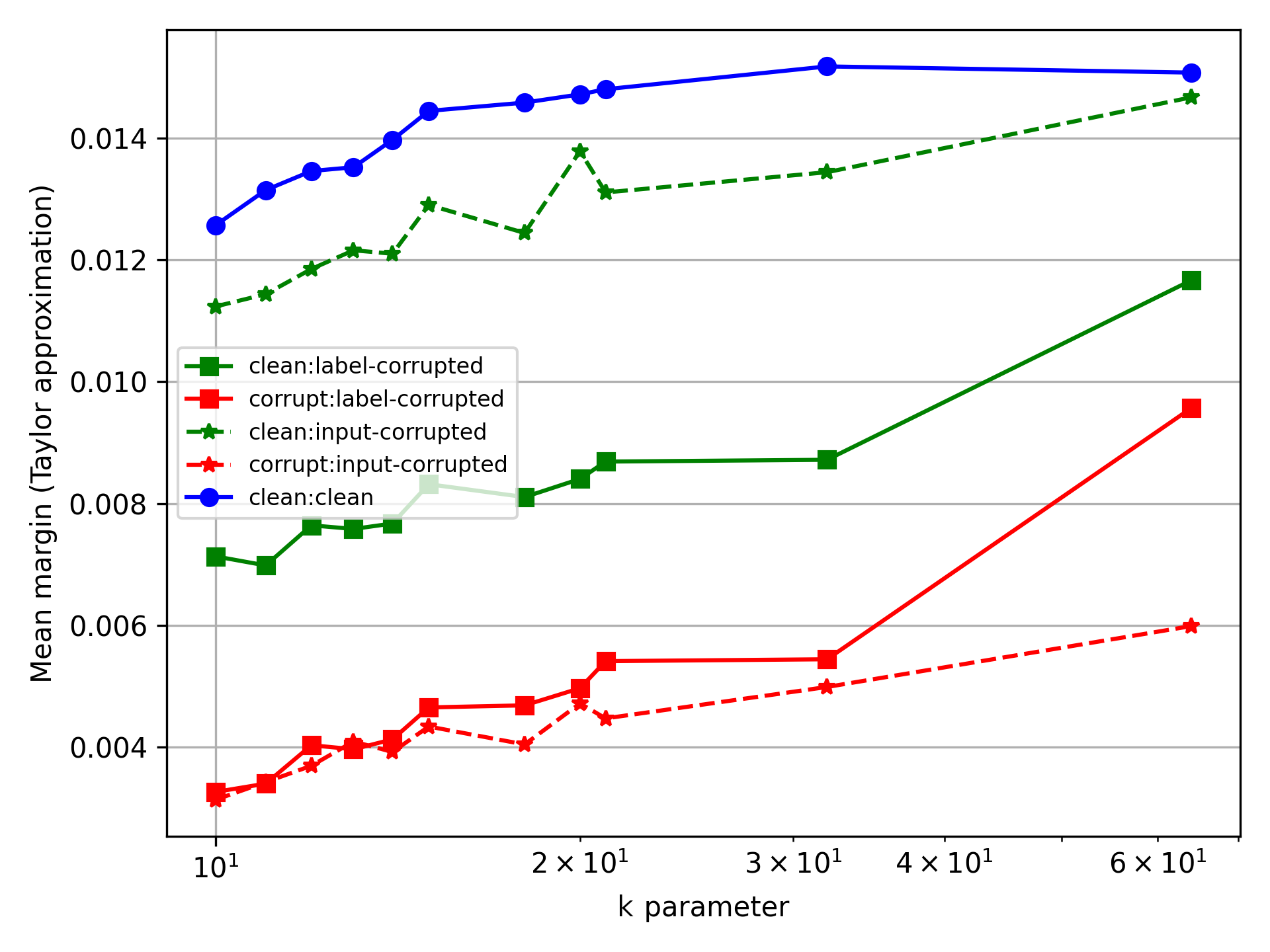}
    \caption[Mean first-order Taylor-approximated hidden margins for MNIST and CIFAR10 models]{Mean first-order Taylor-approximated hidden margins for MNIST models (left) and CIFAR10 models (right). MNIST models have only a single hidden layer, while the margin is averaged over all convolutional layers for CIFAR10. Margins are normalized using Total Variance.}
    \label{fig:noise_mean_margin_taylor_hidden}
\end{figure}

We now measure the margin at each hidden layer for each model. That is, we measure the Taylor-approximated margin at each hidden layer within each model for each sample. For the MNIST models, there is only a single hidden layer. For CIFAR10, we summarize our results following Natekar and Sharma~\cite{rep_based_complexity_pgdl} and average the margin over each convolutional layer. We also normalize the margin of each layer according to the total feature variance at that layer, which is the standard normalization, as explained earlier in Section~\ref{sec:back_hidden_margins}. The results of this are shown in Figure~\ref{fig:noise_mean_margin_taylor_hidden}. 

Firstly, we observe that in both cases the relative ordering of the mean margin for different sample types is mostly the same as for the input space, with a few exceptions. Specifically, we find a deviation in comparison to the input space for the smallest capacity MNIST models where the \textit{corrupt:input-corrupted} margins are slightly smaller than the \textit{clean:label-corrupted} samples. 
In the case of CIFAR10, we find a difference in that the \textit{corrupt:input-corrupted} margins are consistently smaller than the \textit{corrupt:label-corrupted} samples.

In terms of how the margin changes with capacity, we observe that the same trends hold as for the input space for CIFAR10, and the mean margin tends to increase along with an increase in model capacity. However, for MNIST we see the opposite trend -- the mean hidden margin \textit{decreases} as capacity increases.

\begin{figure}[h]
    \centering
    \includegraphics[width=0.65\linewidth]{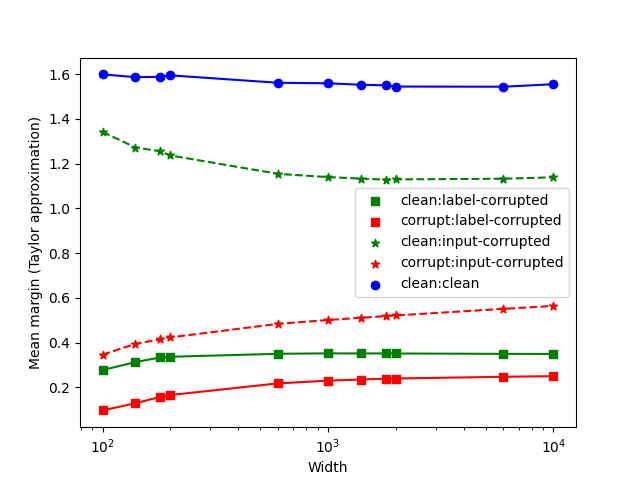}
    \caption[Mean first-order Taylor-approximated hidden margins for MNIST models with no normalization]{Mean first-order Taylor-approximated hidden margins for MNIST models with no normalization.}
    \label{fig:noise_mean_margin_taylor_hidden_mlp_no_norm}
\end{figure}

The reasons for this counter-intuitive observation of decreasing margin are not clear. We believe that this is due to the total variance normalization. In Figure~\ref{fig:noise_mean_margin_taylor_hidden_mlp_no_norm} we show the same results as in Figure~\ref{fig:noise_mean_margin_taylor_hidden} for the MNIST models, but with no normalization of the margin values. 

We find that the previously observed decrease in mean margin mostly disappears. Rather, we observe that the mean margins tend to remain mostly constant after a certain capacity. This suggests that hidden margins can be unreliable for certain analyses, as it is difficult to compare margins measured at spaces of different dimensionality. In the following chapter we discuss this in more detail.



\section{Discussion and conclusion}
\label{sec:noise_discussion}

In this chapter, we have investigated the relationship between margin measurements and sample noise and specifically focused on the implications on generalization. Let us first summarize the key observation we have made. We also indicate the section where the relevant results can be found:

\begin{itemize}
    \item For all models and sample types considered, mean input margins increase as model capacity grows (Section~\ref{sec:noise_mean_margin}).
    \item The margins of label-corrupted and input-corrupted samples are consistently smaller than the margins of their clean counterparts (Section~\ref{sec:noise_mean_margin}).
    \item The introduction of label corruption leads to a reduction in the margins of clean samples within the same model, whereas input corruption does so to a much lesser extent (Section~\ref{sec:noise_mean_margin}).
    \item Label corruption causes many samples (clean and corrupt) to have reduced upper bounds to their margins (max margin) for MNIST, while less so for CIFAR10 (Section~\ref{sec:noise_analysis_label_corruption}). Similarly, Gaussian input corruption causes an increase in the upper bound (Section~\ref{sec:noise_analysis_input_corruption_one}).
    \item Samples have much smaller margins than their upper bound would suggest (by an order of magnitude; Section~\ref{sec:noise_analysis_label_corruption}).
    \item There appears to be a (weak) relationship between each sample's upper bound and its margin in the case of clean and label-corrupted samples, although this need not necessarily be the case (see previous point; Section~\ref{sec:noise_analysis_label_corruption}).
    \item Total variance normalized hidden margins largely follow the same trends, with the exception of MNIST MLPs, where the margin decreases as capacity grows (Section~\ref{sec:noise_hidden_margins}).
\end{itemize}

There are several implications of these observations. First, these findings imply that mislabeled samples can significantly reduce the adversarial robustness of DNNs. Given that the presence of label-corrupted samples leads to a significant decrease in the margins of clean samples, we can conclude that these models are more vulnerable to adversarial perturbations. Seeing that many real-world datasets contain mislabeled samples~\cite{imagenet_noisy_labels,o2u_noisy_labels}, this is an important consideration. This suggests that a valuable first step in improving the robustness and performance of models would be to ensure a clean training set. Similarly, it appears that increasing a model's capacity also increases its adversarial robustness. However, it is not clear whether this would always be the case if different hyperparameters besides layer width are also varied between models (we have not investigated this here).

Second, given the differences between the margins of different sample types, we can ask whether the global average margin metrics, which are often used to predict or promote generalization, are sound.
Since margin sizes vary significantly between different sample types, we can conclude that an average margin will only work to compare two models if the training samples learned by the two models contain an approximately equal proportion of samples with these different margin behaviors. If a small set of samples are averaged over, this could become a problem. If the models to be compared have varying training set performance or were not trained on exactly the same training set, this could become an even more significant problem. It can also be argued that margin-based generalization predictors are more sensitive to off-manifold noise than to on-manifold noise. That is because the small \textit{corrupt:label-corrupted} margins rightly indicate the poor generalization of label-corrupted models. However, the small \textit{corrupt:input-corrupted} margins erroneously also indicate poor generalization.

In addition to providing an extensive comparison of the way in which different types of margin change with increased capacity, we explore some of the possible reasons for the observed behavior. Specifically, we find that the reduced margins accompanying label corruption (i.e. on-manifold corruption) are partly a result of the max margin (closest sample of a different class) being reduced by label corruption for a large portion of the training set. 
For input corruption (i.e. off-manifold corruption), we speculate that the reduced margins come from a lack of incentive to increase these margins, since the input-corrupted samples are remote from other training samples. That said, these hypotheses require further investigation. At present, we are not able to confidently answer how exactly label corruption leads to a reduction in margin, or why some samples have much smaller margins than others.

It is also necessary to touch on the discrepancies observed between several of the results for the MNIST and CIFAR10 models. This comparison is somewhat difficult, as we are not only comparing the margins of different datasets but also different architectures (MLPs vs. CNNs). We suspect that the difference in the underlying nature and difficulty of each dataset is the main contributor, and the input space of each is modeled in some distinct fashion that complicates direct comparison. In addition to this, it is likely that the distinct inductive bias of each architecture also plays a unique role in how different types of samples are modeled (and their subsequent margin behavior). However, our main findings remain consistent between these two tasks.

In conclusion, we investigated margins and sample noise in a controlled setting. In the following chapter, we investigate the use of margin measurements for generalization prediction in a more general case.

%% file: chapters/ch4_hidden_improv.tex
 
\let\cite\parencite

\lhead{}
\rhead{}
\chapter{Evaluating input and hidden margins on PGDL}
\label{chap:ch4} 
\afterpage{\lhead{\ifthenelse{\thepage>0}
       {\it Chapter \thechapter }
      }
\rhead{\ifthenelse{\thepage>0}
       {\it \let\uppercase\relax\leftmark}
      }}
\underline{ \hspace{\textwidth} } 
\textit{``The morning passed so quickly, it was time for them to meet \\
It was 20 past 11 when they walked out in the street \\
Folks were watching from the windows, everybody held their breath \\
They knew this handsome ranger was about to meet his death''} \\ - Marty Robbins, \textit{Big Iron}, Verse 6\\
\underline{ \hspace{\textwidth} }

\section{Introduction}
\label{sec:hidden_intro}

In this chapter we do a careful analysis of input and hidden margins on the PGDL tasks. 
In the previous chapter, we analyzed input margins in a very controlled setting, where the only variation between models was that of layer width. We observe that input margins are predictive of generalization in this setting. However, we now turn our attention to the more general case, where several hyperparameters and training dynamics are varied across a group of models. This allows us to determine how predictive input margins are of generalization in general, and where this metric fails. 

In terms of hidden margin, we also investigated these in a very controlled setting in Section~\ref{sec:noise_hidden_margins}. However, various others have conducted investigations in other settings. As explained in Section~\ref{sec:back_hidden_margins}, hidden margins were originally defined and used in the context of numerical generalization prediction. Specifically, summary statistics from first-order Taylor-approximated hidden margin distributions are used with a linear model. This linear model is fitted to some training set of models, and the quality of the prediction is then evaluated on some held-out set of models. While the original results reported by Jiang et al.~\cite{predict_gen_margin} (on the DEMOGEN~\cite{demogen_online_description} dataset, recall Section~\ref{sec:back_demogen}) were highly convincing, the PGDL tasks provide an additional test bed to independently assess the performance of hidden margins. Furthermore, we wish not only to analyze how well hidden margins perform when paired with a linear model, but also as a standalone complexity measure for generalization ranking. While several authors have explored this~\cite{rep_based_complexity_pgdl,optimal_transport}, none have done so in great depth and several questions remain: 
\begin{itemize}
    \item How does generalization prediction performance compare between different hidden layer selections?
    \item On which hyperparameter variations do hidden margins succeed/fail?
    \item Can hidden margins be improved if a more accurate estimation method is used?
\end{itemize}

The remainder of this chapter is structured as follows. We first turn our attention to the numerical prediction case, in Section~\ref{sec:hidden_lin_reg}.  In this section, we evaluate both input and hidden margins on the PGDL dataset paired with a linear regression model in the numerical prediction setting. Following this, in Section~\ref{sec:hidden_comp_measure}, we consider using these margins as a standalone complexity measure for generalization ranking and answer the previously posed questions. We perform several ablation analyses here. This includes comparing methods of selecting hidden layers and examining which hyperparameter variations result in failure cases. 
Following this, we turn our attention to the precision of the Taylor approximation used to measure these margins in Section~\ref{sec:hidden_beyond_approximations}. In this section, we establish how accurate the Taylor approximation is and whether the predictive power of these measurements can be improved by using more precise methods.

Note that, since input and hidden margins are established methods, we do not distinguish between the development set and test set of PGDL tasks throughout this chapter. Our goal here is not to compete with other complexity measures, but rather to analyze these methods in greater detail.

\section{Linear regression model}
\label{sec:hidden_lin_reg}

As a first step, we evaluate hidden margins on the PGDL dataset in the context of how they were originally used~\cite{predict_gen_margin}. Accordingly, the goal of this section is to fit a linear model on summary statistics of the Taylor-approximated hidden margin distributions. This model is fit using a group of models for which the true generalization gap is known, and is then evaluated on a held-out set of models. We do the same for input margins, which also serves as a baseline with which to compare hidden margins.
Our experimental setup closely follows that of Jiang et al.~\cite{predict_gen_margin}, and we highlight any divergence from their method. 

As an overview, we do the following:
\begin{enumerate}
    \item Calculate the Taylor-approximated margins at three equally spaced hidden layers and the input layers for all the models of a given PGDL task.
    \item From each margin distribution for each selected layer, we extract five summary statistics that characterize the distribution.
    \item For a subset of the models within the task, we fit a linear model using these summary statistics to predict the generalization of each model.
    \item We then evaluate how well this model predicts the generalization of a held-out set of models, using the coefficient of determination ($R^2$).
\end{enumerate}

Formally, given some  margin-based signature $\boldsymbol{\theta}_f$ for a model $f$, we predict the true generalization $y$ of $f$ using  
\begin{equation}
\hat{y} = \boldsymbol{\alpha}^T \phi(\boldsymbol{\theta}_f) + b  
\end{equation}
where $\boldsymbol{\alpha}$ is the coefficients of the linear model, and thus of the same dimensionality as the input $\boldsymbol{\theta}_f$, and $b$ a bias term. Here, $\phi()$ is some element-wise transform of the signatures, and the predicted generalization is given by $\hat{y}$. 

To obtain the margin signature, $\boldsymbol{\theta}$, we extract five summary statistics from the total variance normalized (recall Section~\ref{sec:back_hidden_margins}) margin distributions of some selected layers. Here, the margin is calculated using the previously mentioned first-order Taylor approximation (recall Sections \ref{sec:back_hidden_margins} and \ref{sec:noise_hidden_margins}) for all training samples. 
For each selected layer's distribution, we follow Jiang et al.~\cite{predict_gen_margin} precisely: we extract the first, second, and third quartile ($Q_1$, $Q_2$, and $Q_3$) values, and then also include the upper and lower fences of this distribution. These are found at $Q_3 + 1.5IQR$ and $Q1 - 1.5IQR$, where the interquartile range (IQR) is given by $IQR = Q_3 - Q_1$. We do this for the first, middle, and last convolutional layers of each model. Here `middle' refers to the layer that is halfway between the first and last, and is selected as the layer whose index is equal to the floor of half the total number of hidden layers. We also experiment with including both the input and hidden layers. These summary statistics of these layer's distribution are then concatenated, such that $\boldsymbol{\theta} \in \mathbb{R}^{d}$ provides a model-wide signature, where $d = 20$ or $d = 15$ depending on whether the input is included or not. Tasks $1$ and $8$ contain models with only two convolutional layers; for these, we drop the middle convolutional layer, reducing the dimensionality to $d=15$ and $d=10$ with and without the input layer, respectively. When only the input layer is used, $d=5$ in all cases. In terms of the transformation, $\phi$, we use a simple element-wise natural logarithmic transform, which was also shown in \cite{predict_gen_margin} to perform better than not using a transform.

Given that there is no predefined `training set' and `test set' of models for each task within the context of the PGDL challenge~\footnote{Recall that the PGDL train and test set refers to a split of the different tasks, not a split of models within each task.}, we must create our own by dividing the models within each task into two groups. For each task, we shuffle the model set and use k-fold cross-validation ($k = 3$) to produce three pairs of \textit{training} and \textit{test} model sets, each constituting a $66/33$ split of the model set. This is a departure from Jiang et al.~\cite{predict_gen_margin} where $k = 10$. Fewer folds are required for the smaller model sets of the PGDL tasks; otherwise, the test model set might consist of a very limited number of models. We repeat this for five random shuffles of the model set in order to perform a more comprehensive evaluation. This is again a departure from \cite{predict_gen_margin}, who only used a single shuffle.

Given these model sets and a signature for each model, we then fit a linear regression model on the training set of models and then evaluate the accuracy of its predictions on the test set of models. We follow Jiang et al.~\cite{predict_gen_margin} and use the coefficient of determination ($R^2$) (recall Section~\ref{sec:back_coefficient_of_determination}) for this evaluation.

In Table~\ref{tab:hidden_r_squared_all_signatures} we show the results for the eight tasks of the PGDL challenge. We show the mean coefficient of determination across all folds and shuffles. We also include the standard deviation of these $15$ coefficient of determination measurements. These results point to several interesting findings. First, it is clear that for most tasks, `Hidden only' margins perform much better than the input margin baseline, showing improved performance on six out of the eight tasks. It is further evident that including the input along with the hidden layers (`Hidden w/ input') does not make a large difference. In this case, only five tasks show better results than the input baseline and there is instability in the Task $9$ predictions. This is a clear indication that the predictive
power of these measurements stems from the margin distributions of the hidden layers. 
\begin{table} [H]
\centering
\caption[$R^2$ for linear regression models predicting generalization on the PGDL tasks]{Mean coefficient of determination for a linear regression model numerically predicting generalization on all PGDL tasks using five summary statistics from each selected layer's margin distribution. The coefficient of determination is calculated on a held-out set as the average of three folds $\times$ five random shuffles of the models in each task. $\pm$ indicates the standard deviation of the coefficient of determination across the three folds $\times$ five random shuffles. Number of train/test models indicates the number of models in the training set and test set per fold.}
\label{tab:hidden_r_squared_all_signatures}
\begin{tabular}{cccccc} 
\hline
\multirow{2}{*}{\textbf{Task}} & \multirow{2}{*}{\begin{tabular}[c]{@{}c@{}}\textbf{Number}\\\textbf{ of train}\\\textbf{ models}\end{tabular}} & \multirow{2}{*}{\begin{tabular}[c]{@{}c@{}}\textbf{\textbf{Number}}\\\textbf{\textbf{of test}}\\\textbf{\textbf{models}}\end{tabular}} & \multicolumn{3}{c}{\textbf{Coefficient of Determination} ($R^2$)} \\ 
\cline{4-6}
 &  &  & \begin{tabular}[c]{@{}c@{}}Input\\ only\end{tabular} & \begin{tabular}[c]{@{}c@{}}Hidden \\ w/ input\end{tabular} & \begin{tabular}[c]{@{}c@{}}Hidden \\ only\end{tabular} \\ 
\hline
1 & 64 & 32 & 0.602 $\pm$ 0.057 & 0.936 $\pm$ 0.017 & \textbf{ 0.939 $\pm$ 0.015} \\
2 & 36 & 18 & 0.862 $\pm$ 0.074 & 0.950 $\pm$ 0.023 & \textbf{ 0.957 $\pm$ 0.021} \\
4 & 64 & 32 & 0.777 $\pm$ 0.074 & 0.909 $\pm$ 0.037 & \textbf{ 0.931 $\pm$ 0.020} \\
5 & 42 & 22 & \textbf{0.350 $\pm$ 0.44} & -1.154 $\pm$ 3.392 & -0.846 $\pm$ 2.748 \\
6 & 64 & 32 & -0.034 $\pm$ 0.186 & \textbf{0.702 $\pm$ 0.079} & 0.687 $\pm$ 0.071 \\
7 & 32 & 16 & \textbf{0.061 $\pm$ 0.201} & -0.385 $\pm$ 1.025 & -0.270 $\pm$ 0.704 \\
8 & 42 & 22 & -0.129 $\pm$ 0.216 & 0.143 $\pm$ 0.381 & \textbf{0.219 $\pm$ 0.266} \\
9 & 21 & 11 & -0.272 $\pm$ 0.966 & -2214.825 $\pm$ 7954.918 & \textbf{0.238 $\pm$ 0.501} \\
\hline
\end{tabular}
\end{table}
\begin{table}[H]
\centering
\caption[Same results as in Table~\ref{tab:hidden_r_squared_all_signatures} when using only the last two summary statistics of each margin distribution rather than all five]{Same results as in Table~\ref{tab:hidden_r_squared_all_signatures} when using only the last two summary statistics of each margin distribution rather than all five.}
\label{tab:hidden_r_squared_two_signatures}
\begin{tabular}{cccccc} 
\hline
\multirow{2}{*}{\textbf{Task}} & \multirow{2}{*}{\begin{tabular}[c]{@{}c@{}}\textbf{Number}\\\textbf{ of train}\\\textbf{ models}\end{tabular}} & \multirow{2}{*}{\begin{tabular}[c]{@{}c@{}}\textbf{\textbf{Number}}\\\textbf{\textbf{of test}}\\\textbf{\textbf{models}}\end{tabular}} & \multicolumn{3}{c}{\textbf{Coefficient of Determination} ($R^2$)} \\ 
\cline{4-6}
 &  &  & \begin{tabular}[c]{@{}c@{}}Input\\ only\end{tabular} & \begin{tabular}[c]{@{}c@{}}Hidden \\ w/ input\end{tabular} & \begin{tabular}[c]{@{}c@{}}Hidden \\ only\end{tabular} \\ 
\hline
1 & 64 & 32 & 0.128 $\pm$ 0.076 & \textbf{ 0.926 $\pm$ 0.016} & 0.926 $\pm$ 0.017 \\
2 & 36 & 18 & 0.864 $\pm$ 0.052 & \textbf{ 0.969 $\pm$ 0.012} & 0.927 $\pm$ 0.029 \\
4 & 64 & 32 & 0.787 $\pm$ 0.067 & 0.903 $\pm$ 0.037 & \textbf{0.906 $\pm$ 0.037} \\
5 & 42 & 22 & 0.512 $\pm$ 0.487 & \textbf{0.685 $\pm$ 0.331} & 0.634 $\pm$ 0.379 \\
6 & 64 & 32 & -0.128 $\pm$ 0.193 & \textbf{0.759 $\pm$ 0.06} & 0.717 $\pm$ 0.073 \\
7 & 32 & 16 & 0.169 $\pm$ 0.210 & 0.332 $\pm$ 0.410 & \textbf{0.381 $\pm$ 0.377} \\
8 & 42 & 22 & -0.067 $\pm$ 0.186 & 0.280 $\pm$ 0.274 & \textbf{0.300 $\pm$ 0.165} \\
9 & 21 & 11 & -0.251 $\pm$ 0.216 & \textbf{0.795 $\pm$ 0.087} & 0.670 $\pm$ 0.171 \\
\hline
\end{tabular}
\end{table}
One also observes that there are several tasks on which all three measures perform very poorly: Tasks $5$, $7$, $8$, and $9$. We suspect that this is likely due to the limited number of models available and that the linear regressors severely overfit the training data. In the extreme case of Task $9$, there are nearly as many input signatures ($d=20$ for the `Hidden w/ input' variant) as training samples ($21$ train models). To remedy this, we repeat this analysis, but reduce the number of input signatures. This is done by restricting the distribution signatures for each layer to only the third quartile and upper fence, such that $d=2$, $d=8$, and $d=6$ for the 'Input only', 'Hidden w/ input', and 'Hidden only' columns, respectively.  See Appendix~\ref{app:hidden_lin_model_coeffs} for the rationale behind the use of the last two signatures. The results of this analysis are shown in Table~\ref{tab:hidden_r_squared_two_signatures}.

One observes that reducing the number of input signatures increases the performance on the aforementioned problematic tasks, and does not seem to alter the performance much on the others. Furthermore, these results also show that including the input layer does not have a large effect on performance.  Similarly, only considering the input layer performs much worse, on average, than the methods that include hidden margin distributions. Finally, it is also clear that for Tasks $7$ and $8$, none of the methods perform well. This is an interesting observation -- it appears that for these tasks, input or hidden margins are not very predictive, and highlights that these metrics can fail in certain scenarios. Note that this is not solely due to limited training data, as hidden margins perform well for other tasks with a similarly small number of models (e.g. Tasks $2$ and $9$). Instead, we believe this failure to be due to the rather small variation in test accuracy between the models within these tasks. Recall Table~\ref{tab:pgdl_task_overview} in Section~\ref{sec:back_pgdl}.


In summary:
\begin{enumerate}
    \item Input margins are generally not predictive of generalization in this setting.
    \item Hidden margins are more predictive, in general, although there are tasks for which this measure fails. Furthermore, we observe a large variation in performance between different tasks.
    \item We observe worse results on the PGDL dataset (in some cases) compared to the results originally reported by Jiang et al.~\cite{predict_gen_margin} on the DEMOGEN dataset. 
\end{enumerate}

In the following section, we consider the case of \textit{ranking} generalization using input and hidden margins.

\section{Margins as complexity measure}
\label{sec:hidden_comp_measure}

In this section, we investigate the case of using margin measurements as a standalone complexity measure, i.e. without the use of a training set of models and a linear regression model. Where the goal in the previous section was to \textit{numerically} predict generalization, the goal here is to \textit{rank} each model within a set according to its generalization, in accordance with the original goal of the PGDL dataset.

As mentioned earlier, several authors have investigated hidden margins from this perspective~\cite{rep_based_complexity_pgdl,optimal_transport}, but several questions remain (see Section~\ref{sec:hidden_intro}).

In Section~\ref{sec:hidden_hidden_layer_selection} we do an analysis of which hidden layers to consider for the PGDL tasks. We find that the selection of hidden layer can have a large effect on performance and that these measurements fail in certain cases. Following this, in Section~\ref{sec:hidden_failure_cases} we determine which model hyperparameters result in margins failing to accurately rank generalization. In addition, we include input margins in all of these evaluations as a baseline.

\subsection{Hidden layer selection}
\label{sec:hidden_hidden_layer_selection}

Without the use of a linear regression model, there is, of course, the question of which hidden layers to use to characterize a given set of models. As explained in Section~\ref{sec:back_hidden_margins}, Natekar and Sharma~\cite{rep_based_complexity_pgdl} simply take the average of the mean margin over all layers within a network. On the other hand, Chuang et al.~\cite{optimal_transport} consider the median of only the first or last hidden layer. Furthermore, recall that Jiang et al.~\cite{predict_gen_margin} originally considered summary statistics from three equally spaced layers (as we did in the previous section). In this section, we experiment with all these hidden layer selection variations. This allows us to determine how robust hidden margins are to these different selections and also which selection performs best in general for the tasks considered here. We also determine how well input margins perform as a standalone complexity measure -- to our knowledge, no such evaluation has been previously done on input margins.

To evaluate margin measurements as a complexity measure, we make use of the Taylor-approximated margin distributions calculated for all training samples earlier (those used in Section~\ref{sec:hidden_lin_reg}). We then extract a single summary statistic from each distribution and evaluate the predictive performance on the PGDL tasks. In this case, we summarize each distribution in the most interpretable way and simply use the mean, as done by Natekar and Sharma~\cite{rep_based_complexity_pgdl}.
Two metrics are used to evaluate how well the model ranking of each complexity measure aligns with the true ranking given by the generalization of each model: 1) Kendall's rank correlation~\cite{kendall_corr}, and 2) Conditional Mutual Information (CMI)~\cite{pgdl_overview}. Each metric serves a distinct purpose: 1) Kendall's rank correlation provides an easily interpretable score, with values between $-1$ (perfect disagreement) and $1$ (perfect agreement). 2) The CMI metric is more robust to spurious correlations and should be a better indication of whether there is a high probability of a causal relationship, with values between $0$ (no relationship) and $100$ (strong probability of a causal relationship). Recall our explanation of these metrics in Sections \ref{sec:back_kendall_rank} and \ref{sec:back_conditional_mutual_info}. Also, recall the details of how we use each of these metrics, as explained in Section~\ref{sec:back_use_of_evaluation_metrics}. Specifically, we calculate the Kendall's rank correlation between the mean margin and \textit{test accuracy}. On the other hand, we calculate the CMI between the negative of the mean margin and \textit{generalization gap}.

We show the resulting Kendall's rank correlation and CMI for each hidden layer selection and input margins in Tables \ref{tab:hidden_layer_comparison_mean_kendall} (Kendall) and \ref{tab:hidden_layer_comparison_mean_cmi} (CMI). Each column under `Hidden' indicates which hidden layers are used. `First' and `Last' correspond to only using the mean margin of the first or last convolutional layer, respectively. `Equally spaced' indicates taking the average of the mean margin over the first, last, and middle convolutional layers. `All' indicates taking the average of the mean margin over all the convolutional layers.

\begin{table}[H]
\caption[Comparison of hidden layer selection methods for hidden margins on the PGDL tasks (Kendall's)]{Kendall's rank correlation between mean hidden margin and test accuracy using different hidden layer selections.}
\centering
\label{tab:hidden_layer_comparison_mean_kendall}
\begin{tabular}{@{}cccccc@{}}
\toprule
\multirow{3.5}{*}{\textbf{Task}} & \multirow{3.5}{*}{\textbf{Input}} & \multicolumn{4}{c}{\textbf{Hidden}} \\ \cmidrule(l){3-6} 
 &  & First & Last & \begin{tabular}[c]{@{}c@{}}Equally\\ spaced\end{tabular} & All \\ \midrule
1 & 0.0244 & 0.5772 & \textbf{0.829} & 0.7961 & 0.7821 \\
2 & 0.6841 & 0.6981 & 0.7135 & 0.7778 & \textbf{0.8309} \\
4 & 0.6256 & \textbf{0.7975} & 0.1066 & 0.1781 & 0.2724 \\
5 & 0.3958 & \textbf{0.5357} & 0.0069 & 0.1230 & 0.1319 \\
6 & -0.1351 & \textbf{0.4427} & 0.2365 & 0.2637 & 0.2839 \\
7 & 0.3215 & 0.3623 & 0.3179 & 0.3250 & \textbf{0.3925} \\
8 & -0.1223 & -0.0616 & \textbf{0.2078} & 0.0924 & 0.1809 \\
9 & 0.1573 & \textbf{0.7097} & 0.3871 & 0.4194 & 0.4516 \\ \midrule
Average & 0.2439 & \textbf{0.5077} & 0.3507 & 0.3719 & 0.4158 \\ \bottomrule
\end{tabular}
\end{table}

\begin{table}[H]
\caption[Comparison of hidden layer selection methods for hidden margins on the PGDL tasks (CMI)]{Conditional Mutual Information (CMI) between mean hidden margin and generalization gap using different hidden layer selections.}
\centering
\label{tab:hidden_layer_comparison_mean_cmi}
\begin{tabular}{@{}cccccc@{}}
\toprule
\multirow{3.5}{*}{\textbf{Task}} & \multirow{3.5}{*}{\textbf{Input}} & \multicolumn{4}{c}{\textbf{Hidden}} \\ \cmidrule(l){3-6} 
 &  & First & Last & \begin{tabular}[c]{@{}c@{}}Equally\\ spaced\end{tabular} & All \\ \midrule
1 & 00.06 & 09.17 & \textbf{36.52} & 27.35 & 29.55 \\
2 & 06.09 & \textbf{37.14} & 06.91 & 18.88 & 33.37 \\
4 & 15.05 & \textbf{35.08} & 01.21 & 00.97 & 00.78 \\
5 & 10.54 & \textbf{18.54} & 00.11 & 01.13 & 01.54 \\
6 & 00.57 & \textbf{04.24} & 04.22 & 02.90 & 01.36 \\
7 & 01.47 & 05.04 & 05.02 & 04.77 & \textbf{05.62} \\
8 & 00.72 & 00.34 & \textbf{01.02} & 00.24 & 00.82 \\
9 & 00.55 & \textbf{23.74} & 01.58 & 03.31 & 04.21 \\ \midrule
Average & 04.38 & \textbf{16.66} & 07.07 & 07.44 & 09.66 \\ \bottomrule
\end{tabular}
\end{table}

There are several observations to be made from these results. First, it is clear that input margins are not predictive of generalization on average. We observe that for most tasks, there is little relationship between large input margins and generalization, and this is the case for both evaluation metrics. In fact, for Tasks $6$ and $8$, input margins are slightly negatively correlated with generalization (when considering Kendall's rank correlation). This finding is aligned with previous work that analyzes the relationship between adversarial robustness and generalization~\cite{robustness_tradeoff,robustness_tradeoff_two, robustness_tradeoff_three} (recall Section~\ref{sec:back_input_margins}). That said, there does appear to be a stronger relationship between input margins and generalization, for Tasks $2$ and $4$. However, the Kendall's rank correlation for these two tasks can be misleading: Note that for Task $2$, input margins and hidden margins on the first layer achieve a similar Kendall's rank correlation ($0.6841$ and $0.6981$, respectively). On the other hand, we see a large discrepancy between these two metrics when considering CMI ($6.09$ and $37.14$ for input and hidden, respectively). This indicates that the `high' correlation achieved by input margins for this task is perhaps spurious, and the relationship is possibly less causal in nature.

In terms of hidden margins, it is clear that the selection of hidden layers can have a large effect. One observes that performance can vary greatly for a single task depending on which hidden layers are considered. However, all four selection methods outperform input margins on average.  It is also clear that, for the tasks considered here, using only the first layer performs best overall, for both Kendall's rank correlation and CMI. The `Equally spaced' variant is also of particular interest. Recall that in the previous section we used the distribution signatures from three equally spaced layers. This resulted in high $R^2$ values for most tasks. However, here, the equally spaced variant has poor performance. It appears that the linear regression model is able to learn an appropriate weighting to combine the signatures across the different layers, which is not captured by simply averaging over the mean margin values for the purpose of ranking.

This analysis highlights a central issue with using hidden margins as an indicator of generalization -- the metric is not robust across networks of varying architecture, and results can vary much per task depending on the selection of hidden layers. We do a further analysis of these failure cases in the following section.

\subsection{Where do margins fail?}
\label{sec:hidden_failure_cases}

In the previous section, we evaluated margin measurements using summarising metrics (Kendall's rank correlation and CMI). Here, we do a more fine-grained investigation of how margins perform per hyperparameter variation. To this end, we calculate the \textit{granulated Kendall's coefficient} as introduced in Section~\ref{sec:back_granulated_kendall}. We do this for all hyperparameter variations, for each task, and for both hidden and input margins. For hidden margins, we use the first hidden layer, motivated by our findings in Section~\ref{sec:hidden_hidden_layer_selection} which show it to be the best-performing layer selection method. All other details are precisely as explained in the previous section; we simply change the evaluation metric. The granulated Kendall's coefficient for each hyperparameter variation is shown in Tables \ref{tab:hidden_hidden_granulated} and \ref{tab:hidden_input_granulated} for hidden and input margins, respectively. As a reminder, recall that a high granulated Kendall's coefficient for a specific hyperparameter indicates that the metric performs well when \textit{only} that hyperparameter is varied.

Let us first describe each hyperparameter column shown: 
\begin{itemize}
    \item `Learning rate', `Batch size', `Weight decay', and `Dropout probability' are self-explanatory.
    \item `Depth': the number of convolutional layers, or alternatively the number of convolutional blocks, for example, the number of VGG blocks (conv-conv-pool configuration) for the VGG architectures of Task $1$.
    \item `Width': the number of channels of a convolutional layer.
    \item `Num dense': the number of dense, or fully connected, layers at the end of the convolutional stage.
    \item `Reverse': whether the number of channels (Width) of each convolutional layer increases or decreases at each subsequent layer.
\end{itemize}
Finally, the `$\mu(\psi)$' column is the average of the granulated Kendall's coefficient over each hyperparameter for each task (recall Section~\ref{sec:back_granulated_kendall}). 

Note that there is some variation between tasks in the way each hyperparameter is varied due to differences in architecture. For example, for Task $6$, the `Width' is varied by setting all convolutional layers to either $256$ or $512$ channels, while for Tasks $1$ and $8$ the size of the last layer is varied between $256$ and $512$ with the earlier layers respectively smaller. Due to this, not all hyperparameters are directly comparable across tasks, we mark these columns with a \textdagger. Refer to Jiang et al.~\cite{pgdl_overview} for additional details on how each hyperparameter is varied within a task. Finally, the cells marked with a `-' indicate that this hyperparameter was not varied for the task in question.

\begin{table}[H]
\caption[Granulated Kendall's coefficient for hidden margins on the PGDL tasks]{Granulated Kendall's coefficient for hidden margins (1st layer) for all PGDL tasks and hyperparameter variations. `\textdagger' Indicates a hyperparameter variation that is not directly comparable across tasks.}
\label{tab:hidden_hidden_granulated}
\begin{adjustbox}{max width=\textwidth}
\begin{tabular}{@{}cccccccccc@{}}
\toprule
\textbf{Task} & \textbf{\begin{tabular}[c]{@{}c@{}}Learning\\ rate\end{tabular}} & \textbf{\begin{tabular}[c]{@{}c@{}}Batch \\ size\end{tabular}} & \textbf{\begin{tabular}[c]{@{}c@{}}Weight\\ decay\end{tabular}} & \multicolumn{1}{c}{\textbf{\begin{tabular}[c]{@{}c@{}}Dropout\\ probability\end{tabular}}} & \textbf{Depth}\textdagger & \textbf{Width}\textdagger & \multicolumn{1}{c}{\textbf{\begin{tabular}[c]{@{}c@{}}Num\\ dense\end{tabular}}\textdagger} & \multicolumn{1}{c}{\textbf{Reverse}} & \multicolumn{1}{c}{\textbf{$\mu(\psi)$}} \\ \midrule
1 & - & 0.625 & 0.292 & 0.417 & 0.958 & -0.083 & 0.375 & - & 0.431 \\
2 & - & 0.593 & 0.926 & 0.963 & 0.926 & - & - & - & 0.852 \\
4 & 0.833 & 0.583 & 0.250 & - & 0.750 & 0.833 & - & 1.000 & 0.708 \\
5 & 0.563 & 0.500 & 0.688 & - & 0.563 & 0.375 & - & 0.500 & 0.531 \\
6 & 0.833 & 0.833 & 1.000 & -0.532 & -0.271 & 0.250 & - & - & 0.352 \\
7 & - & 0.083 & 0.083 & -0.083 & 0.750 & \textbf{-} & 0.530 & - & 0.273 \\
8 & 0.188 & -0.063 & 0.000 & 0.313 & -0.500 & 0.000 & - & - & -0.010 \\
9 & \textbf{-} & 0.500 & 1.000 & 0.375 & 0.500 & 0.375 & - & - & 0.550 \\ \midrule
\multicolumn{1}{l}{Average} & \multicolumn{1}{l}{0.604} & \multicolumn{1}{l}{0.457} & 0.530 & 0.242 & \multicolumn{1}{l}{0.460} & \multicolumn{1}{l}{0.292} & 0.453 & 0.750 & 0.461 \\ \bottomrule
\end{tabular}
\end{adjustbox}
\end{table}

\begin{table}[H]
\caption[Granulated Kendall's coefficient for input margins on the PGDL tasks]{Granulated Kendall's coefficient for input margins for all PGDL tasks and hyperparameter variations. `\textdagger' Indicates a hyperparameter variation that is not directly comparable across tasks. }
\label{tab:hidden_input_granulated}
\begin{adjustbox}{max width=\textwidth}
\begin{tabular}{@{}cccccccccc@{}}
\toprule
\textbf{Task} & \textbf{\begin{tabular}[c]{@{}c@{}}Learning\\ rate\end{tabular}} & \textbf{\begin{tabular}[c]{@{}c@{}}Batch\\ size\end{tabular}} & \textbf{\begin{tabular}[c]{@{}c@{}}Weight\\ decay\end{tabular}} & \textbf{\begin{tabular}[c]{@{}c@{}}Dropout\\ probability\end{tabular}} & \textbf{Depth}\textdagger & \textbf{Width}\textdagger & \textbf{\begin{tabular}[c]{@{}c@{}}Num\\ dense\end{tabular}}\textdagger & \textbf{Reverse} & \textbf{$\mu(\psi)$} \\ \midrule
1 & - & -0.396 & 0.250 & 0.208 & 0.292 & 0.333 & 0.292 & - & 0.163 \\
2 & - & -0.148 & 0.111 & 0.926 & 1.000 & - & - & - & 0.472 \\
4 & 0.583 & 0.292 & -0.083 & - & 0.604 & 0.917 & - & 1.000 & 0.552 \\
5 & 0.313 & 0.438 & 0.438 & - & 0.500 & 0.313 & - & 0.313 & 0.385 \\
6 & -0.625 & 0.042 & 0.292 & -0.319 & -0.292 & 0.750 & - & - & -0.025 \\
7 & - & 0.083 & -0.083 & -0.083 & 0.750 & - & 0.136 & - & 0.161 \\
8 & -0.375 & -0.375 & 0.000 & 0.625 & -0.313 & -0.063 & - & - & -0.083 \\
9 & - & -0.875 & 0.750 & -0.500 & 0.500 & 0.375 & - & - & 0.050 \\ \midrule
Average & -0.026 & -0.117 & 0.209 & 0.143 & 0.380 & 0.438 & 0.214 & 0.657 & 0.209 \\ \bottomrule
\end{tabular}
\end{adjustbox}
\end{table}

There are several interesting observations contained in Tables \ref{tab:hidden_hidden_granulated} and \ref{tab:hidden_input_granulated}. Let us first consider the hidden margin results. We observe that, strangely, even for hyperparameters that should be comparable across different tasks, the performance of hidden margins can greatly vary. For example, consider the `Weight decay' column: one observes that hidden margins can accurately capture differences in performance between models that vary this hyperparameter for Tasks $2$, $6$, and $9$. However, the hidden margin measure then fails in this regard for Tasks $1$, $4$, $7$, and $8$. Furthermore, surprisingly, one observes that a variation in depth does not appear to be an issue for Tasks $1$ and $2$ -- one would expect hidden margins to fail in this regard, given that we rely solely on analyzing the first hidden layer here. In the case of input margins, the results are similar: some hyperparameter variations can be accurately accounted for by input margins for some tasks, but never in the general case.

These observations lead to an interesting conclusion: The failure (and success!) of margin measurements is not determined by variations in specific hyperparameters alone -- their performance depends on the architectural family and dataset (i.e. task) of the models considered. Furthermore, there are also interactions between different hyperparameter selections which aren't captured by the granulated Kendall's coefficient metric. These interactions could also contribute to discrepancies in performance we observe.

\section{Beyond Taylor}
\label{sec:hidden_beyond_approximations}

In this chapter, we have hitherto only considered the first-order Taylor approximation for estimating margins in both the input and hidden space for the purpose of generalization prediction. As explained earlier, in Section~\ref{sec:back_hidden_margins}, this approximation is widely used. In this section, we consider whether the predictive power of these margin measurements can be improved if a more precise method is used to estimate the distance to the decision boundary. To the best of our knowledge, no such analysis has been previously considered in the literature. 

\subsection{Modifying DeepFool}
\label{sec:hidden_modifying_deepfool}

In Chapter~\ref{chap:ch3} we made use of a constrained optimization formulation to accurately measure margins. However, when considering the PGDL challenge, such a technique is not feasible: there are simply too many models to consider ($550$) for such a computationally expensive analysis. Furthermore, for hidden margins, the dimensionality of the search space is enormous in comparison to the input space, and the computational expense increases significantly. For example, we find that the output of the largest convolutional layers contains $512k$ dimensions, which is much larger than the $3k$ dimensions of the input space. Thus, to alleviate this computational burden, we instead leverage a technique from the field of adversarial attacks: DeepFool~\cite{deepfool}. 

DeepFool is an adversarial attack algorithm that attempts to find the smallest possible perturbation such that a model misclassifies a given sample. The algorithm is essentially an iterative variant of the Taylor approximation, where the sample is stepped in the direction of the decision boundary after the margin is estimated, as explained in Section~\ref{sec:back_input_margins}. 
We modify the DeepFool algorithm so that it functions as a margin measuring method, as delineated in Algorithm~\ref{alg:deepfool_margin}. The notation is the same as used earlier in Section~\ref{sec:noise_margin_formulation}.

Here follows a short explanation of the key steps: 
\begin{enumerate}

    \item \textbf{Line 1}:
    The sample $\mathbf{x}$ is classified by the model $f$, and the highest predicted class is identified (class $i$). The sample on the decision boundary $\mathbf{\hat{x}}$ is initialized as the original sample $\mathbf{x}$. Note that, at this point, $\mathbf{\hat{x}}$ is not a boundary point.
    \item \textbf{Lines 3 - 6}: 
    The output margin $o_j$ is calculated for each class $j$ that is not $i$. In addition, the difference in gradient between the logits of $i$ and $j$ w.r.t. to the input features for the point $\mathbf{\hat{x}}$. This difference in gradient is denoted as $\mathbf{w}_j$. 
    \item \textbf{Line 7}: 
    The class ($l$) with the smallest Taylor-approximated margin for point $\hat{\mathbf{x}}$ is identified. 
    \item \textbf{Lines 8 - 9}: 
    The sample $\hat{\mathbf{x}}$ is updated by stepping in the direction of the smallest approximated margin, where the step is scaled by the learning rate $\gamma$.
    \item \textbf{Lines 10}: 
    $\hat{\mathbf{x}}$ is clipped so that its feature values are within the lower and upper bounds of the search space.
    \item \textbf{Lines 11 - 13}: 
    The equality violation $v$ and distance $d$ are calculated for the new $\mathbf{\hat{x}}$.
\end{enumerate}
 \begin{algorithm}[H]
    \caption{DeepFool margin calculation}
    \label{alg:deepfool_margin}
    \textbf{Input}: Sample $\mathbf{x}$, classifier $f$\\
    \textbf{Parameter}: Stopping tolerance $\delta$, Learning rate $\gamma$, Maximum iterations $max$\\
    \textbf{Output}: Distance~$d_{best}$, Equality violation~$v_{best}$ 
    
    \begin{algorithmic}[1] 
        \STATE $\mathbf{\hat{x}} \leftarrow \mathbf{x},
        i \leftarrow \argmax f_k(\mathbf{x}), d \leftarrow 0
        , v_{best} \leftarrow \infty, c \leftarrow 0$
        
        \WHILE{$c < max$}
        \FOR{$j \neq i$}
        \STATE $o_j \leftarrow f_i(\mathbf{\hat{x}}) - f_j(\mathbf{\hat{x}}$)
        \STATE $\mathbf{w}_j \leftarrow \nabla f_i(\mathbf{\hat{x}}) - \nabla f_j(\mathbf{\hat{x}})$
        \ENDFOR
        \STATE $l \leftarrow \argmin_{j \neq i}\frac{|o_j|}{||\mathbf{w_j}||_{2}}$
        \STATE $\mathbf{r} \leftarrow \frac{o_l}{||\mathbf{w}_l||_{2}^2} \mathbf{w}_l$
        \STATE $\mathbf{\hat{x}} \leftarrow \mathbf{\hat{x}} + \gamma\mathbf{r}$
        \STATE $\mathbf{\hat{x}} \leftarrow$ clip $(\mathbf{\hat{x}})$
        \STATE $j \leftarrow \argmax_{k \neq i} f_{k}(\mathbf{\hat{x}})$
        \STATE $v \leftarrow |f_i(\mathbf{\hat{x}}) - f_j(\mathbf{\hat{x}})|$
        \STATE $d \leftarrow ||\mathbf{x} - \mathbf{\hat{x}}||_{2}$
        \IF{$v \geq v_{best}$ \OR $|d - d_{best}| <  \delta$}
            \STATE \textbf{return} $d_{best}, v_{best}$
        \ELSE
            \STATE $v_{best} \leftarrow v$
            \STATE $d_{best} \leftarrow d$
            \STATE $c \leftarrow c + 1 $
        \ENDIF
        \ENDWHILE
        \STATE \textbf{return} $d_{best}, v_{best}$
    \end{algorithmic}
\end{algorithm}

This process is then repeated until some stopping criterion is met, and the algorithm returns the margin $d_{best}$ and equality violation $v_{best}$ for the point that achieved the smallest equality violation, that is, the point closest to some decision boundary. The term `equality violation' refers to the difference between the two highest class predictions, i.e. $f_i(\mathbf{\hat{x}}) - f_j(\mathbf{\hat{x}})$ (the violation of Equation~\ref{eq:eq_constraint} discussed earlier in Section~\ref{sec:noise_margin_formulation}). The equality violation is an indication of how close the point is to a decision boundary, and should ideally be near zero for the final point, as that would indicate the two top predicted classes have approximately equal outputs.

As stopping criteria, we terminate the search when any of the following occurs:
1) the equality violation grows ($v \geq v_{best}$), 2) the margin converges, i.e. the distance changes less than a given tolerance ($|d - d_{best}| < \delta$), or 3) some maximum number of iterations have elapsed ($c \geq max$). Note that we find that the first condition never happens in practice (the equality violation always decreases), but it is a useful safeguard against unexpected behavior.

This algorithm is similar to how DeepFool was originally defined, although there are key differences:
\begin{enumerate}
    \item Learning rate: The original DeepFool does not make use of a learning rate to scale the step size, that is, it is equivalent to using $\gamma = 1.0$.
    \item Stopping criteria: DeepFool simply terminates its optimization when the classification of $\mathbf{\hat{x}}$ changes, as such it does not rely on a stopping tolerance $\delta$.
    \item Step direction: Notice that the output margin $o_l$ on line $8$ is not necessarily positive, as it will be negative once $\mathbf{\hat{x}}$ crosses the decision boundary into a classification region of a different class. This implies that the point $\mathbf{\hat{x}}$ can `step back' in the opposite direction if it crosses the decision boundary. The original DeepFool implementation does not allow any steps in the opposite direction, which means that $o_l$ on line $8$ would be $|o_l|$.
    \item Clipping: The original DeepFool algorithm does not mention clipping the feature values to stay within the bounds of the dataset. 
\end{enumerate}

While the formulation described in Algorithm~\ref{alg:deepfool_margin} is much more computationally efficient than the constrained optimizer we have employed earlier, this can be further improved by batching. That is, one can calculate the margins for a large batch of samples in parallel, which allows one to effectively utilize GPUs for the computation of the gradient calculations described on line $5$. 
As such, when referring to Algorithm~\ref{alg:deepfool_margin}, note that it is calculated in a batched manner such that the \textit{average} distance ($d$) across the batch is used to determine when the optimization terminates (lines $13$ and $14$). That said, we keep track of the equality violation ($v$) of each sample individually, and always return the margin for each sample that achieved the smallest violation throughout the optimization.  

\subsection{Verifying DeepFool margin calculation (Algorithm~\ref{alg:deepfool_margin})}
\label{sec:hidden_verifying_deepfool_alg}

Before we evaluate how well input or hidden margins perform when measured using Algorithm~\ref{alg:deepfool_margin}, we first verify its accuracy and establish the effect of its hyperparameters on performance. For the sake of brevity, we shall occasionally refer to any margin calculations performed using Algorithm~\ref{alg:deepfool_margin} as `DeepFool' or `DF' followed by the specific hyperparameters employed in each case, if appropriate. Furthermore, we restrict our verification experiments to the input space for computational efficiency.

As a test bed, we select $10$ models from Task $1$ of varying generalization ability. In the following subsections, we establish the effect of each hyperparameter in Algorithm~\ref{alg:deepfool_margin}. In each case, we calculate the input space margins for $500$ randomly selected training samples for each of these models using Algorithm~\ref{alg:deepfool_margin} with different hyperparameter setups. As a baseline, we also calculate the input space margins using the constrained optimization formulation employed earlier in Chapter~\ref{chap:ch3} (recall Section~\ref{sec:noise_margin_formulation}), meaning the augmented Lagrangian combined with the CCSAQ optimizer.

\subsubsection{Effect of learning rate}

First, we establish what effect the learning rate $\gamma$ has on the algorithm's performance, without considering any stopping criteria. To this end, we disable the distance tolerance stopping criterion on lines $14$ and $15$ of Algorithm~\ref{alg:deepfool_margin} and simply terminate the optimization after $100$ steps ($max=100$). We choose $100$ because we empirically find that the margin for each sample has converged (the distance no longer changes) long before this point. We do this for $4$ different learning rates, $\gamma \in \{0.25, 0.50, 0.75, 1.00\}$. 

\begin{figure}[h]
    \centering
    \includegraphics[width=0.49\linewidth]{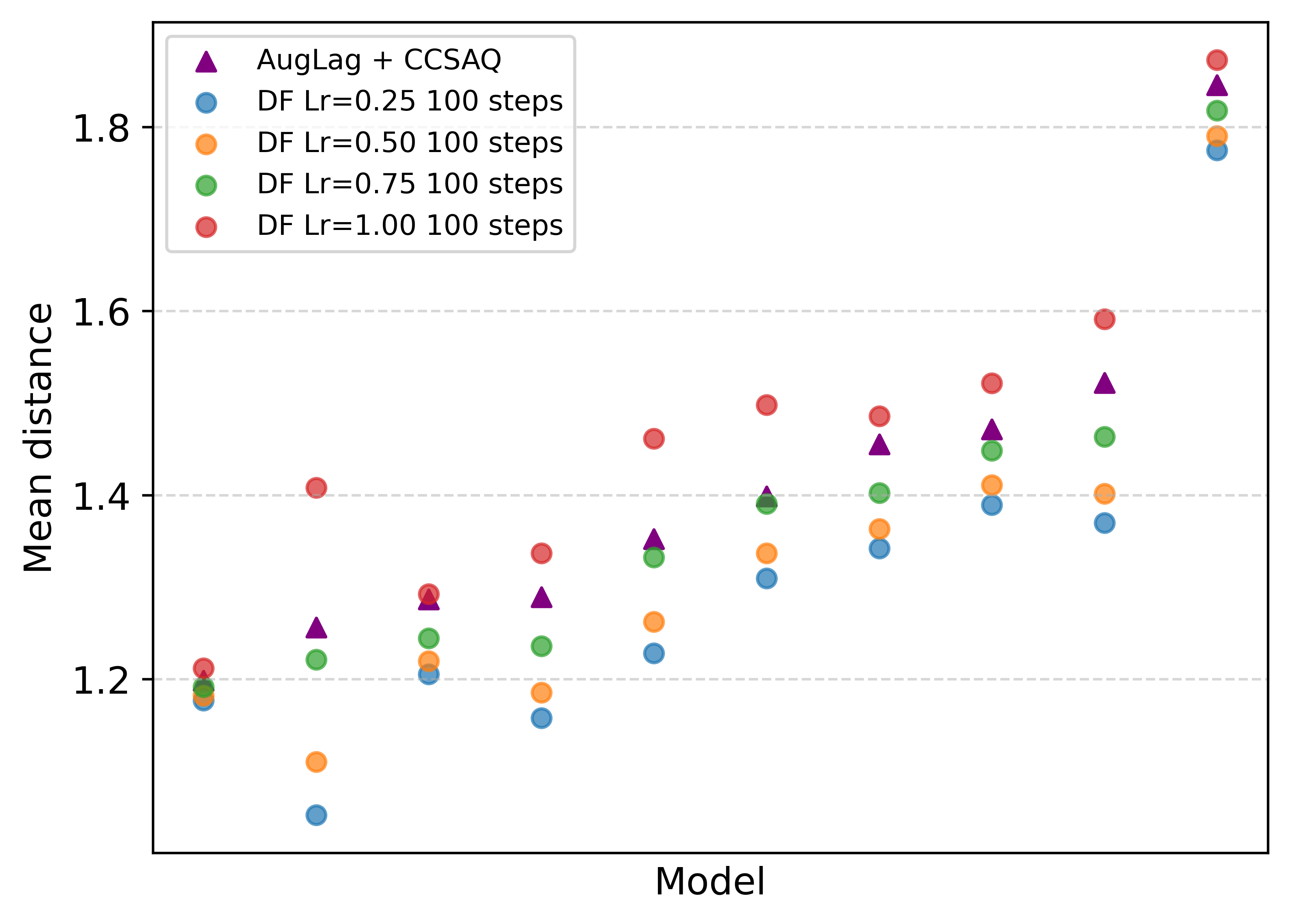}
    \includegraphics[width=0.49\linewidth]{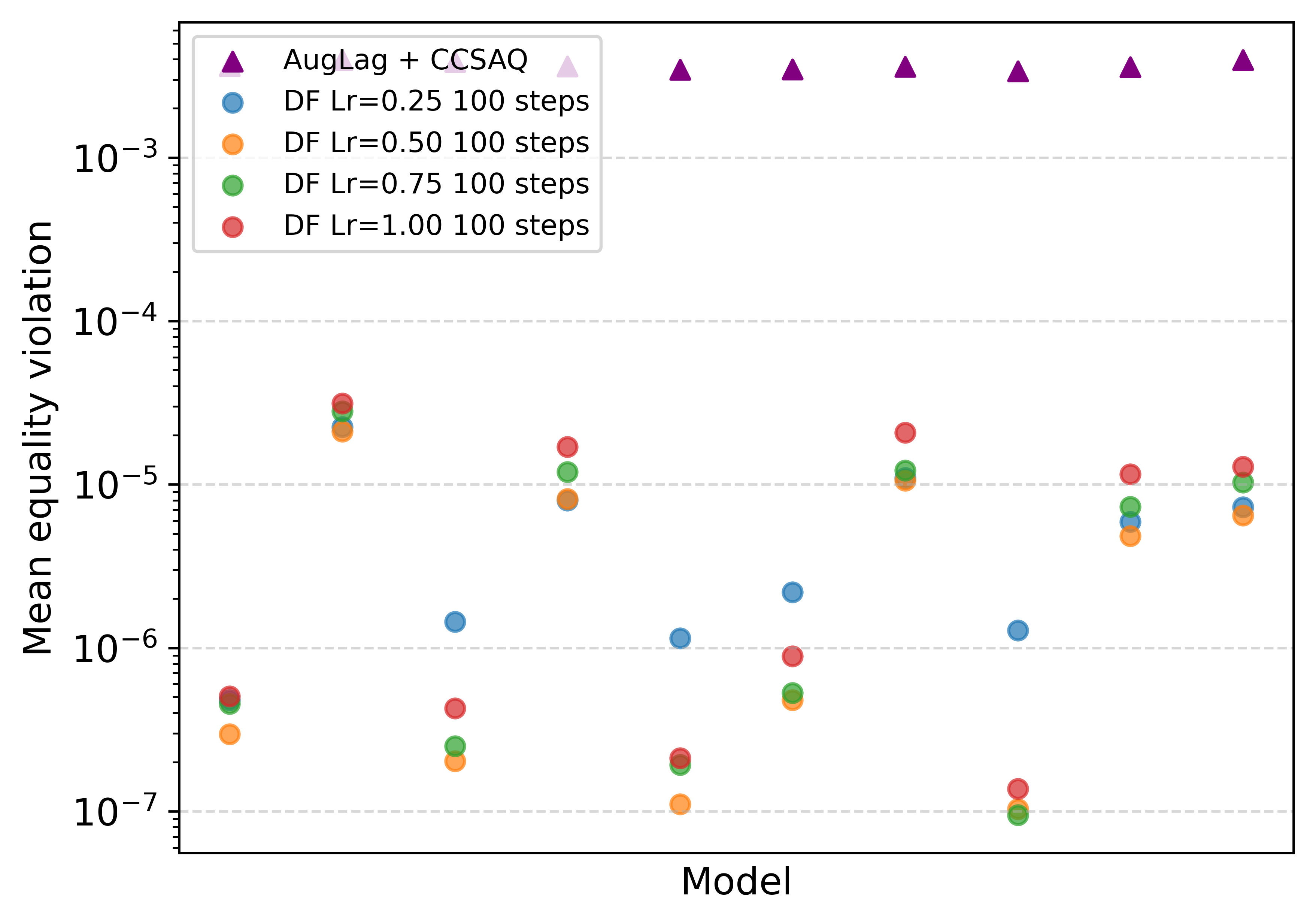}
    \caption[Learning rate comparison for the modified DeepFool algorithm]{Comparison using different learning rates when calculating input margins using the modified DeepFool algorithm ($DF$) for $10$ models from Task $1$. Each batch of samples is optimized for $100$ steps. Left: Mean distance per model. Right: Mean equality violation per model (logarithmic scale). `AugLag + CCSAQ' indicates using a constrained optimization formulation, included as baseline.}
    \label{fig:hidden_mean_margin_and_viol_lr_comparison_no_tol}
\end{figure}

In Figure~\ref{fig:hidden_mean_margin_and_viol_lr_comparison_no_tol}, left, we compare the mean distance ($d_{best}$) found for each of the $10$ models, while on the right we show the average equality violation ($v_{best}$) for the points found. Note that the models are ordered in ascending order of mean distance, using the baseline mean distance for each model.

Several interesting observations are shown in Figure~\ref{fig:hidden_mean_margin_and_viol_lr_comparison_no_tol}. Firstly, on the left, we observe that our modified DeepFool algorithm can outperform the constrained optimization baseline when using any $\gamma < 1.0$, that is, we can find smaller distances. It is also clear that, the smaller the learning rate, the smaller the distance. On the right of Figure~\ref{fig:hidden_mean_margin_and_viol_lr_comparison_no_tol} we also observe that the equality violations are very small in all cases, meaning the points found are very close to the decision boundary. 

\subsubsection{Effect of small distance tolerance}

\begin{figure}[h]
    \centering
    \includegraphics[width=0.49\linewidth]{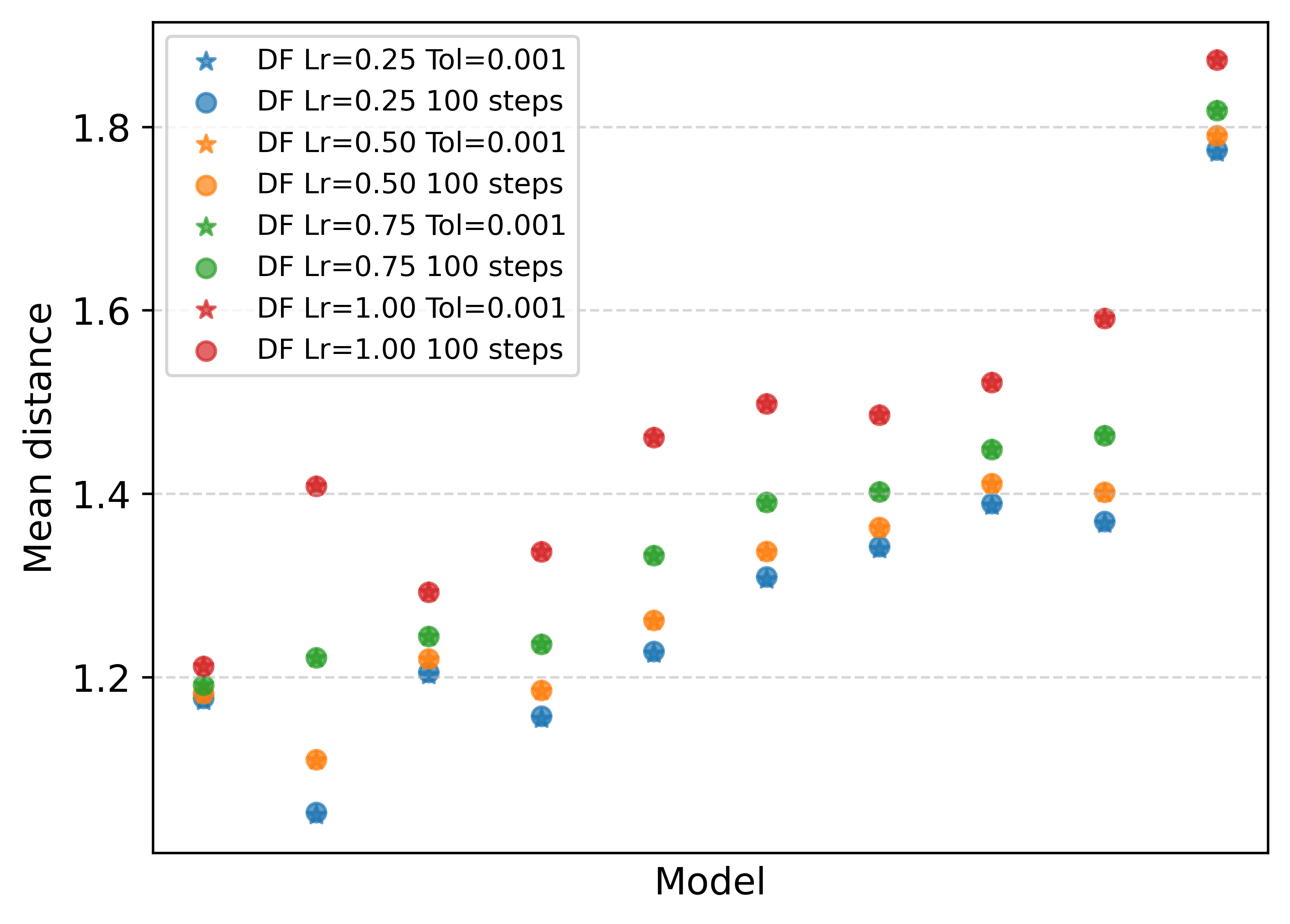}
    \includegraphics[width=0.49\linewidth]{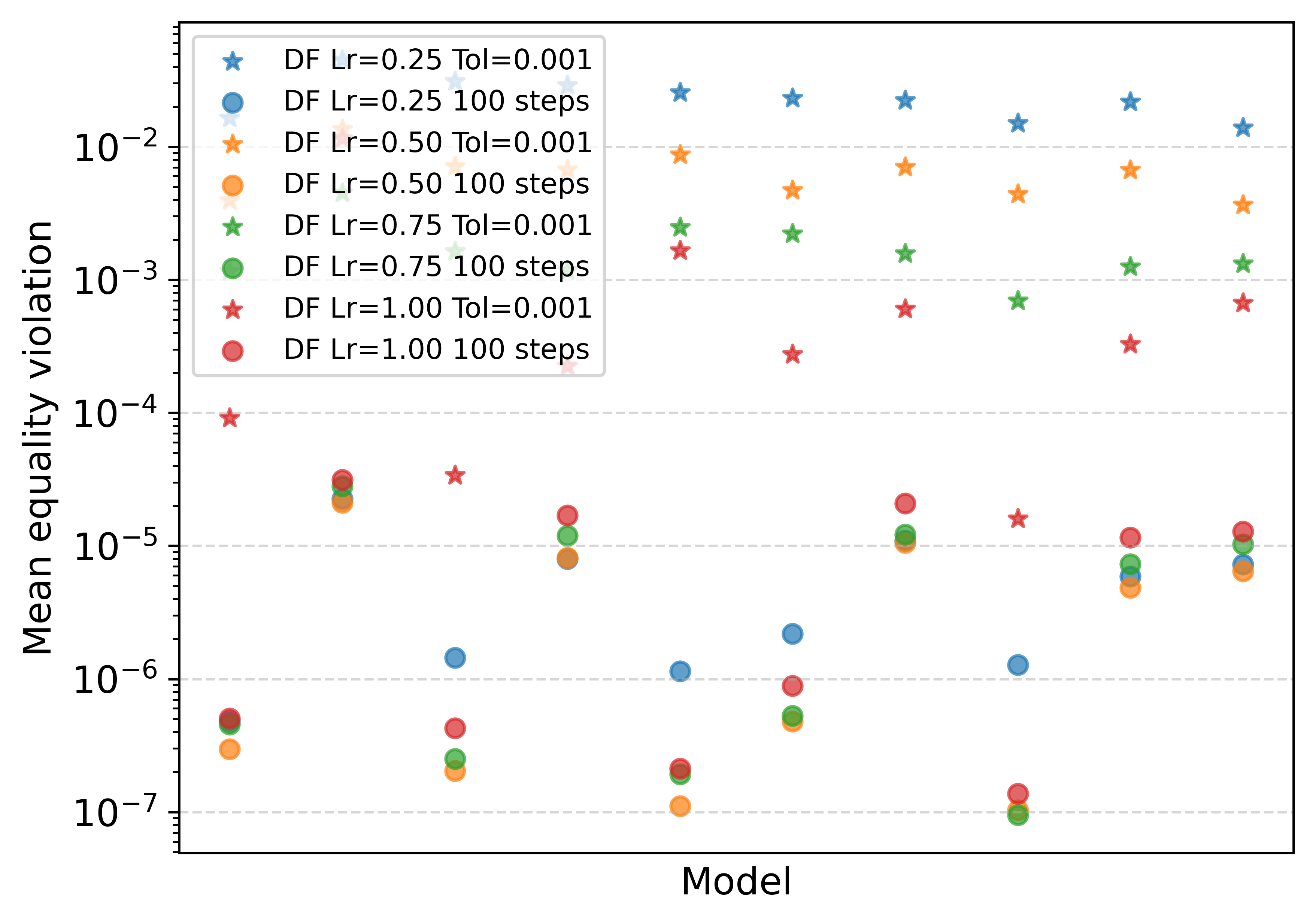} \\
    \includegraphics[width=0.49\linewidth]{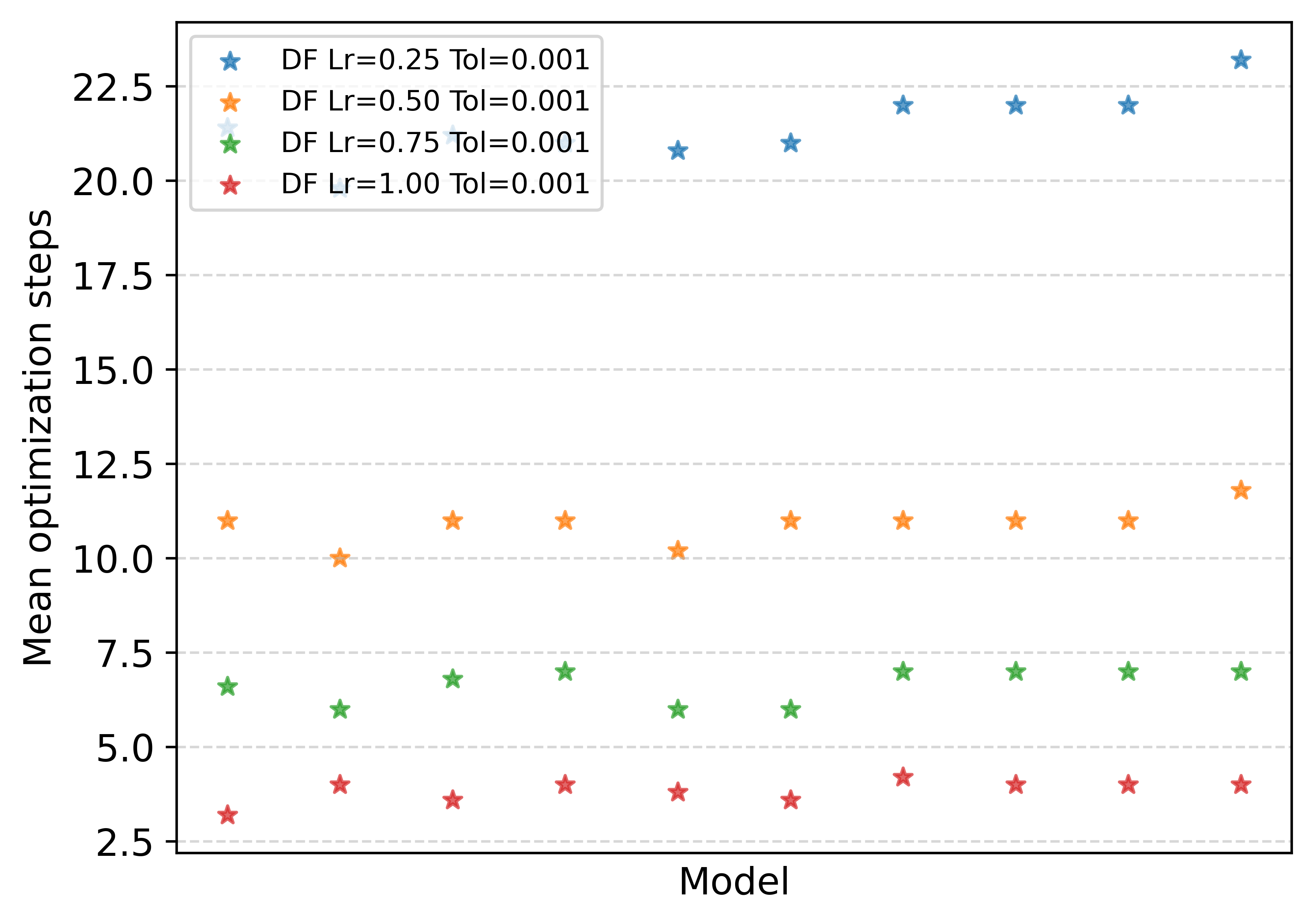}
    \caption[Small distance tolerance versus no tolerance comparison for the modified DeepFool algorithm]{Comparison of using no distance tolerance to a small tolerance ($\delta=0.001$) when calculating input margins using the modified DeepFool algorithm ($DF$) for $10$ models from Task $1$, for $4$ different learning rates. Top left: Mean distance per model. Top right: Mean equality violation per model (logarithmic scale). Bottom: Mean number of optimization steps.}
    \label{fig:hidden_mean_margin_and_steps_small_tol_comparison}
\end{figure}

Let us now consider what effect the distance tolerance ($\delta$) has on the final distance and equality violations. We repeat the previous analysis for all $4$ learning rates, but now we use a very small distance tolerance of $\delta=0.001$ as stopping criterion. In Figure~\ref{fig:hidden_mean_margin_and_steps_small_tol_comparison} we compare these with the previous results using no tolerance. We show the mean distance (top left), equality violations (top right), and also the average number of optimization steps before the optimization terminates (bottom) for the $\delta=0.001$ variants.

First, consider the mean distance on the top left of Figure~\ref{fig:hidden_mean_margin_and_steps_small_tol_comparison}. We observe that the distances are virtually identical when comparing those found with no stopping tolerance (circles) to those with a small tolerance (stars) for the varying learning rates. In fact, they are plotted directly on top of each other (best seen zoomed in). 

On the top right of Figure~\ref{fig:hidden_mean_margin_and_steps_small_tol_comparison} it is clear that the $\delta=0.001$ stopping tolerance (stars) results in points with higher equality violations compared to their no tolerance counterparts (circles). It is also evident from the bottom of Figure~\ref{fig:hidden_mean_margin_and_steps_small_tol_comparison} that the $\delta=0.001$ methods terminate in far fewer than $100$ optimization steps. This suggests that both methods find almost identical points near the decision boundary. The additional optimization steps after the stopping criterion is reached simply `fine tune' these points to further reduce the equality violation, which does not noticeably affect the distance. 

When considering the number of optimization steps (bottom of Figure~\ref{fig:hidden_mean_margin_and_steps_small_tol_comparison}), we observe that smaller learning rates require a larger number of steps before the optimization terminates.  From these observations, we can conclude that while smaller learning rates lead to finding points near the decision boundary closer to the original sample, they are also more computationally expensive. This implies that choosing a learning rate comes with a trade-off between performance (small distances) and computational cost.

\subsubsection{Effect of larger distance tolerance}

\begin{figure}[h]
    \centering
    \includegraphics[width=0.49\linewidth]{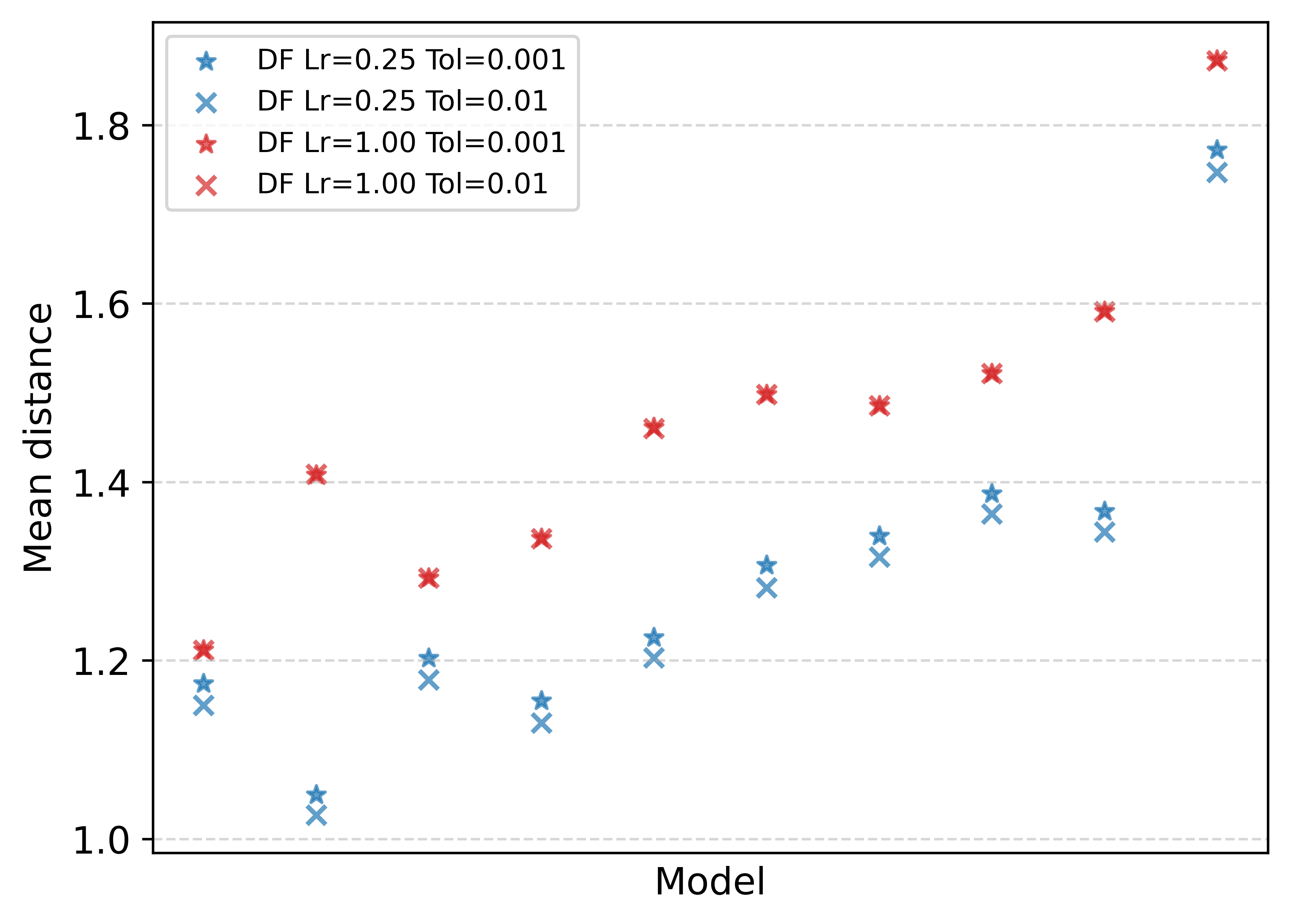}
    \includegraphics[width=0.49\linewidth]{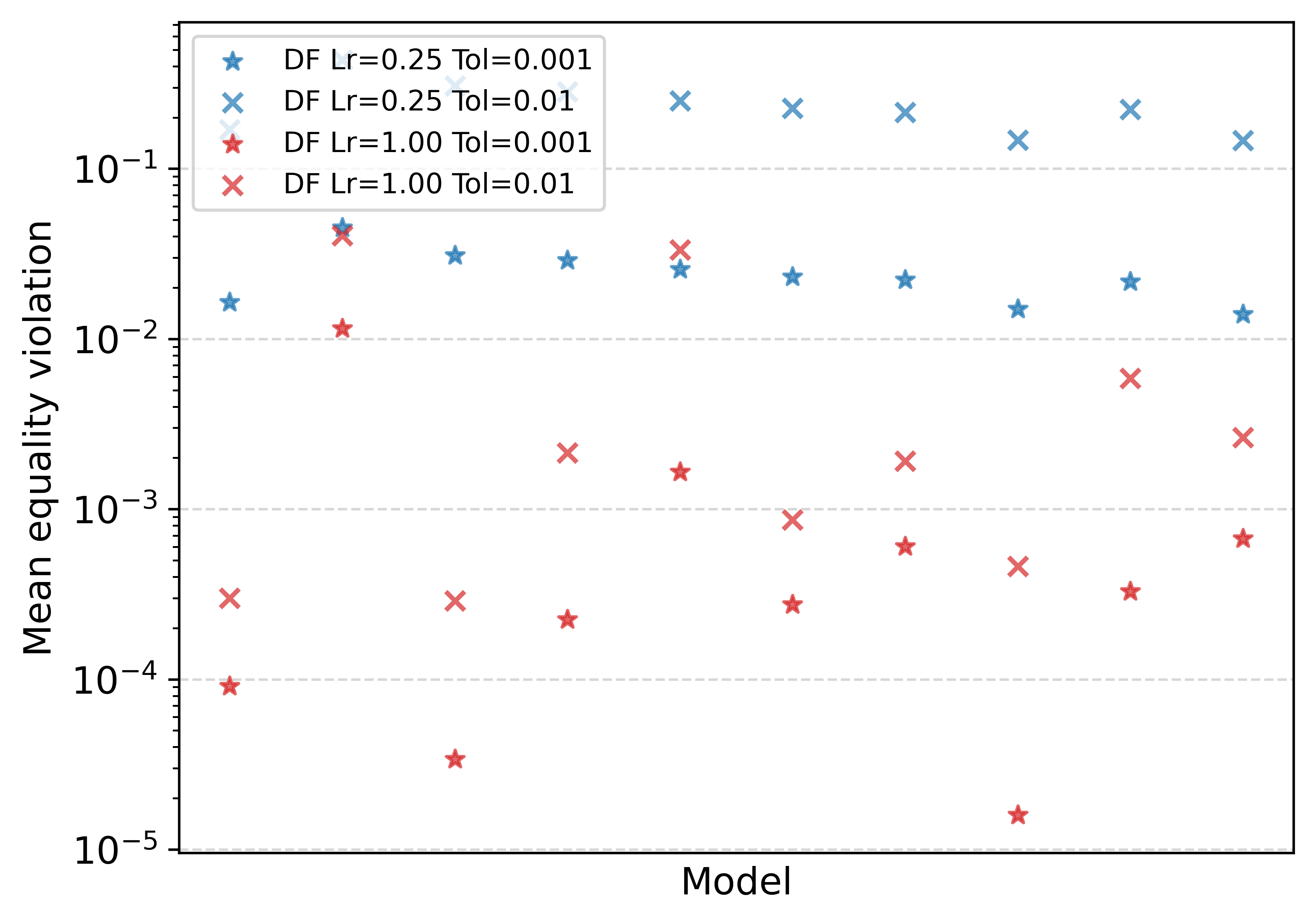} \\
        \includegraphics[width=0.49\linewidth]{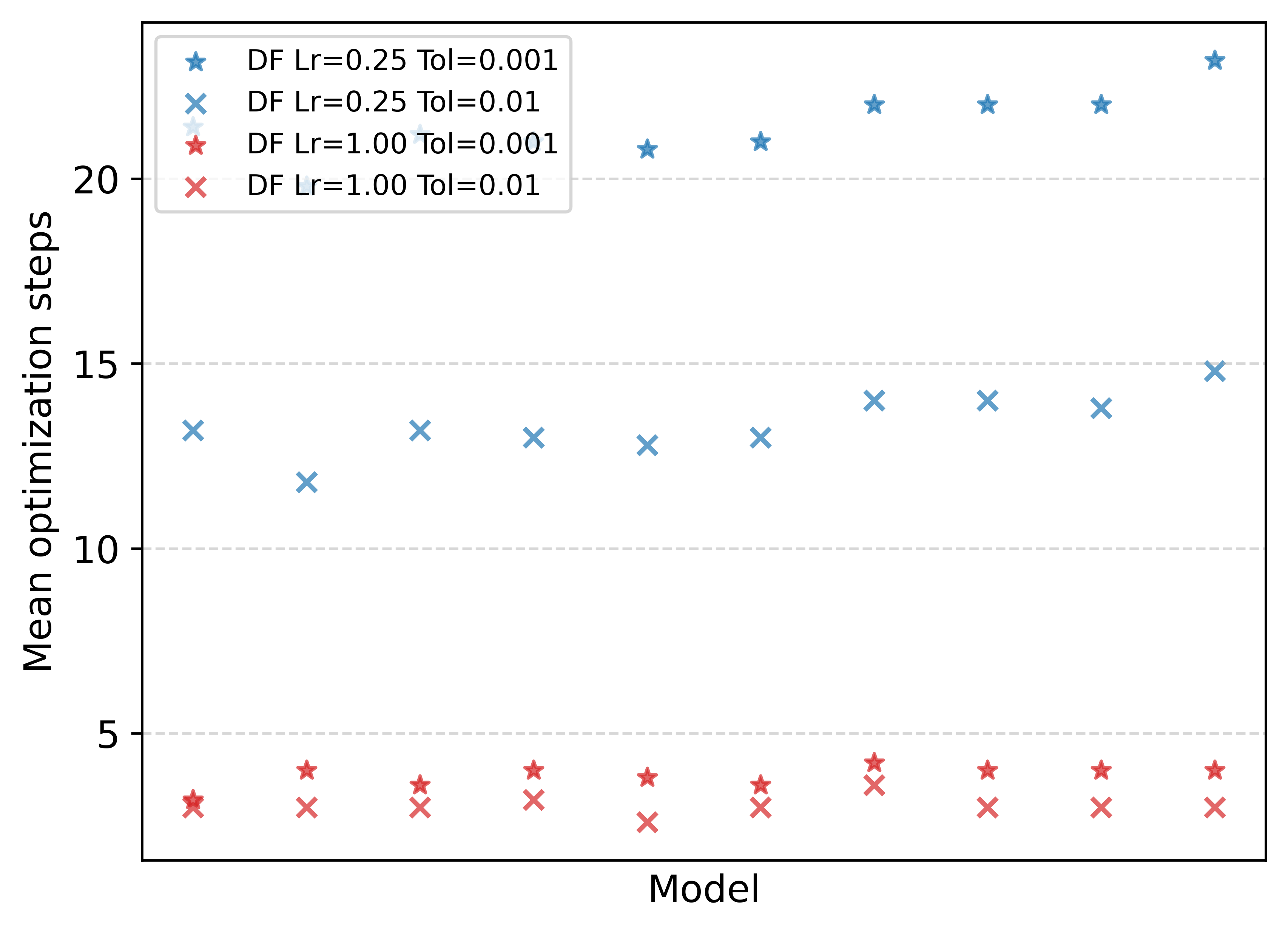} 
    \caption[Small distance tolerance versus large tolerance comparison for the modified DeepFool algorithm]{Comparison of using a small and large distance tolerance when calculating input margins using the modified DeepFool algorithm ($DF$) for $10$ models from Task $1$, for $2$ different learning rates. Top left: Mean distance per model. Top right: Mean equality violation per model. Bottom: Mean number of optimization steps.}
    \label{fig:hidden_mean_margin_and_viol_comparison_small_and_large_tolerance}
\end{figure}

Given these observations, we can now consider whether the number of optimization steps can be further reduced by using a larger distance tolerance $\delta$. Furthermore, we establish what effect this has on the distance and equality violations of the points found.  We select the smallest and largest learning rates ($\gamma=0.25$ and $\gamma=1.0$, respectively) and recalculate the margins using a larger stopping distance tolerance of $\delta=0.01$. In Figure~\ref{fig:hidden_mean_margin_and_viol_comparison_small_and_large_tolerance} we compare the two different distance tolerance methods by considering the mean distance (top left), equality violation (top right) and number of optimization steps (bottom).

Let us first consider the mean distance on the left of Figure~\ref{fig:hidden_mean_margin_and_viol_comparison_small_and_large_tolerance}. For the large learning rate, we observe that the increased tolerance has very little effect. The mean distance per model of the $\delta=0.001$ and $\delta=0.01$ variants is virtually identical (red stars and crosses). For the smaller learning rate, we observe that the distances are slightly smaller for the larger tolerance method ($\delta=0.01$), although the difference is essentially constant between all $10$ models. On the top right of Figure~\ref{fig:hidden_mean_margin_and_viol_comparison_small_and_large_tolerance} we also observe that, as expected, the larger distance tolerance results in findings points with larger equality violations. However, we observe at the bottom of Figure~\ref{fig:hidden_mean_margin_and_viol_comparison_small_and_large_tolerance} that the number of optimization steps is greatly reduced for the small learning rate ($\gamma=0.25$) when using the larger distance tolerance ($\delta=0.01$). On the other hand, for the large learning rate ($\gamma=1.0$) the difference is minimal.

Given these results, we can ask whether using a small learning rate and a large tolerance ($\gamma=0.25$ and $\delta=0.01$) is sufficient for the purpose of generalization ranking. While this method does result in finding higher equality violations, the difference in distance compared to using a smaller tolerance is minimal. Furthermore, this difference appears to be consistent, which implies that it will not influence the final ranking of the models. 

\begin{figure}[h]
    \centering
    \includegraphics[width=\linewidth]{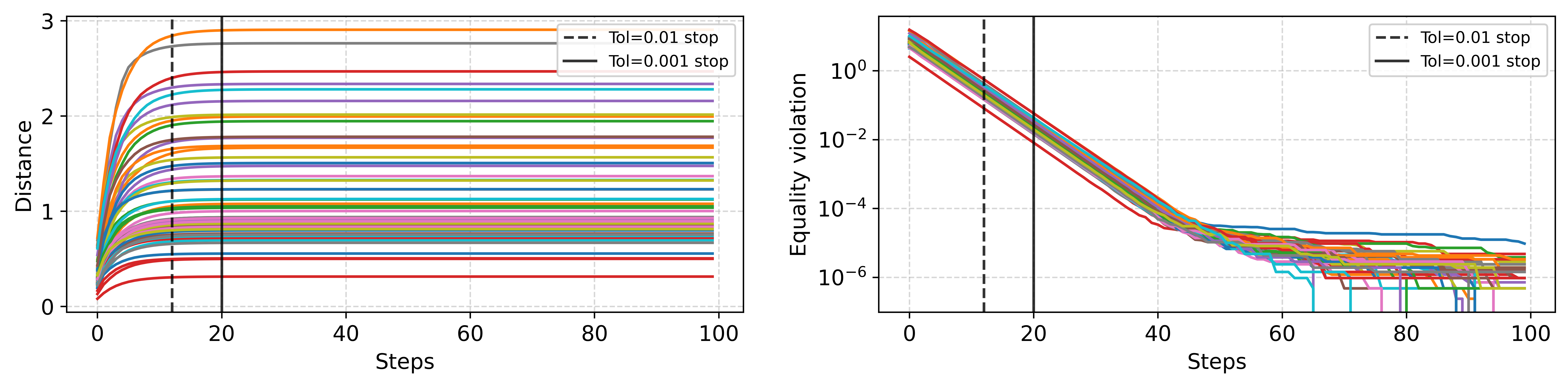} \\
    \includegraphics[width=\linewidth]{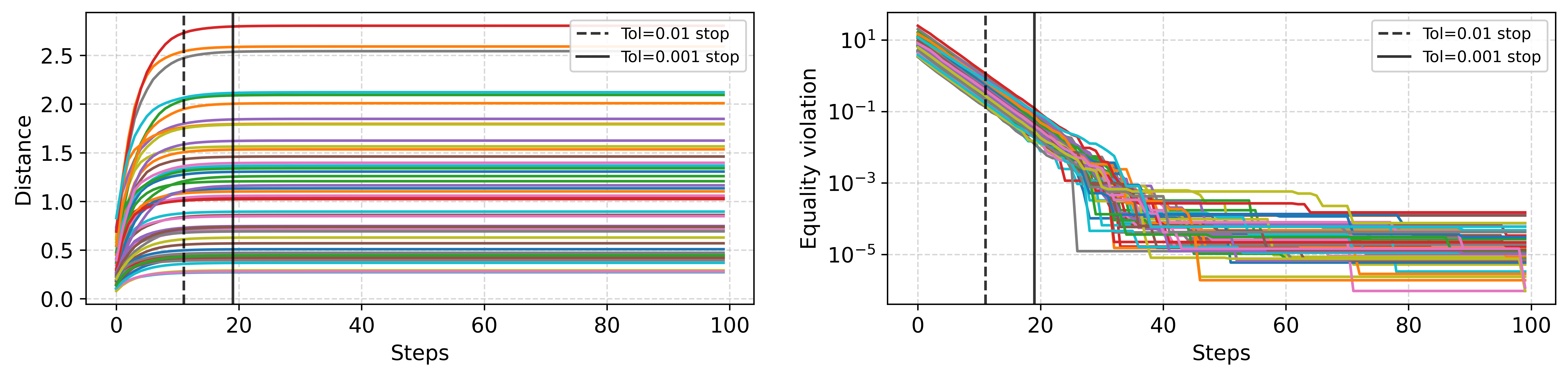}
    \caption[Distance and equality violation as a function of optimization steps for the modified DeepFool algorithm]{Distance (left) and equality violation (right, logarithmic scale) for a batch of $50$ samples as a function of the number of optimization steps for $2$ models (top and bottom) from PGDL Task $1$ using a learning rate of $\gamma=0.25$ when calculating input margins. Vertical lines indicate when the stopping criteria are met for a distance tolerance of $\delta=0.01$ (black line, dashed) and $\delta=0.001$ (black line, solid).}
    \label{fig:hidden_mean_margin_and_viol_comparison_single_batch}
\end{figure}

This is perhaps better illustrated by considering an alternative visualization. In Figure~\ref{fig:hidden_mean_margin_and_viol_comparison_single_batch} we show the distance and equality violation for a single batch ($50$) of samples during optimization as a function of the number of steps. We do this for two randomly selected models from Task 1, using $\gamma=0.25$. We also show the point at which the distance stopping criterion is met for both $\delta=0.01$ and $\delta=0.001$ as a vertical black line (dashed and solid, respectively).

One observes that this reaffirms the previous observation: after the stopping criterion is met for $\delta=0.01$, the equality violations keep decreasing; however, the distance for each sample is not noticeably affected past this point. Despite the higher equality violations of the points found using this method, we still consider these sufficiently close to the decision boundary. As such, we will refer to these points as points `on the decision boundary' when employing Algorithm~\ref{alg:deepfool_margin}.
%

\subsubsection{Additional verification}

We repeat a similar analysis to the one shown in Figure~\ref{fig:hidden_mean_margin_and_viol_comparison_small_and_large_tolerance} for $10$ models from Task $2$ and $4$ (separately) and find that the same trends hold. One caveat is that for the models in Tasks $4$ and $5$, the input data is normalized such that all features values are in the range of $[0, 1]$, whereas the other tasks are trained on $z$-normalized data. This implies that the margins for the models in these two tasks are an order of magnitude smaller in general due to the small scale and, as such, require a smaller stopping tolerance for the same comparative performance. See Appendix~\ref{app:hidden_additional_verification} for additional details and results on Tasks $2$ and $4$.

\subsubsection{Summary of hyperparameter effects}

In summary, we observe the following about the effects of the different hyperparameter choices of Algorithm~\ref{alg:deepfool_margin}. For the models considered for Task $1$:
\begin{itemize}
    \item Smaller learning rates ($\gamma$) result in greater precision. That is, smaller learning rates lead to finding closer points on the decision boundary. However, smaller learning rates also require more optimization steps.
    \item A very small distance tolerance ($\delta=0.001$) finds essentially the same points as when using no distance tolerance (in this case, optimizing for $100$ steps). For this tolerance, optimization also terminates in many fewer than $100$ optimization steps.
    \item A larger distance tolerance ($\delta=0.01$) combined with a small learning rate ($\gamma=0.25$) finds points with higher equality violations than when using a small distance tolerance ($\delta=0.001$). That said,  this is permissible for the purpose of model ranking, as the difference in distance is constant between models and also rather small. Furthermore, using a larger distance tolerance requires fewer optimization steps than using a small distance tolerance.
\end{itemize}

We find that the same trends hold for models in Task $2$. For models in Task $4$, we find that they require smaller distance tolerances for the same behavior. In the following section we compare Algorithm~\ref{alg:deepfool_margin} with the first-order Taylor approximation.

\subsection{Comparison to Taylor}
\label{sec:hidden_df_to_taylor_comp}

In this section, our aim is to determine how accurate the Taylor approximation is at estimating input and hidden margins by comparing them to the margins found using Algorithm~\ref{alg:deepfool_margin}. This allows us to not only determine how accurate the approximation is in general, but also whether a more accurate method would alter the ranking of models for the purpose of generalization prediction. While both methods are approximate in nature, it is important to note some key differences. Recall that the first-order Taylor approximation, as used here, is only an estimation of the \textit{margin} -- it does not provide one with a tangible point on the decision boundary. This implies that the equality violation of the estimated point is never checked to determine whether it is sufficiently close to the decision boundary. On the other hand, our modified DeepFool algorithm explicitly finds a point, and the equality violation is verified to determine whether the point is sufficiently close to the decision boundary. Furthermore, given that DeepFool utilizes the Taylor approximation to determine the direction in which the sample is stepped (during the search), one can conclude that DeepFool should result in a better approximation in all cases. 

As a first step, we compare the size of the mean margin for the $10$ models of Task $1$ and $4$ found with the two methods in question for both the input and hidden space. As we are mainly concerned with determining the accuracy of the Taylor approximation, we use the most accurate formulation of Algorithm~\ref{alg:deepfool_margin} previously employed. That is, we use a very small learning rate ($\gamma=0.25$) and no stopping tolerance, we then select the point with the smallest equality violation after $100$ optimization steps. Recall that this variant outperforms the constrained optimization baseline and is thus the most accurate method available (recall Figures \ref{fig:hidden_mean_margin_and_viol_lr_comparison_no_tol} and \ref{fig:hidden_mean_margin_and_steps_small_tol_comparison}).  We do this for $500$ training samples (as done previously) for each model in both cases. Note: We find that all points calculated using Algorithm~\ref{alg:deepfool_margin} have a maximum equality violation of $10^{-4}$. Therefore, we can say with certainty that these points are very close to the decision boundary.

\begin{figure}[h]
    \centering
    \includegraphics[width=0.49\linewidth]{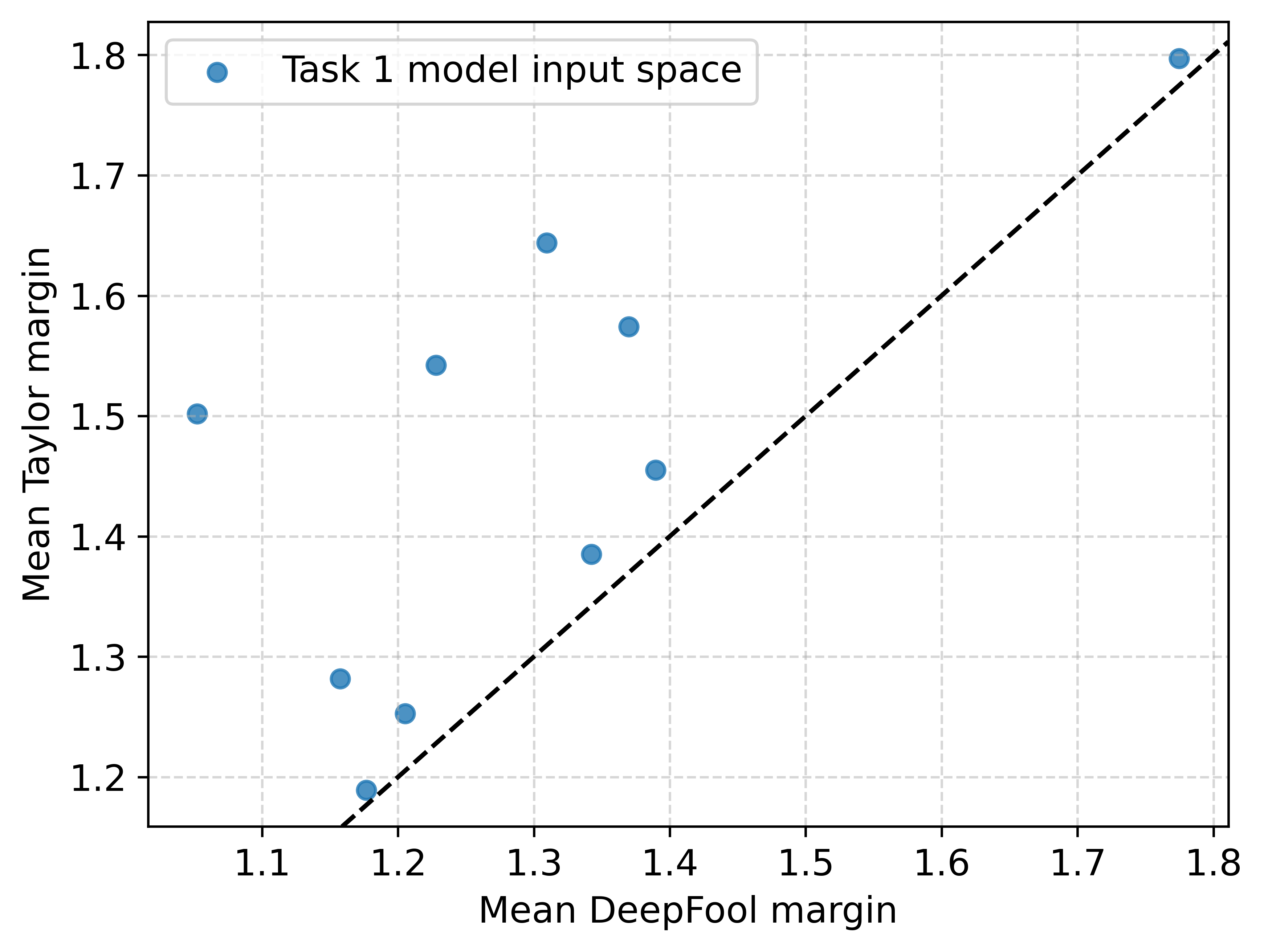}
    \includegraphics[width=0.49\linewidth]{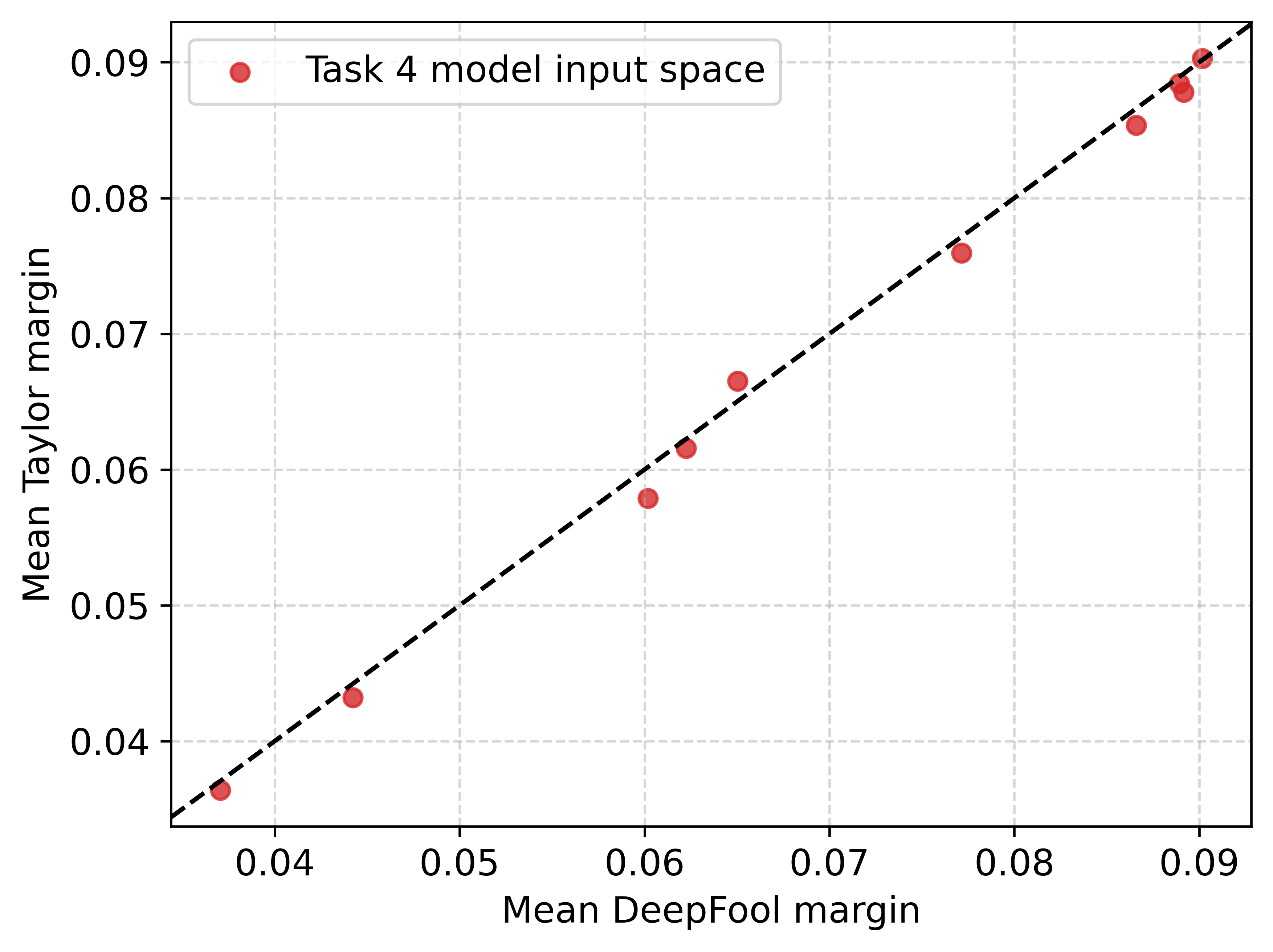}
    \includegraphics[width=0.49\linewidth]{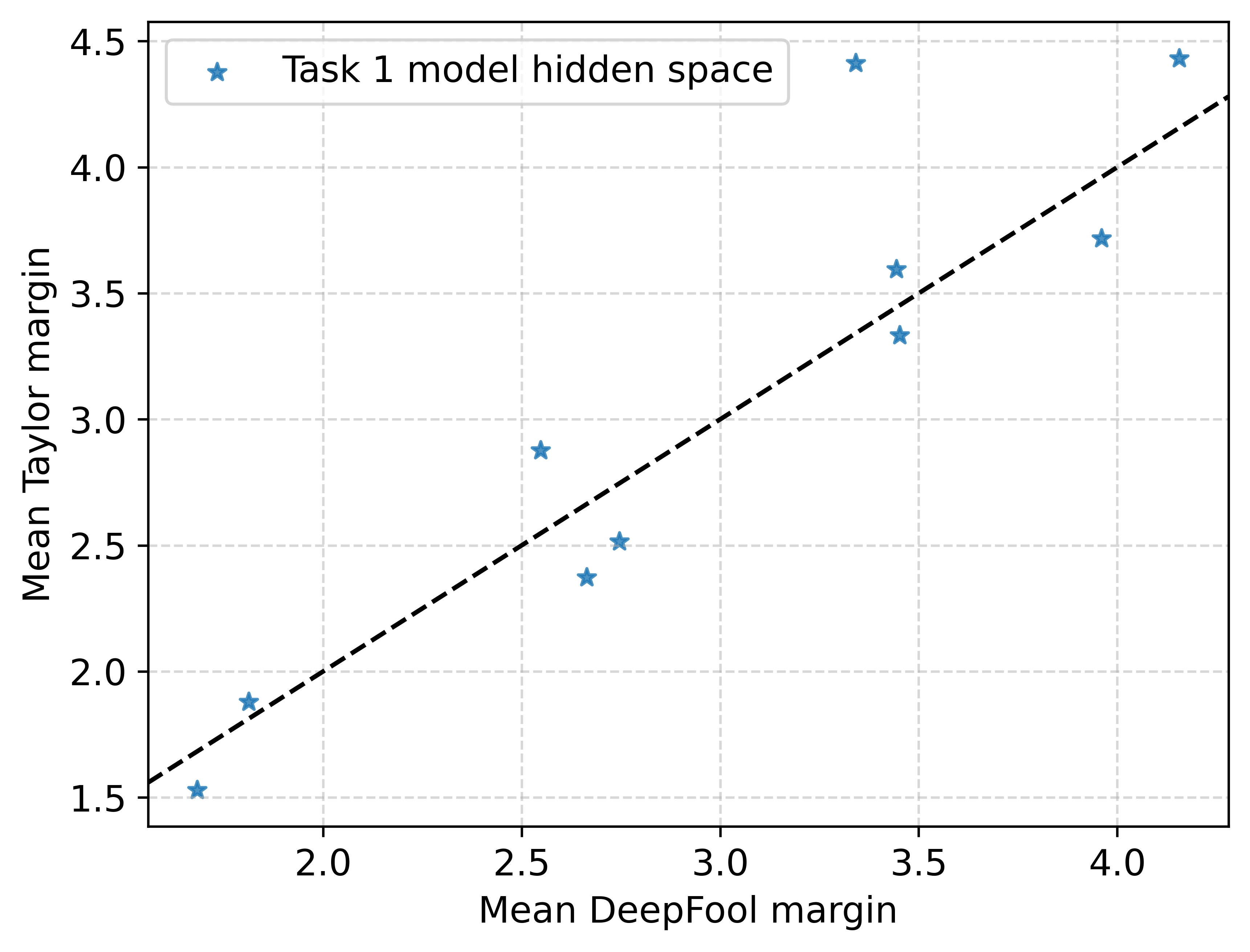}
    \includegraphics[width=0.49\linewidth]{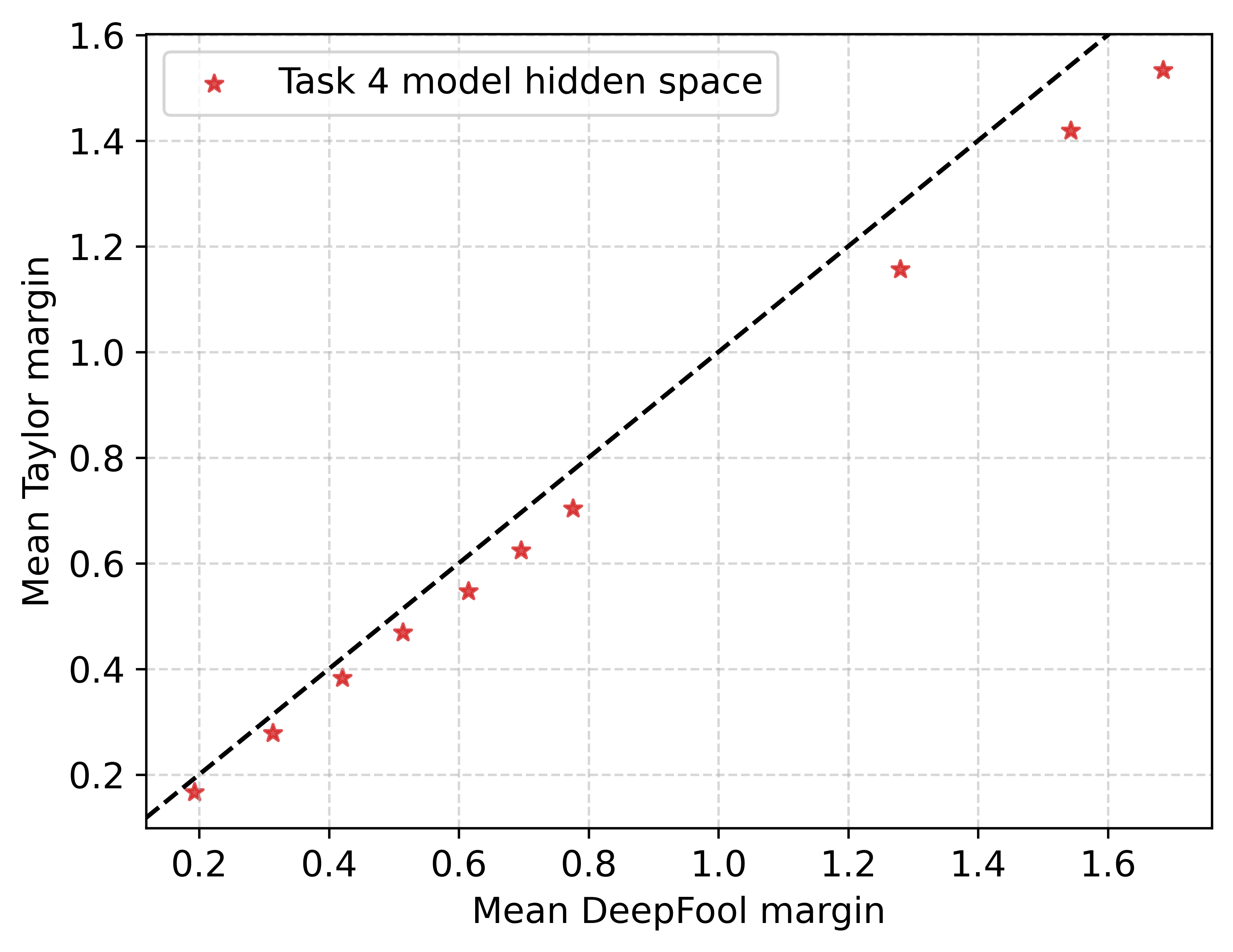}
    \caption[Mean Taylor versus DeepFool margin for $10$ models from Task $1$ and $4$]{Mean Taylor versus DeepFool margin for $10$ models of Task 1 (left) and 4 (right). Top: Input space. Bottom: First convolutional layer hidden space.}
    \label{fig:hidden_mean_margin_taylor_v_deepfool_input_and_hidden_task_1}
\end{figure}

Figure~\ref{fig:hidden_mean_margin_taylor_v_deepfool_input_and_hidden_task_1} shows the mean DeepFool and Taylor margin on the $x$-axis and $y$-axis, respectively, for both Task $1$ (left) and $4$ (right) for both input space (top) and hidden space (bottom, first convolutional layer). In addition, we show the identity line ($y=x$), which means that points above this line indicate a Taylor overestimation of the margin, and points below this line an underestimate, relative to the DeepFool margins.

One observes that there is a distinct difference in behaviour between the two tasks considered. For the input space, the Taylor approximation is an overestimate of the margin (above the identity line) for all $10$ models of Task $1$. On the other hand, the approximation appears to be highly accurate for Task $4$, and the Taylor approximation is only a slight over- or underestimate, in general. The hidden space results are somewhat similar: we observe that the Taylor approximation is less accurate for Task $1$ than $4$, and can severely over- or underestimate the margin, while Task $4$ appears to be more accurate with a more consistent slight underestimate.

\begin{figure}[h]
    \centering
    \includegraphics[width=0.49\linewidth]{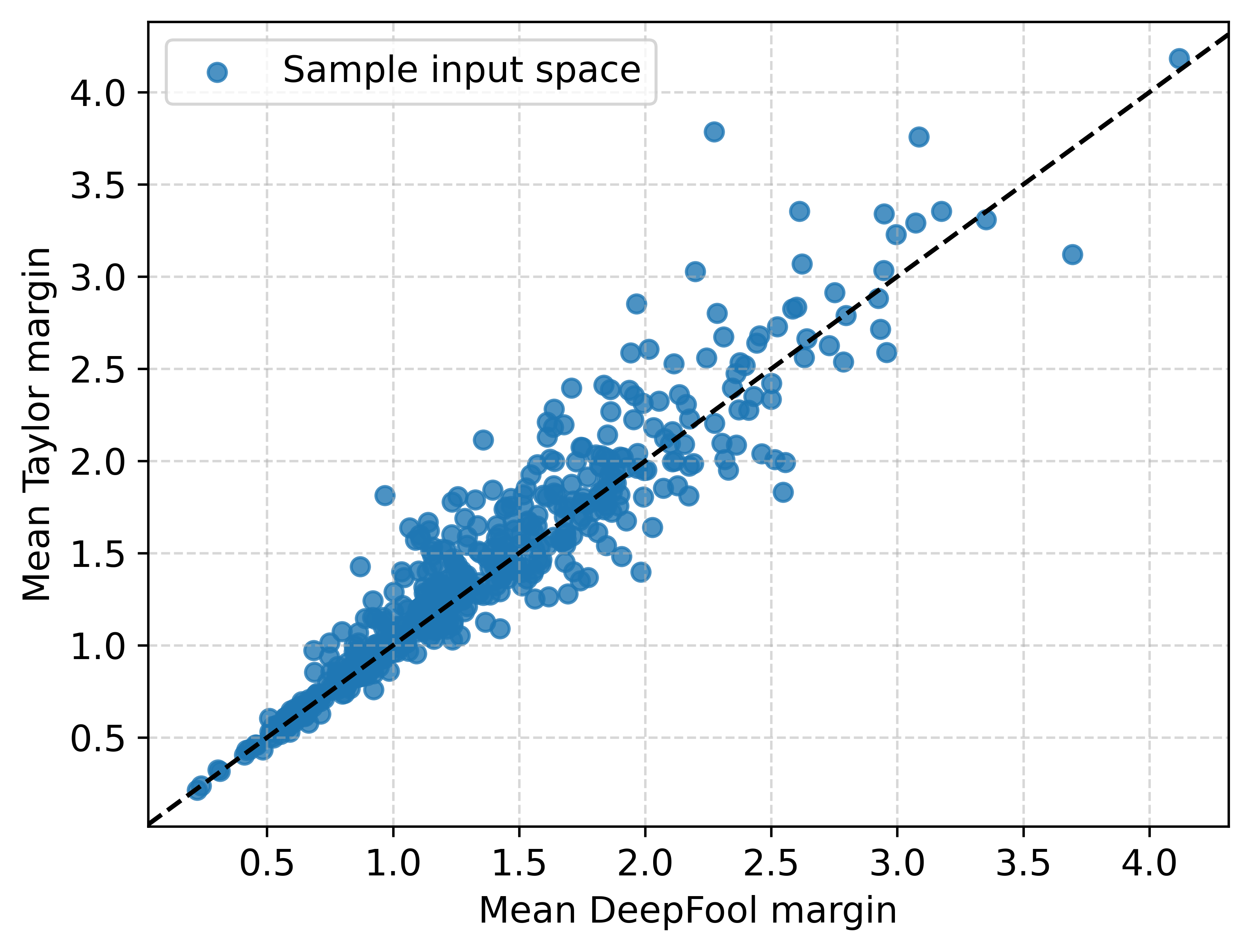}
    \includegraphics[width=0.49\linewidth]{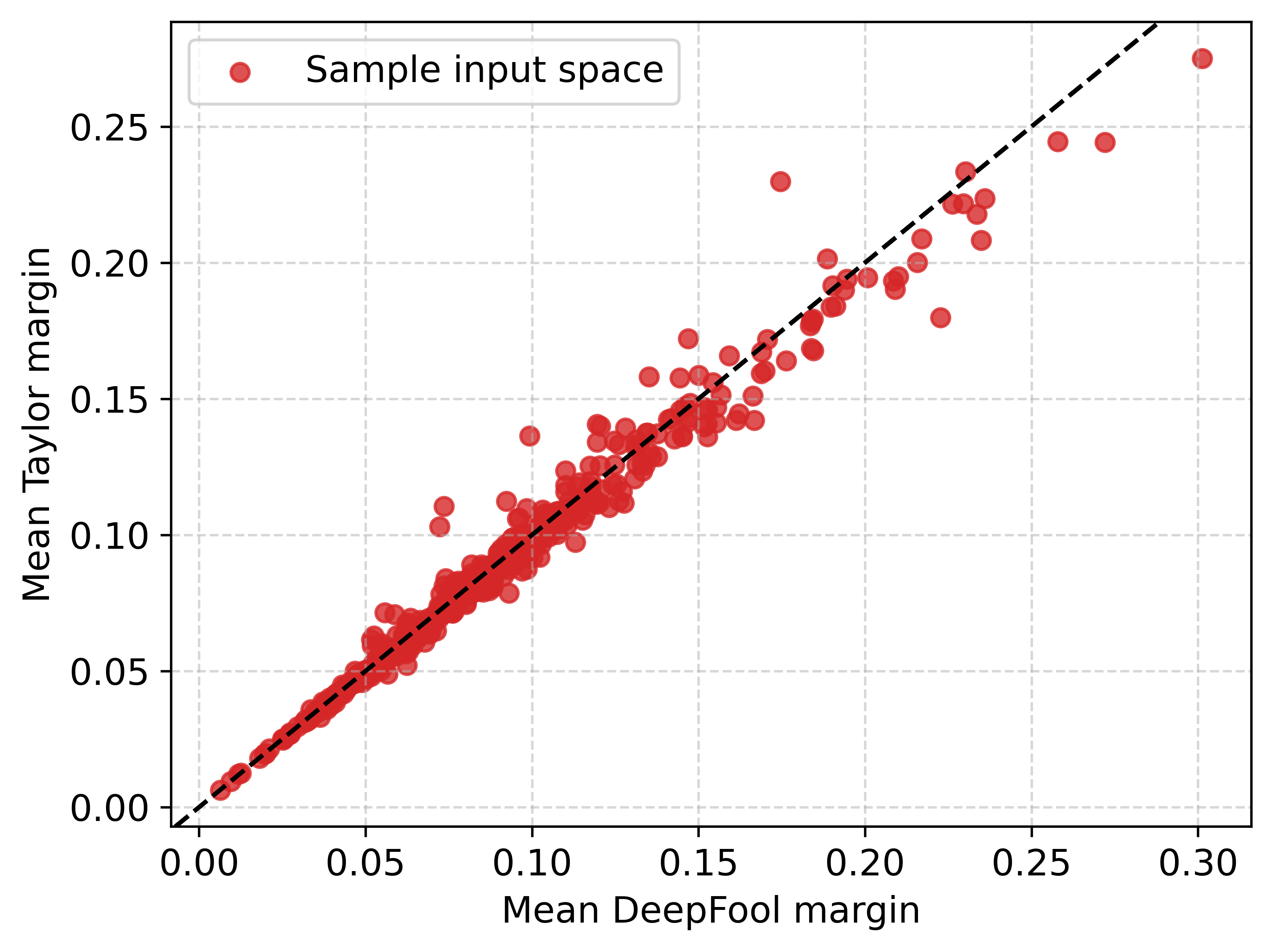}
    \includegraphics[width=0.49\linewidth]{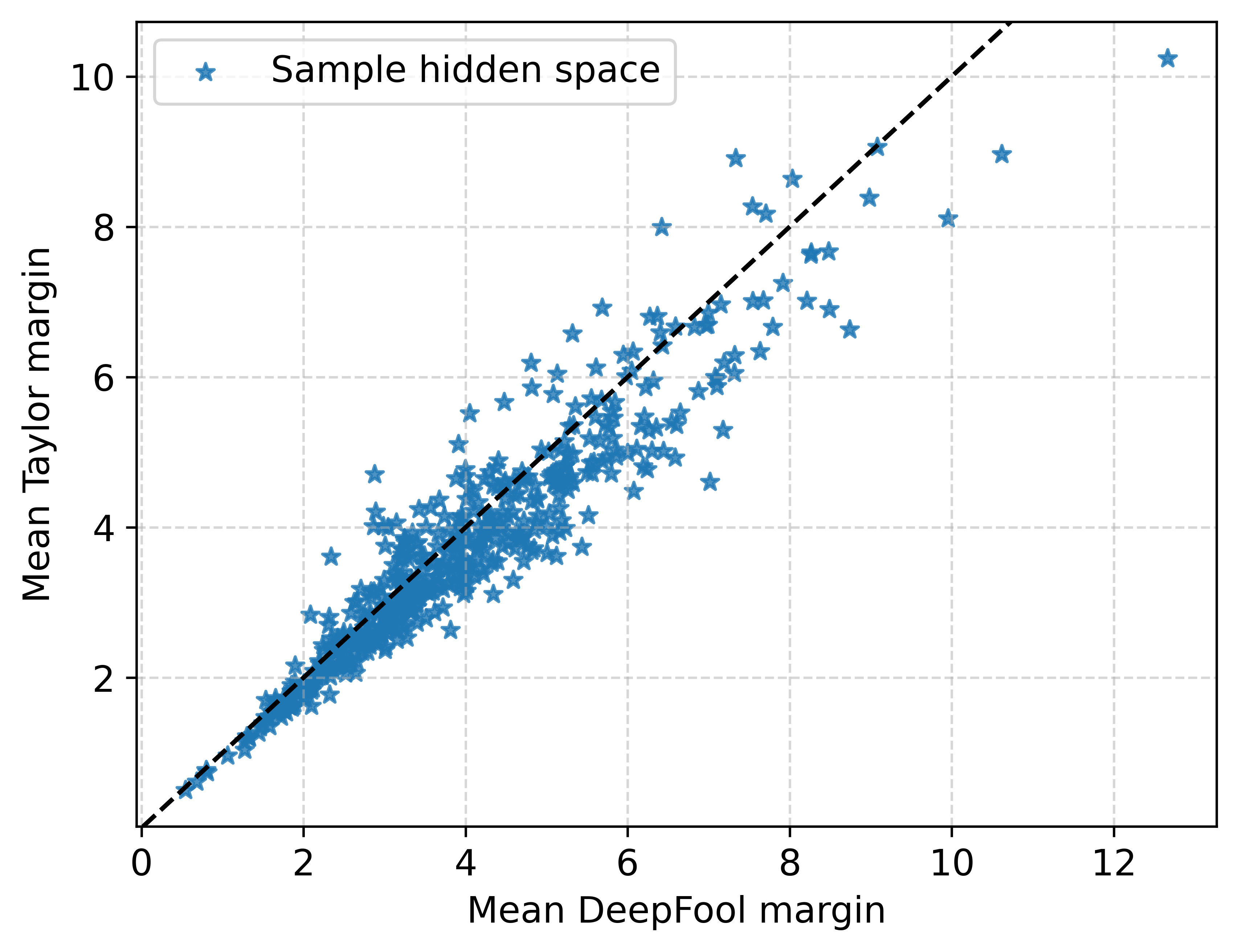}
    \includegraphics[width=0.49\linewidth]{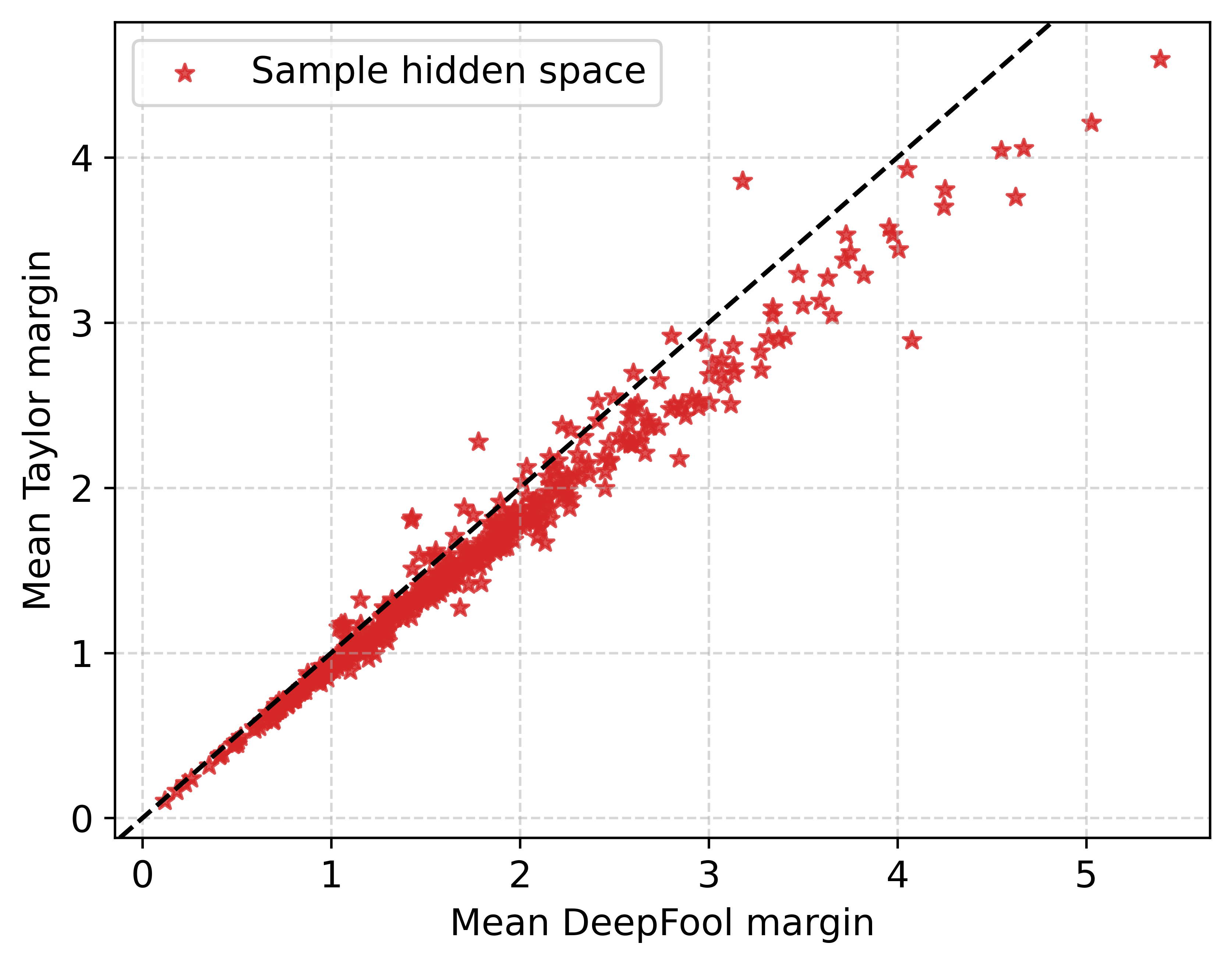}
    \caption[Taylor versus DeepFool margin (per sample) for a model from Task $1$ and $4$]{Taylor versus DeepFool margin for $500$ samples for a model from Task 1 (left) and 4 (right). Top: Input space. Bottom: First convolutional layer hidden space.}
    \label{fig:hidden_taylor_v_deepfool_margin_per_sample}
\end{figure}

Let us consider a more granulated view of this behaviour. In Figure~\ref{fig:hidden_taylor_v_deepfool_margin_per_sample} we show a per-sample comparison of the margins for a model from Task $1$ (left) and $4$ (right) with the largest input space margin from the $10$ models considered previously. We once again show both the input (top) and hidden (bottom) spaces.

One observes that in all cases the Taylor approximation is accurate for samples with very small (DeepFool) margins. Furthermore, one observes that as the DeepFool margin grows, the Taylor approximation tends to be less accurate. This makes sense: the further the `true' point on the decision boundary is from the linear region that contains the training sample, the less accurate one would expect the linear Taylor approximation to be. These results are somewhat contradictory to the findings of Yousefzadeh and O'Leary~\cite{constrained_optim_investigating}: in the case of MLPs, they also find that the Taylor approximation is only accurate for samples with very small margins. However, they find that the approximation is a consistent (and severe) underestimate of the margin for samples with larger margins. Conversely, we observe that in many cases the Taylor approximation can also be an overestimate.

From these results, we can conclude that the Taylor approximation is more accurate in some cases than in others and does not behave consistently. We find that it can be either an underestimate or an overestimate of the DeepFool margin. Furthermore, we find that this behavior differs between models and tasks. Therefore, one would expect that the more accurate DeepFool method would affect the final ranking of the models for the purpose of generalization prediction. 

To further verify these results, we repeat the same analysis using a larger number of samples and a small distance tolerance. We find that these results remain consistent. See Appendix~\ref{app:hidden_taylor_v_deepfool} for details.

\subsection{PGDL results}
\label{sec:hidden_df_pgdl_results}

We use Algorithm~\ref{alg:deepfool_margin} and calculate the DeepFool margin for both the input space and the first convolutional layer representations for all models of each task. We restrict our analysis to $5\ 000$ randomly sampled training samples for computational reasons, which should provide a decent estimate of the mean margin for the purpose of model ranking. See Appendix~\ref{app:hidden_num_samples} for an investigation of the effect of number of samples for both the Taylor and DeepFool estimation methods. 

For both the input and hidden space, we make use of a learning rate of both $\gamma=0.25$ and $\gamma=1.0$. This allows us to determine whether the more accurate margin approximations of the smaller learning rates perform better when predicting generalization. The stopping tolerance is set to $\delta=0.01$ throughout, with the exception of Task $4$ and $5$ which uses $\delta=0.001$ for the input space calculations (recall our explanation at the end of Section~\ref{sec:hidden_verifying_deepfool_alg}). It is important to note that for the hidden space, the optimization is much more computationally expensive. This is due to the large dimensionality of these representations, which implies a greater number of gradient calculations. All of these margin calculations combined take approximately $14$ days on a single Nvidia $A30$. Most of these (about $10$ days) are spent on the hidden margin calculations.

Table~\ref{tab:hidden_hidden_input_taylor_v_deepfool} shows a comparison of the Kendall's rank correlation 
between the mean margin and generalization for each task using the Taylor and DeepFool methods, for both input and hidden margins. Similarly, Table~\ref{tab:hidden_hidden_input_taylor_v_deepfool_cmi} shows the corresponding CMI values. Note that we restrict the Taylor analysis to the same $5\ 000$ training samples to minimize variability between results. (See Appendix~\ref{app:hidden_num_samples} for additional details on what effect the number of samples has on the Taylor-approximated margins.) 

\begin{table}[H]
\centering
\caption[Comparison of Taylor and DeepFool calculated input and hidden margins on the PGDL tasks (Kendall's)]{Kendall's rank correlation between mean margin and test accuracy using the Taylor approximation and DeepFool algorithm in the input and hidden space for all PGDL tasks.}
\label{tab:hidden_hidden_input_taylor_v_deepfool}
\begin{tabular}{@{}cccc|ccc@{}}
\toprule
\multirow{2}{*}{\textbf{Task}} & \multicolumn{3}{c|}{\textbf{Input}} & \multicolumn{3}{c}{\textbf{Hidden}} \\ \cmidrule(l){2-7} 
 & Taylor & \begin{tabular}[c]{@{}c@{}}DeepFool \\ $\gamma = 1.00$\end{tabular} & \begin{tabular}[c]{@{}c@{}}DeepFool \\ $\gamma = 0.25$\end{tabular} & Taylor & \begin{tabular}[c]{@{}c@{}}DeepFool\\ $\gamma = 1.00$\end{tabular} & \begin{tabular}[c]{@{}c@{}}DeepFool\\ $\gamma = 0.25$\end{tabular} \\ \midrule
1 & 0.0265 & \textbf{0.0322} & -0.1643 & 0.5794 & \textbf{0.6412} & 0.6281 \\
2 & 0.6841 & \textbf{0.6883} & 0.6576 & \textbf{0.7037} & 0.6883 & 0.6771 \\
4 & 0.6251 & \textbf{0.6462}& 0.6436& 0.7958 & 0.7949 & \textbf{0.7993} \\
5 & 0.3571& \textbf{0.4544}& 0.4355& \textbf{0.5427} & 0.5278 & 0.5308 \\
6 & \textbf{-0.1351} & -0.1641 & -0.2356 & 0.4427 & 0.4814 & \textbf{0.5029} \\
7 & 0.3215 & \textbf{0.3606} & 0.2913 & 0.3623 & \textbf{0.4139} & 0.3819 \\
8 & \textbf{-0.1233} & -0.1322 & -0.1938 & -0.0656 & -0.0209 & \textbf{-0.0159} \\
9 & 0.1573 & \textbf{0.1613} & 0.0685 & \textbf{0.7097} & \textbf{0.7097} & 0.6935 \\ \midrule
Average & 0.2392& \textbf{0.2558}& 0.1879& 0.5088 & \textbf{0.5300} & 0.5247 \\ \bottomrule
\end{tabular}
\end{table}

\begin{table}[H]
\centering
\caption[Comparison of Taylor and DeepFool calculated input and hidden margins on the PGDL tasks (CMI)]{Conditional Mutual Information (CMI) between mean margin and generalization gap using the Taylor approximation and DeepFool algorithm in the input and hidden space for all PGDL tasks.}
\label{tab:hidden_hidden_input_taylor_v_deepfool_cmi}
\begin{tabular}{@{}cccc|ccc@{}}
\toprule
\multirow{2}{*}{\textbf{Task}} & \multicolumn{3}{c|}{\textbf{Input}} & \multicolumn{3}{c}{\textbf{Hidden}} \\ \cmidrule(l){2-7} 
 & Taylor & \begin{tabular}[c]{@{}c@{}}DeepFool \\ $\gamma = 1.00$\end{tabular} & \begin{tabular}[c]{@{}c@{}}DeepFool \\ $\gamma = 0.25$\end{tabular} & Taylor & \begin{tabular}[c]{@{}c@{}}DeepFool\\ $\gamma = 1.00$\end{tabular} & \begin{tabular}[c]{@{}c@{}}DeepFool\\ $\gamma = 0.25$\end{tabular} \\ \midrule
1 & 00.07 & 00.12 & \textbf{01.76} & 09.40 & \textbf{09.60} & 08.41 \\
2 & 06.12 & 05.39 & \textbf{06.34} & \textbf{37.74} & 36.07 & 34.05 \\
4 & 14.95 & \textbf{17.79}& 17.40& 34.73 & 34.35 & \textbf{35.23} \\
5 & 08.46 & \textbf{13.71}& 13.16& \textbf{19.11} & 17.32 & 17.90 \\
6 & 00.57 & 00.90 & \textbf{02.14} & 04.24 & 05.25 & \textbf{06.23} \\
7 & 01.47 & 01.05 & \textbf{02.51} & \textbf{05.04} & 03.14 & 01.84 \\
8 & 00.70 & \textbf{00.77} & 00.58 & \textbf{00.36} & 00.26 & 00.25 \\
9 & 00.29 & 00.15 & \textbf{01.67} & 23.74 & \textbf{27.27} & 24.50 \\ \midrule
Average & 04.08 & 04.99& \textbf{05.70}& \textbf{16.80} & 16.66 & 16.05 \\ \bottomrule
\end{tabular}
\end{table}

One would expect a more accurate margin estimation method to produce better generalization prediction results. Specifically, for these three estimation methods, one would expect the ordering of the results from worst to best to be as follows: Taylor, DeepFool with $\gamma=1.0$, DeepFool with $\gamma=0.25$.  However, given Tables \ref{tab:hidden_hidden_input_taylor_v_deepfool} and \ref{tab:hidden_hidden_input_taylor_v_deepfool_cmi}, we make two important deductions: 

\begin{enumerate}
    \item Using a margin-finding method that is less of an approximation, in the sense that it actually finds a point on the decision boundary, does not necessarily lead to better generalization prediction.
    \item Refining the more exact margin finding method to be more accurate, in the sense that it finds closer boundary points, does not necessarily lead to a better predictor of the generalization performance.
\end{enumerate}

In support of the first deduction, observe that neither of the two DeepFool methods greatly improves the predictive performance over the Taylor-approximated margins. For any task, using input or hidden margins, the maximum improvement we found is $0.0973$ Kendall's rank correlation at task $5$ (DeepFool with $\gamma = 1.00$ as measured on input margins). Our maximum improvement of $5.25$ with respect to CMI was also found for this method and task. 

In support of the second deduction, note that the more accurate DeepFool method ($\gamma = 0.25$) does not perform much better than when using a larger learning rate. The maximum improvement by using $\gamma = 0.25$ as opposed to $\gamma = 1.00$ is $0.0215$ Kendall's rank on Task $6$ using hidden margins and $1.64$ CMI on Task $1$ using input margins.

These improvements are relatively small, even though we highlight the best instances (above). This leads to a very interesting conclusion -- the predictive power of these margin-based complexity measures is not strongly linked to how accurately the margin is estimated. This strongly suggests that the failure cases we have observed throughout this chapter are not primarily the result of measurement noise.

\section{Discussion and conclusion}

Throughout this chapter, we have analyzed input and hidden margins for the purpose of generalization prediction on the PGDL dataset. Here we summarize our main observations and findings.

\begin{enumerate}
    \item When a training set of models with known accuracies for a specific task is available, a linear model trained using hidden margin summary statistics can perform well in numerically predicting generalization on a held-out set of models ($R^2$ of between $0.634$ and $0.927$ for six tasks). However, this technique fails in other cases ($R^2$ of $0.300$ and $0.381$ for the remaining two tasks). In contrast, input margins used in a similar setting perform poorly for most tasks ($R^2$ between $-0.251$ and $0.512$ for six tasks) with moderate success in limited cases ($R^2$ of $0.787$ and $0.864$ for the remaining two tasks). See Table~\ref{tab:hidden_r_squared_two_signatures} in Section~\ref{sec:hidden_lin_reg}. This implies that in this setting 1) input margins are not generally predictive of generalization, and 2) there are certain cases where hidden margins are not predictive of generalization either. 
    \item Input margins show a similar trend to that above when used as a standalone complexity measure for the purpose of generalization ranking (no training set of models or linear model). We find low Kendall's rank correlation values of between $-0.1351$ and $0.3958$ for six tasks and moderately better values of $0.62566$ and  $0.6841$ for the remaining two tasks. The average Kendall's rank correlation over all tasks is only $0.2439$. The same trends hold for the CMI metric: between $0.06$ and $6.09$ for six tasks, and better values of $10.54$ and $15.05$ for the remaining two tasks. The average CMI over all tasks is $4.38$. See Tables \ref{tab:hidden_layer_comparison_mean_kendall} and \ref{tab:hidden_layer_comparison_mean_cmi} in Section~\ref{sec:hidden_hidden_layer_selection}. The main conclusion of this and the previous observation is that input margins are not a reliable or robust complexity measure: there does not seem to be a strong link between large input margins and generalization.
    \item In the same setting as above, the choice of which hidden layers to select when considering hidden margins has a very large effect on performance. The average Kendall's rank correlation over all tasks is between $0.3507$ and $0.5077$ for the four different hidden layer selection methods. The corresponding CMI values range from $7.07$ to $16.66$. Furthermore, for the tasks considered here, using only the mean hidden margin found at the first layer performs best overall (with an average Kendall's rank correlation of $0.5077$ and CMI of $16.66$). However, for this layer selection method, the values per task vary greatly (between $-0.0616$ and $0.7975$ for Kendall's rank correlation and between $00.34$ and $37.14$ for CMI). Still, all hidden layer selection methods outperform input margins for both evaluation metrics on average. See Tables \ref{tab:hidden_layer_comparison_mean_kendall} and \ref{tab:hidden_layer_comparison_mean_cmi} in Section~\ref{sec:hidden_hidden_layer_selection}. The primary implication of these observations is that hidden margins are brittle as a complexity measure. These measurements rely on an almost arbitrary hidden layer selection, which has a very large effect on prediction performance, and one cannot determine beforehand which selection method will work best for a given task.
    \item When calculating granulated Kendall's coefficient per hyperparameter variation for each task, we find that each margin-based measure fails/succeeds on different hyperparameters for each task. This indicates that the performance of these metrics (when predicting generalization) is more tied to the architectural family and dataset of the models considered than any specific hyperparameter variations. See Tables \ref{tab:hidden_hidden_granulated} and \ref{tab:hidden_input_granulated} in Section~\ref{sec:hidden_failure_cases}.
    \item Our modified DeepFool algorithm, Algorithm~\ref{alg:deepfool_margin}, is able to find smaller margins than the constrained optimization baseline, when using an appropriate learning rate $\gamma < 1$. Furthermore, the accuracy of Algorithm~\ref{alg:deepfool_margin} (the ability to find smaller margins) increases as the learning rate decreases. However, smaller learning rates also require a larger number of optimization steps and are therefore more computationally expensive. See Figures \ref{fig:hidden_mean_margin_and_viol_lr_comparison_no_tol}, \ref{fig:hidden_mean_margin_and_steps_small_tol_comparison}, and \ref{fig:hidden_mean_margin_and_viol_comparison_small_and_large_tolerance} in Section~\ref{sec:hidden_verifying_deepfool_alg}. These observations allow us to select an appropriate margin-finding protocol for the analysis that follows.
    \item When comparing the margins found with the first-order Taylor approximation and Algorithm~\ref{alg:deepfool_margin} in the input and hidden space, we find that the Taylor approximation is only accurate in estimating margins for samples with very small (DeepFool) margins. Furthermore, it tends to be less reliable as the (DeepFool) margin increases. See Figures \ref{fig:hidden_mean_margin_taylor_v_deepfool_input_and_hidden_task_1} and \ref{fig:hidden_taylor_v_deepfool_margin_per_sample} in Section~\ref{sec:hidden_df_to_taylor_comp}. This implies that the inaccuracy of the Taylor approximation can influence model ranking when margins are used as a complexity measure. 
    \item Comparing the generalization prediction performance of margins calculated using the two methods above, we find that the more accurate method does not greatly influence performance on the PGDL tasks. We find a maximum improvement for input margins of $0.0973$ Kendall's rank correlation and $5.25$ CMI. For hidden margins, the largest improvement is $0.0618$ Kendall's rank correlation and $3.53$ CMI. See Tables \ref{tab:hidden_hidden_input_taylor_v_deepfool} and \ref{tab:hidden_hidden_input_taylor_v_deepfool_cmi} in Section~\ref{sec:hidden_df_pgdl_results}. This implies that the reason that input or hidden margins fail in certain cases is not due to poor approximations of the margin.
\end{enumerate}

We conclude that input margins often fail as generalization predictors where hidden margins succeed; however, hidden margins require more effort to calibrate correctly. Furthermore, the failure of these methods is not due to `measurement noise'. That is, they do not fail due to poor approximations of the true shortest distance to the decision boundary.

Informed by these findings, in the next chapter we endeavor to improve input margins so that they can compete with (and exceed) the predictive performance of hidden margins while maintaining their insensitivity to the hyperparameter choices that plague the hidden margin method (e.g. choice of hidden layers, normalization, and computational cost).

%% file: chapters/ch5_constrained.tex
\let\cite\parencite

\lhead{}
\rhead{}
\chapter{Input margins can predict generalization too}
\label{chap:ch5}
\afterpage{\lhead{\ifthenelse{\thepage>0}
       {\it Chapter \thechapter }
      }
\rhead{\ifthenelse{\thepage>0}
       {\it \let\uppercase\relax\leftmark}
      }}
\underline{ \hspace{\textwidth} } 
\textit{``There was 40 feet between them when they stopped to make their play \\
And the swiftness of the ranger is still talked about today \\
Texas Red had not cleared leather 'fore a bullet fairly ripped \\
And the ranger's aim was deadly with the big iron on his hip''}  \\ - Marty Robbins, \textit{Big Iron}, Verse 7\\
\underline{ \hspace{\textwidth} }

\section{Introduction}

In the previous chapter, we conducted an extensive investigation of input and hidden margins for the purpose of generalization prediction on the PGDL dataset. We identify key limitations of these complexity measures: On the one hand, margins measured at the hidden representations appear to be predictive in some cases, although performance can vary greatly across tasks depending on the selection of hidden layer. Furthermore, these measurements require additional normalization due to differences in scale and dimensionality between different networks. However, despite such normalization, we find that it is still difficult to compare networks with different architectural setups (e.g. variations in depth, layer width, and dimensionality) and the measure remains inaccurate in many cases. In contrast, input margins have no such concerns and require no normalization, as all models share the same input space. However, input margins are simply not sufficiently predictive of generalization in the general case.

In this chapter, we address these issues by developing a new margin-based complexity measure, which we refer to as \textit{constrained margins}. This metric is also based on the input space and, therefore, enjoys all the benefits that entails, but will be shown to be more predictive of generalization in general. Note that since our proposed margin metric is also based on the input space, we shall henceforth refer to input margins as we have used them previously as `standard input margins' in order to prevent confusion. Furthermore, the name `constrained margins' should not be confused with the margins calculated using constrained optimization in Chapter~\ref{chap:ch3} -- here we propose a completely different measurement.

The chapter is structured as follows: we first describe our intuition and theoretical approach in Section~\ref{sec:constrained_theoretical_approach}. Following this, in Section~\ref{sec:constrained_results}, we compare our metric with standard input and hidden margins, as well as with other complexity measures. In Section~\ref{sec:constrained_analysis} we analyze the new metric in more detail. Finally, in Section~\ref{sec:constrained_adv_dirs}, informed by our new complexity measure, we investigate adversarial examples.

\section{Theoretical approach}
\label{sec:constrained_theoretical_approach}

This section provides a theoretical overview of the proposed complexity measure. We first explain our intuition surrounding classification margins (Section~\ref{sec:constrained_intuition}), before mathematically formulating constrained margins (Section~\ref{sec:constrained_constrained_margin_formulation}), and then finally describing an approximation of this formulation (Section~\ref{sec:constrained_approximation_formulation}).

\subsection{Intuition}
\label{sec:constrained_intuition}

Let us recall the intuition behind large margins: a correctly classified training sample with a large margin can have more varied feature values, potentially due to noise, and still be correctly classified. However, as we have shown throughout Chapter~\ref{chap:ch4}, standard input margins are generally not predictive of generalization. This observation is supported by the literature on adversarial robustness, where it has been shown that adversarial retraining (which increases standard input margins) can negatively affect generalization~\cite{robustness_tradeoff,robustness_tradeoff_three} (recall Section~\ref{sec:back_input_margins}). 

Stutz et al.~\cite{stutz2019disentangling} provide a plausible reason for this counter-intuitive observation: Through the use of Variational Autoencoder GANs (VAE-GANS) they show that the majority of adversarial examples leave the class-specific data manifold of the attacked sample. They offer an intuitive example of black border pixels in the case of MNIST images, which are zero for all training samples. Samples found on the decision boundary which manipulate these border pixels have a zero probability under the data distribution and do not lie on the underlying manifold. 

We leverage this intuition and argue that any input margin measure that relates to generalization should measure distances along directions that do not rely on spurious features in the input space. The intuition is that, while nearby decision boundaries exist for virtually any given training sample, these nearby decision boundaries are likely in directions which are not inherently useful for test set classification, i.e. they diverge from the underlying data manifold.

More specifically, we argue that margins should be measured in directions of `high utility', that is, directions that are expected to be useful for characterizing a given dataset, while ignoring those of lower utility. In our case, we approximate these directions by defining high-utility directions as directions which explain a large amount of variance in the data. We extract these using Principal Component Analysis (PCA). While typically used as a dimensionality reduction technique, PCA can be interpreted as learning a locally linear low-dimensional approximation of the underlying data  manifold~\cite{hinton1997modeling}. In this way, the PCA manifold identifies subspaces that are believed to contain the latent variables that are truly relevant to the underlying data distribution, which the out-of-sample data is assumed to also be generated from.
In the following section, we formalize such a measure.

\subsection{Constrained margins}
\label{sec:constrained_constrained_margin_formulation}

Let us recall the classical definition of an input margin (as expressed earlier in Section~\ref{sec:noise_margin_formulation}), before adapting it for our purpose. 

Let $f~:~X\rightarrow \mathbb{R}^{|N|}$ denote a classification model with a set of output classes $N = \{1\ldots n\}$, and $f_k(\mathbf{x})$ the output logit of the model for input sample $\mathbf{x}$ and class $k$.  

For a correctly classified input sample $\mathbf{x}$, the objective is to find the closest point $\mathbf{\hat{x}}$ on the decision boundary between the true class $i$ (where $i = \argmax_k (f_k(\mathbf{x}))$) and another class $j \neq i$.
Formally, $\mathbf{\hat{x}}$ is found by solving the constrained minimization problem:
\begin{equation}
\label{eq:constrained_objective_func}
\argmin_{\mathbf{\hat{x}}} ||\mathbf{x} - \mathbf{\hat{x}}||_{2}
\end{equation}
such that
\begin{equation}
\label{eq:constrained_eq_constraint}
    f_i(\mathbf{\hat{x}}) = f_j(\mathbf{\hat{x}}) 
\end{equation}
\begin{equation}
\label{eq:constrained_bound_constraints}
    \mathbf{\hat{x}} \in [l, u]^{\dim(\hat{\mathbf{x}})}
\end{equation}
with $l$ and $u$ the lower and upper bounds of $\mathbf{\hat{x}}$, respectively, and $i$ and $j$ as above.

The margin is then given by the Euclidean distance between the input sample, $\mathbf{x}$, and its corresponding sample on the decision boundary, $\mathbf{\hat{x}}$.
We now adapt this definition to define a `constrained margin'.  
Let the set $P = \{\mathbf{p}_1, \mathbf{p}_2, ..., \mathbf{p}_m\}$ denote the first $m$ principal component vectors of the training dataset, that is, the $m$ orthogonal principal components which explain the most variance.  
Such principal components are straightforward to extract by calculating the eigenvectors of the empirical covariance matrix of the normalized training data.  The data is normalized in the same way as prior to model training, but additionally scaled such that each feature has a mean of $0$ where necessary.

We now restrict $\mathbf{\hat{x}}$ to any point consisting of the original sample $\mathbf{x}$ plus a linear combination of these (unit length) principal component vectors, that is, for some coefficient vector $\mathbf{B} = [\beta_1, \beta_2, ..., \beta_m]$
\begin{equation}
\label{eq:constrained_hat_def}
    \mathbf{\hat{x}} \triangleq \mathbf{x} + \sum_{i=1}^{m} \beta_{i}\mathbf{p}_{i}
\end{equation}
Substituting $\mathbf{\hat{x}}$ into the original objective function of Equation~\ref{eq:constrained_objective_func}, the new objective becomes
\begin{equation}
\label{eq:constrained_final_objective}
      \min_{\mathbf{B}} || \sum_{i=1}^{m} \beta_{i}\mathbf{p}_{i} ||_{2}
\end{equation}
such that Equation~\ref{eq:constrained_eq_constraint} is approximated within a certain tolerance and Equation~\ref{eq:constrained_bound_constraints} holds.
For this definition of margin, the search space is constrained to a lower-dimensional subspace spanned by the principal components with point $\mathbf{x}$ as origin, and the optimization problem then simplifies to finding a point on the decision boundary within this subspace. This brings us to the name of the method: constrained margins.
In restricting the search space, we ensure that boundary samples that rely on spurious features (that is, in directions of low utility) are not considered viable solutions to Equation~\ref{eq:constrained_objective_func}. Note that this formulation does not take any class labels into account for identifying high-utility directions.

\begin{figure}[h]
        \centering
        \includegraphics[width=\linewidth]{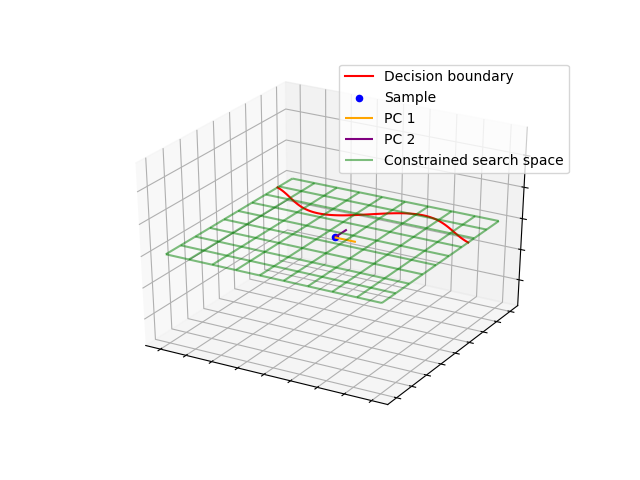} 
        \caption[Toy illustration of a constrained margin search space]{Toy illustration of a constrained margin search space in three dimensions with two principal components (PC 1 and PC 2).}
        \label{fig:constrained_visualization_margin}
\end{figure}

In Figure~\ref{fig:constrained_visualization_margin} we show a toy illustration of the constrained margin search space. This figure depicts a sample within a three dimensional input space, and a two dimensional subspace spanned by the top principal components that is searched within, with the sample as origin. Any points on the decision boundary not on the depicted plane would not factor into the margin search. 

\subsection{Approximating constrained margins}
\label{sec:constrained_approximation_formulation}

While it is possible to solve the CMP expressed in Equation~\ref{eq:constrained_final_objective} using a constrained optimizer, this can be prohibitively expensive for a large number of models and samples. 

Therefore, in line with our investigation of standard input and hidden margin in Chapter~\ref{chap:ch4}, we approximate the solution by adapting the previously mentioned first-order Taylor approximation, which greatly reduces the computational cost.  Consider a sample $\mathbf{x}$ classified as class $i$ by a neural network $f$. The Taylor approximation of the constrained margin $d_{i,j}(\mathbf{x})$ between classes $i$ and $j$ when using an $L_2$ norm is given by
\begin{equation}
\label{eq:constrained_meaningful_taylor_approx}
  d_{f,(i,j)}{(\mathbf{x})} = \frac{f_i(\mathbf{x}) - f_j(\mathbf{x})}
  {||~[~\nabla_{\mathbf{x}} f_i(\mathbf{x}) - \nabla_{\mathbf{x}} f_j(\mathbf{x})~]~\mathbf{P}_m^T||_2}
\end{equation}
where $\mathbf{P}_m$ is the $m \times n$ matrix formed by the top $m$ principal components with $n$ input features and $i$ is the highest predicted class. The derivation of Equation~\ref{eq:constrained_meaningful_taylor_approx} is included in Appendix~\ref{app:constrained_derivation}.

While the Taylor approximation provides a relatively inexpensive estimate of the margin, we have shown earlier that the accuracy of this estimate can vary. Although the estimation accuracy of standard input margins did not affect the predictive performance much, we cannot make this assumption w.r.t. constrained margins, and therefore consider both Taylor and DeepFool. Therefore, we also adapt our modified DeepFool algorithm (recall Algorithm~\ref{alg:deepfool_margin}) to measure constrained margins. The DeepFool constrained margin calculation is described by Algorithm~\ref{alg:manifol_margin}.

\begin{algorithm}[H]
    \caption{DeepFool constrained margin calculation}
    \label{alg:manifol_margin}
    \textbf{Input}: Sample $\mathbf{x}$, classifier $f$, principal components $\mathbf{P}_m$\\
    \textbf{Parameter}: Stopping tolerance $\delta$, Learning rate $\gamma$, Maximum iterations $max$\\
    \textbf{Output}: Distance~$d_{best}$, Equality violation~$v_{best}$ 
    
    \begin{algorithmic}[1] 
        \STATE $\mathbf{\hat{x}} \leftarrow \mathbf{x},
        i \leftarrow \argmax f_k(\mathbf{x}), d \leftarrow 0
        , v_{best} \leftarrow \infty, c \leftarrow 0$
        
        \WHILE{$c < max$}
        \FOR{$j \neq i$}
        \STATE $o_j \leftarrow f_i(\mathbf{\hat{x}}) - f_j(\mathbf{\hat{x}}$)
        \STATE $\mathbf{w}_j \leftarrow [\nabla f_i(\mathbf{\hat{x}}) - \nabla f_j(\mathbf{\hat{x}})] \mathbf{P}_m^T$
        \ENDFOR
        \STATE $l \leftarrow \argmin_{j \neq i}\frac{|o_j|}{||\mathbf{w_j}||_{2}}$
        \STATE $\mathbf{r} \leftarrow \frac{|o_l|}{||\mathbf{w}_l||_{2}^2} \mathbf{w}_l \mathbf{P}_m$
        \STATE $\mathbf{\hat{x}} \leftarrow \mathbf{\hat{x}} + \gamma\mathbf{r}$
        \STATE $\mathbf{\hat{x}} \leftarrow$ clip $(\mathbf{\hat{x}})$
        \STATE $j \leftarrow \argmax_{k \neq i} f_{k}(\mathbf{\hat{x}})$
        \STATE $v \leftarrow |f_i(\mathbf{\hat{x}}) - f_j(\mathbf{\hat{x}})|$
        \STATE $d \leftarrow ||\mathbf{x} - \mathbf{\hat{x}}||_{2}$
        \IF{$v \geq v_{best}$ \OR $|d - d_{best}| <  \delta$}
            \STATE \textbf{return} $d_{best}, v_{best}$
        \ELSE
            \STATE $v_{best} \leftarrow v$
            \STATE $d_{best} \leftarrow d$
            \STATE $c \leftarrow c + 1 $
        \ENDIF
        \ENDWHILE
        \STATE \textbf{return} $d_{best}, v_{best}$
    \end{algorithmic}
\end{algorithm}

The algorithm is very similar to the previous one, but now additionally contains modifications for a principal component subspace transformation. To extract the DeepFool constrained margin for some sample $\mathbf{x}$, the Taylor approximation of the margin in the lower-dimensional principal component subspace is calculated between the true class $i$ and all other classes $j$, individually. The smallest lower-dimensional subspace perturbation is then transformed back to the original feature space. This perturbation is then scaled by a set learning rate and added to the original sample. This process is repeated until the distance changes less than a given tolerance compared to the previous iteration. Note that the dimensionality of the sample $\mathbf{\hat{x}}$ is never reduced -- only the search for a perturbation is restricted to the lower-dimensional principal component subspace.

\section{Results}
\label{sec:constrained_results}

We investigate the extent to which constrained margins are predictive of generalization by comparing the new method with current alternatives. 
In Section~\ref{sec:constrained_experimental_setup} we describe our experimental setup. Following this, we do a careful comparison between our metric and existing techniques based on standard input and hidden margins in Section~\ref{sec:constrained_marg_comp}. Finally, we compare with others' complexity measures in Section~\ref{sec:constrained_main_pgdl_results}.

\subsection{Experimental setup}
\label{sec:constrained_experimental_setup}

We use the PGDL dataset to assess the predictive performance of our proposed constrained margin complexity measure.

To compare constrained margins with standard input and hidden margins, we use both Kendall's rank correlation and CMI as performance metrics, as done earlier in Section~\ref{sec:hidden_hidden_layer_selection}. 
To compare constrained margins with the published results of other complexity measures, we rely on CMI only (the only metric that is generally reported on in the literature).

For all margin-based measures, our indicator of generalization (complexity measure) is the mean margin over $5\ 000$ randomly selected training samples, or alternatively the maximum number available for tasks with fewer than $5\ 000$ training samples. Only correctly classified samples are considered, and the same training samples are used for all models of the same task. We compare our method using both the first-order Taylor approximation, as well as the more precise DeepFool-algorithm, for all margin measurements. 

For \textbf{standard input margins} (Input) and \textbf{hidden margins} (Hidden) we make use of the values reported earlier in Tables \ref{tab:hidden_hidden_input_taylor_v_deepfool} and \ref{tab:hidden_hidden_input_taylor_v_deepfool_cmi}. We use the Taylor-approximated values as is, while for the DeepFool variants, we select the values from the learning rate variation that achieved the highest CMI on average. This implies that we use the $\gamma=0.25$ version for standard input margins and $\gamma=1.0$ for hidden margins.  Note that this biases this comparison in favor of standard input and hidden margins, as one cannot know beforehand which learning rate will perform best. This is done so that we can compare constrained margins against the strongest possible margin baselines for these tasks.

Our \textbf{constrained margin} complexity measure (Constrained) is obtained using Algorithm~\ref{alg:manifol_margin}, although we once again implement this in a batched manner, as explained earlier at the end of Section~\ref{sec:hidden_modifying_deepfool}. We select the exact same hyperparameters as previously used for the other margin measurements. That is, in all experiments, a single learning rate ($\gamma = 0.25$) and max iterations ($max = 100$) are used. The same tolerance ($\delta = 0.01$) is used for all tasks, except Tasks $4$ and $5$, where we use a smaller tolerance ($\delta = 0.001$), as also done for standard input margins. (Recall our explanation at the end of Section~\ref{sec:hidden_verifying_deepfool_alg}.)

An important consideration is the number of principal components used for each dataset. We select this by plotting the explained variance (of the training data) per principal component in decreasing order on a logarithmic scale and applying the elbow method. Specifically, we use the Kneedle algorithm from Satopaa et al.~\cite{kneedle_elbow_method} to select the elbow. This results in a very low-dimensional search space, ranging from three to eight principal components for the seven unique datasets considered. See Appendix~\ref{app:constrained_num_components} for more information on the selection of the number of principal components.

In order to prevent biasing our metric to the PGDL test set (Tasks $6$ to $9$) we did not perform any tuning or development of the complexity measure using these tasks, nor do we tune any hyperparameters per task. The choice of principal component selection algorithm was made after a careful analysis of Tasks $1$ to $5$ only; see additional details in Appendix~\ref{app:constrained_num_components}.

\subsection{Margin complexity measures}
\label{sec:constrained_marg_comp}

In Tables \ref{tab:constrained_margin_complexities_kendall} and \ref{tab:constrained_margin_complexities_cmi} we show the Kendall's rank correlation and CMI, respectively, obtained when ranking models according to constrained margin, standard input margins, and hidden margins using both Taylor and DeepFool. 
\begin{table}[H]
\centering
\caption[Predictive performance comparison of standard input, hidden, and constrained margins on the PGDL tasks (Kendall's)]{Kendall's rank correlation between mean margin and test accuracy for constrained, standard input, and hidden margins for the PGDL dataset.}
\label{tab:constrained_margin_complexities_kendall}
\begin{tabular}{@{}cccc|ccc@{}}
\toprule
\multirow{2}{*}{\textbf{Task}} & \multicolumn{3}{c|}{\textbf{Taylor}} & \multicolumn{3}{c}{\textbf{DeepFool}} \\ \cmidrule(l){2-7} 
 & Constrained & Input & Hidden 1st & Constrained & Input & Hidden 1st \\ \midrule
1 & \textbf{0.6991} & 0.0265 & 0.5794 & \textbf{0.8053} & -0.1643 & 0.6412 \\
2 & \textbf{0.8281} & 0.6841 & 0.7037 & \textbf{0.8616} & 0.6576 & 0.6883 \\
4 & 0.6966 & 0.6251 & \textbf{0.7958} & 0.6633 & 0.6436 & \textbf{0.7949} \\
5 & 0.2381 & 0.3571 & \textbf{0.5427} & 0.2242 & 0.4355 & \textbf{0.5278} \\
6 & \textbf{0.6753} & -0.1351 & 0.4427 & \textbf{0.8017} & -0.2356 & 0.4814 \\
7 & \textbf{0.4192} & 0.3215 & 0.3623 & \textbf{0.5115} & 0.2913 & 0.4139 \\
8 & \textbf{0.3419} & -0.1233 & -0.0656 & \textbf{0.5994} & -0.1938 & -0.0209 \\
9 & \textbf{0.7258} & 0.1573 & 0.7097 & \textbf{0.8145} & 0.0685 & 0.7097 \\ \midrule
Average & \textbf{0.5780} & 0.2392 & 0.5088 & \textbf{0.6602} & 0.1879 & 0.5300 \\ \bottomrule
\end{tabular}
\end{table}

\begin{table}[H]
\centering
\caption[Predictive performance comparison of standard input, hidden, and constrained margins on the PGDL tasks (CMI)]{Conditional Mutual Information between mean margin and generalization gap for constrained, standard input, and hidden margins for the PGDL dataset.}
\label{tab:constrained_margin_complexities_cmi}
\begin{tabular}{@{}cccc|ccc@{}}
\toprule
\multirow{2}{*}{\textbf{Task}}& \multicolumn{3}{c|}{\textbf{Taylor}} & \multicolumn{3}{c}{\textbf{DeepFool}} \\ \cmidrule(l){2-7} 
 & Constrained & Input & Hidden 1st & Constrained & Input & Hidden 1st \\ \midrule
1 & \textbf{23.77} & 00.07 & 09.40 & \textbf{39.49} & 01.76 & 09.60 \\
2 & \textbf{43.37} & 06.12 & 37.74 & \textbf{50.63} & 06.34 & 36.07 \\
4 & 22.18 & 14.95 & \textbf{34.73} & 21.41& 17.40& \textbf{34.35} \\
5 & 05.42 & 08.46 & \textbf{19.11} & 04.80& 13.16& \textbf{17.32} \\
6 & \textbf{10.65} & 00.57 & 04.24 & \textbf{30.73} & 02.14 & 05.25 \\
7 & \textbf{12.91} & 01.47 & 05.04 & \textbf{13.20} & 02.51 & 03.14 \\
8 & \textbf{03.70} & 00.70 & 00.36 & \textbf{13.35} & 00.58 & 00.26 \\
9 & \textbf{18.61} & 00.29 & 23.74 & \textbf{51.46} & 01.67 & 27.27 \\ \midrule
Average & \textbf{17.58} & 04.08 & 16.80 & \textbf{28.13}& 05.70& 16.66 \\ \bottomrule
\end{tabular}
\end{table}

We observe that constrained margins outperform the other margin measurements on average, and this is clear for both metrics considered. Let us break down the individual observations:

There is a large performance gap between constrained and standard input margins: an increase from $0.2392$ to $0.5780$ average rank correlation is observed by constraining the margin search for the Taylor approximation, and an increase from $0.1869$ to $0.6602$ for DeepFool. This performance gap is even more evident when considering the CMI results: we note an increase from $4.08$ to $17.58$ (Taylor) and $5.70$ to $28.13$ (DeepFool).  This strongly supports our initial intuitions. 

In the case of hidden margins, performance compared to constrained margins is more competitive; however, constrained margins still outperform hidden margins on six out of eight tasks, as well as on average. For Taylor-approximated margins, we observe an average Kendall's rank correlation of $0.5088$ for hidden margins versus $0.5780$ for constrained margins. Similarly, for DeepFool, $0.5300$ (hidden) versus $0.6602$ (constrained). The CMI results are also better for constrained margins. For hidden versus constrained: $16.80$ versus $17.58$ (Taylor), and $16.66$ versus $28.13$ (DeepFool). Furthermore, since we select the best-performing layer selection method, the comparison is biased in favor of hidden margins, as there is no method at present to determine a priori which layer selection will perform best for a given task.  Given that our constrained margin measurement is limited to the input space, there are multiple advantages: 1) no normalization is required, as all models share the same input space, and 2) the method is more robust when comparing models with varying topology, as no specific layers need to be selected.

Finally, it is clear that the more accurate DeepFool method (Algorithm~\ref{alg:manifol_margin}) significantly improves the predictive power of constrained margins, while this is not the case for standard input and hidden margins.

\subsection{Other complexity measures}
\label{sec:constrained_main_pgdl_results}

To further assess the predictive power of constrained margins, we compare our method (DeepFool variant) to the reported CMI scores of several other complexity measures. We compare against three solutions from the winning team~\cite{rep_based_complexity_pgdl}, as well as the best solutions from two more recent works~\cite{optimal_transport,schiff2021predicting}, where that of Schiff et al.~\cite{schiff2021predicting} has the highest average test set performance we are aware of. We do not compare against pre-trained GANs~\cite{gan_pgdl}, for the reasons expressed in Section~\ref{sec:back_pgdl_complexity_measures_explanation}. The original name of each method is kept. Of particular relevance are the $MM$ and $AM$ columns, which are hidden margins applied to Mixup and Augmented samples, as well as $k$V-Margin and $k$V-GN-Margin which are output and hidden margins with $k$-Variance normalization, respectively. Refer back to Sections \ref{sec:back_hidden_margins} and \ref{sec:back_output_margins} for a full description of these techniques. 

\begin{table}[H]
\caption[Predictive performance comparison of  constrained margins versus other complexity measures on the PGDL tasks (CMI)]{Conditional Mutual Information (CMI) scores for several complexity measures on the PGDL dataset. Acronyms: $DBI$=Davies Bouldin Index, $LWM$=Label-wise Mixup, $MM$=Mixup Margins, $AM$=Augmented Margins, $kV$=$k$-Variance, $GN$=Gradient Normalized, $Gi$=Gi score, $Mi$=Mixup. Test set average is the average over tasks $6$ to $9$. \textdagger Indicates a margin-based measure.}
\label{tab:constrained_pgdl_results}
\centering
\begin{adjustbox}{max width=\textwidth}
\begin{tabular}{cccccccc}
\hline
\multicolumn{1}{c}{\multirow{2}{*}{\textbf{Task}}} & \multicolumn{3}{c}{\textbf{Natekar and Sharma}} & \multicolumn{2}{c}{\textbf{Chuang et al.}} & \textbf{Schiff et al.} & \textbf{Ours} \\
\multicolumn{1}{c}{} & DBI*LWM & MM\textdagger & AM\textdagger & \begin{tabular}[c]{@{}c@{}}$k$V-\\ Margin 1st\textdagger \end{tabular} & \begin{tabular}[c]{@{}c@{}}$k$V-GN-\\ Margin 1st\textdagger\end{tabular} & \begin{tabular}[c]{@{}c@{}}PCA\\ Gi\&Mi\end{tabular} & \begin{tabular}[c]{@{}c@{}} Constrained\\ Margin\textdagger \end{tabular} \\ \hline
1 & 00.00 & 01.11 & 05.73 & 05.34 & 17.95 & 00.04 & \textbf{39.49}\\
2 & 32.05 & 47.33 & 44.60 & 26.78 & 44.57 & 38.08 & \textbf{50.63}\\
4 & 31.79 & 43.22 & \textbf{47.22} & 37.00 & 30.61 & 33.76 & 21.41\\
5 & 15.92 & \textbf{34.57} & 22.82 & 16.93 & 16.02 & 20.33 & 04.80\\ \hline
6 & \textbf{43.99} & 11.46 & 08.67 & 06.26 & 04.48 & 40.06 & 30.73\\
7 & 12.59 & \textbf{21.98} & 11.97 & 02.07 & 03.92 & 13.19 & 13.20\\
8 & 09.24 & 01.48 & 01.28 & 01.82 & 00.61 & 10.30 & \textbf{13.35}\\
9 & 25.86 & 20.78 & 15.25 & 15.75 & 21.20 & 33.16 & \textbf{51.46} \\ \hline
\begin{tabular}[c]{@{}l@{}}Test set \\ average\end{tabular} & 22.92 & 13.93 & 09.29 & 06.48 & 07.55 & 24.18& \textbf{27.19}\\ \hline
\end{tabular}
\end{adjustbox}
\end{table}

The results of this comparison are shown in Table~\ref{tab:constrained_pgdl_results}.
One observes that constrained margins achieve competitive scores, and in fact, outperform all other measures on four out of eight tasks. Furthermore,  the new method achieves the highest average accuracy on the test set of tasks (Tasks $6$ to $9$). 

It is also important to note that the \textit{MM} and \textit{AM} columns show that hidden margins can be improved in some cases if they are measured using the representations of Mixup or augmented training samples. That said, these methods still underperform on average in comparison to constrained input margins, which do not rely on any form of data augmentation.

\section{Additional analysis}
\label{sec:constrained_analysis}

In this section, we do a further analysis of constrained margins. First, we investigate how the performance of constrained margins changes when lower utility subspaces are considered (Section~\ref{sec:constrained_high_versus_low_util}). Following this, we make a qualitative comparison between constrained and standard input margin boundary points in Section~\ref{sec:constrained_qualitative_comparison}. Finally, we discuss limitations of the constrained margin method in Section~\ref{sec:constrained_limitations}.

\subsection{High to lower utility}
\label{sec:constrained_high_versus_low_util}

In this section, we determine whether `useful' directions are indeed required, or whether constraining the margin search in other ways would give a similar effect. To this end, we examine how high-utility directions compare to those of lower utility when calculating constrained margins.

\begin{figure}[h]
        \centering
        \includegraphics[width=0.49\linewidth]{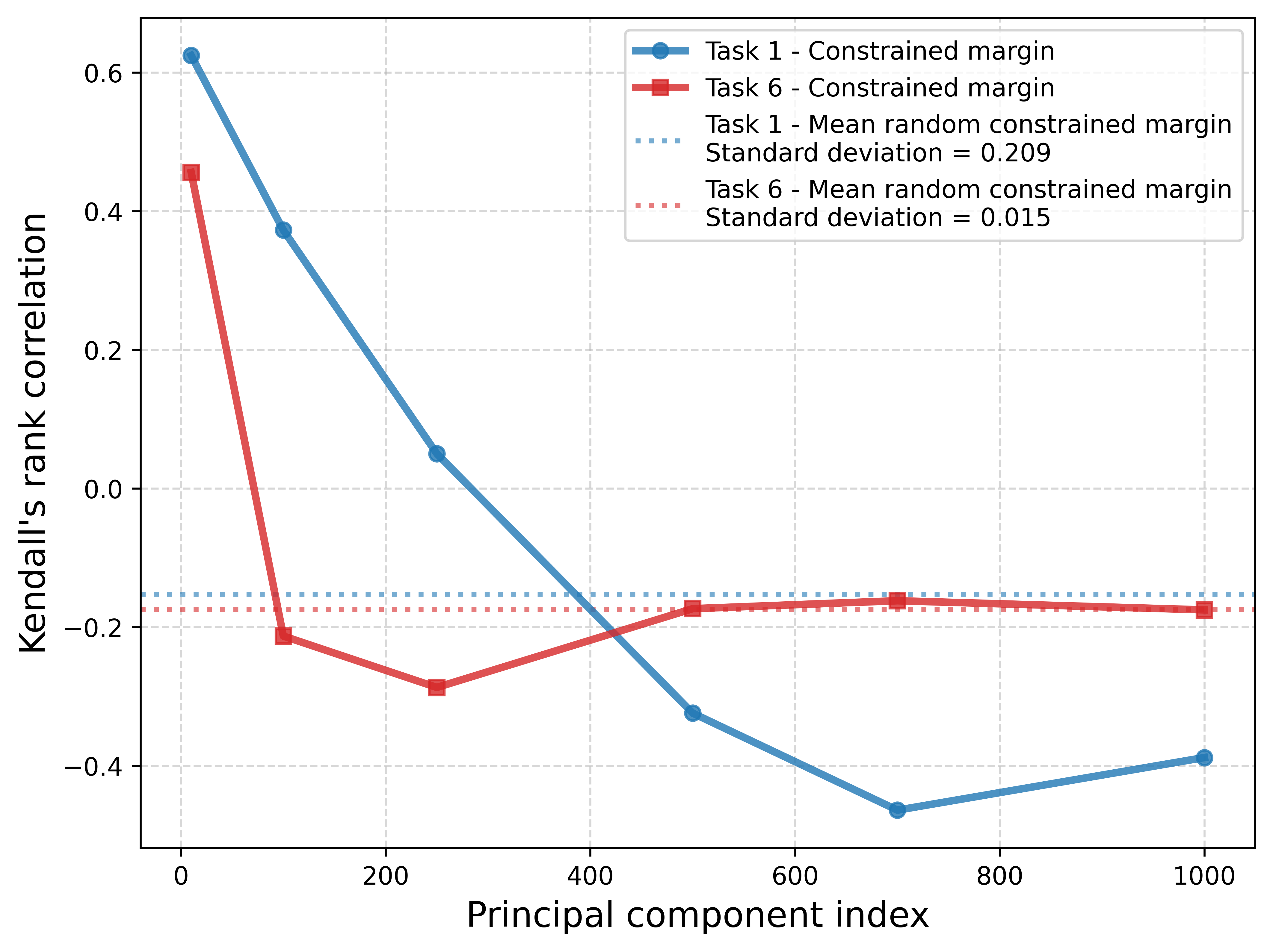}
        \includegraphics[width=0.49\linewidth]{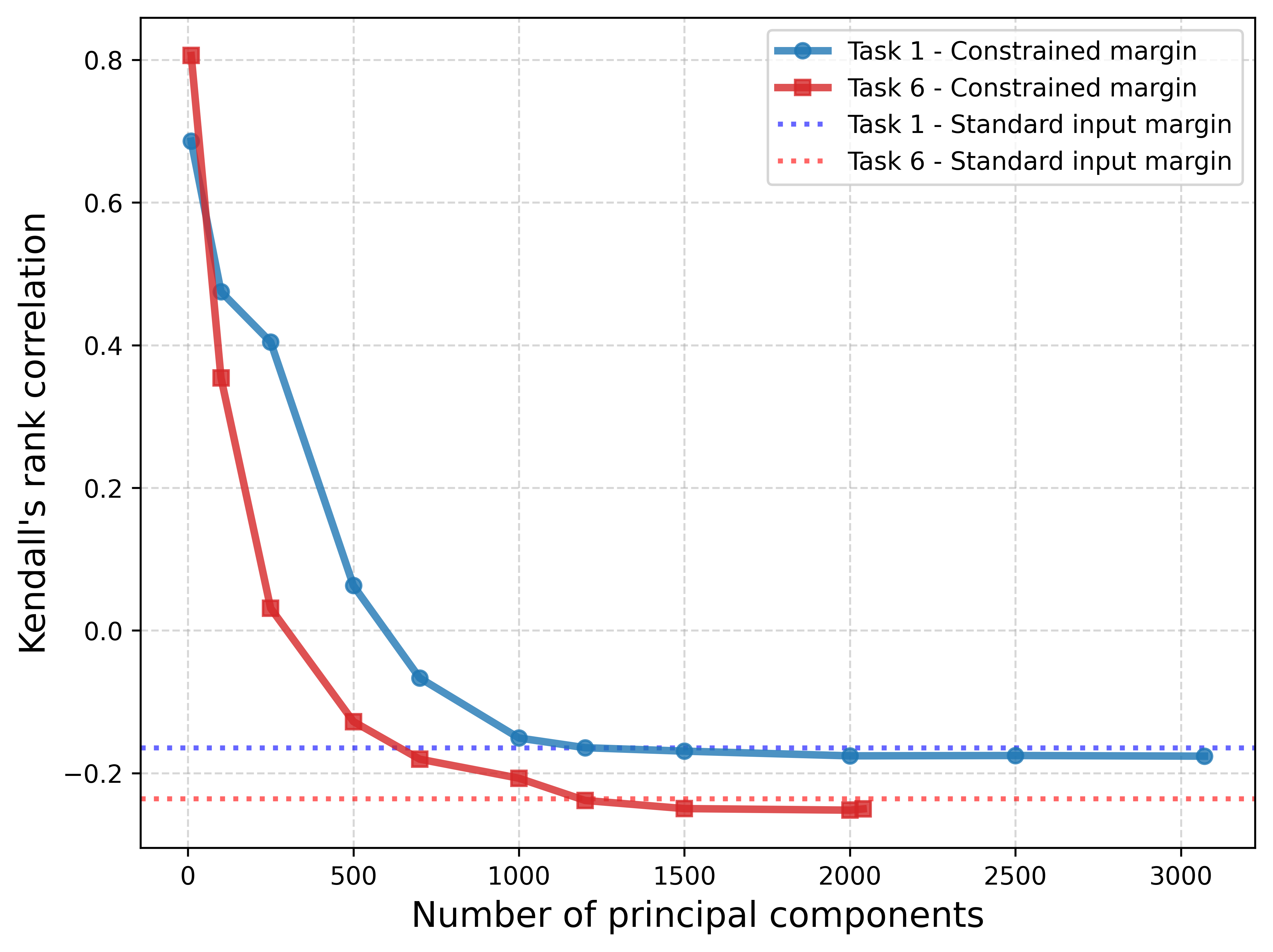} 
        \caption[Comparison of predictive performance of constrained margins using two different ways of selecting principal components for Tasks $1$ and $6$]{Comparison of predictive performance (Kendall's rank correlation) of constrained margins using two different ways of selecting principal components for Tasks $1$ and $6$. Left: Performance comparison of subspaces spanned by $10$ principal components of decreasing utility. The $x$-axis indicates the first component in each set of principal components. Dashed horizontal lines indicate the performance of a subspace spanned by $10$ randomly chosen principal components, averaged over $10$ random seeds. Right: Performance comparison when selecting an increasing number of components. Dashed horizontal lines indicate the performance of standard input margins.}
        \label{fig:constrained_high_v_low_task_1_6}
\end{figure}

We calculate the mean constrained margin using two different ways of selecting the principal components:
\begin{itemize}
    \item We select subsets of $10$ contiguous principal components in descending order of explained variance. For example, we calculate the constrained margins using components $1$ to $10$, then $100$ to $109$, etc. This allows us to calculate the distance to the decision boundary using $10$ dimensional subspaces of decreasing utility. As a baseline for this analysis, we also calculate the mean constrained margin using a subspace spanned by $10$ randomly selected principal components. For example, principal components $\{3, 6, 45, 774, 906, 933, 1855, 2938, \\ 2957, 3068\}$. Our final baseline value is given by the average of the predictive performance over $10$ such random selections. 
    \item We select an increasing number of principal components, again in decreasing order of explained variance. For example, we select principal components $1$ to $10$, then $1$ to $50$, and so on and calculate the constrained margins for each selection. For this analysis, we use the predictive performance of the standard input margin as baseline, as that should be equivalent to considering all the available principal components.
\end{itemize}

For these analyses, we select two tasks where there is a large difference between the performance of constrained margins and standard input margins: Tasks $1$ and $6$. We make use of the DeepFool variant, as used earlier. Furthermore, to ease the computational burden, we limit our margin calculations to $1\ 000$ randomly selected training samples per model. 

Figure~\ref{fig:constrained_high_v_low_task_1_6}, left, shows the resulting Kendall's rank correlation for each subset of principal components indexed by the first component in each set (principal component index) for Task $1$ and $6$. Additionally, the right shows the Kendall's rank correlation when using an increasing number of principal components for these two tasks.

Let us first discuss the left of Figure~\ref{fig:constrained_high_v_low_task_1_6}. We observe that the first principal components lead to margins that are more predictive of generalization, and then a strong decrease in predictive power when considering later principal components. Additionally, we observe that constrained margins calculated using the first principal components perform much better than the randomly selected principal component baseline (dashed blue and red lines, for Task $1$ and $6$, respectively). Conversely, the later principal components reach \textit{negative} correlations and perform worse than this baseline for Task $1$, while for Task $6$ the later components and baseline show approximately equal performance. We also note that after the point shown here (index $1\ 000$), we find that the mean margin increases as DeepFool struggles to find points on the decision boundary within the bound constraints. Due to this, it is difficult to draw any conclusions from an investigation of the lower-ranked principal components.

On the right of Figure~\ref{fig:constrained_high_v_low_task_1_6} we observe that predictive performance decreases as more principal components are considered for the constrained margin calculation. Eventually, the predictive performance converges to that of standard input margins (dashed blue and red lines, for Task $1$ and $6$, respectively) which one would expect. This further confirms our intuitions: as the size of the search space is increased, the performance of the margin measurement decreases.

Both of these results strongly support the idea that the use of the highest utility directions is a necessary aspect of constrained margin measurements.

\subsection{Qualitative comparison}
\label{sec:constrained_qualitative_comparison}

\begin{figure}[h]
        \centering
        \includegraphics[width=\linewidth]{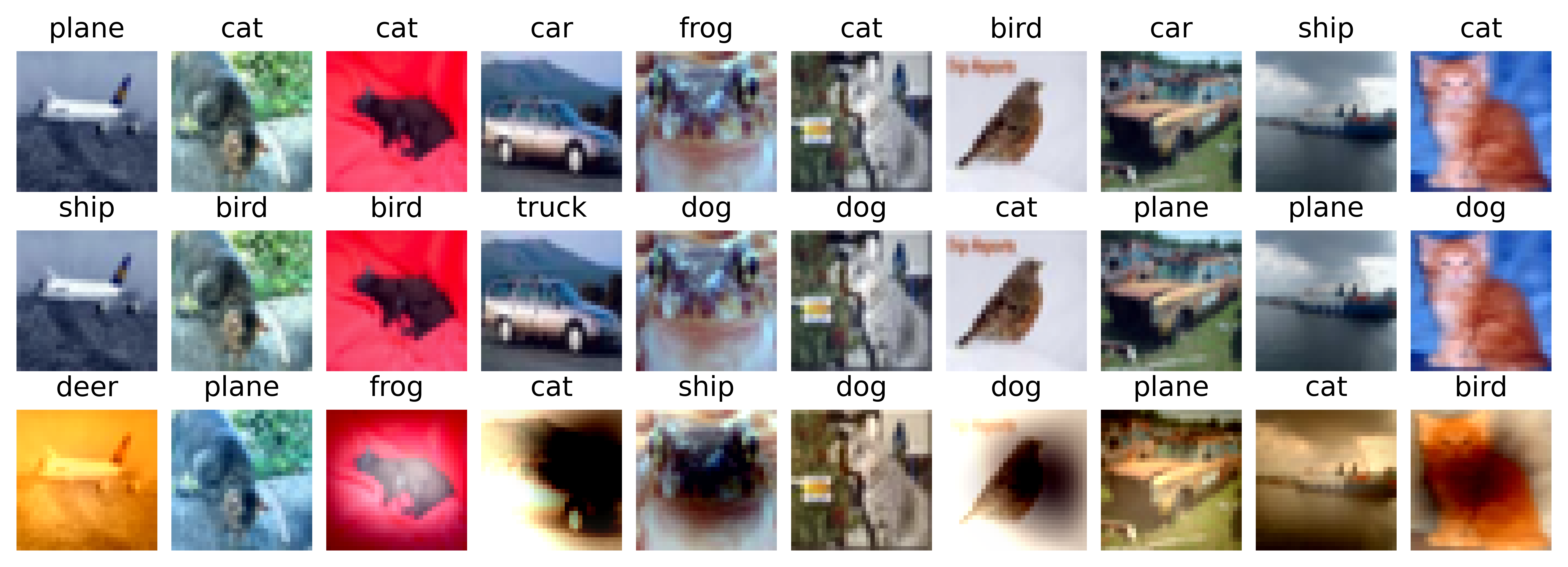} 
        \caption[Points found on the decision boundary when calculating standard input and constrained margins for a model from Task $1$, CIFAR10]{Points found on the decision boundary when calculating standard input and constrained margins for a well generalizing model from Task $1$, CIFAR10. Top: original training sample. Middle: standard input margin boundary point. Bottom: constrained margin boundary point. Text above each image indicates its classification label (top) or the class to which it is changed (middle and bottom).}
        \label{fig:constrained_task_1_qualitative_comparison}
\end{figure}

\begin{figure}[h]
        \centering
        \includegraphics[width=\linewidth]{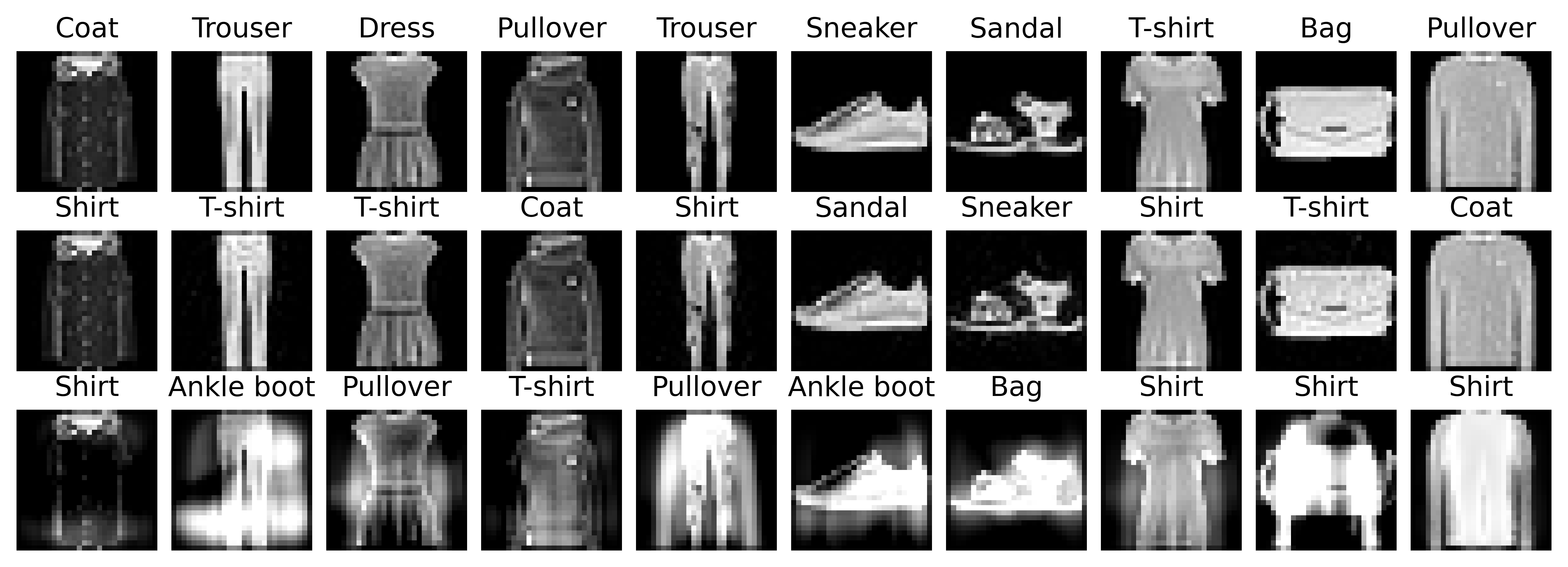} 
        \caption[Points found on the decision boundary when calculating standard input and constrained margins for a model from Task $8$, FMNIST]{Points found on the decision boundary when calculating standard input and constrained margins for a well generalizing model from Task $8$, FMNIST. Top: original training sample. Middle: standard input margin boundary point. Bottom: constrained margin boundary point. Text above each image indicates its classification label (top) or the class to which it is changed (middle and bottom).}
        \label{fig:constrained_task_8_qualitative_comparison}
\end{figure}

In this section, we perform a qualitative comparison of standard input margin and constrained margin boundary points. That is, we visually inspect the points found on the decision boundary with both methods. We randomly select $10$ training samples from two datasets and visualize 1) the original training samples, 2) boundary points found when calculating standard input margins, and 3) boundary points when calculating constrained margins. To this end, we select the model with the best generalization performance from Tasks $1$ (CIFAR10) and $8$ (FMNIST), to provide us with results for two dissimilar datasets. This is shown in Figure~\ref{fig:constrained_task_1_qualitative_comparison} for Task $1$ and Figure~\ref{fig:constrained_task_8_qualitative_comparison} for Task $8$. We also show the original labels for the samples (top row), and the class to which the sample is perturbed (middle and bottom rows) above each image.

For both datasets, we observe that standard input margins find points on the decision boundary that are visually identical to the original training samples, meaning that the perturbation is completely imperceptible. This is to be expected as these boundary points are essentially adversarial examples, and this behavior is well documented. On the other hand, one observes that constrained margins result in points that are more altered, and the difference in comparison to the original sample is clear.  

For the CIFAR10 results (Figure~\ref{fig:constrained_task_1_qualitative_comparison}) the constrained boundary points appear to `erase' or `black out' certain key sections within the image. In other cases, the color of the sample is changed. 
For the FMNIST results (Figure~\ref{fig:constrained_task_8_qualitative_comparison}), we see that in some cases the constrained boundary sample appears to be more visually aligned with another class. For example, one of the `trouser' samples (3rd row, 2nd column) appears to be haunted by a shoe-like apparition, and is changed to class `ankle boot'. Similarly, several of the images appear to receive ghostly sleeves and flip to the `pullover' or `shirt' classes. 

These results indicate that constrained margins are qualitatively different to standard input margins. Given that constrained margins perturb samples in a more visually meaningful way, one would expect that they are more aligned with the underlying model's generalization ability. Furthermore, from the CIFAR10 results we can clearly see that color is very important to CNN architectures for classification, and perhaps suggests something about their underlying inductive bias. Finally, it is also interesting that constrained and standard input margins do not always flip each sample to the same class.

\subsection{Limitations}
\label{sec:constrained_limitations}

It has been demonstrated that our proposed metric performs well and aligns with our initial intuition.  
However, there are also certain limitations that require explanation. We consider the limitations of constrained margins according to 1) specific hyperparameter variations, 2) a hypothetical case where high variance does not correspond to high utility, and 3) specific poor-performing tasks.

\subsubsection{Analysis of hyperparameters}

To determine for which hyperparameters constrained margins fail or succeed, we calculate the granulated Kendall's rank correlation per hyperparameter variation across all PGDL tasks. We do this in the exact same manner as we have done previously for standard input and hidden margins in Section~\ref{sec:hidden_failure_cases}. For this analysis, we make use of the best-performing method, i.e. the DeepFool variant of constrained margins. Furthermore, we use the same mean DeepFool constrained margin values as used earlier for all comparisons in Section~\ref{sec:constrained_results}. The granulated Kendall's rank correlation per hyperparameter is shown in Table~\ref{tab:constrained_constrained_granulated}.

\begin{table}[h]
\caption[Granulated Kendall's coefficient for constrained margins on the PGDL tasks]{Granulated Kendall's coefficient for constrained margins for all PGDL tasks and hyperparameter variations. `\textdagger' Indicates a hyperparameter variation that is not directly comparable across tasks. }
\label{tab:constrained_constrained_granulated}
\begin{adjustbox}{max width=\textwidth}
\begin{tabular}{@{}cccccccccc@{}}
\toprule
\textbf{Task} & \textbf{\begin{tabular}[c]{@{}c@{}}Learning\\ rate\end{tabular}} & \textbf{\begin{tabular}[c]{@{}c@{}}Batch\\ size\end{tabular}} & \textbf{\begin{tabular}[c]{@{}c@{}}Weight\\ decay\end{tabular}} & \textbf{\begin{tabular}[c]{@{}c@{}}Dropout\\ probability\end{tabular}} & \textbf{Depth}\textdagger & \textbf{Width}\textdagger & \textbf{\begin{tabular}[c]{@{}c@{}}Num\\ dense\end{tabular}}\textdagger & \textbf{Reverse} & \textbf{$\mu(\Psi$)} \\ \midrule
1 & - & 0.917 & 0.458 & 0.958 & 0.792 & 0.125 & 0.333 & - & 0.597 \\
2 & - & 0.519 & 0.926 & 0.815 & 0.889 & - & - & - & 0.787 \\
4 & 0.875 & 0.625 & -0.125 & - & 0.271 & 0.792 & - & 0.875 & 0.552 \\
5 & 0.313 & 0.250 & 0.250 & - & -0.063 & 0.500 & - & 0.563 & 0.302 \\
6 & 0.958 & 0.792 & 1.000 & 0.489 & 0.667 & 0.875 & - & - & 0.797 \\
7 & - & 0.000 & 0.500 & 0.500 & 0.500 & - & 0.405 & - & 0.381 \\
8 & 0.563 & 0.688 & 0.125 & 0.563 & 0.813 & 0.250 & - & - & 0.500 \\
9 & - & 1.000 & 1.000 & 0.750 & 0.750 & 0.875 & - & - & 0.875 \\ \midrule
Average & 0.677 & 0.599 & 0.517 & 0.679 & 0.577 & 0.569 & 0.369 & 0.719 & 0.599 \\ \bottomrule
\end{tabular}
\end{adjustbox}
\end{table}

These results show a pattern somewhat similar to those of hidden margins. We observe that, for tasks where constrained margins perform well, they do not perform consistently well for the same hyperparameters. For example, our metric is able to accurately account for a variation in weight decay for models in Task $9$ ($1.000$), but not so for Task $1$ ($0.458$). Similarly, a variation in batch size for Task $1$ is handled well ($0.917$), but less so for Task $2$ ($0.519$). Tasks with poorer overall performance also do not fail in the same ways. Consider a variation in batch size for Task $7$ ($0.000$) versus that of Task $4$ ($0.625$). These observations imply that the success or failure of constrained margins is tied to the architectural family and dataset of the task considered, rather than to a specific hyperparameter variation.

\begin{figure}[h]
        \centering
        \includegraphics[width=0.49\linewidth]{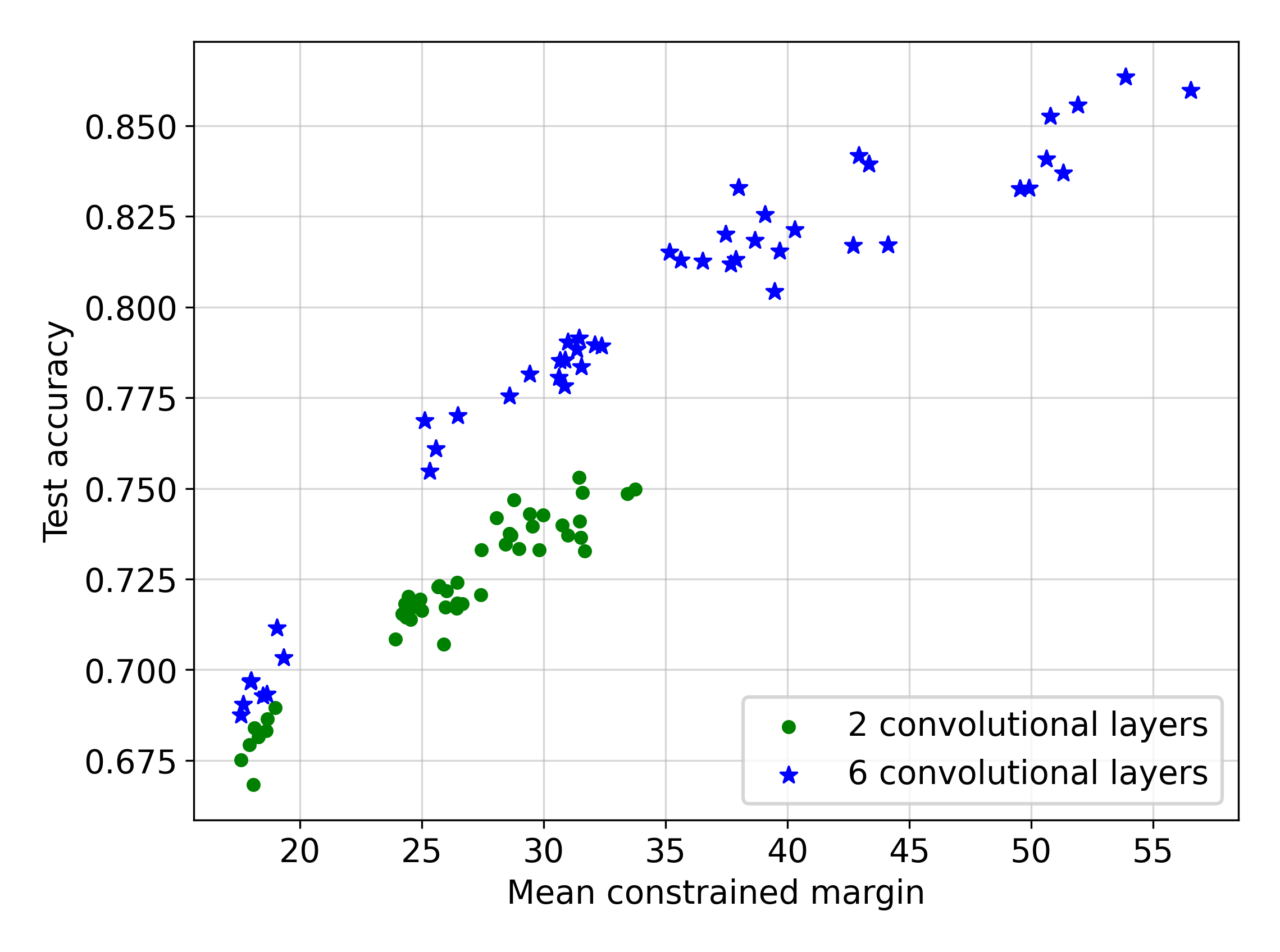}
        \includegraphics[width=0.49\linewidth]{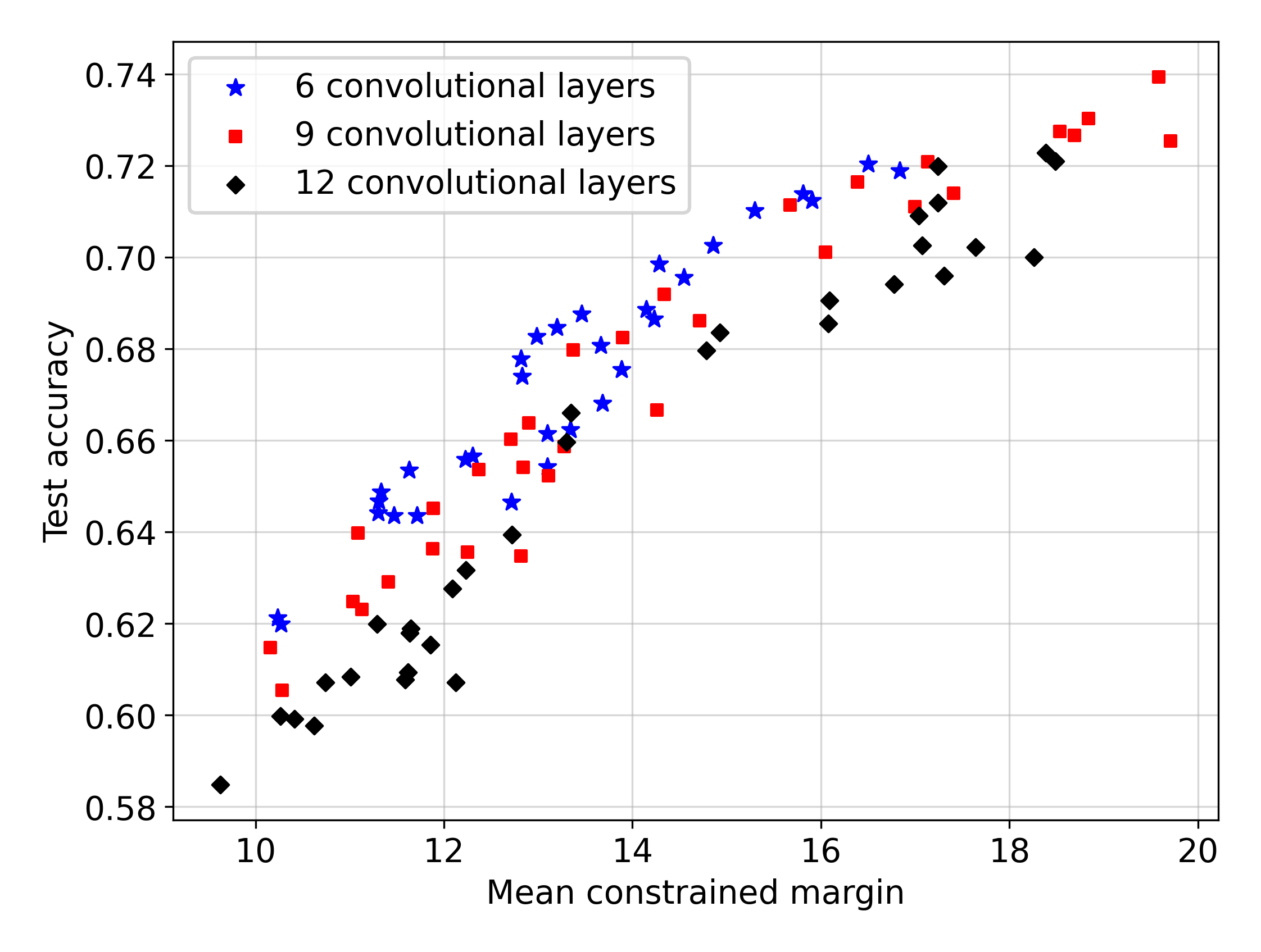}
        \caption[Mean constrained margin versus test accuracy for PGDL Task $1$ and $6$]{Mean constrained margin versus test accuracy for PGDL Task $1$ (left) and $6$ (right).  Left: Models with $2$ (green circle) and $6$ (blue star) convolutional layers from Task $1$. Right: Models with $6$ (blue star), $9$ (red square), and $12$ (black diamond) convolutional layers from Task $6$.}
        \label{fig:constrained_two_v_six_layer}
\end{figure}

It is also important to note that the values expressed in Table~\ref{tab:constrained_constrained_granulated} can be somewhat misleading in the specific case of model depth. This is illustrated in Figure~\ref{fig:constrained_two_v_six_layer}, which shows the mean constrained margin versus test accuracy for Task $1$ (left) and $6$ (right). On the left, we observe that sets of networks with two and six convolutional layers, respectively, each exhibit a separate relationship between margin and test accuracy. This is despite achieving a high rank correlation ($0.792$) for variations in depth.  This discrepancy is not always as strongly present: for Task $6$, all three depth configurations show a more similar relationship, as observed on the right of Figure~\ref{fig:constrained_two_v_six_layer}, although the discrepancy is still present. The same trend holds for several tasks ($1$, $2$, $4$, $6$, $9$).
It appears that shallower networks model the input space in a distinctly different fashion than their deeper counterparts.

We conclude that constrained margins occasionally fail to accurately rank generalization for variations in specific hyperparameters, but this failure is not consistent across different datasets and architectures.

\subsubsection{Can high variance be a poor indicator of high utility?}

We now consider a hypothetical failure case where high-variance directions do not correspond to high-utility directions. Indulge us for the following thought experiment:

Consider a binary linearly separable classification problem. Let ($\mathbf{X}_1$, $\mathbf{Y}$) represent a dataset comprising of samples $\mathbf{x} \in \mathbf{X}_1$ and labels $y \in Y$, with $\mathbf{x} \in \mathbb{R}^d$ and $y \in \{-1, 1\}$, indicating two classes: positive ($y = 1$) and negative ($y = -1$).
For positive labels, the samples $\mathbf{x}$ are drawn from $(1, 0.1\times\mathcal{N}(0,\mathbf{I}_d))$, and similarly for negative labels the samples are drawn from $(-1, 0.1\times\mathcal{N}(0,\mathbf{I}_d))$. That is, the first feature is either $1$ or $-1$ for the two labels, while the other $d - 1$ features are drawn from a Gaussian distribution ($\mathcal{N}$) with a mean of $0$ and standard deviation of $1$, and then multiplied by $0.1$.
Now consider a single neuron $y = \mathbf{w} \cdot \mathbf{x}$ where its hyperplane is given by $\mathbf{w} = (1, 0, ..., 0)$. This neuron generalizes perfectly for any data drawn from these distributions.
The constrained margin in this case would provide a good estimate of how well the classes are separated by this hyperplane. 

\begin{figure}[h]
        \centering
        \includegraphics[width=0.49\linewidth]{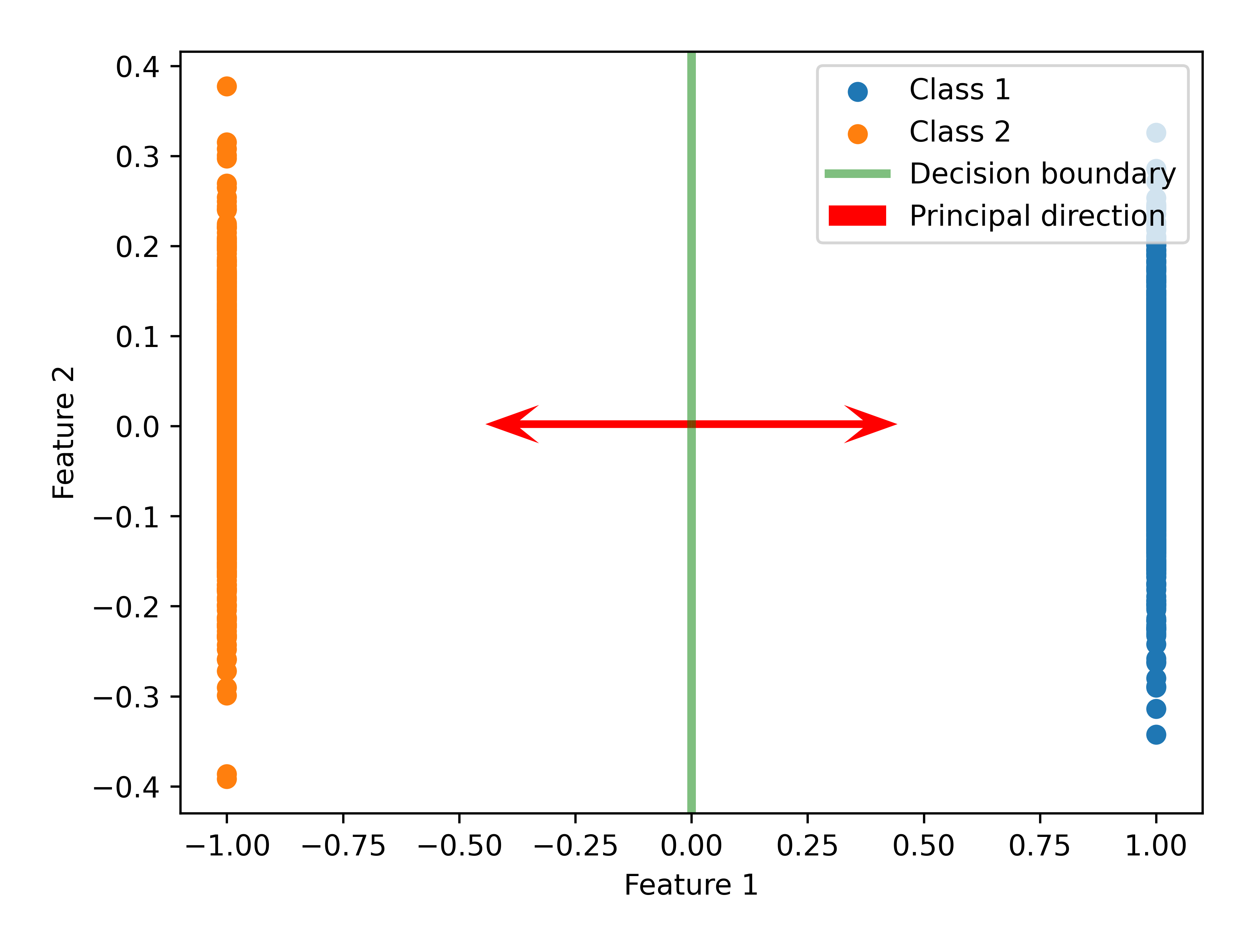}
        \includegraphics[width=0.49\linewidth]{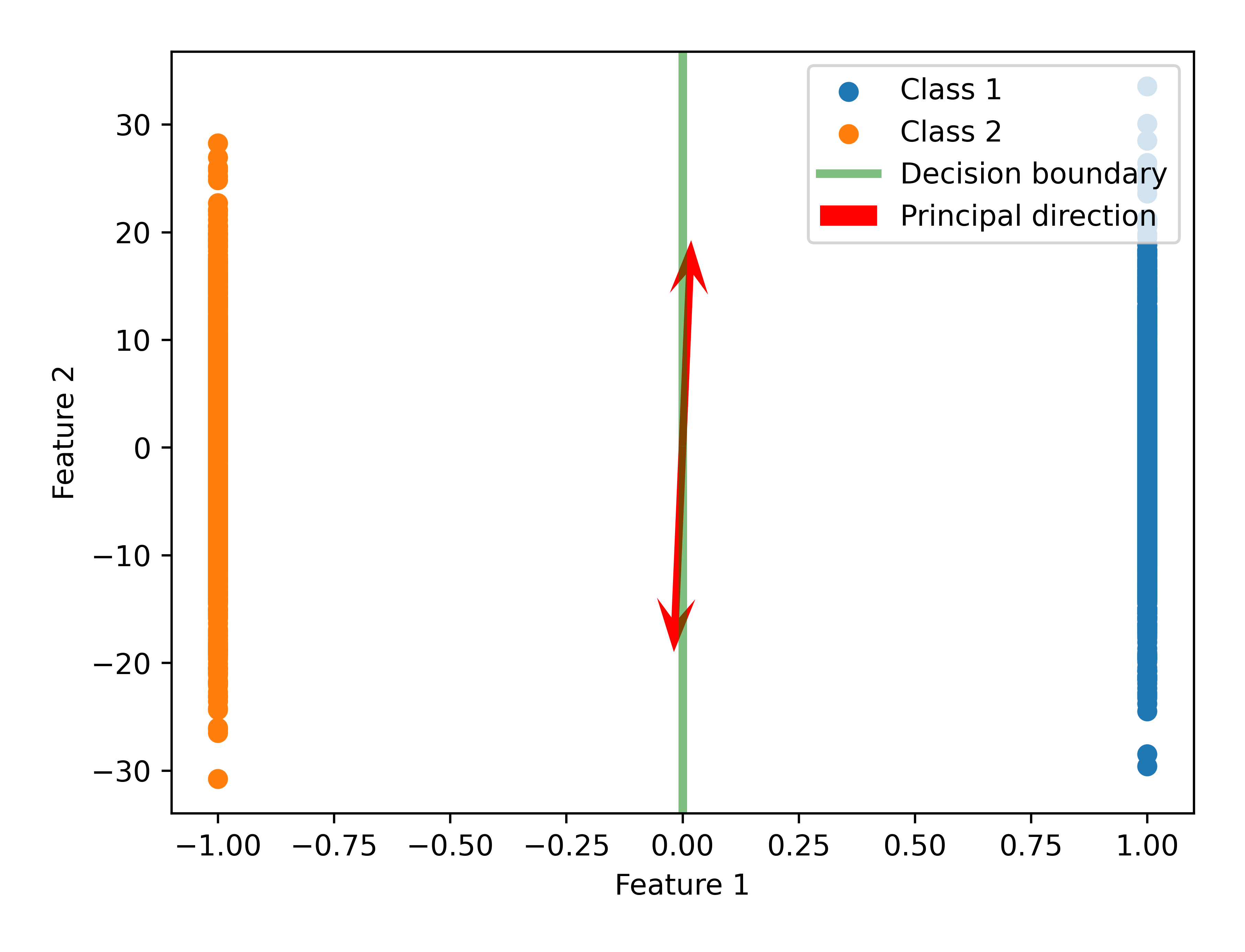}
        \caption[Illustration of a hypothetical failure case of constrained margins]{Two dimensional example of the datasets $\mathbf{X}_1$ (left) and  $\mathbf{X}_2$ (right) along with a separating decision boundary. The principal direction indicates the first principal component of each dataset, meaning the direction in which the constrained margin will be measured from each sample to the decision boundary.}
        \label{fig:constrained_thought_experiment}
\end{figure}

On the other hand, consider an almost identical dataset ($\mathbf{X}_2$, $\mathbf{Y}$) where the only difference is that the samples $\mathbf{x}$ are drawn from $(1, 10\times\mathcal{N}(0,\mathbf{I}_d))$ and $(-1, 10\times\mathcal{N}(0,\mathbf{I}_d))$ for positive and negative labels, respectively, separated by the same hyperplane.
For $\mathbf{X}_2$, given that there is much greater variance within a class than between classes, the constrained margin for any sample would be exceptionally large, and would not be a good indicator of class separation. However, the standard input margin would be the same (margin of $1$) for both $\mathbf{X}_1$ and $\mathbf{X}_2$. We show a two dimensional example of each dataset and its principal component in Figure~\ref{fig:constrained_thought_experiment}.

From this thought experiment, it is clear that scenarios can be constructed where directions of high variance provide a misleading proxy for following the data manifold. However, as evidenced by the general strong performance of constrained margins, this scenario does not typically present itself in natural image datasets.

\subsubsection{Poor-performing tasks}

We now consider tasks where constrained margins have poor performance. Specifically, we focus on the three tasks where constrained margins achieve the lowest Kendall's rank/CMI in comparison to its performance on the others: Tasks $5$, $7$, and $8$.

We first consider the worst-performing task. Constrained margins have by far the worst performance on Task $5$, when compared to itself on other tasks or to other complexity measures on the same task. This is an especially interesting failure case, as Tasks $4$ and $5$ are nearly identical, yet constrained margins perform much better on Task $4$ than on Task $5$ ($21.41$ versus $4.80$ CMI for DeepFool calculated constrained margins). Specifically, both tasks consist of fully convolutional networks trained on the CINIC10~\cite{CINIC} dataset. However, there is a key difference between these two: models in Task $4$ are trained with batch normalization, while those in Task $5$ are trained without. 

It is not made clear which of the other PGDL tasks are trained with or without batch normalization~\cite{pgdl_overview}. In the PGDL challenge setup, if a model is trained with batch normalization, the normalization weights are `merged' with the model's parameters. This implies that one cannot tell if batch normalization is used by inspecting the model's architecture.\footnote{This was ascertained through personal correspondence with Yiding Jiang~\cite{jiang_personal_correspondence}, one of the challenge organizers. Unfortunately, he does not recall which tasks have batch normalization or not.} Due to this, we cannot investigate the difference that batch normalization makes on the other tasks of the PGDL challenge.
Furthermore, we note another idiosyncrasy of Task $5$: we find that Taylor-approximated constrained margins achieve a peak performance (Kendall's rank correlation of $0.40$) when using $270$ principal components for the calculation. On the other hand, constrained margins achieve peak performance on the other development set tasks using two to five principal components. See Appendix~\ref{app:constrained_num_components} for more details. The reason for this is unclear. We refer back to this in Chapter~\ref{chap:conclusion}.

We now turn our attention to Tasks $7$ and $8$. In comparison to the performance of constrained margins on the other tasks, these two seem problematic. However, note that compared to the other complexity measures in Table~\ref{tab:constrained_pgdl_results}, constrained margins have the highest CMI for Task $8$ and the second highest for Task $7$. This suggests that these tasks are inherently difficult when it comes to generalization prediction. This makes sense, as the models in these two tasks have the smallest variation in test accuracy between them. Recall Table~\ref{tab:pgdl_task_overview} in Section~\ref{sec:back_pgdl}. This suggests that current complexity measures (including ours) are not sensitive enough to account for such slight variations in performance between models.



Despite these limitations, we note that the resulting scatter plots of mean constrained margin versus test accuracy for all three of these tasks do not show points in the lower right (large margin but low generalization) or upper left (small margin but high generalization) quadrants. It is therefore possible that a larger mean constrained margin is always beneficial to a model's generalization, even though it is not always fully descriptive of its performance. 


\section{Variance and adversarial perturbations}
\label{sec:constrained_adv_dirs}

The question of `why do adversarial examples exist?' is an open problem in the field of adversarial machine learning. Many hypotheses have been put forward, yet the question remains mostly unanswered after a decade of research~\cite{dimpled_manifold}. An exhaustive discussion of the many hypotheses and ideas that have been investigated surrounding the existence of adversarial examples is outside our scope. Rather, we provide a brief summary of influential work. For further details, see relevant reviews on this topic such as \cite{adv_reasons_review_1} and \cite{adv_reasons_review_2}.

Goodfellow et al.~\cite{goodfellow_adversarial} attribute the existence of adversarial examples to the high linearity of DNNs. They state that this is in constrast to many initial speculations, which attributed the existence of adversarial examples to high \textit{non-linearity}. Their support for this linearity hypothesis is motivated by the success of the very simple fast-gradient sign method (FGSM)~\cite{goodfellow_adversarial} for generating adversarial examples. This method simply identifies a direction which maximizes the model's loss w.r.t. the attacked sample. The sample is then perturbed (using a set size) in that direction. They argue that such a simple method only works due to the linearity of the model.  Some time later, Goodfellow states that adversarial perturbations are likely in the directions of a different class's centroid~\cite{goodfellow_centroid_lecture}.\footnote{Note that here we cite a statement made during a lecture (as done in~\cite{dimpled_manifold}). Perhaps we need not take this single remark too seriously.} 

Tanay et al.~\cite{tanay_adv_manifold} argue that the linearity hypothesis is not sufficient to explain adversarial examples. Specifically, they show that 1) linear classifiers are often not susceptible to adversarial attacks, and 2) if adversarial examples are found in the linear case, they are visually very different from those found for DNNs (it is difficult to consider such samples `adversarial'). Instead, they motivate that adversarial examples are caused by a decision boundary close to the data manifold. Stutz et al.~\cite{stutz2019disentangling}, whom we have referred to earlier in Section~\ref{sec:constrained_intuition}, empirically show that adversarial examples leave the class manifold of the target sample (through the use of VAE-GAN approximated manifolds). However, they further show that on-manifold adversarial examples also exist, which they simply identify as generalization errors of the classification model. 

In line with above, Shamir et al.~\cite{dimpled_manifold} recently put forward the `dimpled manifold model'. This framework speaks to the shape of the manifold. In essence, they argue that different class samples cause opposing forces during training, which results in decision boundaries tightly `hugging' the data manifold in some regions. In other words, that small `dimples' are formed by the DNN's decision boundaries. 

Ilyas et al.~\cite{adversarial_not_bugs_Ilyas} take a feature-centric view, and argue that adversarial examples exist due to `non-robust features'. Specifically, a non-robust feature is some pattern contained within the data distribution that is highly predictive, yet not interpretable to humans. They provide extensive empirical evidence and are, in fact, able to show that a model trained \textit{solely} on adversarial examples (with the corresponding adversarial labels!) still generalizes on a standard, non-adversarial test set. This is a remarkable finding which implies that the model is able to generalize from the non-robust features alone. Furthermore, this also aligns well with our intuitions of spurious features in the input space (recall Section~\ref{sec:constrained_intuition}).

In addition to these theories, many authors have also found that adversarial examples are highly transferable across different models~\cite{universal_perturbations,goodfellow_adversarial,curvature_universal_relation,adversarial_not_bugs_Ilyas}. Specifically, a perturbation that fools one model is often able to fool another. This is usually referred to as \textit{adversarial transferability}.

In this section, we do not attempt to specifically challenge or confirm any of these arguments. Rather, we simply ascertain how adversarial examples are related to the principal component directions of the dataset in question.  
We have shown that input margins are generally not predictive of generalization, although, when measured in directions of high variance, they are. Given this, we now ask `in what directions are adversarial examples found?' We first describe how we investigate this in Section~\ref{sec:constrained_adv_dirs_setup}, before showing our empirical results in Section~\ref{sec:constrained_adv_dirs_results}.

\subsection{Approach}
\label{sec:constrained_adv_dirs_setup}

In this section, we describe how we express adversarial perturbations in terms of their principal component directions and how we visualize the resulting data. Specifically, we determine along which principal component directions samples are most perturbed when generating adversarial examples. 
Let $\mathbf{X} \in \mathbb{R}^{s \times n}$ denote the training data of $s$ samples with $n$ features. In addition, let $\mathbf{x}_i \in \mathbb{R}^n$ be one such sample and $\mathbf{\hat{x}}_i \in \mathbb{R}^n$ its point on the decision boundary corresponding to its standard input margin. Furthermore, let the full principal component matrix of the dataset be given by $\mathbf{P} \in \mathbb{R}^{n \times n}$, i.e. consisting of $n$ principal component vectors $\mathbf{p}  \in \mathbb{R}^n$. We then find the coefficients $\mathbf{B} = [\beta_1, \beta_2, ..., \beta_n]$, such that 
\begin{equation}
\label{eq:constrained_adv_direction_coeffs}
    \mathbf{\hat{x}}_i = \mathbf{x}_i + \sum_{j=1}^{n} \beta_{j}\mathbf{p}_{j}
\end{equation}
where $\mathbf{p}_{j}$ denotes the $j^{th}$ principal component, i.e. the $j^{th}$ row of $\mathbf{P}$.
To do this, we transform $\mathbf{x}_i$ and $\mathbf{\hat{x}}_i$ to the principal component coordinates. We then simply subtract these two vectors from each other, which provides us with the perturbation expressed in terms of the $\beta$-coefficients, which we denote by $\bm{\beta}_{adv}$.  More formally, let the principal component transform $\phi()$ be given by
\begin{equation}
    \phi(\mathbf{x}) = \mathbf{P} \mathbf{x}^T
\end{equation}
assuming $\mathbf{x}$ is a row vector from the matrix $\mathbf{X}$. We can then define our adversarial perturbation in terms of the coefficients of Equation~\ref{eq:constrained_adv_direction_coeffs} by subtracting the transformed version of $\mathbf{x}_i$ and $\mathbf{\hat{x}}_i$ 
\begin{equation}
    \bm{\beta}_{adv_i} =  |\phi(\mathbf{x_i}) - \phi(\mathbf{\hat{x_i}})|
\end{equation}
where $|\cdot|$ is the element-wise absolute value operation and $\bm{\beta}_{adv_i}$ a column vector.
The absolute value is used as we are not concerned with whether the sample is moved in a positive or negative direction with respect to a principal component for this analysis. Instead, we are interested only in the size of the perturbation along that principal component direction. 

We rescale  $\bm{\beta}_{adv_i}$ by dividing by its maximum value such that $\bm{\beta}_{adv_i} \in [0, 1]^n$, ensuring that these coefficients are comparable across different samples. Formally, $\bm{\beta}_{adv_i}$ is updated by
\begin{equation}
    \bm{\beta}_{adv_i} \leftarrow \frac{\bm{\beta}_{adv_i}}{\max(\bm{\beta}_{adv_i})}
\end{equation}
where $\max(\bm{\beta}_{adv_i})$ is the maximum element of $\bm{\beta}_{adv_i}$.
We do this for several ($t$) samples, and then construct a matrix $\mathbf{B}_{adv} \in \mathbb{R}^{t \times n}$ such that
\begin{equation}
    \mathbf{B}_{adv} = 
    \begin{bmatrix}
     \bm{\beta}_{adv_1}^T\\
     \bm{\beta}_{adv_2}^T\\
     ...\\
     ...\\
     \bm{\beta}_{adv_t}^T
\end{bmatrix}
\end{equation}

Armed with such a matrix, we can now determine how much each principal component contributes to the perturbations across all the samples. Specifically, for each component, we add up the values in its respective column, then divide by the total sum of all column sums. Formally, for principal component $j$, its `share of adversarial perturbation', $p_{share_j}$, is calculated as follows:
\begin{equation}
    p_{share_j} = \frac{\sum^{t}_{i=1}\mathbf{B}_{adv_{i, j}}}{\sum_{j=1}^{n}\sum^{t}_{i=1} \mathbf{B}_{adv_{i, j}}}
\end{equation}
here $\mathbf{B}_{adv_{i, j}}$ indicates the $i^{th}$ row and $j^{th}$ column of $\mathbf{B}_{adv}$. This provides a ratio that indicates how much of the total perturbation across all samples is along this principal direction. 

Finally, we can visualize this as a distribution. We plot $p_{share}$ for each principal component in descending order of the explained variance per component. This allows us to visualize how much each principal component contributes to the overall perturbation across samples, and how this changes across the principal component landscape. In the following section we make use of this technique to analyze adversarial perturbations of several models.

\subsection{Results}
\label{sec:constrained_adv_dirs_results}

\begin{figure}[H]
        \centering
       \includegraphics[width=0.49\linewidth]{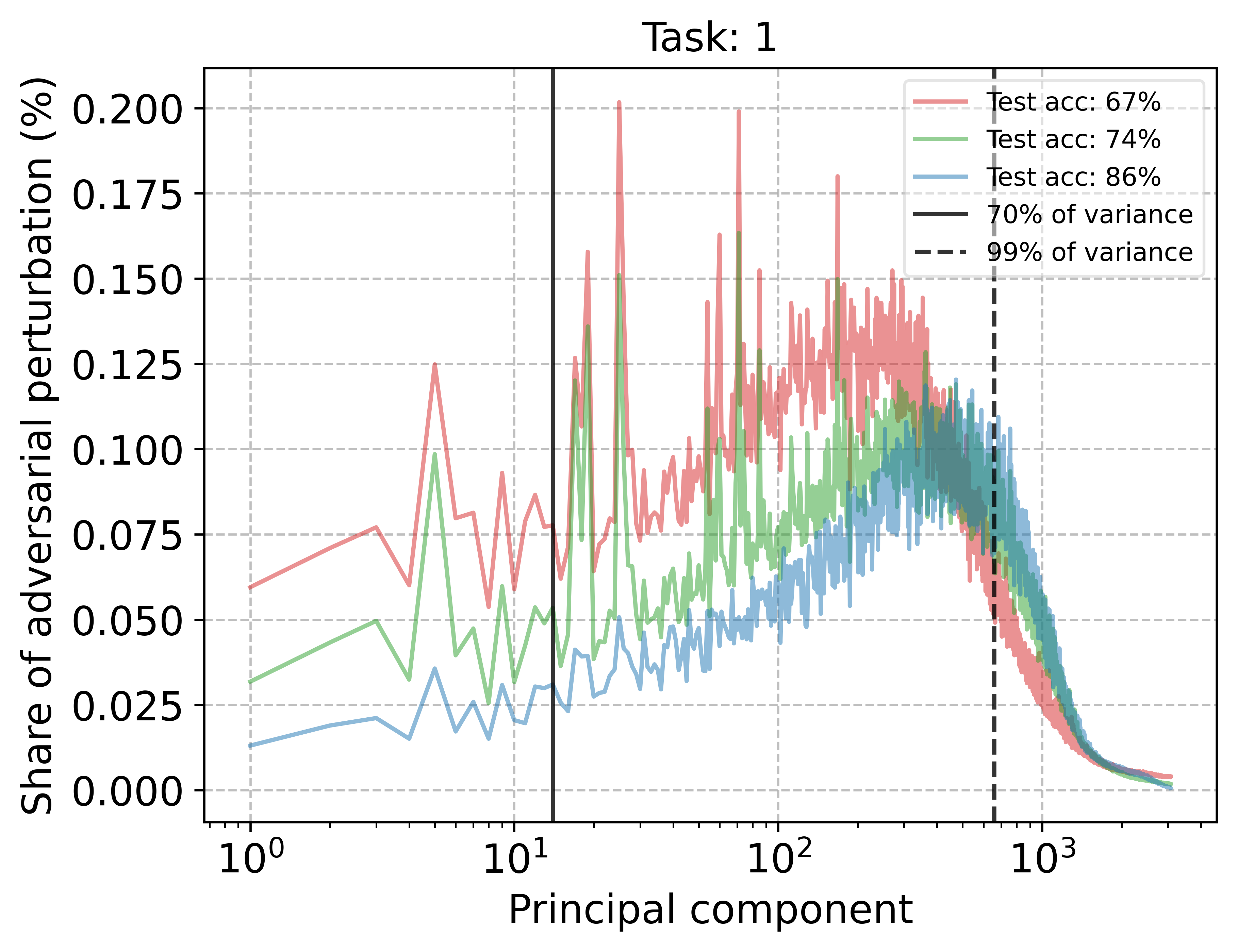}
        \includegraphics[width=0.49\linewidth]{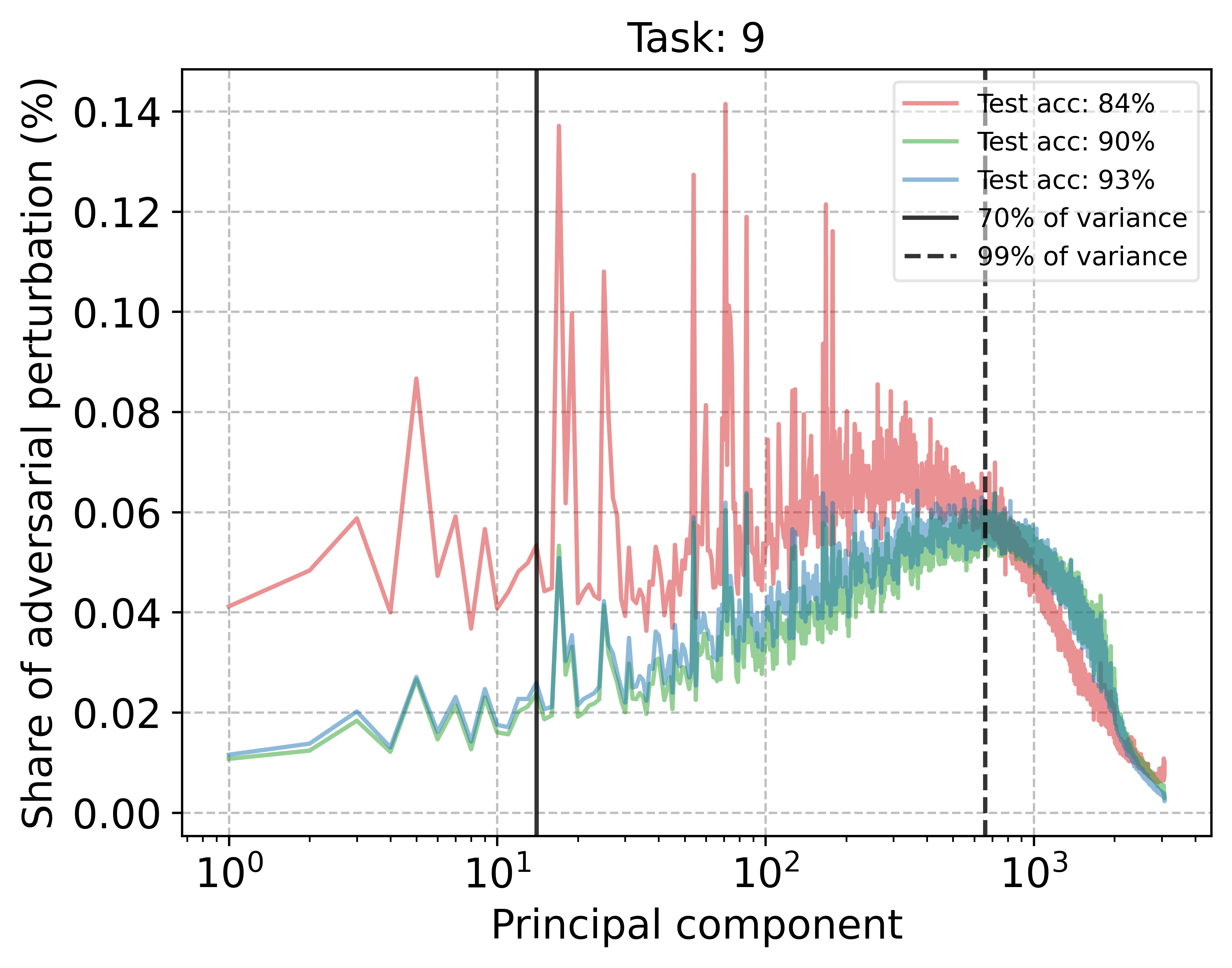} \\
        \includegraphics[width=0.49\linewidth]{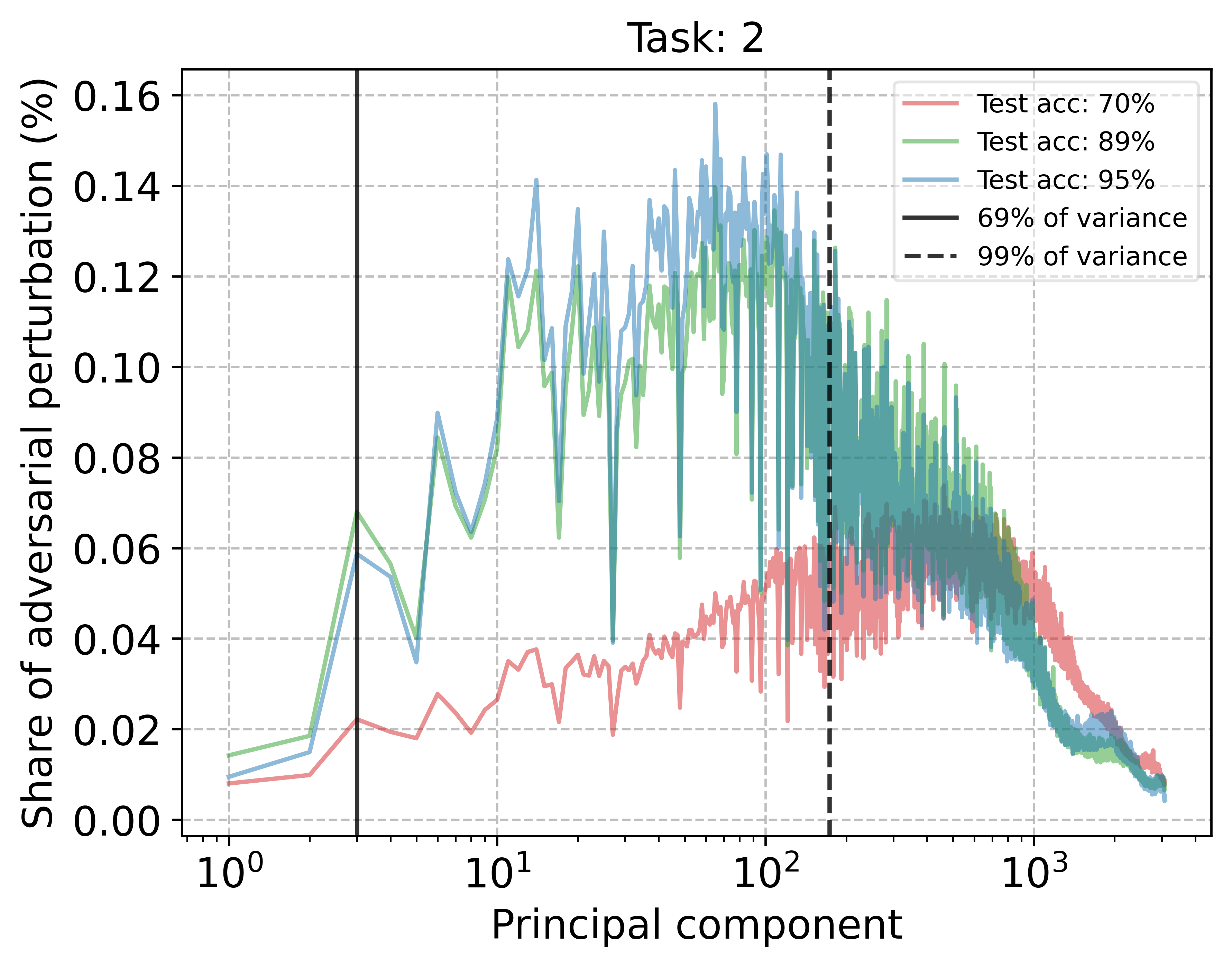}
        \includegraphics[width=0.49\linewidth]{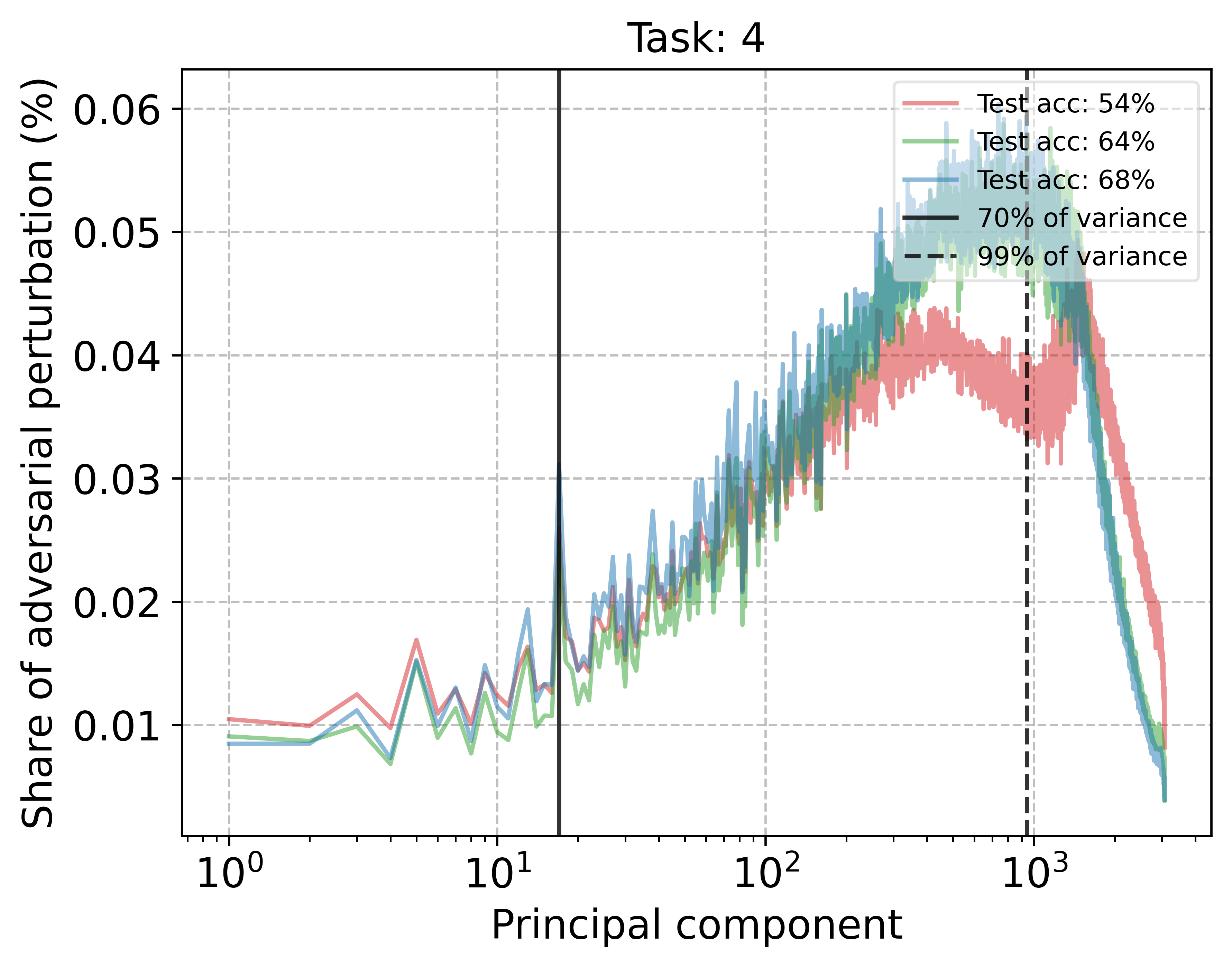}
        \caption[Share of adversarial perturbation per principal component for $3$ models per task, for $4$ tasks]{Share of adversarial perturbation per principal component for $3$ models per task, for $4$ tasks. From top-left to bottom-right, each plot refers to Tasks $1$, $9$, $2$, and $4$. Vertical black lines indicate the point at which the principal components capture $70\%$ (solid) and $99\%$ (dashed) of the variance in the data. Each plot for each task shows the distributions for the best (blue), middle (green), and worst (red) performing models.}
        \label{fig:constrained_adv_directions_model_comparison_lognorm_dists}
\end{figure}

\begin{figure}[H]
        \centering
       \includegraphics[width=0.49\linewidth]{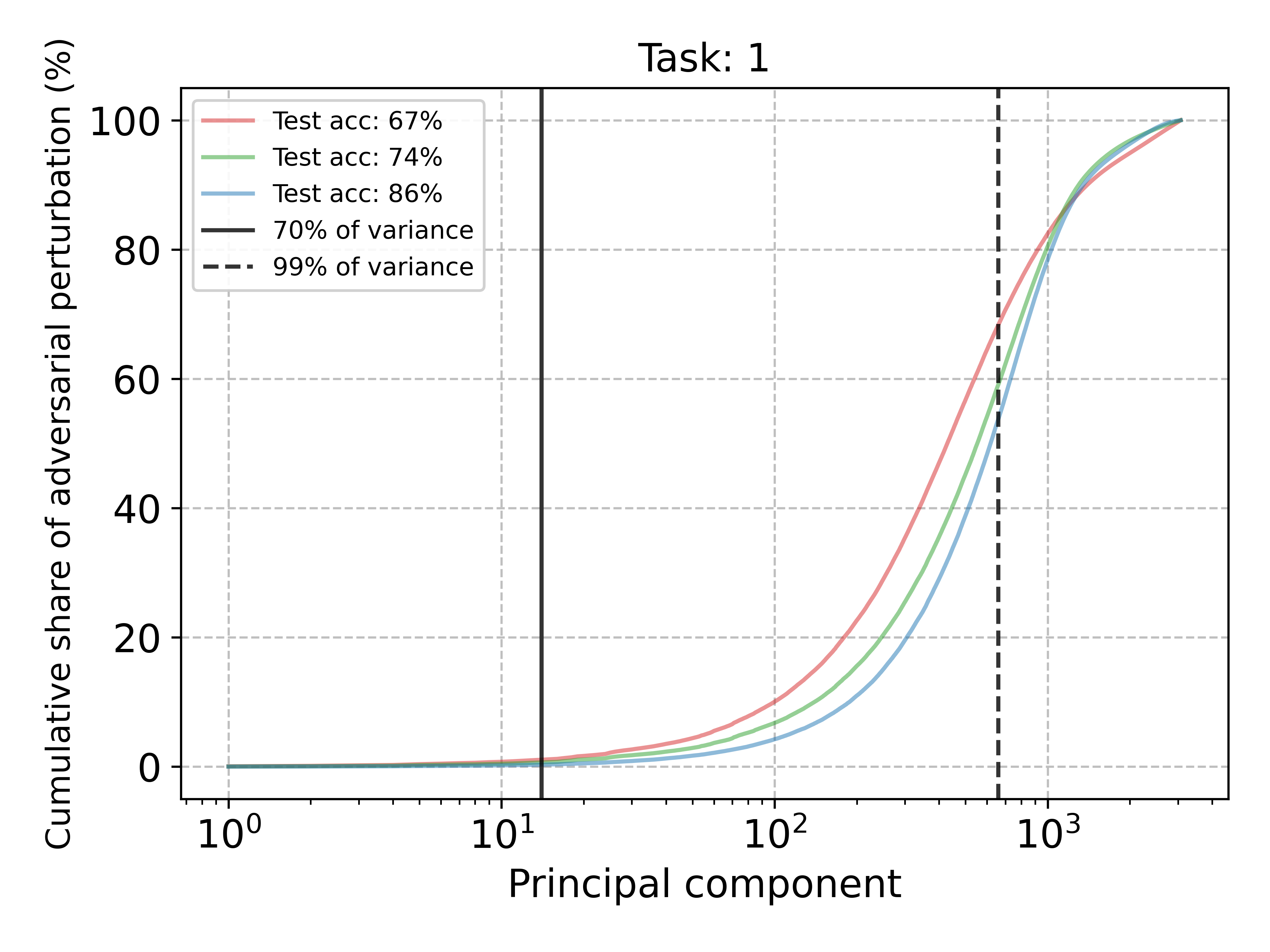}
        \includegraphics[width=0.49\linewidth]{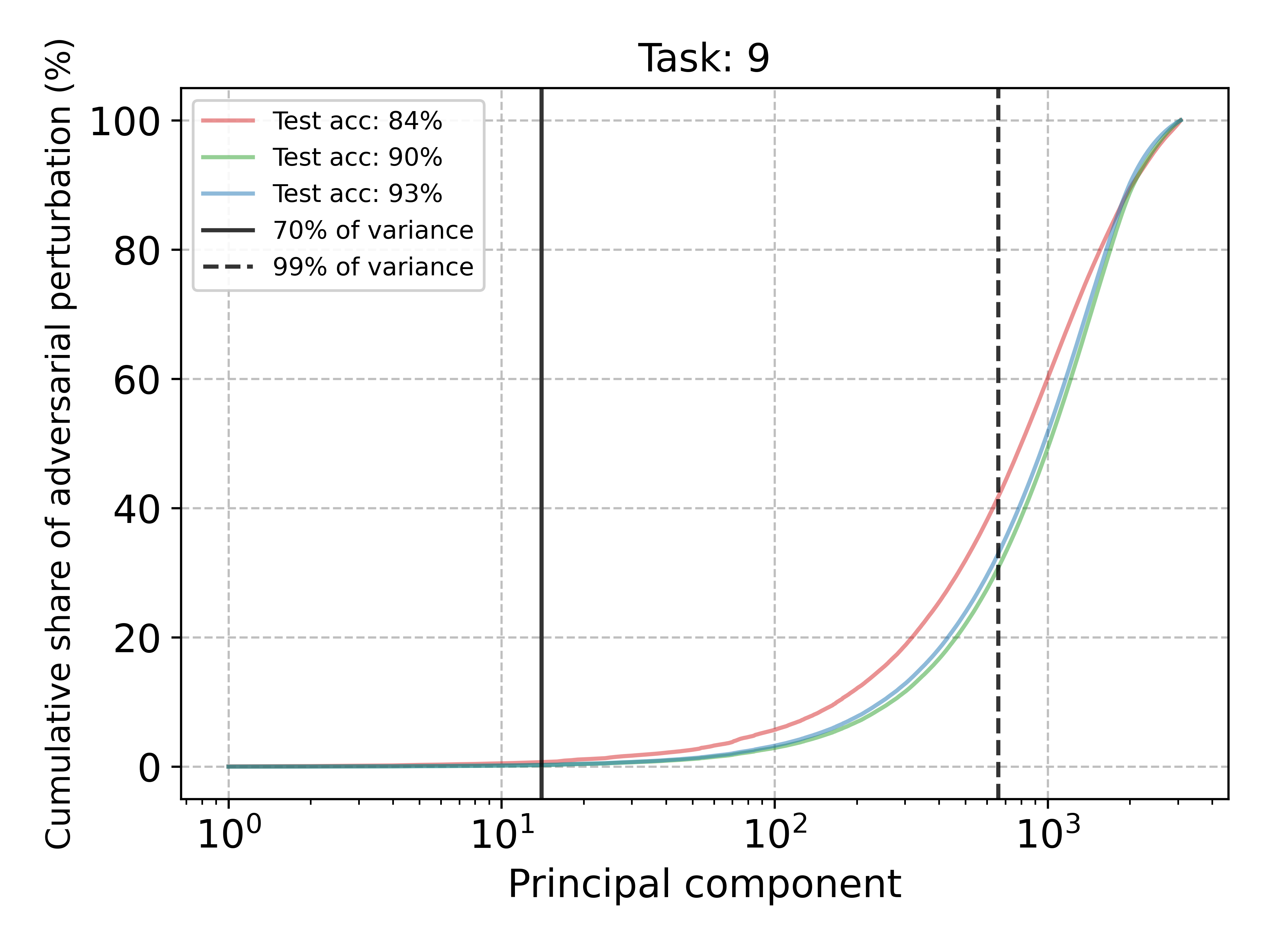} \\
        \includegraphics[width=0.49\linewidth]{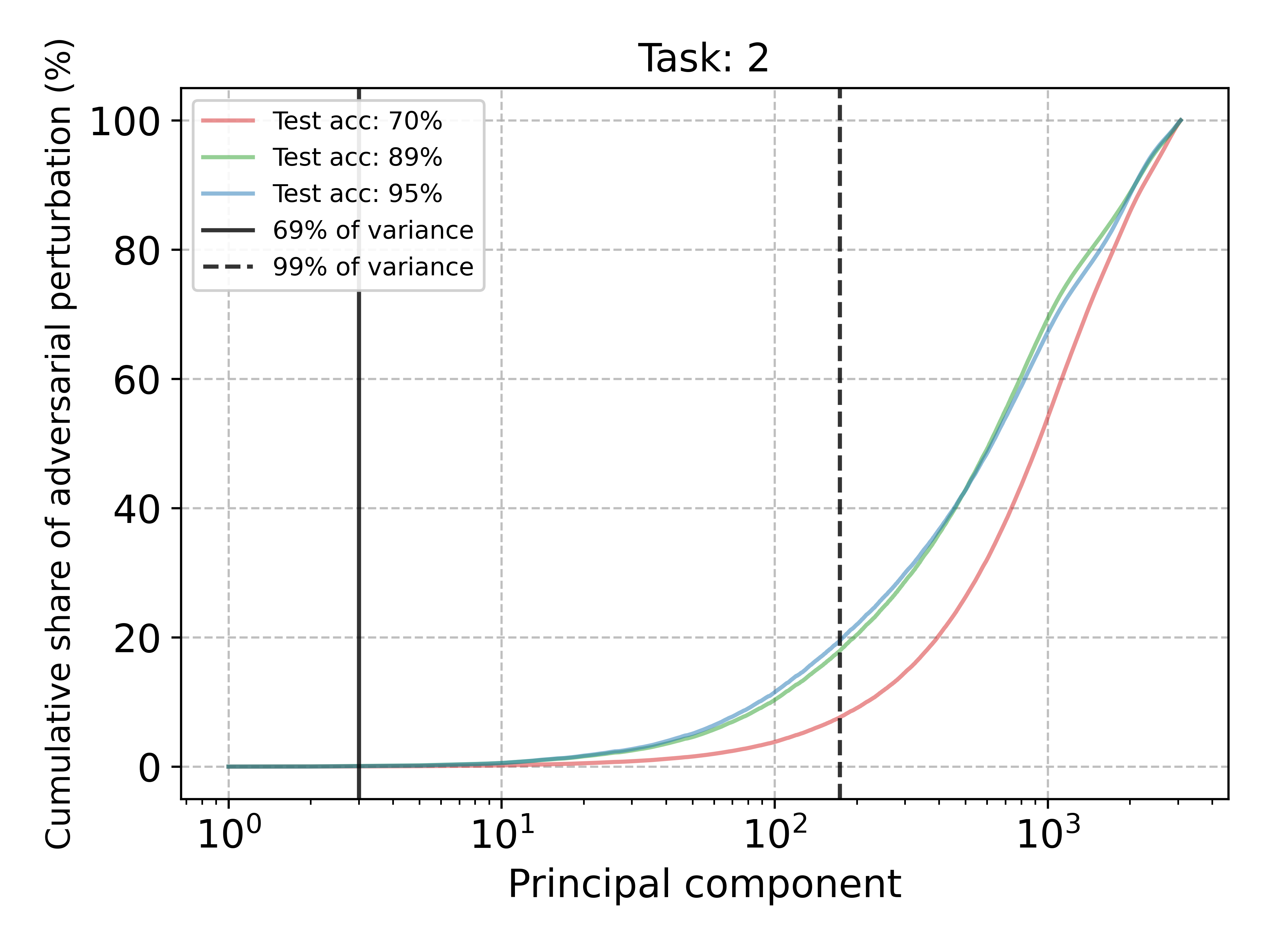}
        \includegraphics[width=0.49\linewidth]{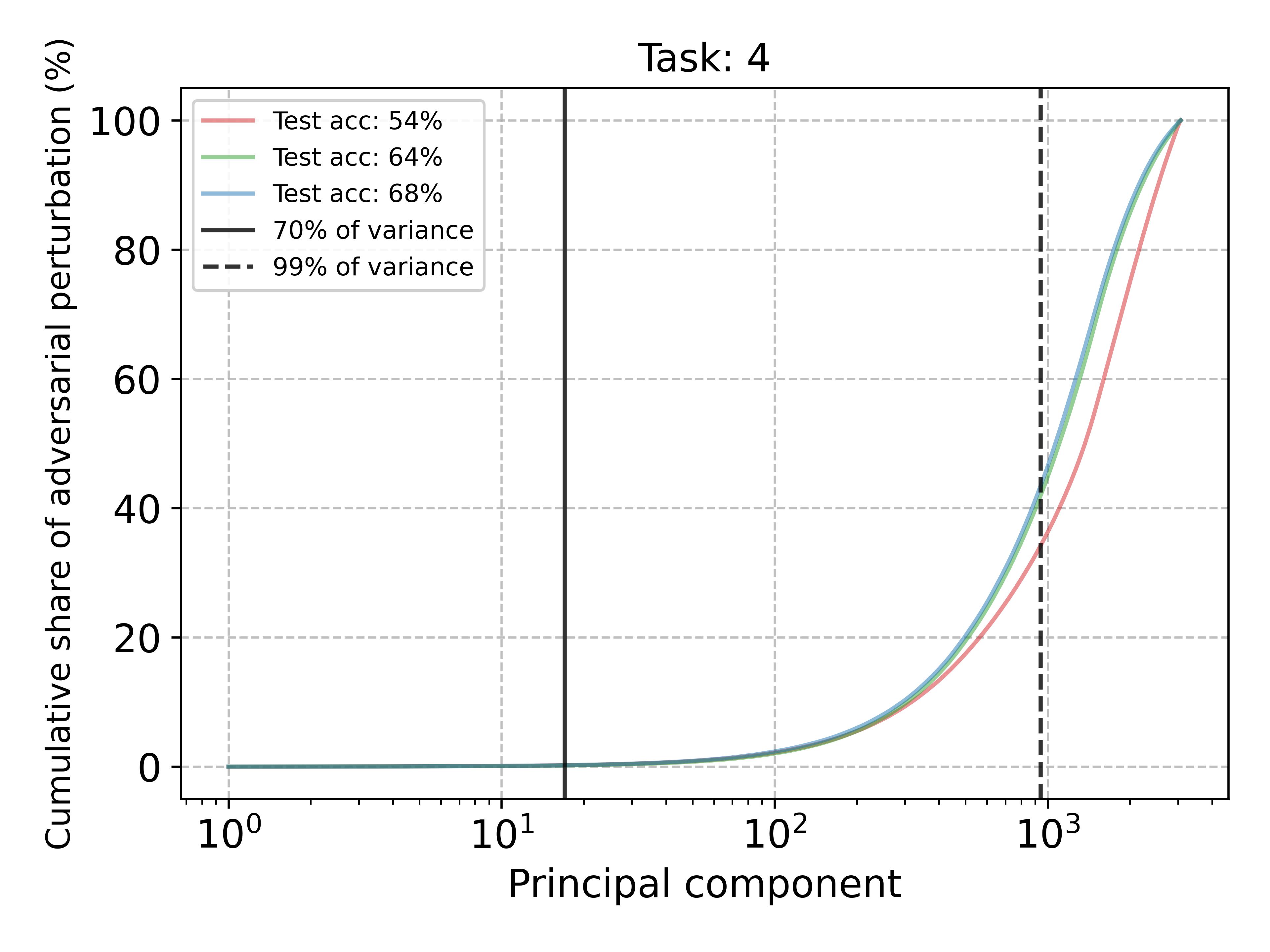}
        \caption[Same results as in Figure~\ref{fig:constrained_adv_directions_model_comparison_lognorm_dists} but visualized as a cumulative distribution]{Same results as in Figure~\ref{fig:constrained_adv_directions_model_comparison_lognorm_dists} but visualized as a cumulative distribution.}
        \label{fig:constrained_adv_directions_model_comparison_lognorm_dist_cumulative}
\end{figure}

For this investigation, we compare three models per task for four different tasks. We limit our analysis to distinct datasets with a large number of samples. More precisely, we calculate the $\mathbf{B}_{adv}$ matrix for the best, worst, and middle generalizing models from Task $1$ (CIFAR10), $2$ (SVHN), $4$ (CINIC10), and $9$ (augmented CIFAR10), each. Here `middle' refers to the model with the median test accuracy in the set of models. We use the standard input margin boundary points of $10\ 000$ samples to calculate $\mathbf{B}_{adv}$ in each case. These boundary points are calculated using the $\gamma=0.25$ variant of Algorithm~\ref{alg:deepfool_margin} (DeepFool). 
The resulting $p_{share}$  distribution for each model from each task is shown in Figure~\ref{fig:constrained_adv_directions_model_comparison_lognorm_dists}. Additionally, we also visualize each $\mathbf{B}_{adv}$ distribution as a cumulative distribution in Figure~\ref{fig:constrained_adv_directions_model_comparison_lognorm_dist_cumulative}, meaning each point shows the total sum of adversarial share up to the specified principal component. Furthermore, we also indicate the points at which the principal components capture $70\%$ and $99\%$ of the variance in the data as vertical lines (solid and dashed, respectively). 


Note that, in the following discussion of Figure~\ref{fig:constrained_adv_directions_model_comparison_lognorm_dists} and \ref{fig:constrained_adv_directions_model_comparison_lognorm_dist_cumulative}, we distinguish between different `groups' of principal components positioned relative to the solid and dashed lines. Specifically, we refer to the principal components to the left of the solid line as the `top principal components', those between the solid and dashed lines as the `middle principal components', and those to the right of the dashed line as the `lowest principal components'. Each group of components collectively explains approximately $70\%$, $29\%$, and $1\%$ of the variance, respectively.  Additionally, to distinguish the observations from the two different distribution visualization methods, we refer to the distributions shown in Figure~\ref{fig:constrained_adv_directions_model_comparison_lognorm_dists} as the `standard distributions' and those shown in Figure~\ref{fig:constrained_adv_directions_model_comparison_lognorm_dist_cumulative} as the `cumulative distributions'. 

We make the following observations for each group of principal components:
\begin{itemize}
    \item \textbf{Top principal components:} We find that the majority of the adversarial perturbations are not along the components that explain the most variance. In all observed cases, it is clear that the standard distributions peak to the right of the solid black lines. Similarly, it is clear from the cumulative distributions that the total share of adversarial perturbation up to the position of the solid black lines are very small (approximately $<1\%$).
    \item \textbf{Middle principal components:} It is clear from both the standard and cumulative distributions that the middle principal components contribute significantly to the adversarial perturbations. We observe that the standard distributions peak between the solid and dashed black lines, and similarly that the cumulative distributions show a sharp increase at these positions. We also observe that the standard distributions show several `spikes', i.e. components that make notably greater contributions to the adversarial perturbations than their neighbours.
    \item \textbf{Lowest principal components:} The standard distributions show that these components do not individually contribute significantly to the adversarial perturbations, i.e. each component does not contribute a large share to the adversarial perturbation on its own. However, the cumulative distributions show that, collectively, these components  contain a very large share of the overall perturbation. We find that the lowest principal components generally contribute between $40\%$ to $80\%$ of the total perturbation, depending on the model and task.
\end{itemize}

In addition to these observations for each group of principal components, we also note the following:
\begin{itemize}
    \item There is some variation in behaviour between tasks. For example, relative to the number of components, the standard distributions are centered more to the left for two models in Task $2$ (bottom-left). Interestingly, the cumulative distributions show that the middle principle components contribute less to the overall perturbation for models in this task compared to the others.
    \item Within each task, the three investigated models have a similarly distributed share of adversarial perturbation (for both the standard and cumulative visualizations). This is with the exception of the worst (red) models in Tasks $9$, $2$, and $4$, which show slightly different behavior.
\end{itemize}

These observations are very interesting. We make three primary conclusions:
\begin{enumerate}
    \item Adversarial perturbations are not along the directions of highest variance, i.e. along the top principal components.
    \item The middle principal components significantly contribute to adversarial perturbations, and have the highest contribution per component in comparison to the top and lowest principal components.
    \item The lowest principal components do not significantly contribute to the adversarial perturbations individually, but due to there being so many, they collectively contribute a large share to the adversarial perturbations.
\end{enumerate}

These conclusions strongly support our constrained margin approach, and shows that standard input margins are not measured in directions of high variance. 

We now consider how these findings align with the multiple theories that have been put forward for the existence of adversarial examples.

\begin{itemize}
    \item \textbf{Perturbation are towards a class centroid}~\cite{goodfellow_centroid_lecture}: One would expect a perturbation that moves towards a class centroid to be along a high-variance direction. However, our findings with respect to the top principal components suggests that this is not the case. 
    \item \textbf{Adversarial examples are off-manifold}~\cite{tanay_adv_manifold,stutz2019disentangling}: If one assumes that the principal components are a good approximation of the underlying manifold, our results support the theory that adversarial examples are off-manifold. We find that the top principal components contribute very little to the adversarial perturbations, i.e. the directions that can be considered on-manifold. This theory of adversarial examples also strongly supports our intuitions surrounding constrained margins (recall Section~\ref{sec:constrained_intuition}).
    \item \textbf{The manifold is dimpled}~\cite{dimpled_manifold}: Our findings do not directly speak to the shape of the manifold. However, given that the middle principal components appear to contribute more individually to adversarial perturbations than the other principal components, it is possible that these are the directions in which the hypothesized `dimples' are found. 
    \item \textbf{Adversarial perturbations rely on non-robust features}~\cite{adversarial_not_bugs_Ilyas}: We cannot confirm this theory, as it is unclear whether these perturbations in the middle and lowest components rely on highly predictive yet uninterpretable features. However, we can speculate that one would expect the high-variance directions to capture features that are more inline with human notions of interpretable features, while those in the middle and lowest components less so. Therefore, the perturbations along the middle and lowest principal components are potentially well aligned with these non-robust features.
    \item \textbf{Adversarial transferability}: Our findings align well with the notion of adversarial transferability. Given that the directions of perturbation are highly similar for the different models (with the exception of the worst generalizing models in some cases), it is natural to assume that the perturbation for one model along these directions would also fool another.
\end{itemize}

In conclusion, we have related adversarial perturbations to the variance of the training data. We find that generally these perturbations are not along the high-variance directions in the input space, and that this finding is somewhat aligned with prior work on the existence of adversarial examples. However, further investigation is necessary to better understand these links.

\section{Conclusion}

We have shown that constraining input margins to high-utility subspaces can significantly improve their predictive power i.t.o. generalization. Specifically, we have used the principal components of the data as a proxy for identifying these subspaces, which can be considered a rough approximation of the underlying data manifold.

Let us summarize our main findings:
\begin{enumerate}
    \item Constraining the search for a point on the decision boundary to a subspace spanned by the highest-variance principal components increases the predictive performance of input margins. When considering the average performance on the PGDL tasks for Taylor-approximated margins in the input space, we see an increase from $4.08$ CMI for standard input margins to $17.58$ CMI for constrained margins. Similarly, for the DeepFool variant, we see an increase from $5.70$ to $28.13$ CMI. In comparison to hidden margins, constrained margins also perform better on average on the PGDL tasks. We observe an average of $16.80$ CMI for Taylor-approximated hidden margins versus $17.58$ for Taylor-approximated constrained margins. For the DeepFool margin complexity measures, we observe a CMI of $16.66$ for hidden margins versus $28.13$ for constrained margins. See Table~\ref{tab:constrained_margin_complexities_cmi} in Section~\ref{sec:constrained_marg_comp}.
    \item The more accurate DeepFool margin-estimation method improves the scores, on average, for constrained margins in comparison to the Taylor approximation. We observe an average CMI of $17.58$ for the Taylor approximation versus $28.13$ for the DeepFool method. Only Task $5$ shows reduced performance with the DeepFool method (a CMI of $5.42$ for Taylor versus $4.80$ DeepFool). See Table~\ref{tab:constrained_margin_complexities_cmi} in Section~\ref{sec:constrained_marg_comp}.
    \item The DeepFool constrained margin complexity measure outperforms all other complexity measures on the PGDL test set, on average. We find an average test set CMI of $27.19$ for constrained margins versus $24.18$ for the closest competitor. See Table~\ref{tab:constrained_pgdl_results} in Section~\ref{sec:constrained_main_pgdl_results}.
    \item The subspace spanned by the highest-variance principal components is a necessary aspect of constrained margins, and predictive performance suffers if the search is constrained in other ways for Tasks $1$ and $6$. See Figure~\ref{fig:constrained_high_v_low_task_1_6} in Section~\ref{sec:constrained_high_versus_low_util}.
    \item Points found on the decision boundary when calculating constrained margins are qualitatively different than those of standard input margins. See Section~\ref{sec:constrained_qualitative_comparison}.
    \item Constrained margins still have several limitations, such as: failures on certain hyperparameters (e.g. depth), potential scenarios where high variance does not correspond to high utility, and tasks on which the metric performs poorly. See Section~\ref{sec:constrained_limitations}.
    \item Adversarial perturbations are generally in the direction of the middle and lowest principal components. Specifically, we observe that the majority of perturbation lies along principal directions which collectively explain approximately $30\%$ of the variance in the data, not the top components which together explain $70\%$ of the variance. See Figures~\ref{fig:constrained_adv_directions_model_comparison_lognorm_dists} and \ref{fig:constrained_adv_directions_model_comparison_lognorm_dist_cumulative} in Section~\ref{sec:constrained_adv_dirs}.
\end{enumerate}

In conclusion, we propose constraining input margins to make them more predictive of generalization in DNNs. It has been demonstrated that this greatly increases the predictive power of input margins and also outperforms hidden margins and several other contemporary methods on the PGDL tasks. This method has the benefits of requiring no per-layer normalization, no arbitrary selection of hidden layers, and does not rely on any form of surrogate test set (e.g. data augmentation or synthetic samples). However, this method is still subject to certain limitations.

In the following chapter, we take a holistic view of the observations and findings made throughout this study. We discuss these findings and others made previously in more detail.



%% file: chapters/ch6_conclusion.tex
\let\cite\parencite

\lhead{}
\rhead{}
\chapter{Conclusion}
\label{chap:conclusion}
\afterpage{\lhead{\ifthenelse{\thepage>0}
       {\it Chapter \thechapter }
      }
\rhead{\ifthenelse{\thepage>0}
       {\it \let\uppercase\relax\leftmark}
      }}
\underline{ \hspace{\textwidth} } 
\textit{''It was over in a moment and the folks had gathered round \\
There before them lay the body of the outlaw on the ground \\
Oh, he might have went on living but he made one fatal slip \\
When he tried to match the ranger with the big iron on his hip''} \\ - Marty Robbins, \textit{Big Iron}, Verse 8\\
\underline{ \hspace{\textwidth} }

\section{Overview}

Throughout this study, we have made many observations. In this chapter, we consolidate these findings, discuss these results, and also speculate on new questions that have been raised. 

The layout is as follows. We first discuss our key findings and their implications in Section~\ref{sec:conclusion_key_findings}. We then identify key questions for future work in Section~\ref{sec:conclusion_dicussion}. Our closing remarks are contained in Section~\ref{sec:conclusion_in_closing}.

\section{Key findings and implications}
\label{sec:conclusion_key_findings}

In this section, we take a holistic view of all of the observations we have made throughout this study. We then discuss the main findings and their implications. 

\subsection{Constrained margins}

In this study, we have observed that standard input margins are generally not predictive of generalization in most cases. We find that this is true in the numerical prediction setting as well as when considering model ranking. 
However, after a single non-linear transformation, these same measurements at the first hidden layer (hidden margins) are better correlated with generalization, in some settings. 
Despite this improved performance, the varying characteristics of these hidden representations (dimensionality, scale, number of layers) make it difficult to compare these margins between different models. 
It was shown that normalization approaches proposed and demonstrated in one setting, may fail in another. 
Therefore, we introduce `constrained margins' which do not present such concerns. We show that by constraining the search for a point on the decision boundary to a principal component subspace centered around a training sample, input margins are generally much more predictive of generalization and a more robust measurement overall.
As models are compared in the input space -- which is identical for all models -- no normalization is required. 

These core observations lead to the following thesis: \textit{Margin measurements require an appropriate distance metric to be predictive of generalization in DNNs.} We know that constraining the search to a warped subspace and using Euclidean distance to measure closeness is equivalent to defining a new distance metric on the original space. This implies that, ultimately, we are seeking a relevant distance metric to measure the closeness of the decision boundary. Understanding the requirements for such a metric remains an open question. However, the positive results achieved with the current PCA-and-Euclidean-based approach of constrained margins have numerous implications.

First, this shows that the data manifold needs to be taken into account for future investigations between margins and generalization. We have provided ample evidence that considering standard input margins alone does not generally elucidate much of a model's performance on unseen data. 

Second, given that the constrained margin metric is limited to the input space, it provides a more human-interpretable way of probing generalization than hidden margin measurements.

Third, we believe that constrained margins have the potential of being a useful regularizer, in line with how other complexity measures have been used in the past. Several authors have developed techniques to maximize standard input~\cite{xu2023exploring} and hidden~\cite{large_margin_dnns} margins during training; however, these have not resulted in improved generalization. Given that constrained margins have a closer relationship with generalization than these techniques, it is possible that constrained margins are well suited for this purpose.

Finally, we have improved upon the current state-of-the-art (w.r.t. the PGDL challenge) for generalization prediction, with all of the benefits that entails. (Recall our list of the many uses of complexity measures in Section~\ref{sec:intro_generalization}.) Additionally, we have done so without relying on generating synthetic data, as is common with the other solutions. This implies that our metric is not subject to the additional complexity introduced by such synthetic data.

\subsection{Measuring margins}

In this study, we investigated numerous ways of measuring margins. Specifically, we investigated using a simple first-order Taylor approximation, an iterative variant of this approximation (DeepFool), and a constrained optimization formulation. We also compared these different methods and identified their advantages and disadvantages.

We show that the first-order Taylor approximation, while relatively inexpensive, can significantly over- or underestimate the margin (recall Section~\ref{sec:hidden_df_to_taylor_comp}). This is in contradiction to earlier views on the matter~\cite{constrained_optim_investigating}. Furthermore, we also modified the DeepFool adversarial attack~\cite{deepfool} into a margin-finding algorithm, Algorithm~\ref{alg:deepfool_margin}, which is able to outperform a constrained optimization baseline (find smaller margins) at a fraction of the computational cost. While this method is based on DeepFool, this algorithm seems to be uniquely adept at accurately estimating margins when a small learning rate is introduced (along with the other changes we make, such as stopping criteria and clipping, see Section~\ref{sec:hidden_verifying_deepfool_alg}). 
We have not compared this to other modern techniques such as the Fast Adaptive Boundary (FAB) attack, but perhaps this finding on the importance of learning rate can lead to the design of improved approximation methods.  This is an important consideration for future investigations of margins and adversarial robustness in DNNs. 

In a somewhat contradictory fashion to that stated above, we also notice that the accuracy of a margin-approximation method does not seem to be very important for the purpose of comparing summary statistics. Recall that at the end of Chapter $3$ (Section~\ref{sec:noise_hidden_margins}) we observe that the mean first-order Taylor-approximated margins give very similar results to the mean constrained optimization margins. Furthermore, we show in Chapter $4$ (Section~\ref{sec:hidden_df_pgdl_results}) that the predictive performance of standard input and hidden margins is not improved when using a more accurate margin-finding method. This implies that, at least for some use cases, one need not be too concerned about the accuracy of the estimation method. This is interesting, as much of the literature surrounding classification margins is primarily concerned with the arms race of developing better margin-finding methods.

\subsection{Adversarial robustness}

Although an investigation of adversarial robustness is not the primary goal of this study, our findings concerning input margins contribute to a better understanding of adversarial examples. 
%
The adversarial robustness of a classification model is measured in different ways throughout the literature. Most commonly, the model's accuracy is evaluated on test set samples perturbed up to a specific perturbation bound~\cite{pgd_attack,deepfool}. This is not how we investigate robustness in this study. Rather, if one assumes that a larger standard input margin, as measured on training data, implies greater adversarial robustness (as assumed in multiple studies~\cite{xu2023exploring,large_margin_dnns,robustness_acc_TRADES}) we can assert the following.

First, we show that there is no consistent positive or negative relationship between adversarial robustness and generalization when comparing different models trained on the same dataset in the PGDL setting. Specifically, consider our findings on the relationship between standard input margins and generalization for the PGDL tasks. In Table~\ref{tab:hidden_hidden_input_taylor_v_deepfool}, for all three approximation methods, we find that for two tasks there is a slight negative correlation between large input margins and test accuracy. On the other hand, for two others there is a stronger positive correlation. For the remaining tasks, there is only a mild positive correlation. These findings may explain the contradictory literature surrounding the topic (recall Section~\ref{sec:back_input_margins}). These results show that the nature of the relationship between robustness and generalization depends on the specific models and dataset being considered -- there seems to be no general rule. That said, the reasons for this inconsistent behavior are unclear.

We also investigate the relationship between sample noise and standard input margins in Chapter~\ref{chap:ch3}. The most significant finding with respect to adversarial robustness is that the presence of label-corrupted samples leads to a reduction in the margins of clean samples in the same model (see Figure~\ref{fig:noise_margin_means_all_models} in Section~\ref{sec:noise_mean_margin}).\footnote{This finding was also observed in two other works~\cite{twice_dd_real_ver,label_noise_adversarial}. These studies were conducted concurrently with our study described in \cite{missing_margin}.} This implies that one could improve a classifier's adversarial robustness by ensuring clean training data. Given that mislabeled samples are commonly found in real-world datasets~\cite{imagenet_noisy_labels,o2u_noisy_labels}, this deserves consideration. 

Furthermore, we also find in Chapter~\ref{chap:ch3} that input-corrupted samples induce unique margin behaviors in the various models considered (see Figure~\ref{fig:noise_margin_means_all_models} in Section~\ref{sec:noise_mean_margin}). This is an important finding, as no previous work has analyzed the relationship between margins and off-manifold samples. Therefore, we contribute to a better understanding of how different types of samples affect the adversarial robustness of DNNs. 

Finally, we also relate adversarial perturbations to the principal components of the dataset (in Section~\ref{sec:constrained_adv_dirs} of Chapter~\ref{chap:ch5}). We find that, mostly, adversarial perturbations are along the middle and lowest principal components. Specifically, we find that for most of the models considered, the directions in which these perturbations occur are very similar: generally those principal directions that explain approximately $30\%$ of the variance in the data, and not the top ranked principal components that explain the most variance. This is an important finding. Principal components are a more interpretable dimensionality reduction (or manifold modelling) technique than those previously considered in the literature for analyzing adversarial examples in relation to the data manifold. This is a useful consideration for further investigations on the existence of adversarial examples. Furthermore, this finding also sheds some light on why adversarial perturbations are highly transferable between different models, as most models (within the same task) seem to be vulnerable in the same principal component directions. 


\section{Future work}
\label{sec:conclusion_dicussion}

In this section, we identify and speculate on remaining questions, as well as new questions that the constrained margin framework now allows us to probe.

\subsection{Improving constrained margins}

We must not forget that the constrained margin complexity measure we have presented is not without limitations.  In Section~\ref{sec:constrained_limitations} we discuss these. We identify the following questions that require consideration.

\begin{itemize}
    \item \textbf{Would a better manifold approximation measure improve performance?} A natural next step is to establish whether the predictive performance of constrained margins can be improved if a more advanced manifold approximation measure is used. This is unfortunately difficult as it is unclear how to combine our approximation methods (the first-order Taylor approximation or DeepFool) with non-linear dimensionality reduction techniques and requires further consideration.
    \item \textbf{What effect does batch normalization have on a model's decision boundaries?} We observe, in Table~\ref{tab:constrained_margin_complexities_kendall}, that constrained margins perform poorly on Task $5$ (trained without batch normalization) but relatively well on Task $4$ (trained with batch normalization). We believe that understanding how batch normalization affects a model's decision boundaries would be the first step in establishing the reason behind this behavior. 
    \item \textbf{What effect does depth have on a model's decision boundaries?} Similar to above, we note in Section~\ref{sec:constrained_limitations} that models of different depth show different behavior when considering the relationship between constrained margins and test accuracy.
\end{itemize}

\subsection{Why are hidden margins more predictive of generalization?}

We have provided a strong argument and evidence for the question, `why are standard input margins not predictive of generalization?' (recall Sections \ref{sec:constrained_intuition} and \ref{sec:constrained_results}). However, we have not answered `why are hidden margins more predictive?' (than standard input margins).  We suspect that a plausible reason for this is that the non-linear transformations of the hidden layers act as a `feature selector'. These transformations are done in such a way that Euclidean measurements in this space are more aligned with the underlying data manifold of the input space. 

To investigate the correspondence between the hidden and input space, one could simply map points found on the decision boundary in the hidden space back to the input space. These points can then be visually inspected to determine whether there is truth to this hypothesis. Very recent work by Youzefsadeh~\cite{yousef_paths} shows that in the case of MLPs, it is possible to map points in the hidden space back to points in the input space. However, we have thus far been unable to reproduce this\footnote{This is preliminary work and not shown in this study.} for the CNNs that we investigated throughout this study. 

Similarly to the question posed above, we have also not answered `why are standard input margins \textbf{sometimes} predictive of generalization'. Recall that we find a stronger correlation between standard input margins and test accuracy for two PGDL tasks (see Table~\ref{tab:hidden_hidden_input_taylor_v_deepfool} in Section~\ref{sec:hidden_df_pgdl_results}) compared to the other tasks. If our view surrounding input margins is considered, i.e. that of spurious features (recall Section~\ref{sec:constrained_intuition}), this finding implies that the model's generalization is well characterized by the model's separation of regions in these directions. Why this is the case for some tasks but not others is an interesting consideration for future work.

\subsection{Why do some samples have smaller margins than others?}

Recall that in Chapter $3$ (specifically, Section~\ref{sec:noise_analysis}) we find a weak relationship between the margin of each sample and its Euclidean distance to the nearest sample of a different class (i.e. its max margin). However, we do not satisfactorily answer why some samples have smaller margins than others. On the other hand, at the end of Chapter $5$ (Section~\ref{sec:constrained_adv_dirs_results}) we find that standard input margins (i.e. adversarial examples) are mostly measured in the directions of the middle and lowest principal components. It is thus possible that the proximity of the samples to each other in these directions has a greater effect on their respective margins than Euclidean distance over the entire space. This is an important consideration for gaining a better understanding of how noisy samples influence the margin, and thus generalization and robustness behavior, of DNNs. We are eager to investigate this hypothesis.

\section{In closing}
\label{sec:conclusion_in_closing}

In conclusion, we have investigated margin-based generalization prediction methods. Our core thesis is that an appropriate distance metric is required to relate margin measurements to generalization. In this study, we have gained a better understanding of the requirements for such a metric.  However, there is still much left unanswered. 

And now, dear reader, our time together comes to an end. It is our hope that you have enjoyed margins as much as we have.

%% file: chapters/appendix.tex
\appendix{}

\chapter{Supplemental content}

\section{Appendix: Chapter 4} 
\label{app:ch4}

This appendix contains additional results and explanations relevant to Chapter~\ref{chap:ch4}.

\subsection{Linear model coefficients}
\label{app:hidden_lin_model_coeffs}

This section is relevant to Section~\ref{sec:hidden_lin_reg}. 
In Table~\ref{tab:hidden_r_squared_all_signatures} we fit linear regressor models using $5$ summary statistics from the margin distributions of each layer. This resulted in some of the models overfitting the training data for some of the tasks. Thereafter, in Table~\ref{tab:hidden_r_squared_two_signatures} we reduced the number of input features by only using two summary statistics: the third quartile ($Q_3$) and upper fence. Here follows the rationale behind selecting these two statistics.

\begin{figure}[H]
    \centering
    \includegraphics[width=0.3\linewidth]{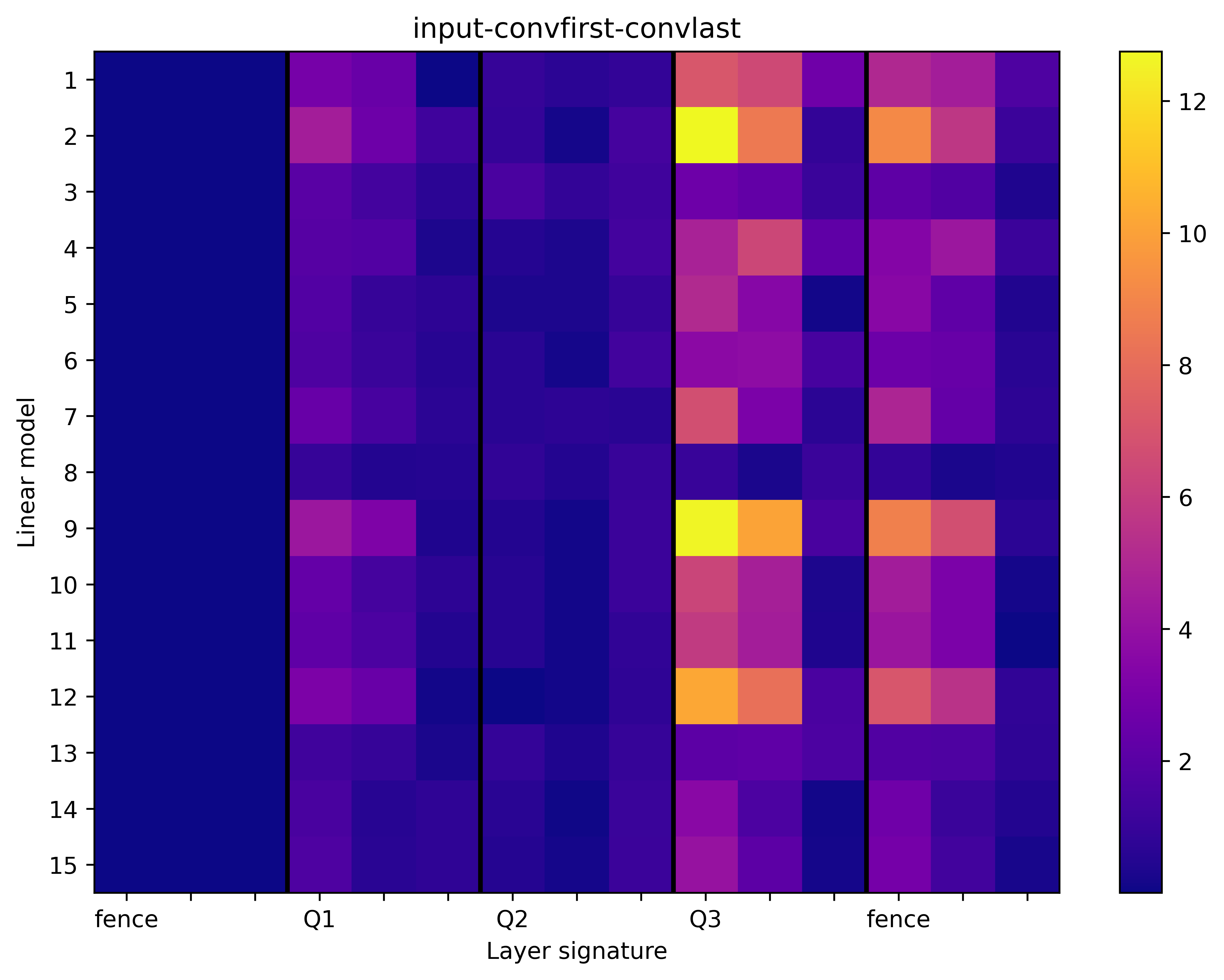}
    \includegraphics[width=0.3\linewidth]{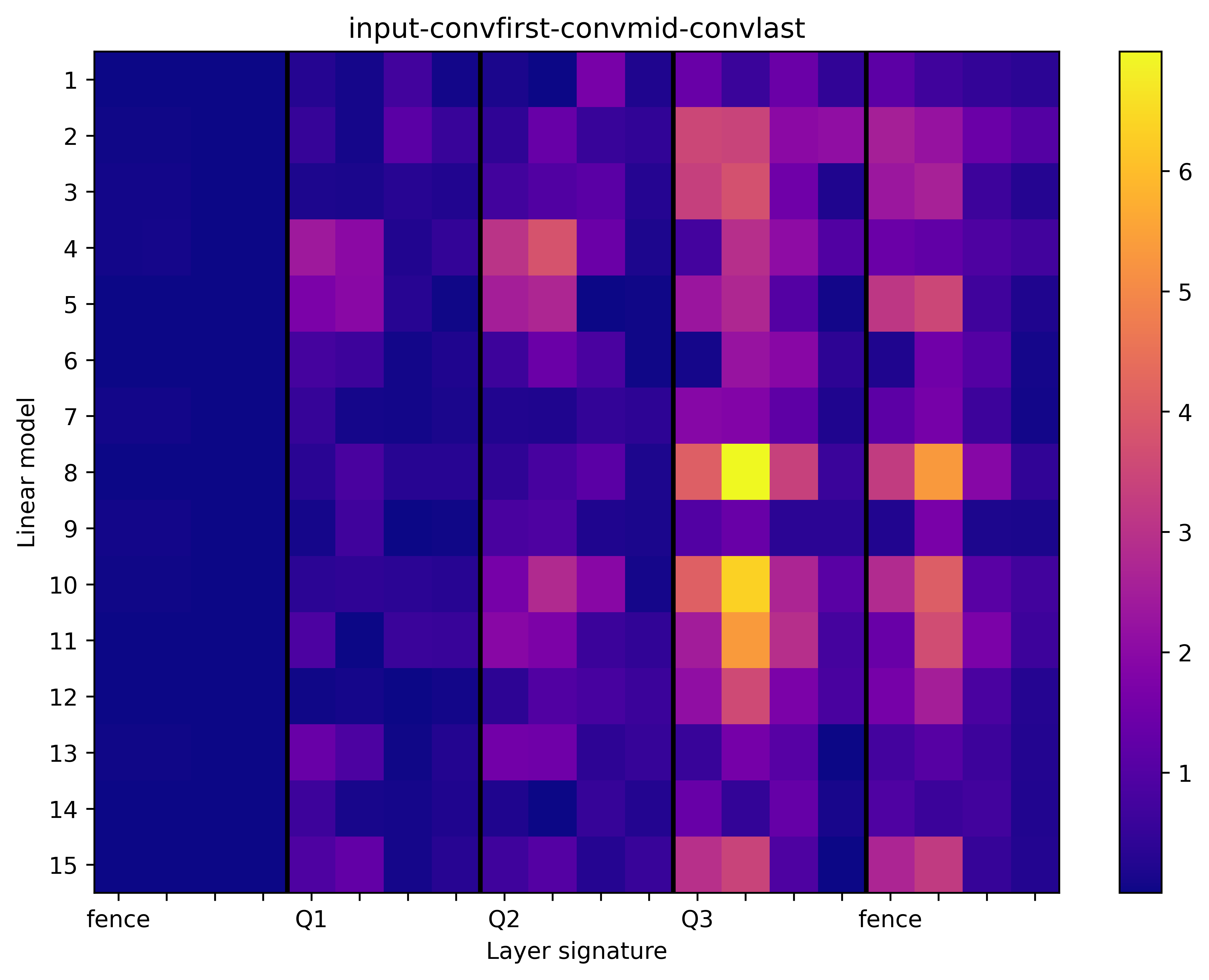}
    \includegraphics[width=0.3\linewidth]{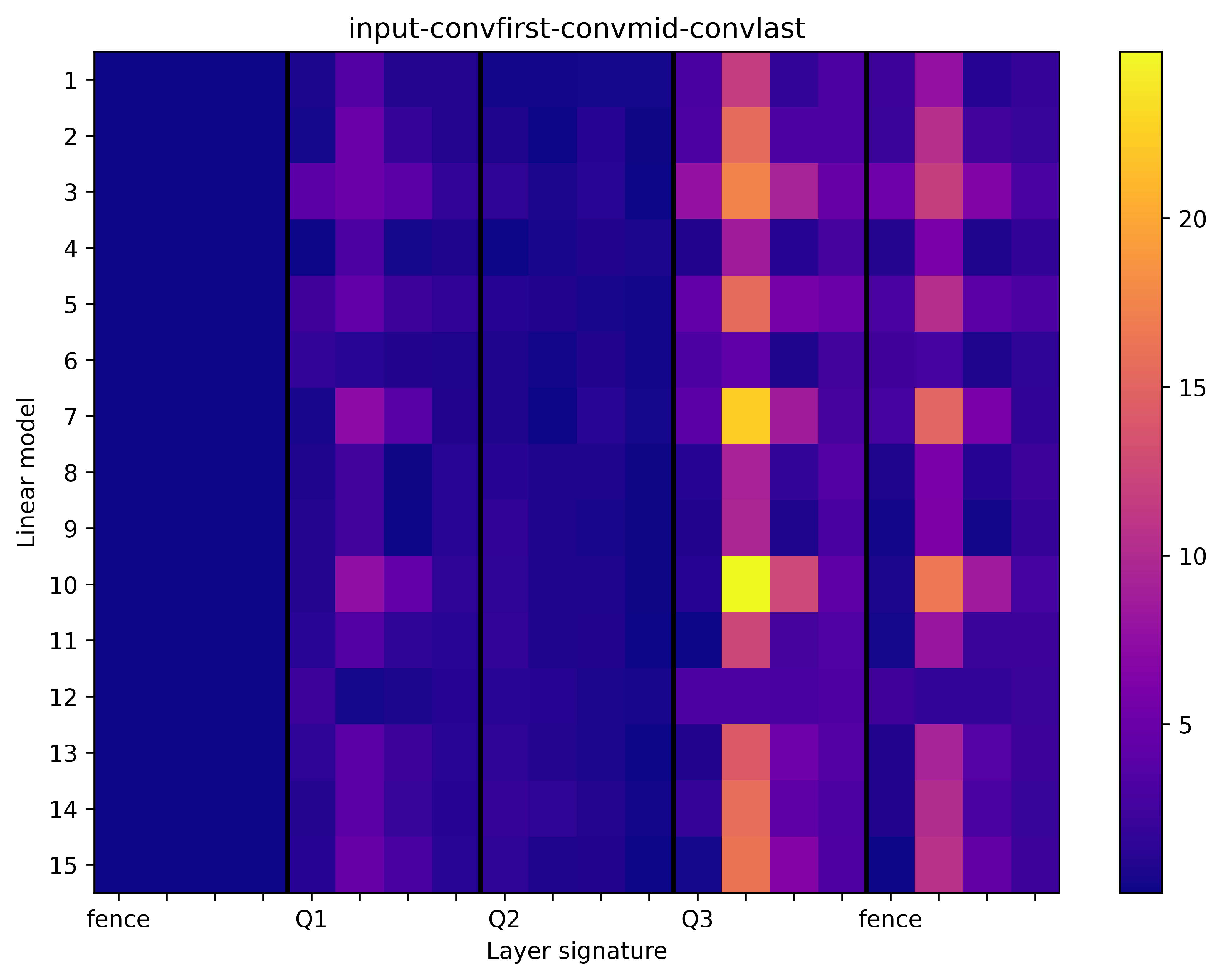}
    \includegraphics[width=0.3\linewidth]{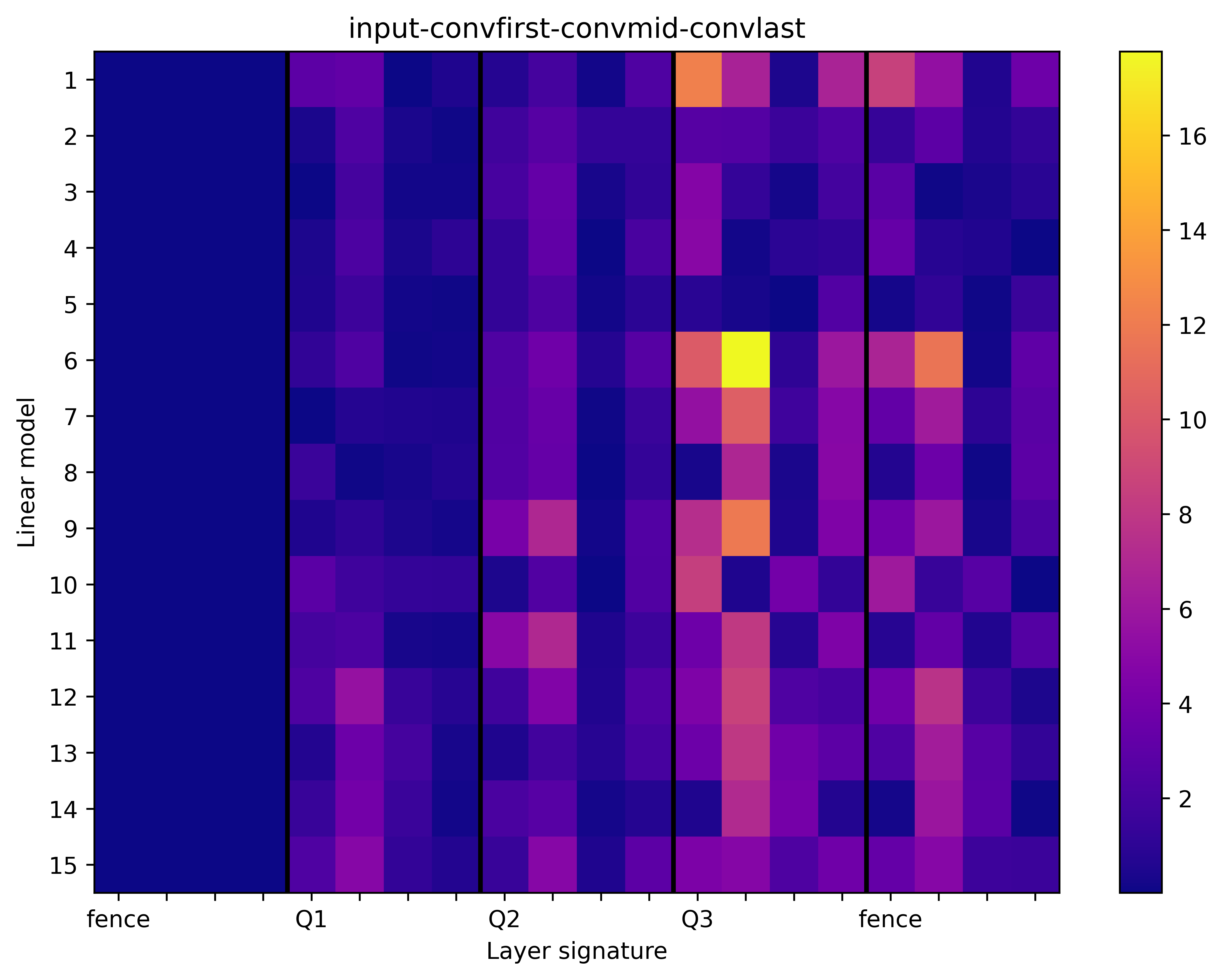}
    \includegraphics[width=0.3\linewidth]{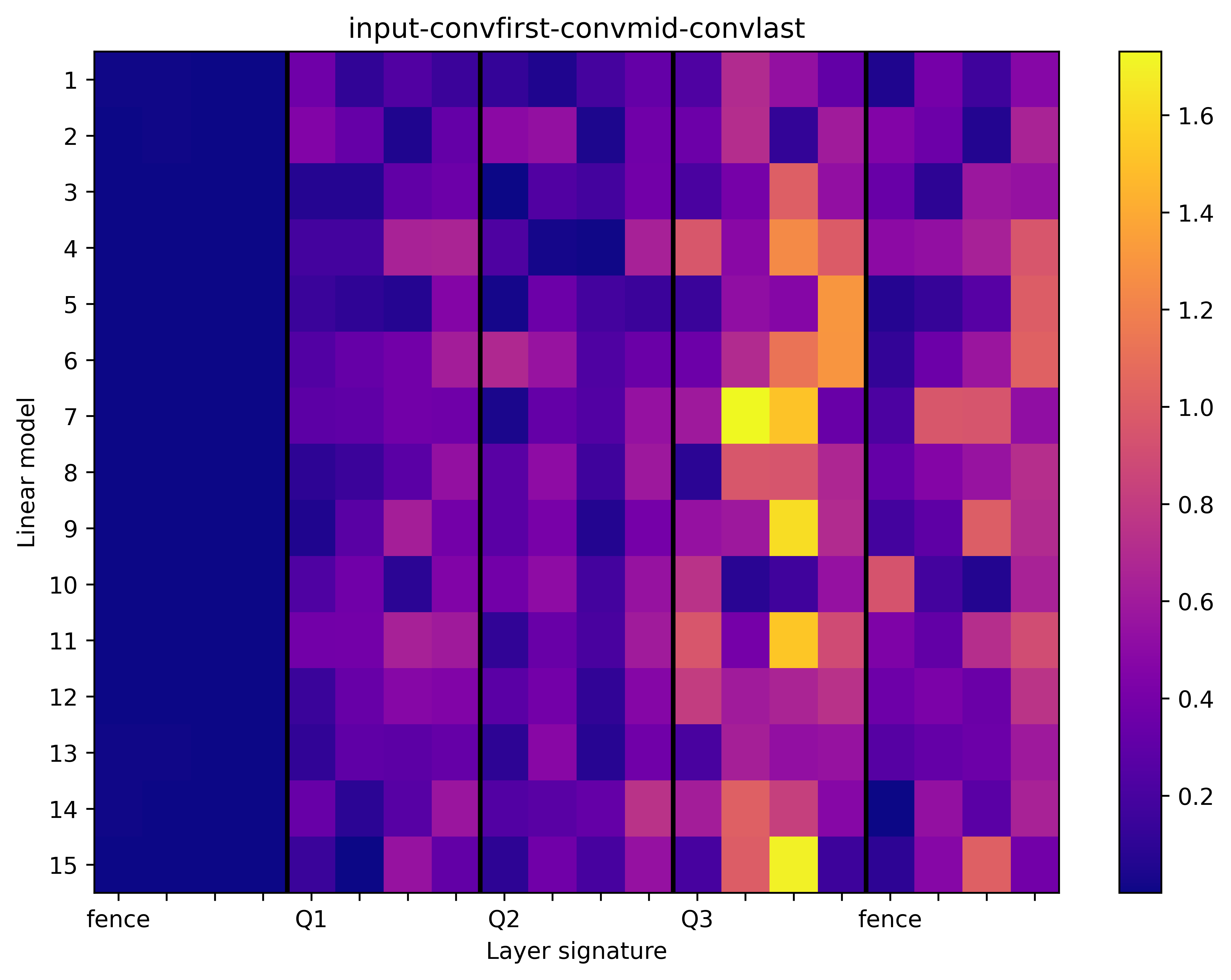}
    \includegraphics[width=0.3\linewidth]{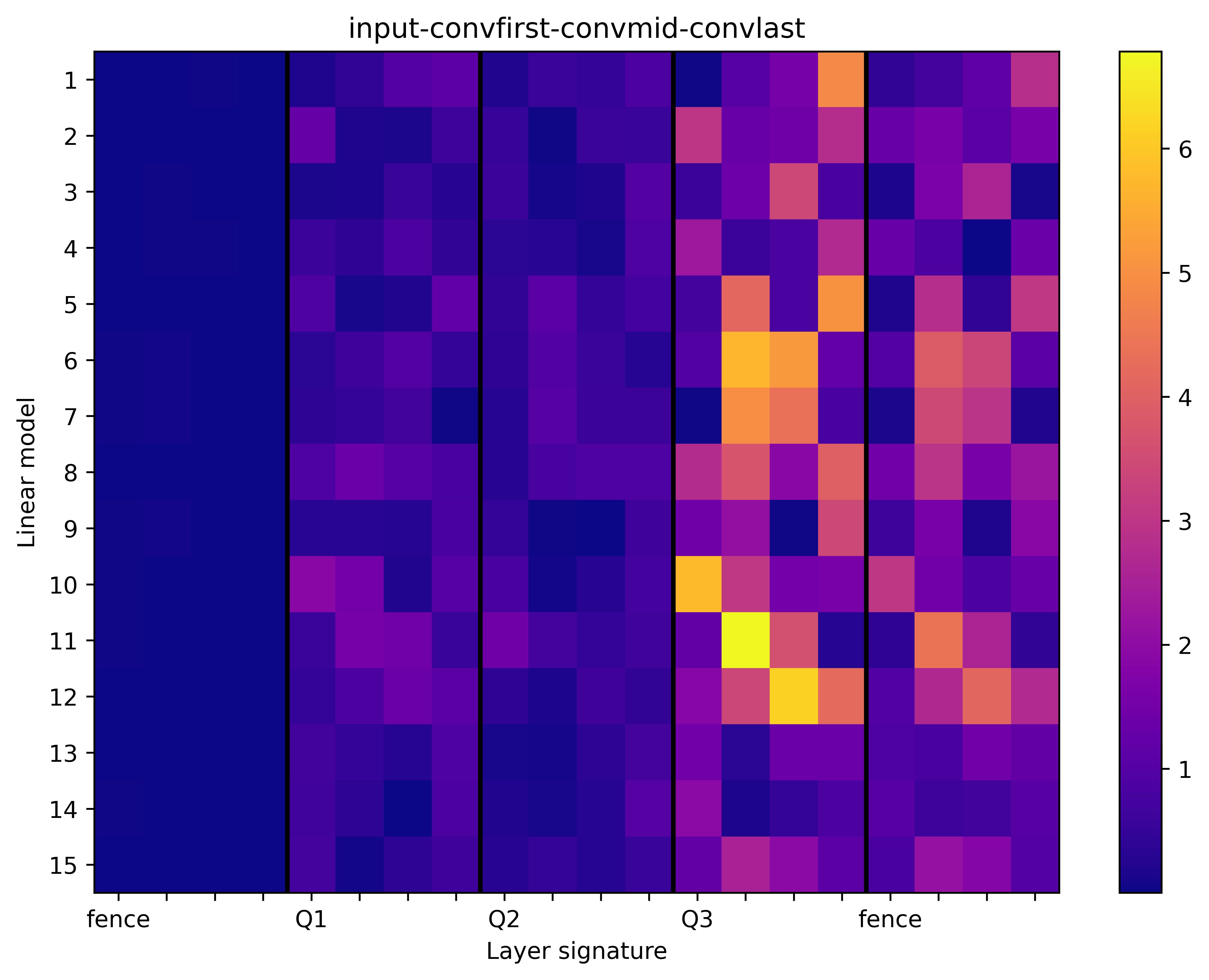}
    \includegraphics[width=0.3\linewidth]{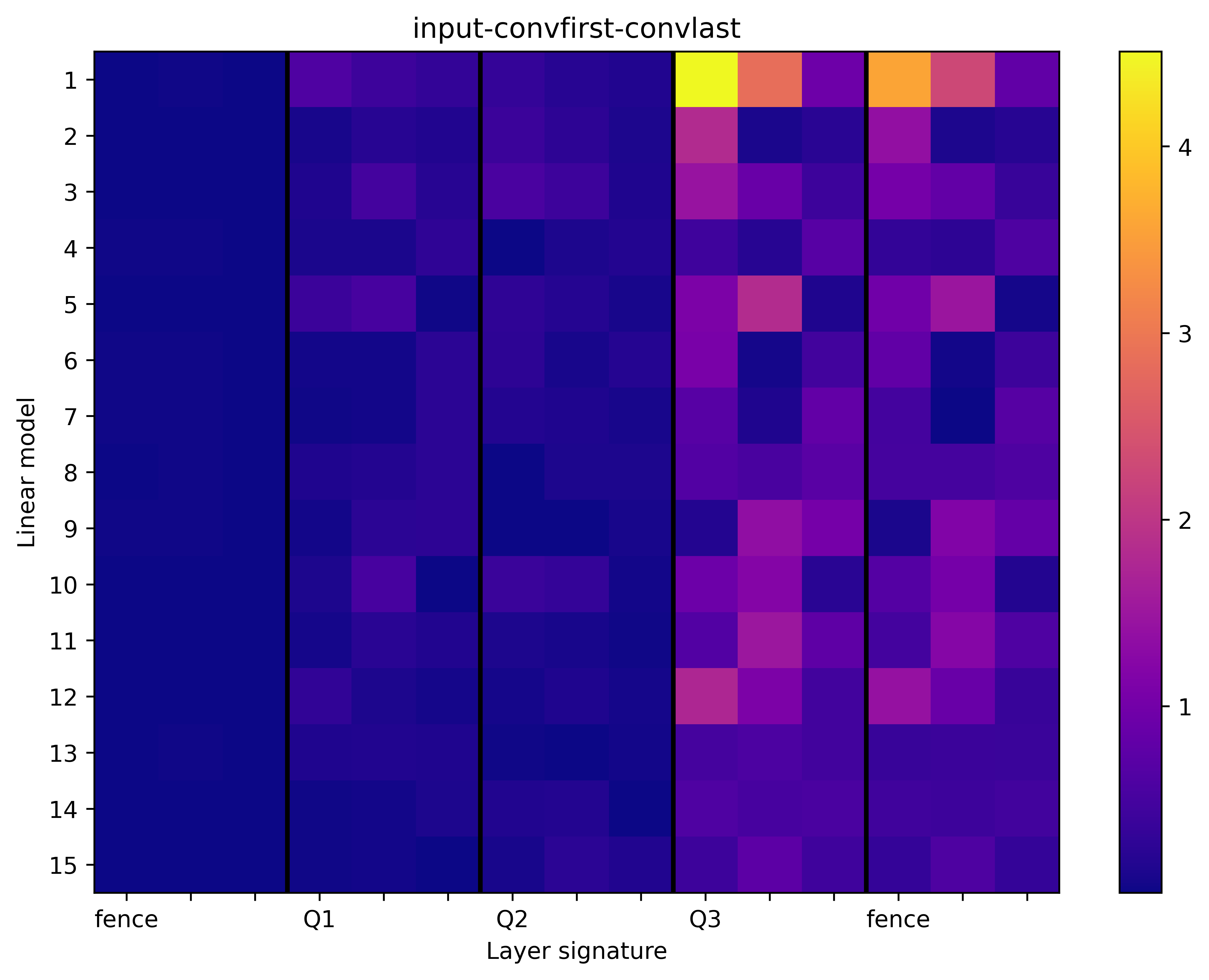}
    \includegraphics[width=0.3\linewidth]{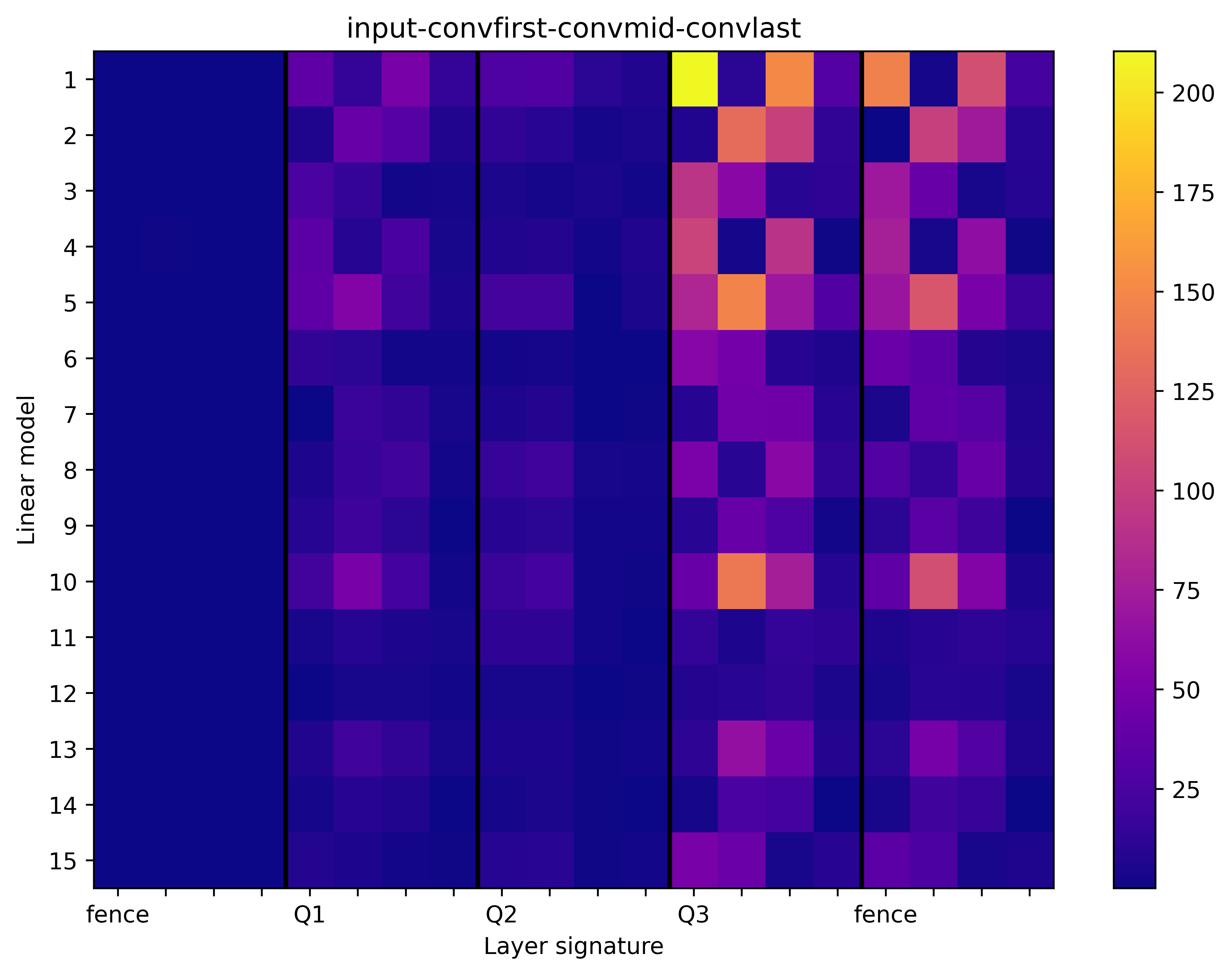}
    \caption[Learnt coefficients of the linear regression models for all the PGDL tasks]{The absolute value of the learnt coefficients of the three-folds $\times$ five-seeds linear regression models for each task in the PGDL set. Models correspond to the `Hidden w/ input' variation in Table~\ref{tab:hidden_r_squared_two_signatures}. Text above each plot indicates the ordering of the layers within each summary statistic's columns. From left to right. Top: Tasks $1$, $2$, $4$. Middle: Tasks $5$, $6$, $7$. Bottom: Tasks $8$, $9$.}
    \label{fig:app_hidden_regression_coeff}
\end{figure}

In Figure~\ref{fig:app_hidden_regression_coeff} we visualize the absolute value of the coefficients of the fitted linear models. Specifically, we show the coefficients of the `Hidden w/ input' variation when using five summary statistics (the middle column of Table~\ref{tab:hidden_r_squared_all_signatures}). We do this for all $15$ linear models of each task, i.e. three-folds $\times$ five-seeds. Each row corresponds to a single linear model. The horizontal ticks show the signature statistics corresponding to each column. The order of layers within each statistic's columns can be read from the plot titles. For example, for Task $1$, the first three columns correspond to the coefficient of the lower fence for the input, first convolutional layer, and last convolutional layer, in that order. The following three columns show the same for the first quartile, in the same order, and so forth. We include black bars between each summary statistic's columns for easier visualization.

We observe that the first three statistics are seldom used to predict generalization, and the majority of the weighting corresponds to the last two ($Q_3$ and upper fence). Therefore, restricting the linear models to only these two summary statistics seems a logical choice.

\subsection{Additional verification of Algorithm~\ref{alg:deepfool_margin}}
\label{app:hidden_additional_verification}

\begin{figure}[h]
    \centering
    \includegraphics[width=0.49\linewidth]{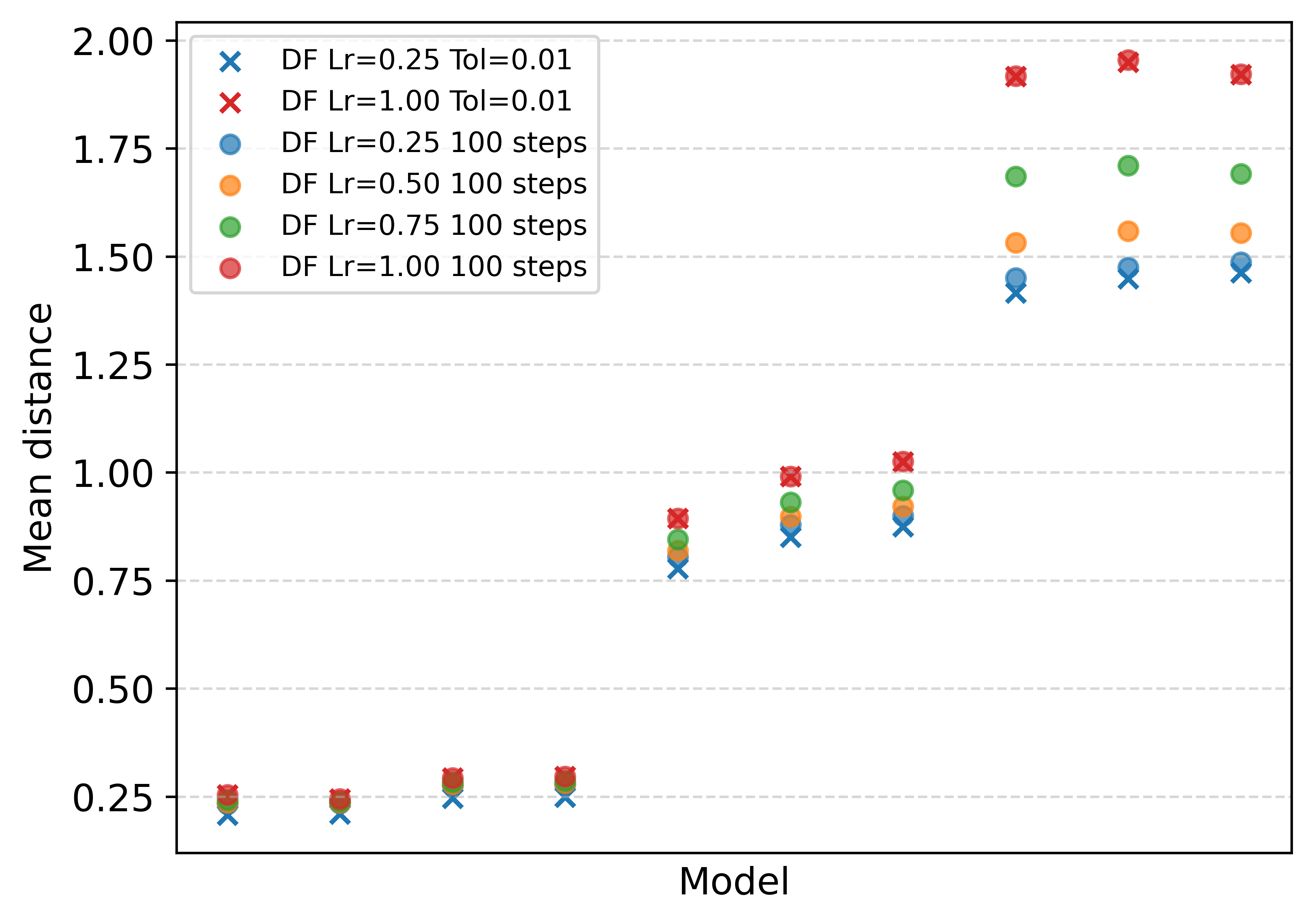}
    \includegraphics[width=0.49\linewidth]{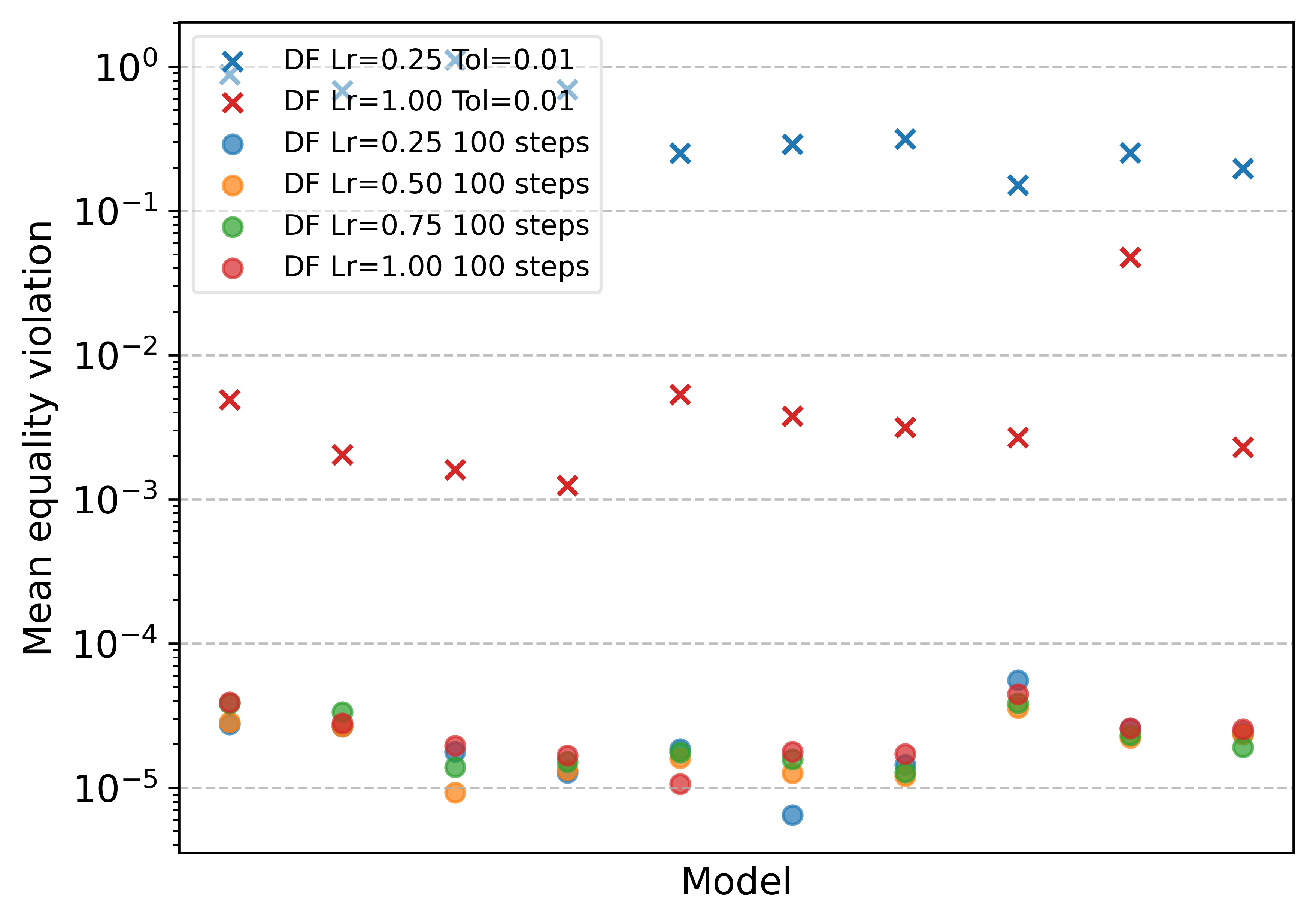}
    \caption[Large distance tolerance versus no tolerance comparison for the modified DeepFool algorithm (Task $2$)]{Comparison between margins and equality violations found using no stopping tolerance and a large tolerance using the modified DeepFool algorithm for $10$ models from Task $2$. Left: Mean margin per model. Right: Mean equality violation per model (logarithmic scale).}
    \label{fig:app_hidden_mean_margin_and_steps_tol_comparison_task_2}
\end{figure}

This section is related to Section~\ref{sec:hidden_verifying_deepfool_alg}. In Figure~\ref{fig:hidden_mean_margin_and_viol_comparison_small_and_large_tolerance} we compared the mean distance, equality violations, and number of optimization steps when calculating input space margins using Algorithm~\ref{alg:deepfool_margin}. Specifically, we established what effect a large tolerance ($\delta=0.01$) has on the margin, in comparison to using a small tolerance ($\delta=0.001$), for Task $1$. Here we repeat a similar analysis for Task $2$. We calculate the margins for $10$ models using no tolerance, i.e. for $100$ optimization steps, and compare this with using a large tolerance of $\delta=0.01$. The mean distance and equality violations for this analysis are shown in Figure~\ref{fig:app_hidden_mean_margin_and_steps_tol_comparison_task_2}.

\begin{figure}[h]
    \centering
    \includegraphics[width=0.49\linewidth]{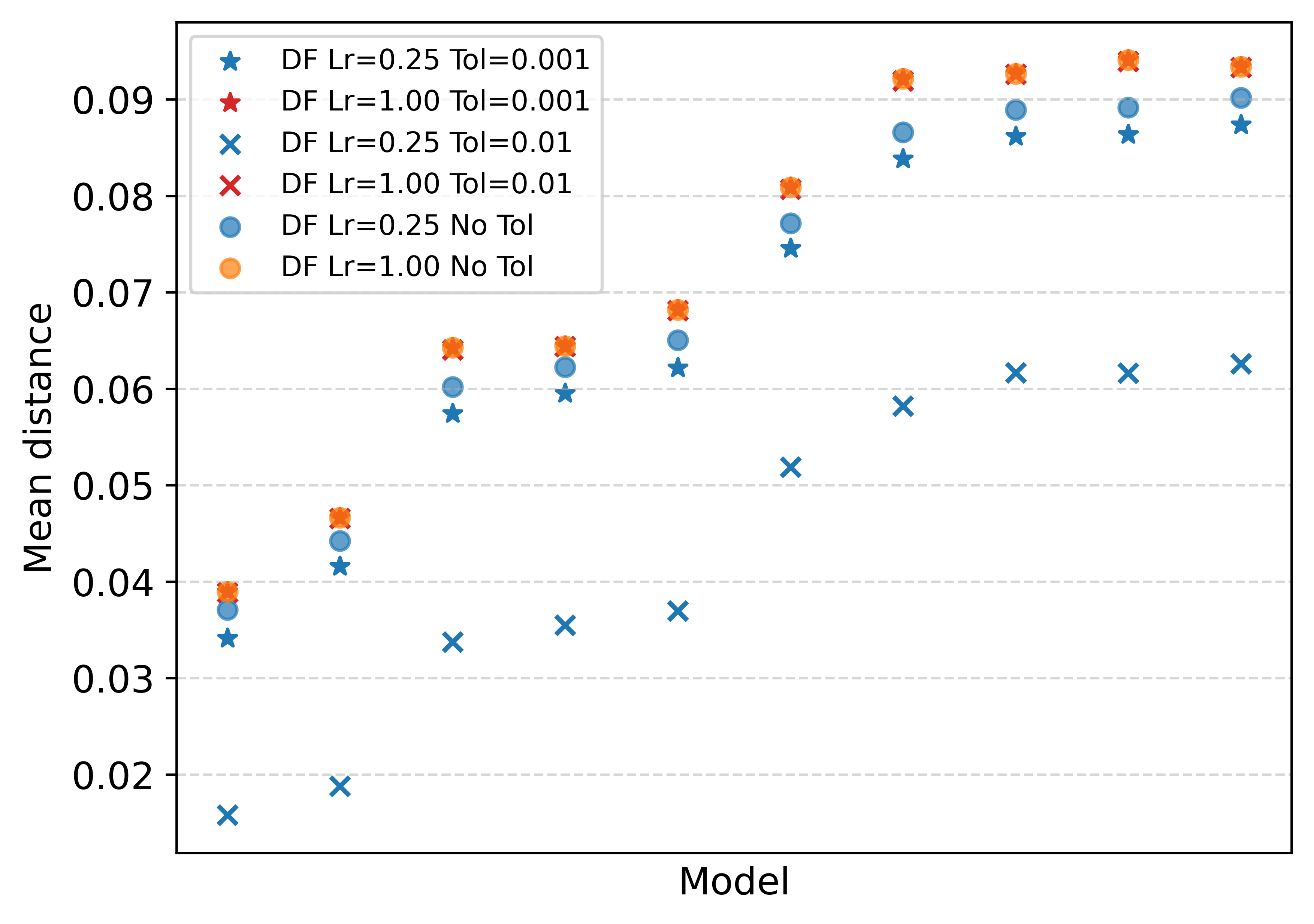}
    \includegraphics[width=0.49\linewidth]{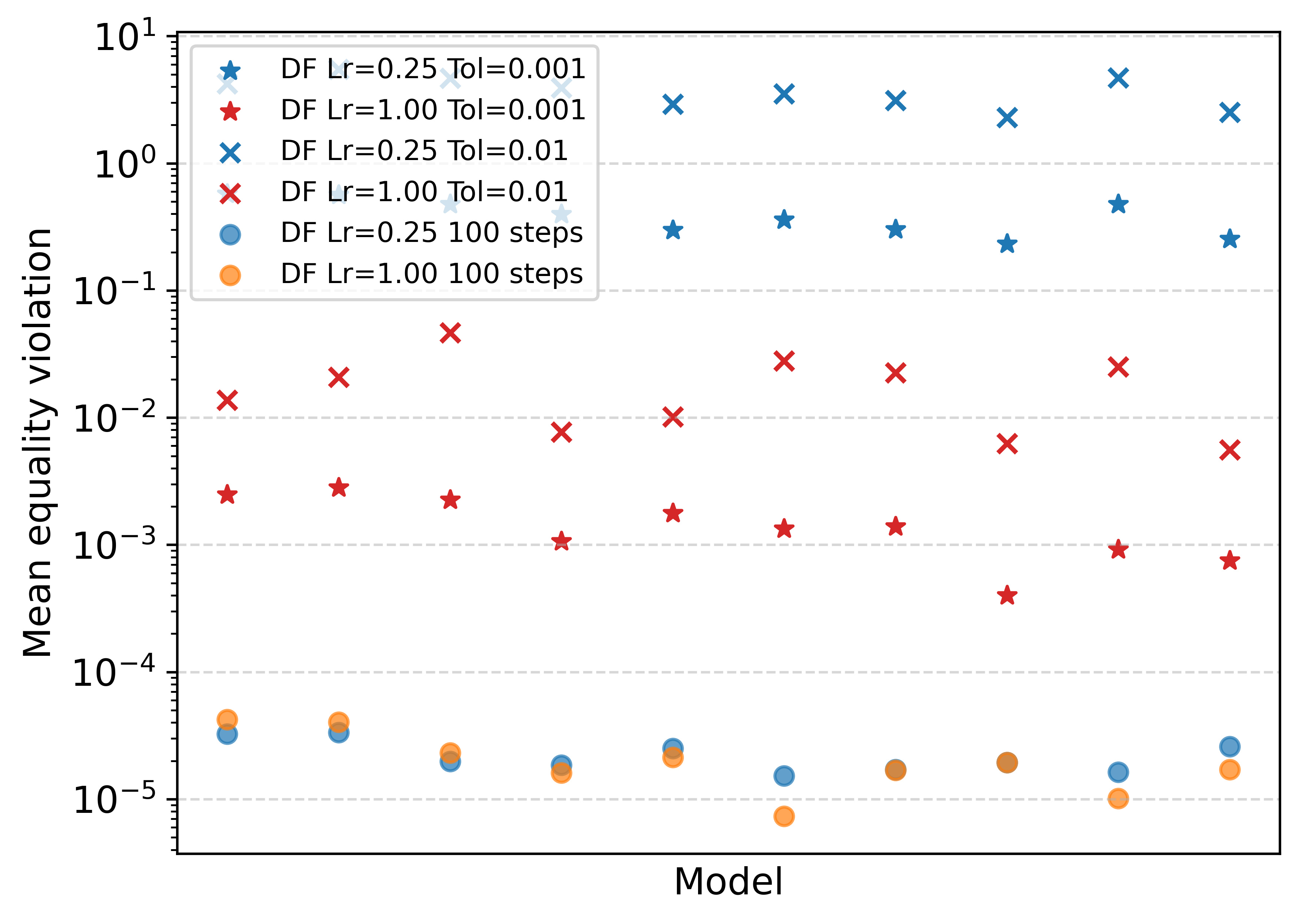}
    \caption[Large distance tolerance versus small tolerance versus no tolerance comparison for the modified DeepFool algorithm (Task $4$)]{Comparison between margins and equality violations found using no stopping tolerance, a large tolerance, and a small tolerance using the modified DeepFool algorithm for $10$ models from Task $4$. Left: Mean margin per model. Right: Mean equality violation per model (logarithmic scale).}
    \label{fig:app_hidden_mean_margin_and_steps_tol_comparison_task_4}
\end{figure}

We observe that the same trends hold that were established earlier. The larger tolerance methods (blue and red crosses) find virtually the same distance as their no tolerance counterparts (blue and red circles), albeit with higher equality violations. Additionally, we again observe that the difference in distance between the two $\gamma=0.25$ variants (blue stars and circles) is virtually constant across the $10$ models.

Let us now also repeat a similar analysis for Task $4$, although now we compare the no tolerance method with both a smaller and larger tolerance, such that $\delta=0.001$ and $\delta=0.01$, respectively. We do this for both the smallest learning rate ($\gamma=0.25$) and largest ($\gamma=1.0$). This is shown in Figure~\ref{fig:app_hidden_mean_margin_and_steps_tol_comparison_task_4}.

First consider the results for the smaller learning rate, $\gamma=0.25$. We observe that the distances are significantly smaller for the $\delta=0.01$ variant in comparison to the no tolerance method (blue crosses, in comparison to blue circles, left). We also observe that this large tolerance results in very high equality violations (blue crosses, in comparison to blue circles, right). This implies the points found are still rather far away from the decision boundary, and the distances should not be considered as `margins'. On the other hand, the smaller tolerance variant ($\delta=0.001$, blue stars) is more in line with what we have observed earlier on the other tasks. It is clear that the $\delta=0.001$ method results in distances that are only slightly smaller and the equality violations are in a more acceptable range. The reason why a smaller tolerance is required here, is that the distances are an order of magnitude smaller than for the other tasks. For example, compare the scale on the $y$-axis on the left of Figures \ref{fig:app_hidden_mean_margin_and_steps_tol_comparison_task_2} and \ref{fig:app_hidden_mean_margin_and_steps_tol_comparison_task_4}). For the $\gamma=1.0$ learning rate, we observe that the tolerance makes no obvious difference to the distance (red circles, stars, and crosses).

We suspect the reason for these small margins are simply that the input features are normalized to be in the range $[0, 1]$ for Tasks $4$ and $5$, while for the other tasks the data is $z$-normalized, and as such have a larger range and scale.

\subsection{Additional Taylor versus DeepFool comparison}
\label{app:hidden_taylor_v_deepfool}

In Section~\ref{sec:hidden_df_to_taylor_comp} we compare the size of the margins found with Taylor and DeepFool for several models from Task $1$ and $4$. For that calculation, we use $500$ samples and no distance tolerance for the DeepFool calculation per model. Here, we repeat the same analysis as found in Figures \ref{fig:hidden_mean_margin_taylor_v_deepfool_input_and_hidden_task_1} and \ref{fig:hidden_taylor_v_deepfool_margin_per_sample}, but make use of $5\ 000$ samples with a small distance tolerance as stopping criterion. Specifically, for $10$ models from Task $1$, we use a distance tolerance of $\delta=0.001$. For $10$ models from Task $4$, we use a distance tolerance of $\delta=0.0001$. Furthermore, we discard points with an equality violation higher than $0.1$. This results in very few points being discarded.

\begin{figure}[h]
    \centering
    \includegraphics[width=0.49\linewidth]{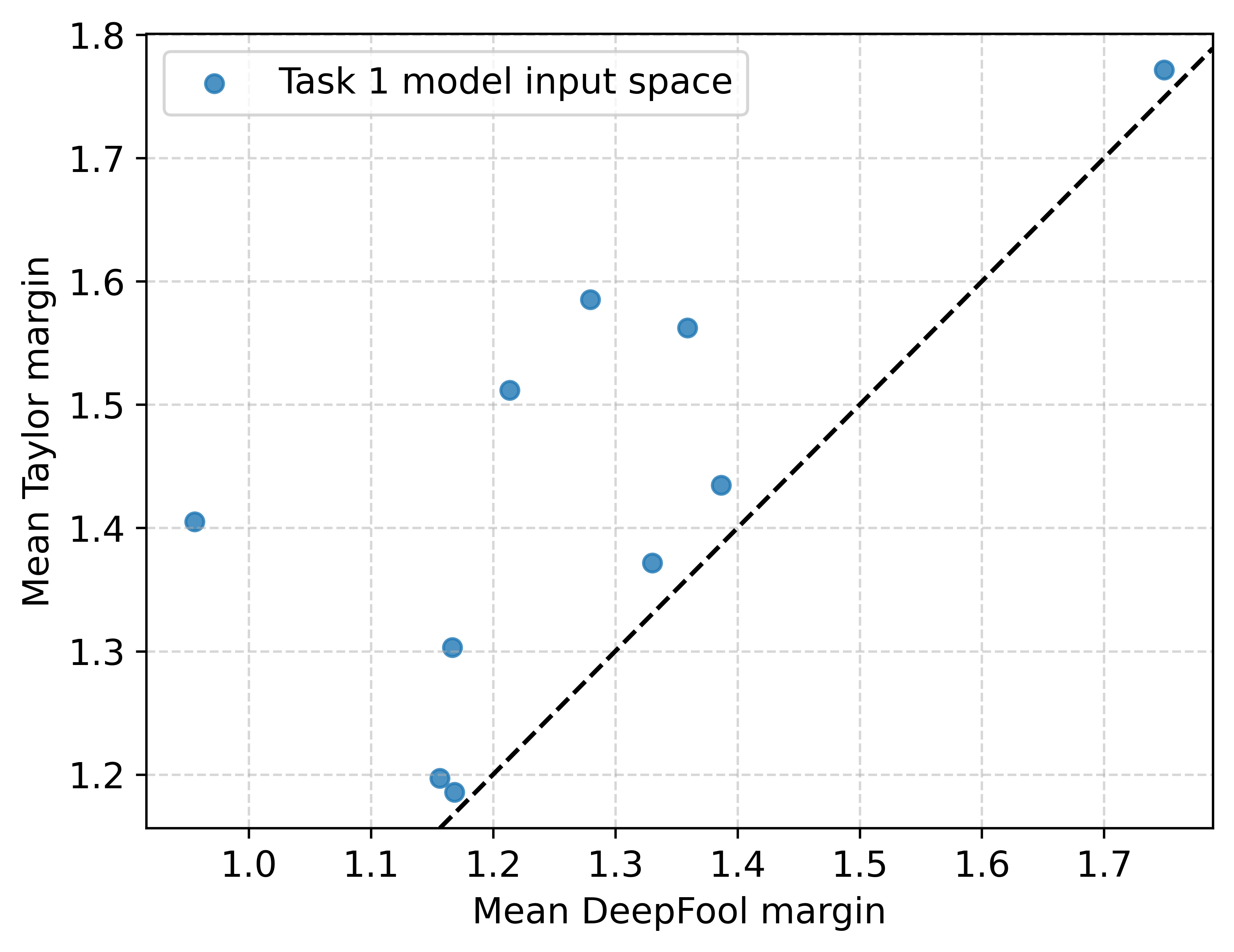}
    \includegraphics[width=0.49\linewidth]{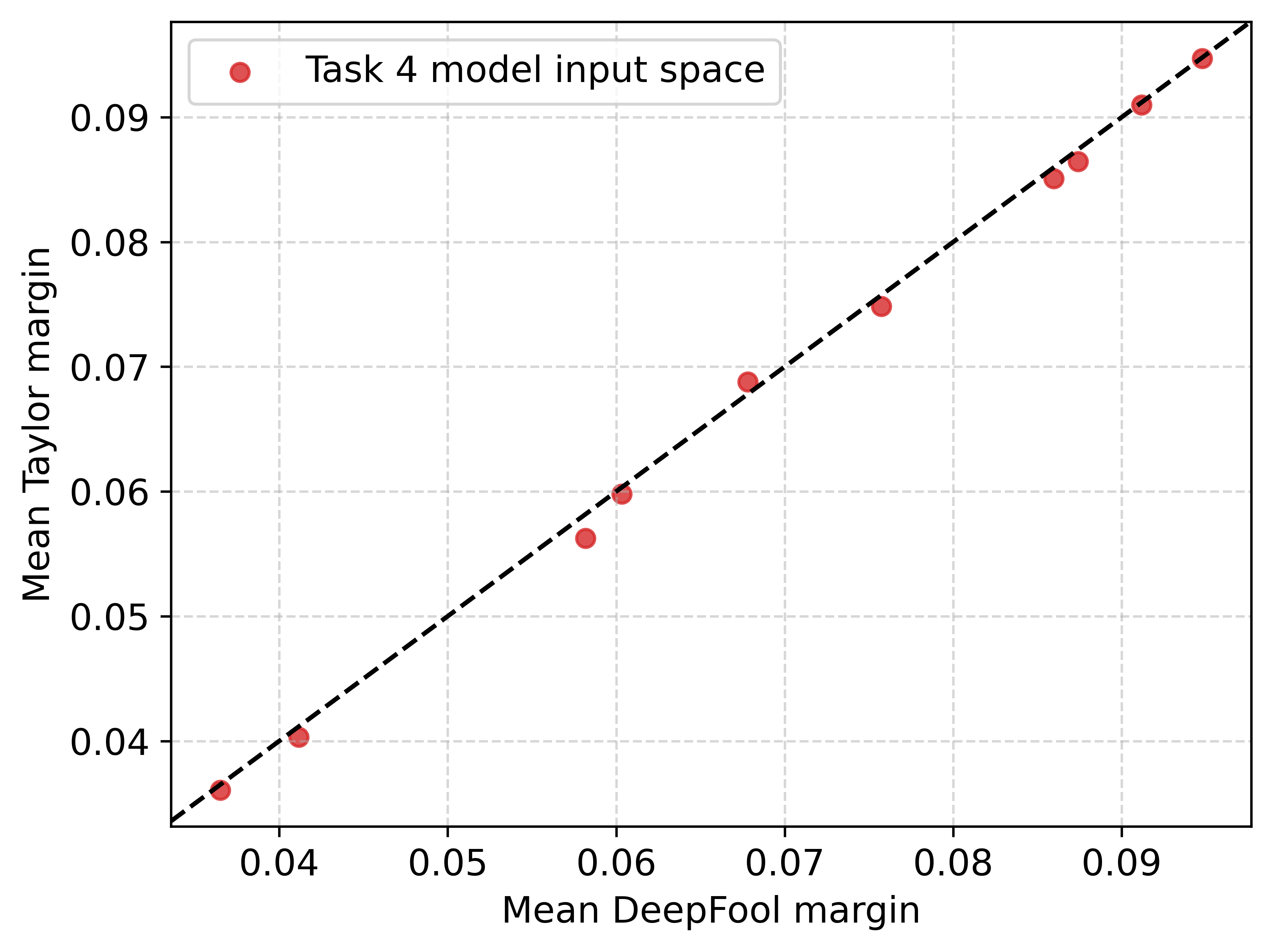}
    \includegraphics[width=0.49\linewidth]{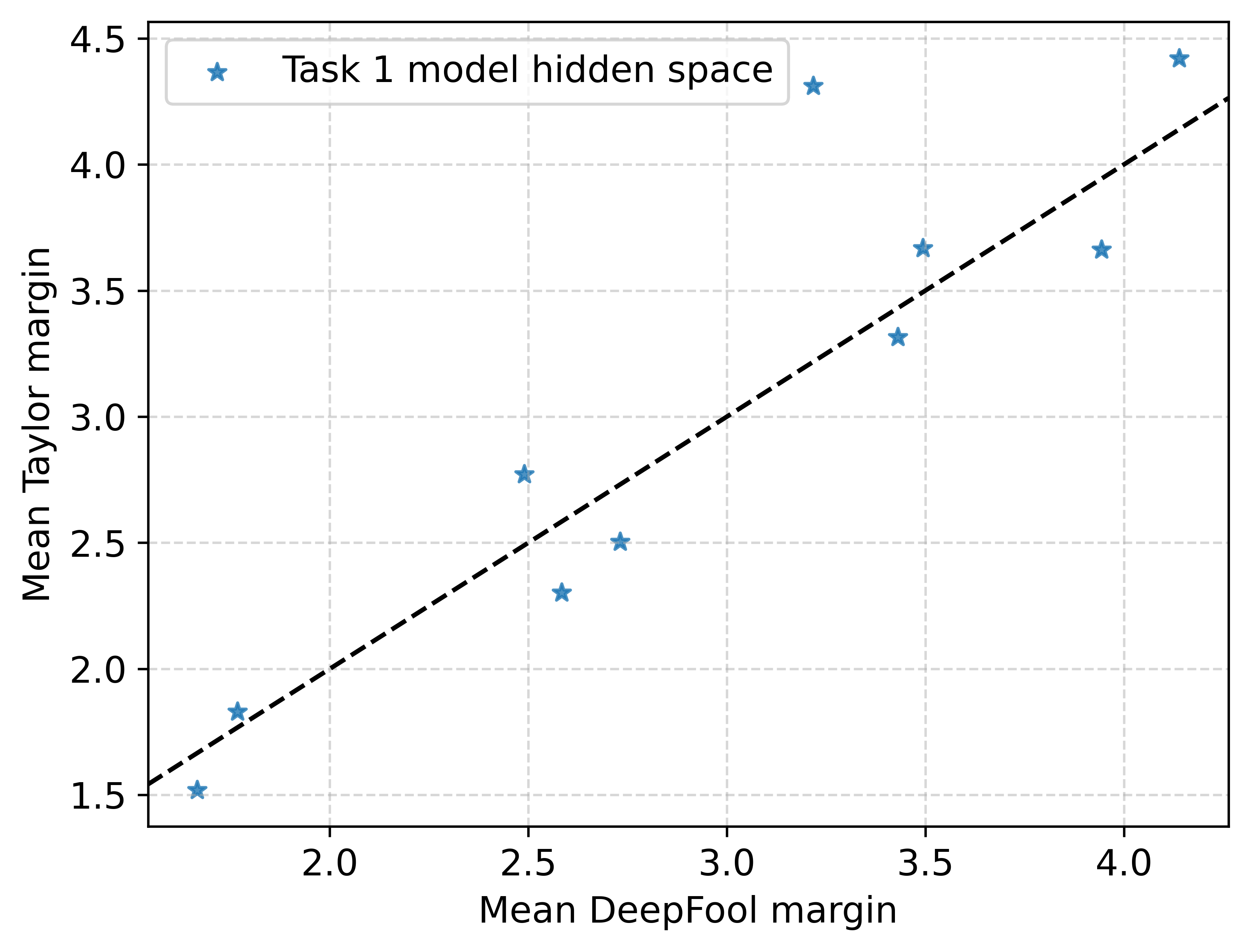}
    \includegraphics[width=0.49\linewidth]{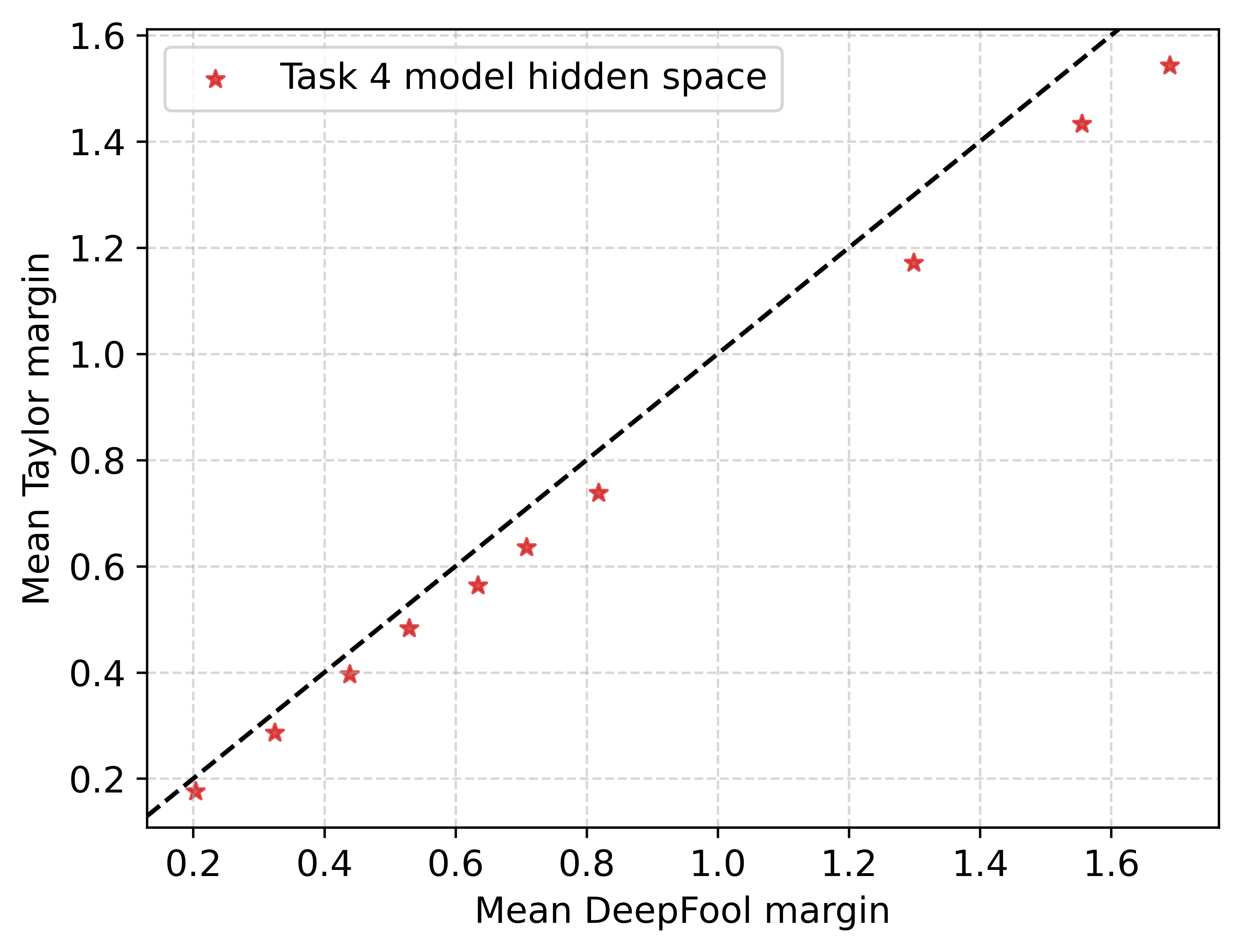}
    \caption[Mean Taylor versus DeepFool margin for $10$ models of Task 1 and 4 using a small distance tolerance]{Mean Taylor versus DeepFool margin for $10$ models of Task 1 (left) and 4 (right) using $5\ 000$ samples and small distance tolerance. Top: Input space. Bottom: First convolutional layer hidden space.}
    \label{fig:hidden_mean_margin_taylor_v_deepfool_input_and_hidden_task_1_with_tol}
\end{figure}

\begin{figure}[h]
    \centering
    \includegraphics[width=0.49\linewidth]{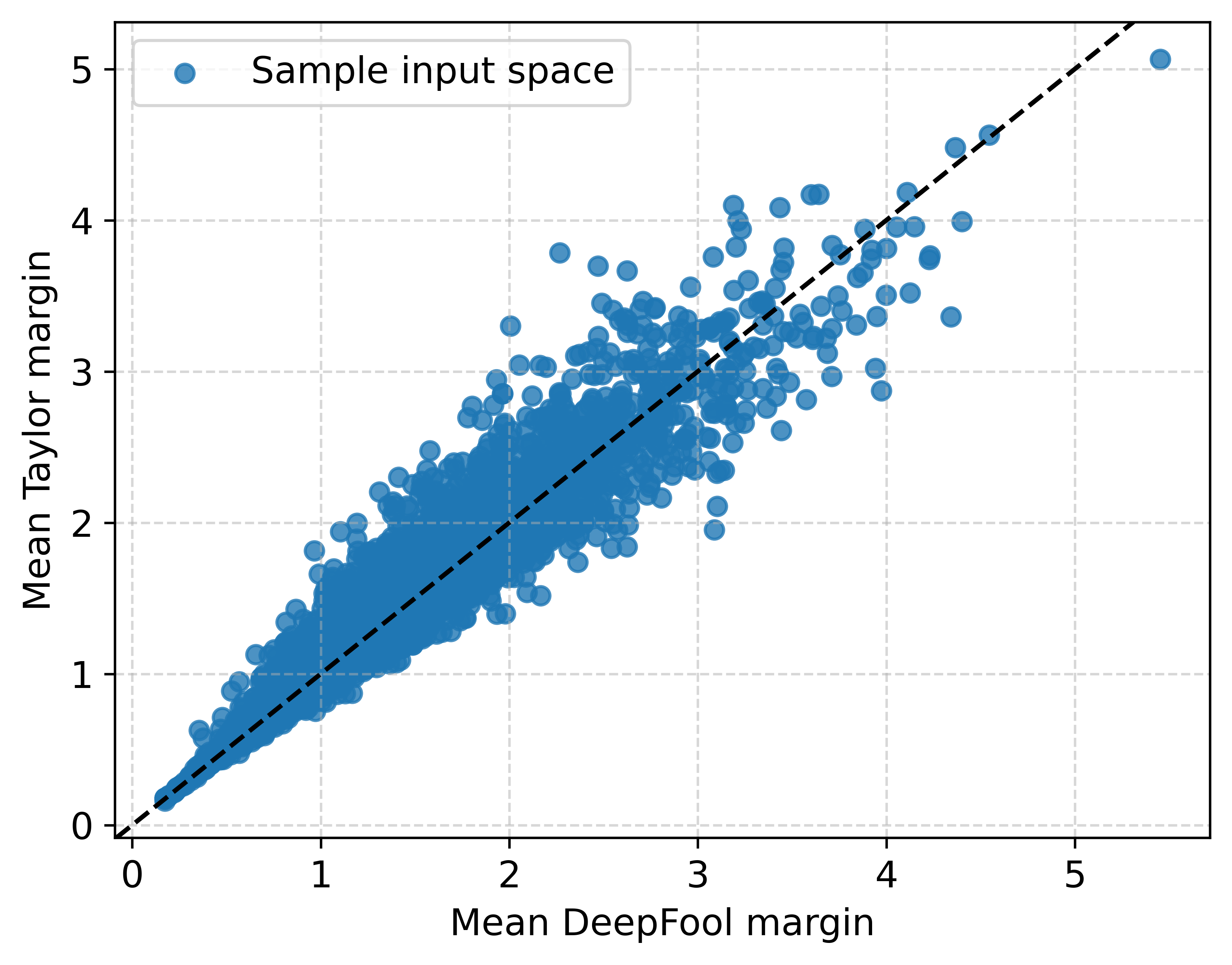}
    \includegraphics[width=0.49\linewidth]{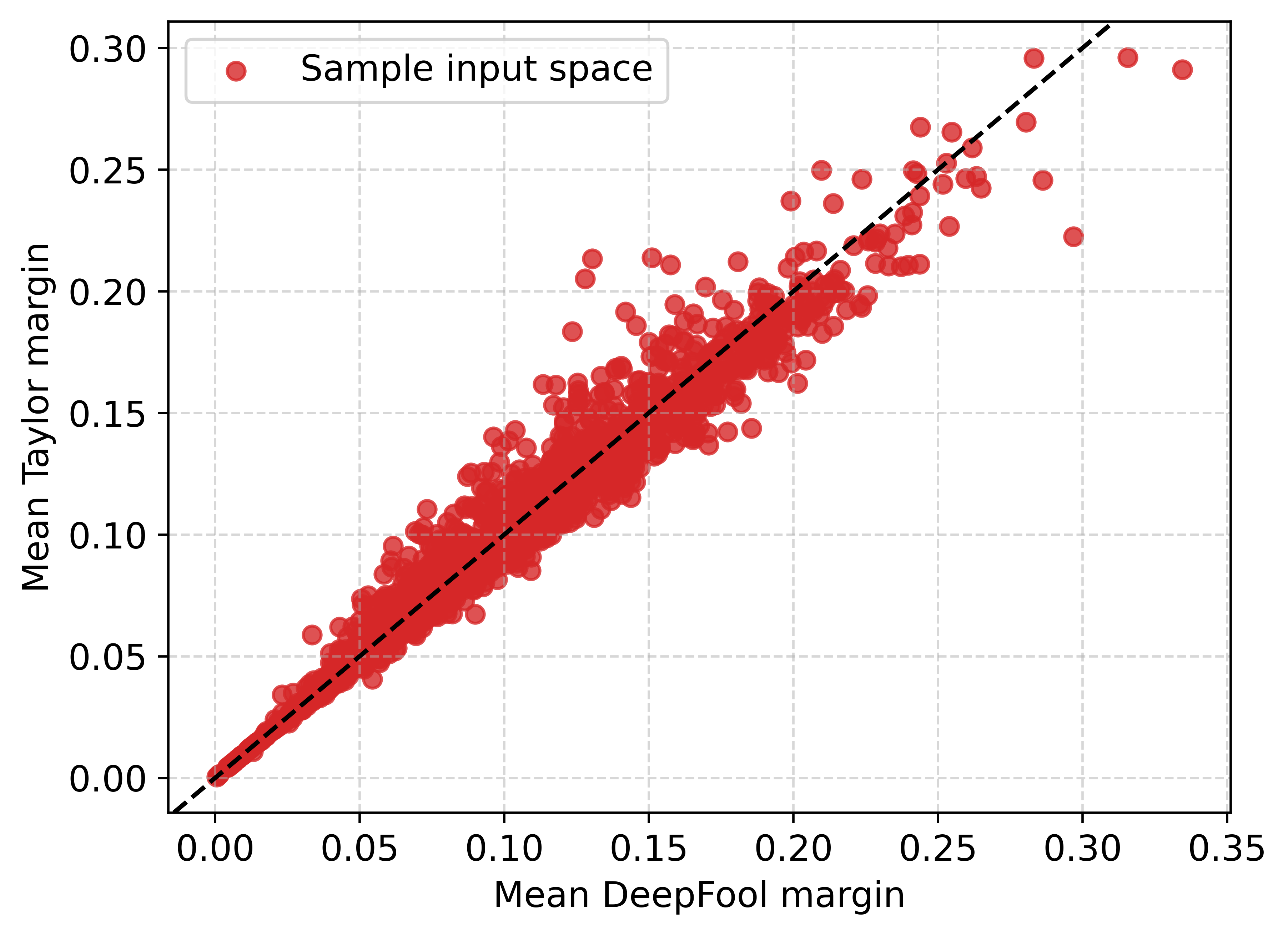}
    \includegraphics[width=0.49\linewidth]{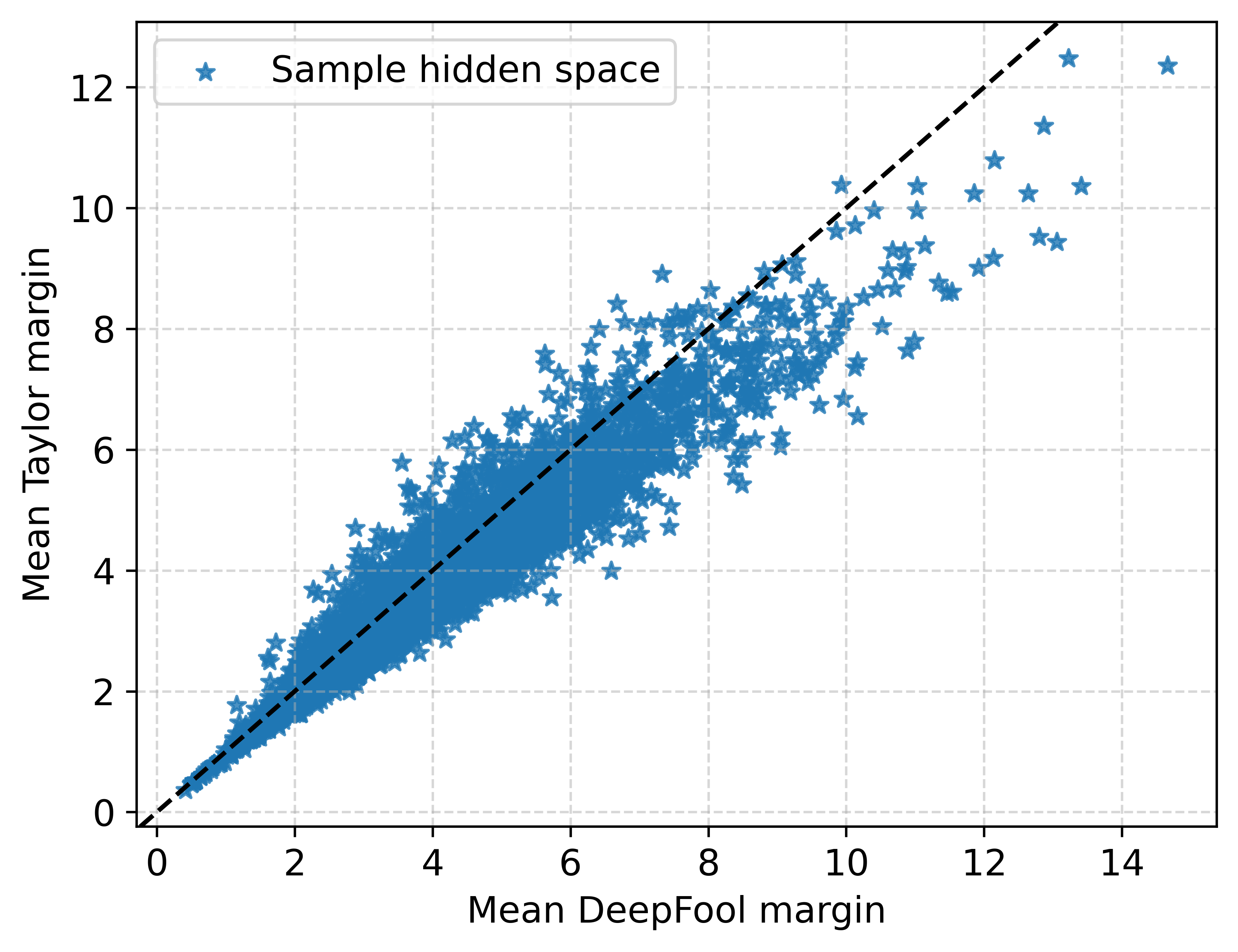}
    \includegraphics[width=0.49\linewidth]{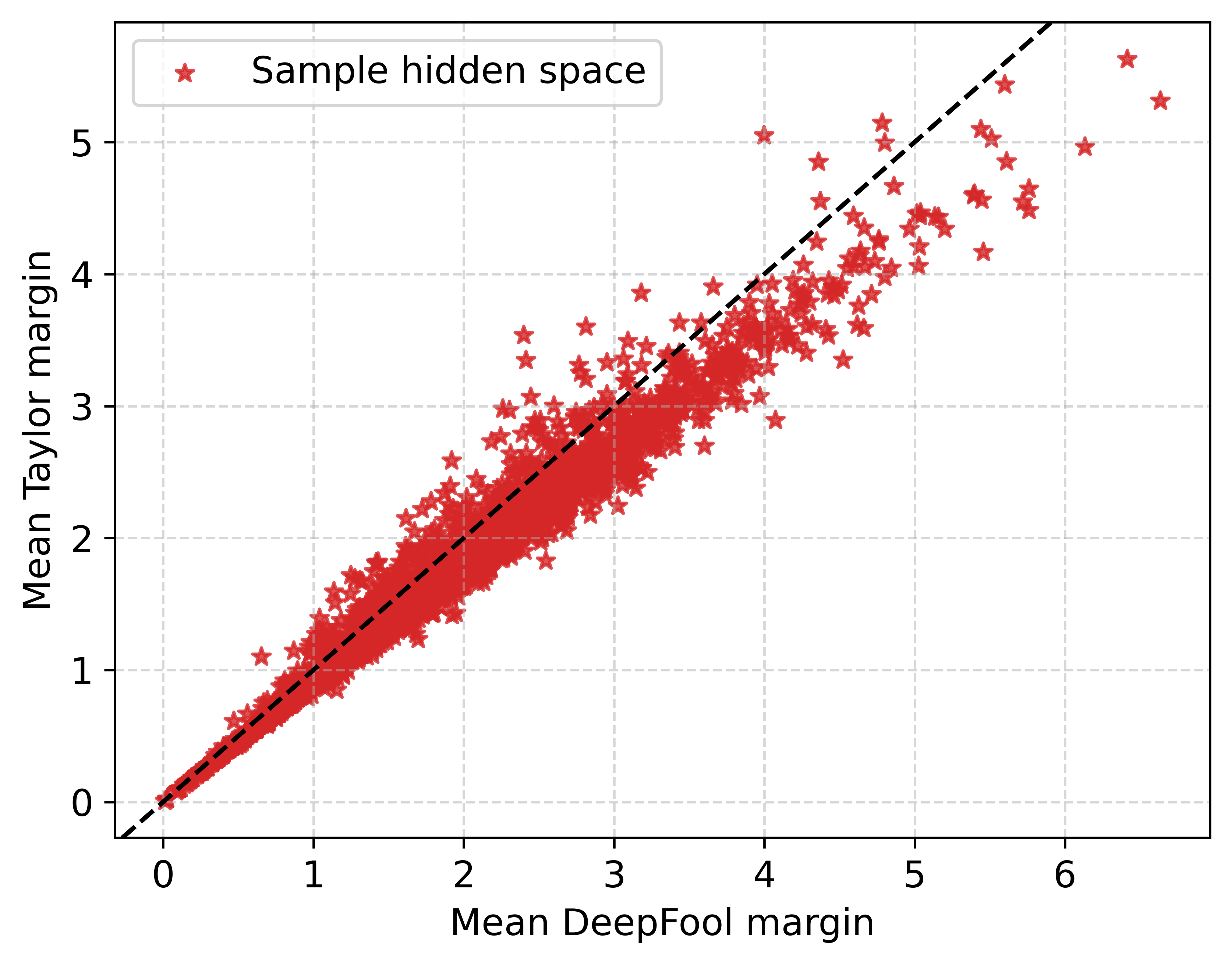}
    \caption[Taylor versus DeepFool margin (per sample) for a model from Task 1 and 4 using a small distance tolerance]{Taylor versus DeepFool margin for $5\ 000$ samples for a model from Task 1 (left) and 4 (right) using a small distance tolerance. Top: Input space. Bottom: First convolutional layer hidden space.}
    \label{fig:hidden_taylor_v_deepfool_margin_per_sample_with_tol}
\end{figure}

We then compare the mean DeepFool and Taylor approximated margins for these models in both the input and hidden space in Figure~\ref{fig:hidden_mean_margin_taylor_v_deepfool_input_and_hidden_task_1_with_tol}.  We also do a per-sample comparison for one model from each task in Figure~\ref{fig:hidden_taylor_v_deepfool_margin_per_sample_with_tol}. All other details are exactly the same as in Section~\ref{sec:hidden_df_to_taylor_comp}.


We observe that these results are highly similar to those shown in Section~\ref{sec:hidden_df_to_taylor_comp}, and all the same observations are present.

\subsection{Number of samples for Taylor and DeepFool margins}
\label{app:hidden_num_samples}

In Section~\ref{sec:hidden_df_pgdl_results}, we use only $5\ 000$ training samples to calculate input and hidden margins with DeepFool for computational efficiency. For the sake of comparison, we limit the Taylor approximation margins to the same number. Prior to Section~\ref{sec:hidden_df_pgdl_results}, we calculate the mean Taylor-approximated margins of all training samples for each model of each task (recall Sections \ref{sec:hidden_lin_reg}, \ref{sec:hidden_hidden_layer_selection}, and \ref{sec:hidden_failure_cases}). Several other authors also restrict the number of training samples for computational efficiency. In terms of Taylor-approximated margins, Natekar and Sharma~\cite{predict_gen_margin} rely only on $500$ randomly sampled training samples, and similarly Chuang et al.~\cite{optimal_transport} use $2\ 000$. Here, we determine what effect the number of samples has on the predictive performance of 1) the first-order Taylor approximation margin measurements and 2) the DeepFool margin measurements.

\begin{figure}[h]
    \centering
    \includegraphics[width=0.49\linewidth]{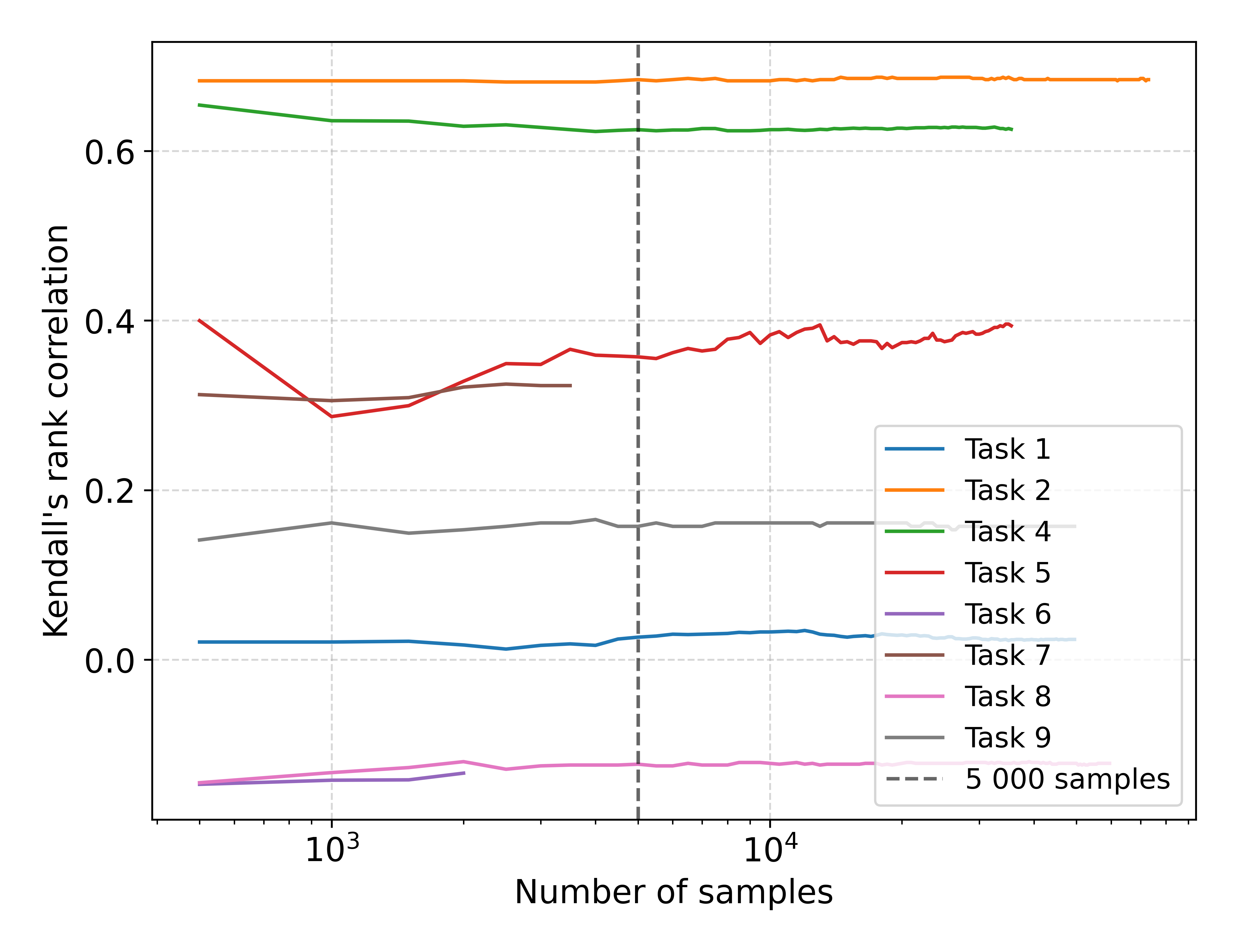}
    \includegraphics[width=0.49\linewidth]{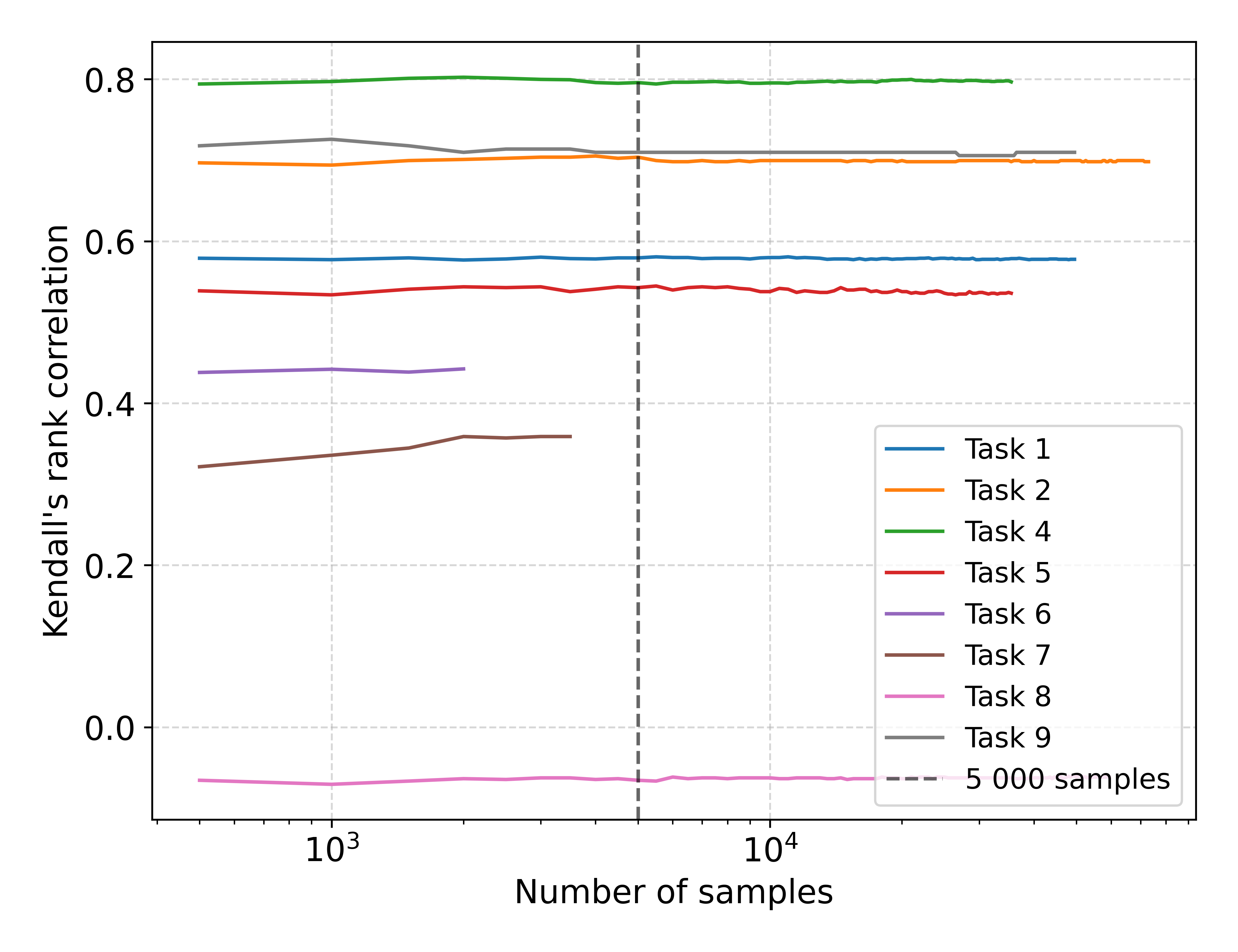}
    \caption[Number of samples versus predictive performance for Taylor input and hidden margins on the PGDL tasks]{Predictive performance of Taylor-approximated input margins (left) and hidden margins (1st layer, right)  as a function of the number of samples for all PGDL tasks. Note the logarithmic $x$-axis.}
    \label{fig:app_hidden_mean_margin_taylor_input_and_hidden_num_samples}
\end{figure}

We start with (1). We calculate the Kendall's rank correlation between the mean Taylor margin and test accuracy for each task using an increasing number of samples. Specifically, we vary the number of samples from $500$ to the maximum available. This is done for both the input and hidden space. For hidden margins, we perform this analysis using the best-performing hidden layer selection method, i.e. considering only the first layer. The result of this analysis is shown in Figure~\ref{fig:app_hidden_mean_margin_taylor_input_and_hidden_num_samples}. Note that we also indicate $5\ 000$ samples with a dashed vertical line, as that is the number of samples we have used for comparison in Section~\ref{sec:hidden_df_pgdl_results}.

It is observed that, in general, these metrics are not very sensitive to the number of samples considered, and only a small set of samples is required for adequate performance (as little as $500$ for some cases, while approximately $5\ 000$ for others, depending on the task). However, we observe that Task $5$, in both cases, shows a much larger variance in its predictive performance. That said, we observe that the use of $5\ 000$ samples very closely resembles using the maximum number of available samples for all the tasks considered. Let us consider the largest difference in rank correlation between using $5\ 000$ samples or the maximum number available across all tasks. For input margins, we find the largest difference of $0.039$ at Task $5$. For hidden margins, the largest difference is $0.006$ for Task $2$). These are rather unremarkable differences in ranking performance. 

\begin{figure}[h]
    \centering
    \includegraphics[width=0.49\linewidth]{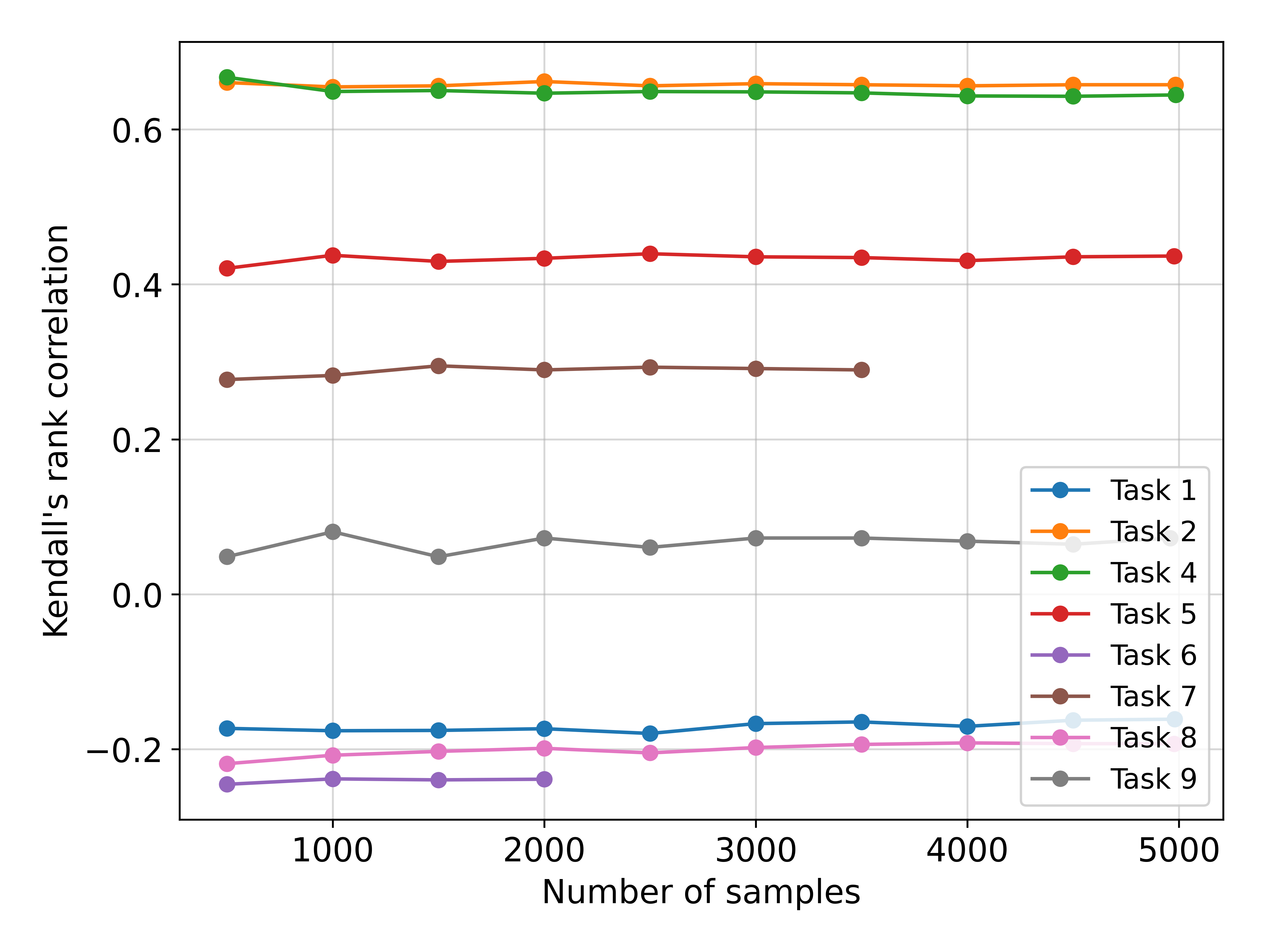}
    \includegraphics[width=0.49\linewidth]{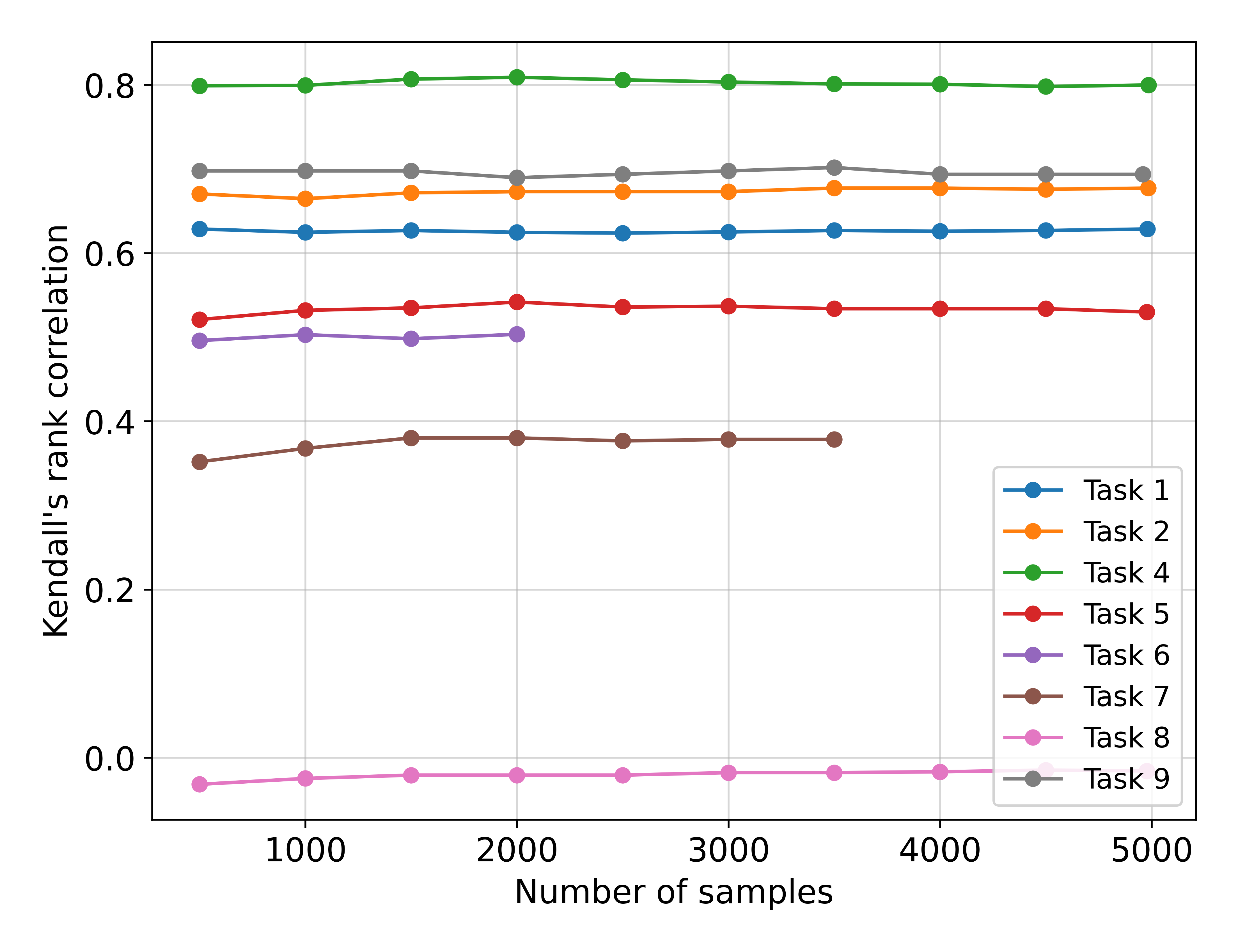}
    \caption[Number of samples versus predictive performance for DeepFool input and hidden margins on the PGDL tasks]{Predictive performance of DeepFool ($\gamma=0.25$) input margins (left) and hidden margins (1st layer, right) as a function of the number of samples for all PGDL tasks.}
    \label{fig:app_hidden_mean_margin_df_input_and_hidden_num_samples}
\end{figure}

We now turn to analysis (2) and repeat the same for the DeepFool margin variants with small learning rates ($\gamma=0.25$, recall our comparison in Table~\ref{tab:hidden_hidden_input_taylor_v_deepfool}). We calculate the Kendall's rank correlation between the DeepFool margin measurements and test accuracy using $500$ to $5\ 000$ samples. Recall that $5\ 000$ is the maximum we are able to use given our computational budget. In Figure~\ref{fig:app_hidden_mean_margin_df_input_and_hidden_num_samples} we show this for both the input space (left) and the hidden space (right).

The results for the DeepFool margins are somewhat similar to those of the Taylor-approximated margins. One observes that the number of samples does not have a large effect. Furthermore, it appears that these measurements are even less sensitive to the number of samples than the Taylor-approximated margins. From this we can conclude that $5\ 000$ samples are likely more than enough for our analysis and comparisons.
 
\section{Appendix: Chapter 5} 
\label{app:ch5}

This appendix contains additional results and explanations relevant to Chapter~\ref{chap:ch5}.

\subsection{Number of principal components}
\label{app:constrained_num_components}

\begin{figure}[h]
        \centering
        \includegraphics[width=0.49\linewidth]{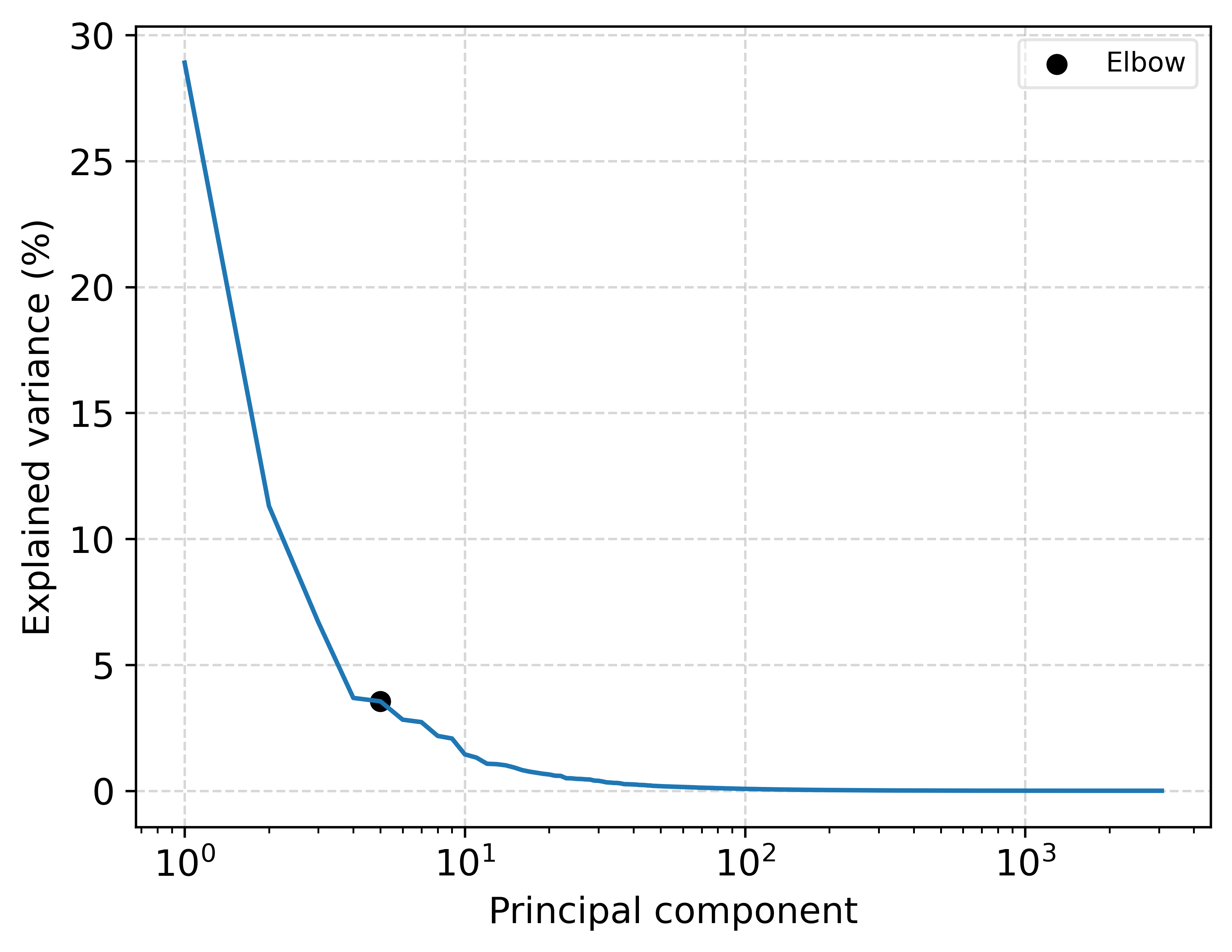}
        \includegraphics[width=0.49\linewidth]{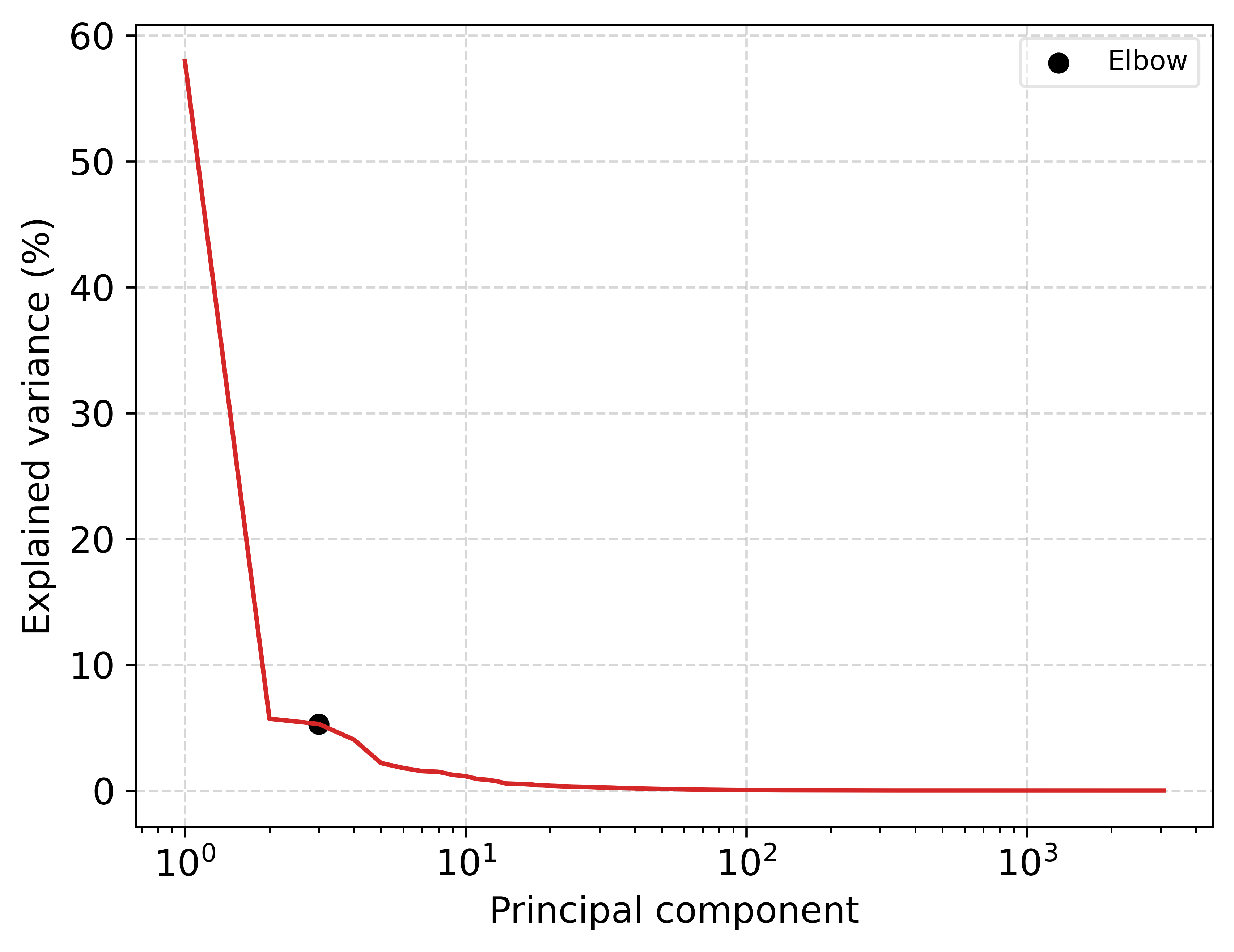}
        \caption[Percentage of explained variance per principal component for Task $1$ and $2$]{Percentage of explained variance per principal component for all principal components. Left: Task $1$ (blue line). Right: Task $2$ (red line). Black dot indicates the position of the elbow selected by the Kneedle algorithm.}
        \label{fig:app_constrained_explained_variance_curves}
\end{figure}

In Section~\ref{sec:constrained_experimental_setup} we explained that the number of principal components for each dataset is selected using an elbow method. More precisely: `We select this by plotting the explained variance (of the training data) per principal component in decreasing order on a logarithmic scale and applying the elbow method. Specifically, we use the Kneedle algorithm from Satopaa et al.~\cite{kneedle_elbow_method} to select the elbow.' Here we show what these curves typically look like. In Figure~\ref{fig:app_constrained_explained_variance_curves} we plot the percentage of variance explained by each principal component, for both the datasets of Task $1$ (left, CIFAR10) and $2$ (right, SVHN). We also indicate the position of the elbow as selected by the aforementioned Kneedle algorithm.

It is quite clear that the first few principal components capture the majority of the variance in the data. Furthermore, the Kneedle algorithm selects the elbow at a position that is well aligned with the human notion of an `elbow'.

\begin{figure}[h]
        \centering
        \includegraphics[width=1.0\linewidth]{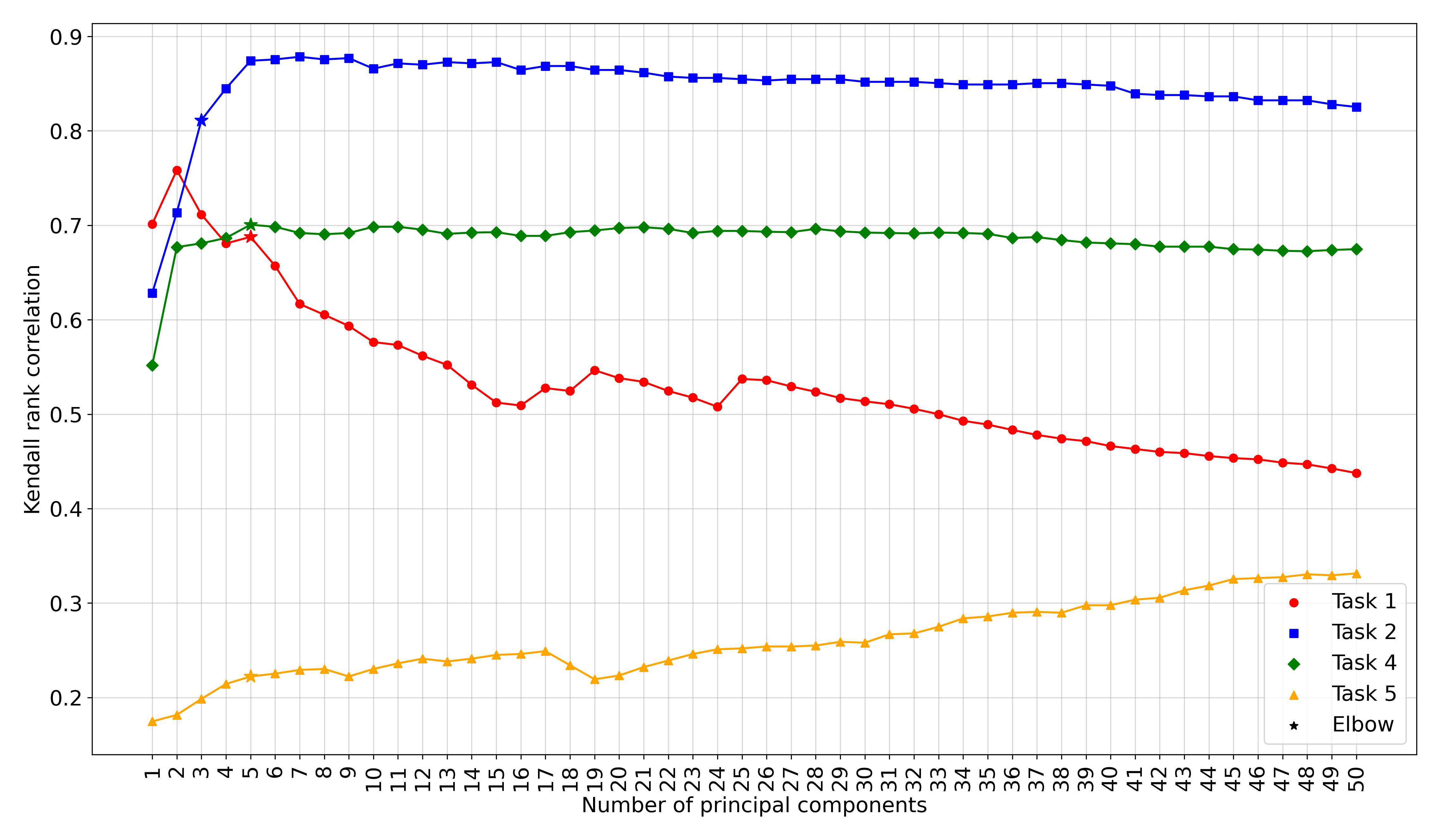}
        \caption[Number of principal components versus predictive performance for constrained margins for Task $1$, $2$, $4$, and $5$]{Predictive performance (Kendall's rank correlation) as a function of the number of principal components for Task 1 (red circles), 2 (blue squares), 4 (green diamonds), and 5 (yellow triangles). The number of principal components reported on per task in the main paper is indicated with a star.}
        \label{fig:app_constrained_effect_of_num_comps}
\end{figure}

Allow us to now show the interaction between the number of principal components and predictive performance of constrained margins. We calculate the mean constrained margin using $1$ to $50$ principal components for all the development set tasks (tasks $1$ to $5$).  We make use of $5\ 000$ samples as done throughout Chapter~\ref{chap:ch5}. However, in this case, the first-order Taylor approximation is used to reduce the computational burden.  The result of this analysis is shown in Figure~\ref{fig:app_constrained_effect_of_num_comps}.  We indicate the number of principal components selected by the Kneedle algorithm for each task with a star.

One observes that the elbow method selects the number of components in a near-optimal fashion for Task $1$, $2$, and $4$.  Furthermore, the optimal number is generally very low, whereafter the correlation decreases.  
Task $5$ (which is the task for which constrained margins produce the lowest performance) behaves in a contrary manner, as the ranking correlation increases as the number of components becomes larger.  We find that it only reaches a maximum rank correlation of $0.4$ at $270$ components (not shown here). In Table~\ref{tab:app_constrained_num_comps_num_samples} we show the number of principal components selected for each dataset of each Task.

\subsection{Number of samples}
\label{app:constrained_num_samples}

\begin{figure}[h]
        \centering
        \includegraphics[width=0.75\linewidth]{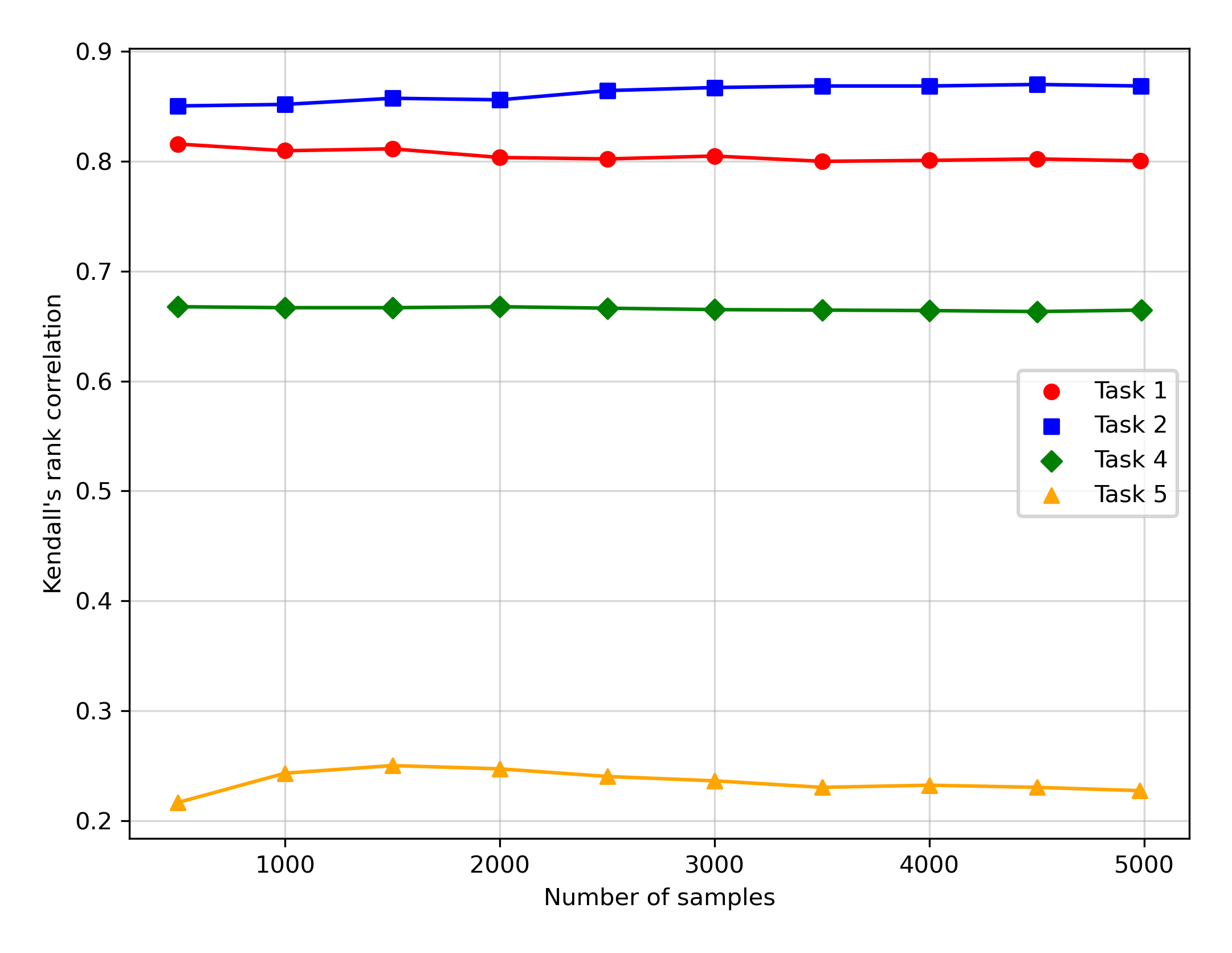}
        \caption[Number of samples versus predictive performance for constrained margins for Task $1$, $2$, $4$, and $5$]{Predictive performance of constrained margins (Kendall's rank correlation) as a function of the number of samples for Task 1 (red circles), 2 (blue squares), 4 (green diamonds), and 5 (yellow triangles).}
        \label{fig:app_constrained_effect_of_num_samples}
\end{figure}

In Section~\ref{sec:constrained_results} we had used $5\ 000$ samples to calculate the mean constrained margin for each task (and the same number for all other margin measurements). It is worth determining what effect the number of samples has on the final performance. In Figure~\ref{fig:app_constrained_effect_of_num_samples} we show the Kendall's rank correlation between mean constrained margin and test accuracy for the development set using $500$ to $5\ 000$ samples (using Algorithm~\ref{alg:manifol_margin}).

One observes that the rank correlation plateaus rather quickly for most tasks, and one can likely get away with only using $500$ to $1\ 000$ samples per model. However, to mitigate any effect that the stochastic selection of training samples can have on the reproducibility of the results, we have chosen to use $5\ 000$ throughout. For the sake of reproducibility, we show the number of principal components selected, as well as the number of samples used for each task in Table~\ref{tab:app_constrained_num_comps_num_samples}. Note that Tasks $6$ and $7$ use the maximum number of samples available.

\begin{table}[H]
\centering
\caption[Number of principal components and samples used for each task to calculate constrained margins]{Number of principal components and samples used for each task to calculate constrained margins. Tasks 6 and 7 use the maximum number of samples available for the dataset.}
\label{tab:app_constrained_num_comps_num_samples}
\begin{tabular}{clcc}
\hline
\textbf{Task} & \textbf{Dataset} & \textbf{Components} & \textbf{Samples} \\ \hline
1 & CIFAR10 & 5 & 5 000 \\
2 & SVHN & 3 & 5 000 \\
4 & CINIC10 & 5 & 5 000 \\
5 & CINIC10 & 5 & 5 000 \\
6 & OxFlowers & 8 & 2 040 \\
7 & OxPets & 3 & 3 680 \\
8 & FMNIST & 4 & 5 000 \\
9 & \begin{tabular}[c]{@{}l@{}}CIFAR10\\ (augmented)\end{tabular} & 5 & 5 000
\end{tabular}
\end{table}




\subsection{Derivation of constrained margins Equation~(\ref{eq:constrained_meaningful_taylor_approx})}
\label{app:constrained_derivation}

In this section we provide the derivation of Equation~\ref{eq:constrained_meaningful_taylor_approx} first presented in Section~\ref{sec:constrained_approximation_formulation}. We use the same notation as used earlier.  We first describe the standard linear approximation of the margin following Huang et al.~\cite{huang2016learning}, before deriving the constrained margin of Equation~\ref{eq:constrained_meaningful_taylor_approx}.

Consider any differentiable vector-valued function $f~:~\mathbf{x}\rightarrow \mathbb{R}^{c}$, with input $\mathbf{x} \in \mathbb{R}^n$.  This function can be approximated by its differential at point $\mathbf{x}$ using
\begin{eqnarray}
\hat{f}({\mathbf x}+{\mathbf d}) & = & f({\mathbf x}) + \mathbf{J} {\mathbf d} \label{eq:lin_approx} 
\end{eqnarray}
where
\begin{equation}
    \mathbf{J} =  
    \begin{bmatrix}
     \nabla_{\mathbf{x}} f_1(\mathbf{x})^T\\
     \nabla_{\mathbf{x}} f_2(\mathbf{x})^T\\
     ...\\
     ...\\
     \nabla_{\mathbf{x}} f_c(\mathbf{x})^T\\
    \end{bmatrix}
\end{equation}

that is, the Jacobian of the output with respect to the input features at point ${\mathbf x}$, where $\nabla_{\mathbf{x}} f_k(\mathbf{x})$ is the gradient vector of the $k^{th}$ output value of $f$ with regard to input $\mathbf{x}$.  
We aim to find the smallest $||{\bf d}||$ for some norm $||.||$ such that $\argmax(f({\bf x})) \neq \argmax(f({\bf x}+{\bf d}))$, or
\begin{eqnarray}
f_j({\mathbf x}+{\mathbf d}) & \geq & f_i({\mathbf x}+{\mathbf d})
\end{eqnarray}
where $f_i$ and $f_j$ indicate the scalar output of $f$ at index $i$ or $j$, respectively, with $i = \argmax (f({\bf x}))$ and $j = \argmax (f({\bf x} + {\mathbf d}))$.
If we approximate $f(.)$ with $\hat{f}(.)$, this implies:
\begin{eqnarray}
f_j({\mathbf x}) + \nabla_{\mathbf{x}} f_j(\mathbf{x}) \cdot {\mathbf d}  & \geq & f_i({\mathbf x}) + \nabla_{\mathbf{x}} f_i(\mathbf{x}) \cdot {\mathbf d} \notag \\
\implies (\nabla_{\mathbf{x}} f_j(\mathbf{x}) - \nabla_{\mathbf{x}} f_i(\mathbf{x})) \cdot {\mathbf d} & \geq & f_i({\mathbf x}) - f_j({\mathbf x})
\end{eqnarray}
Then, as shown in~\cite{huang2016learning}, the minimum $||{\mathbf d}||$ will be at:
\begin{eqnarray}
\label{eq:orig_lin}
||{\mathbf d}|| & = & {f_i({\mathbf x}) - f_j({\mathbf x})\over{||\nabla_{\mathbf{x}} f_j(\mathbf{x}) - \nabla_{\mathbf{x}} f_i(\mathbf{x})||_*}}
\end{eqnarray}
where $||.||$ and $||.||_*$ are dual norms. 
Specifically, if  $||.||$ is the $L_2$ norm, then:
\begin{eqnarray}
\label{eq:std1}
{\mathbf d}  
 =  {{f_i({\mathbf x}) - f_j({\mathbf x})}
\over{||\nabla_{\mathbf{x}} f_j(\mathbf{x}) - \nabla_{\mathbf{x}} f_i(\mathbf{x}) ||^2_2}}(\nabla_{\mathbf{x}} f_j(\mathbf{x}) - \nabla_{\mathbf{x}} f_i(\mathbf{x})) 
\end{eqnarray}
\begin{eqnarray}
\label{eq:std2}
\textrm{and }  ||{\mathbf d}||_2
 =  {{f_i({\mathbf x}) - f_j({\mathbf x})}\over{||\nabla_{\mathbf{x}} f_j(\mathbf{x}) - \nabla_{\mathbf{x}} f_i(\mathbf{x})||_2}}
\end{eqnarray}
Equations \ref{eq:std1} and \ref{eq:std2} provide the standard linear approximation of the margin as used by various authors~(\cite{predict_gen_margin, large_margin_dnns}). 

The derivation process for constrained margins is identical -- it is only the calculation of the Jacobian that differs, as the gradient is calculated with regard to the transformed features rather than the original features. 
Note that the size and direction of the update are calculated with regard to the transformed features but the actual step is given in the original feature space.

Let $\mathbf{P}_m$ be the matrix constructed from the first $m$ principal components as column vectors: 
\begin{eqnarray}
\mathbf{P}_m = [ {\bf p}_1, {\bf p}_2, .... {\bf p}_m]^T
\end{eqnarray}
The new parameterisation ${\bf x}'$ of any point ${\bf x}$ can then be approximated by:
\begin{eqnarray}
\label{eq:mm_project}
{\mathbf x}' \approx \mathbf{P}_m{\mathbf x}
\end{eqnarray}
where ${\mathbf x}$ is a column vector. 
Let $\mathbf{B}_m$ be the pseudoinverse of $\mathbf{P}_m$, that is,
\begin{eqnarray}
{\mathbf x} & \approx & \mathbf{B}_m{\mathbf x'}  \\
\textrm{or } 
x_r & \approx & \sum_{l=1}^m b_{r, l} x'_l 
\end{eqnarray}
where $b_{r,l}$ is the entry at row $r$ and column $l$ in $\mathbf{B}_m$.
%
%
%
Assuming $n$ input features, we can use the existing $c \times n$ Jacobian $\mathbf{J}$, to calculate the new $c \times m$ Jacobian $\mathbf{J}'$ in terms of ${\mathbf x'}$ rather than ${\mathbf x}$, using the chain rule.
The element $j'_{r,l}$ in the new $\mathbf{J}'$ at row $r$ and column $l$ will then be given by:
\begin{eqnarray}
\label{eq:scaled_hessian}
j'_{r,l} 
& = & {{\delta f_r({\bf x})}\over{\delta x'_l}}  \notag \\
& = & {{\delta f_r({\bf x})}\over{\delta x_1}} . {{d x_1}\over{d x'_l}} + ... + {{\delta f_r({\bf x})}\over{\delta x_n}} . {{d x_n}\over{d x'_l}} \notag \\
& = & j_{r,1}b_{1,l} + ... + j_{r,n}b_{n,l} \notag \\
& = & \nabla_{\mathbf{x}} f_r(\mathbf{x}) \cdot {\mathbf b}_l 
\end{eqnarray}
where ${\mathbf b}_l$ is the $l^{th}$ column of $\mathbf{B}_m$, 
and $j_{r,l}$ indicates the element at row $r$ and column $l$ of the original Jacobian $\mathbf{J}$. 
Then each row ${\mathbf j}'_r$ of the new Jacobian in terms of ${\mathbf x'}$ is given by
\begin{eqnarray}
\label{eq:adjusted_h}
{\mathbf j}'_r & = &  \nabla_{\mathbf{x}} f_r(\mathbf{x}) \mathbf{B}_m
\end{eqnarray}
which can be used directly in the adjusted version of Equations \ref{eq:std1} and \ref{eq:std2}, such that
\begin{eqnarray}
 {\mathbf d} &
 =  & {{f_i({\mathbf x}) - f_j({\mathbf x})}\over
 {||{\mathbf j}'_j - {\mathbf j}'_i ||_2^2}}
  ({\mathbf j}'_j - {\mathbf j}'_i)
 \notag \\
 & =  & {{f_i({\mathbf x}) - f_j({\mathbf x})}\over{|| ( \nabla_{\mathbf{x}} f_j(\mathbf{x}) - \nabla_{\mathbf{x}} f_i(\mathbf{x})) \mathbf{B}_m ||_2^2}}  (\nabla_{\mathbf{x}} f_j(\mathbf{x}) - \nabla_{\mathbf{x}} f_i(\mathbf{x})) \mathbf{B}_m 
 \label{eq:constrained_Bm}
 \\
\textrm{and }
 ||{\mathbf d}||_2 
 & =  & {{f_i({\mathbf x}) - f_j({\mathbf x})}\over{|| [ \nabla_{\mathbf{x}} f_j(\mathbf{x}) - \nabla_{\mathbf{x}} f_i(\mathbf{x})] \mathbf{B}_m ||_2}}
 \end{eqnarray}
 
In effect, we start at point ${\bf x}$ (the only point we have a model output for), and then use the gradient in the lower-dimensional space to find the minimal distance $||{\mathbf d}||_2$.

In the case where $\mathbf{P}_n$ is a full rank matrix, $\mathbf{B}_m$ can be constructed directly from $\mathbf{P}_n$.
Since the full $\mathbf{P}_n$, when all components are selected, is orthogonal, $(\mathbf{P}_n)^{-1} = (\mathbf{P}_n)^T$ and
$\mathbf{B}_m$ then equals the first $m$ rows of $(\mathbf{P}_n)^T$ and Equation \ref{eq:constrained_Bm} becomes
\begin{eqnarray}
 {\mathbf d} &
 =  & {{f_i({\mathbf x}) - f_j({\mathbf x})}\over
 {||{\mathbf j}'_j - {\mathbf j}'_i ||_2^2}}
  ({\mathbf j}'_j - {\mathbf j}'_i)
 \notag \\
 & =  & {{f_i({\mathbf x}) - f_j({\mathbf x})}\over{|| ( \nabla_{\mathbf{x}} f_j(\mathbf{x}) - \nabla_{\mathbf{x}} f_i(\mathbf{x})) \mathbf{P}_m^T ||_2^2}}  (\nabla_{\mathbf{x}} f_j(\mathbf{x}) - \nabla_{\mathbf{x}} f_i(\mathbf{x})) \mathbf{P}_m^T 
 \end{eqnarray}
 where $\mathbf{P}_m$ as above.
This more intuitive form is used in much of the discussion of constrained margins, while in practice, the pseudo-inverse is always calculated.